\documentclass{article}

\usepackage[nonatbib,preprint]{neurips_2022}

\usepackage[utf8]{inputenc} 
\usepackage[T1]{fontenc}    
\usepackage{hyperref}       
\usepackage{url}            
\usepackage{booktabs}       
\usepackage{amsfonts}       
\usepackage{nicefrac}       
\usepackage{microtype}      
\usepackage{xcolor}         

\usepackage{amsmath}
\usepackage{bbm}
\usepackage{physics}
\usepackage{graphicx}
\newsavebox{\measurebox}
\usepackage{subcaption}
\usepackage{tikz}
\usetikzlibrary{automata, arrows.meta, positioning}
\usepackage[export]{adjustbox}
\usepackage{float}
\usepackage{multirow}
\usepackage{array,makecell}
\usepackage{wrapfig}
\usepackage{csquotes}
\usepackage{tabularx}
\usepackage{setspace}

\DeclareMathOperator*{\argmax}{arg\,max}

\definecolor{1st}{HTML}{D81B60}
\definecolor{2nd}{HTML}{1E88E5}
\definecolor{3rd}{HTML}{FFC107}
\definecolor{4th}{HTML}{004D40}

\title{GaussianMLR: Learning Implicit Class Significance via Calibrated Multi-Label Ranking}

\author{%
  V. Bugra Yesilkaynak$^*$† \\
  Technical University of Munich \\
  \texttt{bugra.yesilkaynak@tum.de} \\
  \And
  Emine Dari$^*$ \\
  Istanbul Technical University \\
  \texttt{dari18@itu.edu.tr} \\
  \AND
  Alican Mertan \\
  University of Vermont \\
  \texttt{alican.mertan@uvm.edu} \\
  \And
  Gozde Unal \\
  Istanbul Technical University \\
  \texttt{gozde.unal@itu.edu.tr} \\
}

\begin{document}

\maketitle
\def\thefootnote{*}\footnotetext{These authors contributed equally to this work}
\def\thefootnote{†}\footnotetext{Contact author}

\begin{abstract}

Existing multi-label frameworks only exploit the information deduced from the bipartition of the labels into a positive and negative set. Therefore, they do not benefit from the ranking order between positive labels, which is the concept we introduce in this paper. We propose a novel multi-label ranking method: GaussianMLR, which aims to learn implicit class significance values that determine the positive label ranks instead of treating them as of equal importance, by following an approach that unifies ranking and classification tasks associated with multi-label ranking. Due to the scarcity of public datasets, we introduce eight synthetic datasets generated under varying importance factors to provide an enriched and controllable experimental environment for this study. On both real-world and synthetic datasets, we carry out extensive comparisons with relevant baselines and evaluate the performance on both of the two sub-tasks. We show that our method is able to accurately learn a representation of the incorporated positive rank order, which is not only consistent with the ground truth but also proportional to the underlying information. We strengthen our claims empirically by conducting comprehensive experimental studies. Code is available at \href{https://github.com/MrGranddy/GaussianMLR}{\textit{https://github.com/MrGranddy/GaussianMLR}}.

\end{abstract}

\section{Introduction}
Multi-label ranking (MLR) \cite{Zhou2014ATO,Dery2021MultilabelRM,InconsistentRankers} is a supervised learning problem where the objective is to not only classify the labels by their relevancy to the instance, but also to predict a ranking that represents the instance's preferences over chosen relevant labels. This can be considered as a generalization of the two sub-problems: multi-label classification and label ranking. To briefly define, given a set of labels $Y$, multi-label classification bipartites the set into two and associates the instances with the relevant label set $\mathcal{Y} \subseteq Y$. On the other hand, label ranking aims to map instances to a total order over $Y$. With the objectives of these sub-problems combined, multi-label ranking can be applied to any scenario where the expected output is a ranked subset of all possible labels. 

Although the use cases of multi-label ranking have been defined in the literature, most of the public multi-label datasets has ranking information of labels as being a positive or a negative label only.
Ranked datasets specific to MLR, where the label order represents the preference of relevant labels to the instance according to the chosen criteria, are very rare. Therefore, not many studies deal with evaluating the performance of a proposed approach in terms of the ranking order between the predicted positive classes. Rather, multi-label ranking has been used as an approach to solve multi-label classification problems, applied to object recognition \cite{Bucak2009EfficientMR}, image classification \cite{DBLP:journals/corr/LiSL17}, or used in online algorithms \cite{pmlr-v84-jung18a}, where the ranking of predictions are not taken into consideration as a measure. Furthermore, existing algorithms agree with the restraint that, only the partial information that is deduced from the bipartition of the labels is available.


Given the limitations in the field, the motivation for our work is to study multi-label ranking with a novel approach that exploits the order between positive labels, instead of assuming the positive and negative labels of equal importance in their own set. In the following subsection, we highlight our contributions and state the focus of our study in detail.

\subsection{Our Contributions}
\begin{itemize}
    \item  In the field of MLR, we establish the paradigm of exploiting the ranking information between positive labels of an instance, where the main objective is to extract significance values of labels that determine their rank, which may not be feasible to obtain numerically during the labeling process.
    
    
    \item We model the MLR problem as a distribution learning problem based on a probabilistic approach that unifies the bi-partition and ranking of the labels in the same space. This way, we incorporate pairs consisting of positive labels into the optimization in addition to positive and negative label pairs, to reveal a preference relation over a varying number of positive labels.
    
    \item We introduce ranked image datasets generated under different setups with varying importance factors that determine the ranks, which create a controllable environment to test new approaches while facilitating unambiguous interpretations of their performances.
    
    \item We compare our novel framework with different methods related to our approach, and empirically explore and interpret the outcomes, thus setting up a clear set of baselines for the MLR problem.
    
    
\end{itemize}

\section{Related Work}
In the literature, multi-label ranking has been applied to solve classification problems \cite{Bucak2009EfficientMR,DBLP:journals/corr/LiSL17,PositivePairwiseCorrelations,DembczynskiConsistent}, to learn from incompletely or inconsistently labeled data \cite{InconsistentRankers,PositiveUnlabeled}, and has been studied from perspectives of consistency and generalization by \cite{RethinkingWu}. In this section, we review the related works on learning problems associated with multi-label ranking. Due to the misuse of terminology in previous works, we clearly state the differences between concepts and examine the ones in the scope of this paper.

\subsection{Label Ranking}
Preference learning has been long studied in various contexts, and this paper is concerned with learning label preferences where an instance is associated with a finite set of ordered labels, namely label ranking problem \cite{hullermeier_2008,Vembu2010LabelRA}. Label ranking should not be confused with object ranking \cite{Cohen1997LearningTO}, where the aim is to predict the ranking over a predefined class of objects related to the given user query, such as sorting the responses of search engines according to their relevance to the query \cite{KamishimaSurvey}.

Proposed methods for label ranking include pointwise \cite{pointChen,pointSigur,pointToderici}, pairwise \cite{hullermeier_2008,Zhou2014ATO,DBLP:journals/corr/LiSL17,mertan_new_2020} and listwise \cite{listwiseCao,listwiseXia} methods. 
In this paper, we are concerned with pairwise methods. A pairwise method that transforms the problem into a binary classification problem was first introduced in \cite{HarPeled2002ConstraintCF}, as the Constraint Classification framework (CC). The idea behind CC is to build constraints according to the preference relation between labels, where the relation $\lambda_i > \lambda_j$, denoting that label $\lambda_i$ precedes $\lambda_j$ in relevance to the instance, constructs the two constraints $f_i(x) - f_j(x)>0$ being the positive constraint and $ f_j(x) - f_i(x)<0$ being the negative constraint. Then, both constraints are used as training samples of the single classifier to find the suitable weight vector satisfying the constraints. More efficiently, Ranking by Pairwise Comparison (RPC) \cite{hullermeier_2008,Frnkranz2010PreferenceLA} transforms the problem into training a binary classifier for each label pair, producing a number of $K(K-1)/2$ models which is half of the number of constraints of CC, where $K$ is the number of classes. The final ranking is determined after obtaining each model's decision for the given pair and calculating the sum of weighted votes. More recently, the Log-Sum-Exp-Pairwise (LSEP) loss function introduced in  \cite{DBLP:journals/corr/LiSL17}
improves the previous approaches using hinge loss \cite{Gong2014DeepCR,Weston2011WSABIESU} by learning the pairwise comparisons in a smooth and easier way to optimize. The proposed loss function follows the method of pairing one positive and one negative label from the sets constructed according to the ground truth and wraps BP-MLL (Backpropogation for Multi-Label Learning) \cite{ZhangBPMLL} in a logarithmic function with a bias term. The objective of LSEP is to enforce the positive labels to be in higher ranks and negative labels to be in lower ranks by optimizing the loss function while training convolutional neural networks \cite{Krizhevsky2012ImageNetCW}. However, LSEP involves an obligatory ordering of the whole set of labels to be followed by a thresholding that determines which labels are to be discarded, while our GaussianMLR introduces a natural label selection and ranking process.

\subsection{Label Classification}
Classification algorithms can be roughly divided into two categories based on the predicted label count, as multi-class and multi-label classification. Determining the label count of the prediction set is one challenge of multi-label classification, which is not a concern in multi-class classification where instances are associated with a single label only. Although some approaches include setting fixed label counts such as choosing top-k labels or fixed thresholds as a confidence score boundary, forcing a label count by a fixed value is impractical as it ignores the context of the problem at hand. Varying label counts can be realized by setting learnable thresholds or label counts as in \cite{DBLP:journals/corr/LiSL17}, or they can be jointly learned in the label ranking step as proposed in the Calibrated Label Ranking method \cite{10.5555/1567016.1567123}, by using virtual labels as split points which are inserted in the label set before the ranking process. Then, labels in higher ranks compared to the virtual label are included in the relevant set. Compared against all previous related work, our GaussianMLR implicitly introduces a zero-point, i.e. an inherent threshold, to perform binary classification of labels into positives and negatives in the same space that we perform ranking, thus combining both tasks in a unified model.

\section{Problem and Notation}

\subsection{Dataset Definiton}
\label{sec:dataset_notation}
We start with a dataset of $N$ examples, $\{(\textbf{\textit{x}}^{(i)}, R^{(i)})\}^N_{i=1}$ where $\textbf{\textit{x}}^{(i)} \in \mathbb{R}^d$ is a real-valued sample from the input distribution, and $R^{(i)} = \{(y_j, r^{(i)}_j)\}_{j=1}^K$ is the label, where $y_j \in Y$ is a symbolic class representation for an entity that can be semantically present in $\textbf{\textit{x}}^{(i)}$, $Y = \{y_1, y_2, ..., y_K\}$ is the set of $K$ possible classes, and $r^{(i)}_j \in \mathbb{N}$ is the rank of the associated class $y_j $ for $\textbf{\textit{x}}^{(i)}$. It is the case that prior work mostly uses standard multi-label classification datasets defined as $\{(\textbf{\textit{x}}^{(i)}, \mathcal{Y}^{(i)})\}^N_{i=1}$ where $\mathcal{Y}^{(i)} \subseteq Y$ is the set of classes semantically present in $\textbf{\textit{x}}^{(i)}$ and called the \textit{positive} classes, while the rest of the classes are called \textit{negative} classes. This kind of dataset only provides a ranking information between negative and positive classes where positives have a higher rank than negatives, and can be seen as a special case of the former 
definition where $R^{(i)} = \{(y_u, 1) | y_u \in \mathcal{Y}^{(i)}\} \cup \{(y_v, 0) | y_v \notin \mathcal{Y}^{(i)}\} $. In our definition, a negative class will always have rank 0. Throughout our work, our findings are under the fair assumptions: \textbf{(i)} $\textbf{\textit{x}}$ are identically and independently distributed (i.i.d.) in all datasets. \textbf{(ii)} All label/rank pairs $(y_j, r_j^{(i)})$ are conditionally independent given an input $\vb*{x^{(i)}}$.

\subsection{Problem Definition}

We construct the \textit{multi-label ranking} problem, assuming that the two sub-problems, the multi-label classification and label ranking, are independent, i.e., $\mathcal{Y}^{(i)}$ and $\mathcal{B}^{(i)}$ are independent given $\vb*{x}^{(i)}$:
\begin{equation}\label{eqn:mlr_raw}
    P_{\mathcal{Y},\mathcal{B}}(\mathcal{Y}^{(i)}, \mathcal{B}^{(i)}|\textbf{\textit{x}}^{(i)};\theta) = P_\mathcal{Y}(\mathcal{Y}^{(i)}|\textbf{\textit{x}}^{(i)} \theta)P_\mathcal{B}(\mathcal{B}^{(i)}|\textbf{\textit{x}}^{(i)};\phi).
\end{equation}
Thus, we formally define the following likelihood optimization problem, given an instance $\vb*{x}^{(i)}$:
\begin{equation}\label{eqn:mlr}
    \max_{\theta,\phi} P_\mathcal{Y}(\mathcal{Y}^{(i)}|\textbf{\textit{x}}^{(i)};\theta)P_\mathcal{B}(\mathcal{B}^{(i)}|\textbf{\textit{x}}^{(i)};\phi).
\end{equation}
Here $\mathcal{Y}^{(i)}$ is the set of \textit{positive} classes for an instance $\textbf{\textit{x}}^{(i)}$, $P_\mathcal{Y}$ is the parameterized family of probability mass functions of possible positive class sets, parameterized by $\theta$, and conditioned by $\textbf{\textit{x}}^{(i)}$. 
While $\mathcal{B}^{(i)}$ is a \textit{bucket order}\cite{10.1145/1055558.1055568}, which is a class of partial orders allowing \textit{ties}. A partial order is a reflexive, antisymmetric, and transitive binary relation on a set of items, in our case $Y$. $\mathcal{B}^{(i)}$ intuitively partitions labels $y \in Y$ into mutually exclusive bucket of ranks $\langle \mathcal{M}_1^{(i)}, ..., \mathcal{M}_{b^{(i)}}^{(i)} \rangle$ where $b^{(i)}$ is the number of buckets. If two items $y_u, y_v \in \mathcal{M}_k^{(i)}$ are the member of the same bucket then we say they are in tie, meaning they can not be distinguished ordinally and they virtually have the same rank. Formally, for a bucket $\mathcal{M}_k^{(i)}$, ~$y_u, y_v \in \mathcal{M}_k^{(i)} \iff (y_u, y_v) \notin \mathcal{B}^{(i)} \wedge (y_v, y_u) \notin \mathcal{B}^{(i)}$. Here it should be noted that $\mathcal{M}_k^{(i)} \subseteq Y$ is a set and is only introduced to better visualize the bucket orders, $\mathcal{B}^{(i)}$ is just a relation and is enough to define a bucket order. On the other hand, different buckets for an instance $x^{(i)}$ have total ordinal relationship between them, such that: for any two distinct buckets $\mathcal{M}_k^{(i)}, \mathcal{M}_l^{(i)}$, $y_u \in \mathcal{M}_k^{(i)}, y_v \in \mathcal{M}_l^{(i)}$, $(y_u, y_v) \in \mathcal{B}^{(i)} \iff k > l$, i.e. $r_u^{(i)} \geq r_v^{(i)} \iff k > l$, where $r_u^{(i)}$ and $r_v^{(i)}$ are corresponding ranks of $y_u$ and $y_v$. $P_{\mathcal{B}}$ is the parameterized family of probability mass functions of such bucket orders parameterized by $\phi$, conditioned by $\textbf{\textit{x}}^{(i)}$.

The optimization problem in hand can be seen as the joint optimization of two distinct problems, namely: multi-label classification and label ranking.
Both of the terms can be divided into practicable sub-problems.

We can simplify the first likelihood  $P_{\vb{y}}(\vb{y}|\textbf{\textit{x}}^{(i)};\theta)$ by defining the random variable $\vb{y} \in \{0, 1\}^K$ via the random vector:
\begin{align*}
    &\vb{y}_c =
    \begin{cases} 
      1 & y_c \in \mathcal{Y} \\
      0 & y_c \notin \mathcal{Y}
   \end{cases}, &c \in \{1, ..., K\}.
\end{align*}
Here $\mathcal{Y}$ is any random variable distributed by $P_\mathcal{Y}(\mathcal{Y}|\textbf{\textit{x}}^{(i)};\theta)$, then we can model the sub-problem with $K$ binary classification models using Bernoulli distribution:
\begin{equation}
     \max_\theta P_\mathcal{Y}(\mathcal{Y}^{(i)}|\textbf{\textit{x}}^{(i)};\theta) = 
     \max_\theta \prod_{c=1}^K P(\vb{y}_c=1|\textbf{\textit{x}}^{(i)};\theta)^{\mathbb{I}[y_c \in \mathcal{Y}^{(i)}]} P(\vb{y}_c=0|\textbf{\textit{x}}^{(i)};\theta)^{\mathbb{I}[y_c \notin \mathcal{Y}^{(i)}]},
\end{equation}
where $\mathbb{I}[.]$ is the indicator function.

\begin{figure}[ht]
\centering
\resizebox{0.8\textwidth}{!}{
  \begin{subfigure}[b]{0.3\textwidth}
    \begin{tikzpicture} [node distance = 4cm, on grid, auto]
     
    \node (n2) [state] {$\vb{r}_2$};
    \node (n1) [state, above right=2.25cm and 1.5cm of n2] {$\vb{r}_1$};
    \node (n3) [state, below right=2.25cm and 1.5cm of n1] {$\vb{r}_3$};
     
    \path [-stealth]
        (n2) edge [bend left=8]   node  [sloped, anchor=center, above]  {\scalebox{.8}{\tiny$P(\vb{r}_2 \geq \vb{r}_1)$}} (n1)
        (n2) edge [bend right=8]  node  [below]  {\scalebox{.8}{\tiny$P(\vb{r}_2 \geq \vb{r}_3)$}} (n3)
        
        (n1) edge [bend right=8]  node [above right] [sloped, anchor=center, below]  {\scalebox{.8}{\tiny$P(\vb{r}_1 \geq \vb{r}_3)$}} (n3)
        (n1) edge [bend left=8]   node [sloped, anchor=center, below]  {\scalebox{.8}{\tiny$P(\vb{r}_1 \geq \vb{r}_2)$}} (n2)
        
        (n3) edge [bend right=8]  node [sloped, above]  {\scalebox{.8}{\tiny$P(\vb{r}_3 \geq \vb{r}_2)$}}  (n2)
        (n3) edge [bend right=8]  node [sloped, anchor=center, above]  {\scalebox{.8}{\tiny$P(\vb{r}_3 \geq \vb{r}_1)$}} (n1);
    \end{tikzpicture}
    \caption{}\label{fig:first-graph}
  \end{subfigure}
 \hfill
  \begin{subfigure}[b]{0.3\textwidth}
    \begin{tikzpicture} [node distance = 4cm, on grid, auto]
     
    \node (n2) [state] {$\vb{r}_2$};
    \node (n1) [state, above right=2.25cm and 1.5cm of n2] {$\vb{r}_1$};
    \node (n3) [state, below right=2.25cm and 1.5cm of n1] {$\vb{r}_3$};
     
    \path [-stealth]
        (n2) edge [bend left=8]   node  [sloped, anchor=center, above]  {\scalebox{.8}{\tiny$P(\vb{r}_2 \geq \vb{r}_1)$}} (n1)
        (n2) edge [bend right=8]  node  [below]  {\scalebox{.8}{\tiny$P(\vb{r}_2 \geq \vb{r}_3)$}} (n3)

        (n1) edge [thin,densely dotted,bend left=8]   node  [sloped, anchor=center, below]  {}(n2)
        (n1) edge [bend right=8]  node [sloped, anchor=center, below]  {\scalebox{.8}{\tiny$P(\vb{r}_1 \geq \vb{r}_3)$}} (n3)
        
        (n3) edge [thin,densely dotted,bend right=8]  node [sloped, above]  {}  (n2)
        
        (n3) edge [thin,densely dotted,bend right=8]  node [above right] [sloped, anchor=center, below]  {} (n1);
    \end{tikzpicture}
    \caption{}\label{fig:second-graph}
  \end{subfigure}
   \hfill
  \begin{subfigure}[b]{0.3\textwidth}
      \begin{tikzpicture} [node distance = 4cm, on grid, auto]
     
    \node (n2) [state] {$\vb{r}_2$};
    \node (n1) [state, above right=2.25cm and 1.5cm of n2] {$\vb{r}_1$};
    \node (n3) [state, below right=2.25cm and 1.5cm of n1] {$\vb{r}_3$};
     
    \path [-stealth]
        (n2) edge [bend left=8]   node  [sloped, anchor=center, above]  {\scalebox{.8}{\tiny$P(\vb{r}_2 \geq \vb{r}_1)$}} (n1)
        (n2) edge [bend right=8]  node  [below]  {\scalebox{.8}{\tiny$P(\vb{r}_2 \geq \vb{r}_3)$}} (n3)
        
        (n1) edge [thin,densely dotted,bend right=8]  node [above right] [sloped, anchor=center, below]  {} (n3)
        (n1) edge [thin,densely dotted,bend left=8]   node [sloped, anchor=center, below]  {} (n2)
        
        (n3) edge [thin,densely dotted,bend right=8]  node [sloped, above]  {}  (n2)
        (n3) edge [thin,densely dotted,bend right=8]  node [sloped, anchor=center, above]  {} (n1);
    \end{tikzpicture}
    \caption{}\label{fig:third-graph}
  \end{subfigure}
  }
  \caption{ Visual explanation of how a real valued random variable, a total partial order and a bucket order is related. Random vector $\vb{r}$ defines a complete graph as depicted in (a), a total partial order contains magnitude relations for all pairs of the set that it is defined on. (b) shows how a total partial order can be applied on the random vector $\vb{r}$ to calculate its likelihood: $P(\vb{r}_2 \geq \vb{r}_1)P(\vb{r}_1 \geq \vb{r}_2)P(\vb{r}_2 \geq \vb{r}_3)$. Lastly a bucket agrees with a set of total partial orders, since the bucket order depicted in (c) does not specify an ordering between $\vb{r}_1$ and $\vb{r}_3$, both $\langle \vb{r}_2, \vb{r}_1, \vb{r}_3\rangle$ and $\langle \vb{r}_2, \vb{r}_3, \vb{r}_1\rangle$ is valid for (c). Then it is trivial how the elimination process will follow, yielding the likelihood: $P(\vb{r}_2 \geq \vb{r}_1)P(\vb{r}_2 \geq \vb{r}_3)$ for the bucket order shown in (c).}\label{fig:graph}
\end{figure}
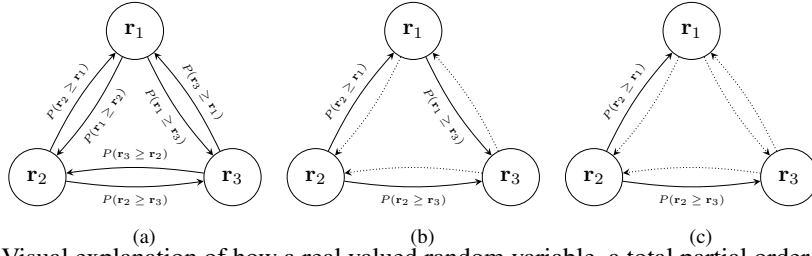

\textbf{Likelihood of a Bucket Order.} We can use a random vector $\vb{r} \sim P_{\vb{r}}(\vb{r}|\phi)$ where $\vb{r} \in \mathbb{R}^K$ and $|Y| = K$ to model the likelihood of a bucket order, where each element of $\vb{r}_i$ corresponds to the significance value of class $y_i$. Let $\vb{r}$ define a weighted directed complete graph $\vb*{G} = (\vb*{V}, \vb*{E}, \vb*{\omega})$ such that, $\vb*{V} = Y$, $\vb*{E} = Y\cross Y$ and $\vb*{\omega}: \vb*{E} \rightarrow \mathbb{R}$. We define the edge weights as $\vb*{\omega}(y_u, y_v) = P(\vb{r}_u \geq \vb{r}_v)$. Let $\mathcal{P}$ be a total partial order defined by $\succ$ such that $\forall y_u, y_v \in Y, (y_u, y_v) \in \mathcal{P}$ or $(y_v, y_u) \in \mathcal{P}$, which describes a unique permutation of the elements. Then we can model the likelihood of $\mathcal{P}$ with independent Bernoulli distributions on the edges, such that for any two nodes $y_u, y_v \in Y$ either $(y_u, y_v) \in \mathcal{P}$ or $(y_v, y_u) \in \mathcal{P}$, let $S$ be the set of unique pairs on $Y$, such that, if $(u, v) \in S$ then $(v,u) \notin S$ : $P(\mathcal{P}) = \prod_{(y_u, y_v) \in S} P(\vb{r}_u \geq \vb{r}_v)^{\mathbb{I}[(y_u, y_v) \in \mathcal{P}]}P(\vb{r}_u < \vb{r}_v)^{\mathbb{I}[(y_u, y_v) \notin \mathcal{P}]}$. $\mathcal{P}$ is defined by $\succ$ relations so we can simplify: $P(\mathcal{P}) = \prod_{(y_u, y_v) \in \mathcal{P}} P(\vb{r}_u \geq \vb{r}_v)$. A bucket order $\mathcal{B}$ agrees with a unique set of total partial orders $S_{\mathcal{P}}$, formally $\forall \mathcal{P} \in S_{\mathcal{P}}, \forall (y_u, y_v) \in \mathcal{B}, (y_u, y_v) \in \mathcal{P}$. Then we can express $P(\mathcal{B}) = \sum_{\mathcal{P} \in S_\mathcal{P}}\prod_{(y_u, y_v) \in \mathcal{P}} P(\vb{r}_u \geq \vb{r}_v)$. For any pair of tied elements $y_u, y_v \in Y, (y_u, y_v) \notin \mathcal{B} \wedge (y_v, y_u) \notin \mathcal{B}$, there will be $(y_u, y_v) \in \mathcal{P}_1$ and $(y_v, y_u) \in \mathcal{P}_2$ and the rest of the elements are the same, where $\mathcal{P}_1, \mathcal{P}_2 \in S_\mathcal{P}$. $P(\vb{r}_u \geq \vb{r}_v) + P(\vb{r}_u < \vb{r}_v) = 1$, grouping and simplifying the summed terms in $P(\mathcal{B})$ formula, we have: $P(\mathcal{B}) = \prod_{(y_u, y_v) \in \mathcal{B}}P(\vb{r}_u \geq \vb{r}_v)$. A visual explanation of the likelihood of a bucket order can be seen in Figure~\ref{fig:graph}.

We parameterize $P_{\vb{r}}(\vb{r}|\textit{\textbf{x}}^{(i)};\phi)$ by $\phi$, and re-write  $P(\mathcal{B})$ in a likelihood maximization:
\begin{equation}
         \max_\phi P_\mathcal{B}(\mathcal{B}^{(i)}|\textbf{\textit{x}}^{(i)};\phi) = \max_\phi \prod_{(y_u, y_v) \in \mathcal{B}^{(i)}} P(\vb{r}_u \geq \vb{r}_v|\textit{\textbf{x}}^{(i)};\phi).
\end{equation}

As a result, the multi-label ranking problem as defined in Equation \eqref{eqn:mlr} can be re-written as:
\begin{equation}\label{eqn:simple_mlr}
    \max_{\theta,\phi} \prod_{c=1}^K P(\vb{y}_c=1|\textbf{\textit{x}}^{(i)};\theta)^{\mathbb{I}[y_c \in \mathcal{Y}^{(i)}]} P(\vb{y}_c=0|\textbf{\textit{x}}^{(i)};\theta)^{\mathbb{I}[y_c \notin \mathcal{Y}^{(i)}]} \prod_{(y_u, y_v) \in \mathcal{B}^{(i)}} P(\vb{r}_u \geq \vb{r}_v|\textit{\textbf{x}}^{(i)};\phi).
\end{equation}

The refined optimization problem in Equation~\eqref{eqn:simple_mlr} sets up the basis for our proposed unified multi-label ranking method. It establishes a probabilistic foundation for the multi-label ranking problem, which we further develop with a Gaussian probability model next.

\section{GaussianMLR}

\subsection{Core Idea}
Multi-label ranking problem can be viewed as both classifying and ranking the classes in a given instance, meaning that we are not only interested in assigning correct classes to an instance, but we also want to measure how relevant these classes are for the given instance.

\textbf{Significance Values.} We start by assuming all $y_j \in \mathcal{Y}^{(i)}$ for a given input $\vb*{x}^{(i)}$ has an underlying significance value $s_j^{(i)} \in \mathbb{R}$, we model these significance values using Gaussian distributions, such that: $s_j^{(i)} \sim \mathcal{N}(\mu_j^{(i)}, \sigma_j^{2(i)})$ where $\mathcal{N}(\cdot,\cdot)$ is a Gaussian distribution, $\mu_j^{(i)}$ is the mean and $\sigma_j^{2(i)}$ is the variance, for a given $\vb*{x}^{(i)}$.

\subsection{Methodology}

Our goal is to obtain a function $f$ such that it produces significance values matching with the ranking information in the ground truth. Let $f: \mathbb{R}^d \rightarrow \mathbb{R}^{2K}$ be a trainable function parameterized by $\zeta$, where the output of the function is the predicted Gaussian distribution parameters $\hat{\mu}^{(i)} \in \mathbb{R}^K$ and $\hat{\sigma}^{2(i)} \in \mathbb{R}^K$ for the input $\textbf{\textit{x}}^{(i)} \in \mathbb{R}^d$ such that $f(\textbf{\textit{x}}^{(i)};\zeta) = [\hat{\mu}^{(i)} \; \hat{\sigma}^{2(i)}]$ and the predicted significance values $\hat{s}_j^{(i)} \sim \mathcal{N}\big(\hat{\mu}_j^{(i)}, \hat{\sigma^2}_j^{(i)}\big)$.

We adapt the optimization problem in Equation \eqref{eqn:simple_mlr} as follows: instead of using a separate binary variable for the classification task, we model the significance values such that for a class ${y}_c$,  $\hat{s}_c^{(i)} \geq 0$ indicates a predicted positive, and $\hat{s}_c^{(i)} < 0$ indicates a predicted negative. The reformulated optimization problem reads:
\begin{equation}
    \label{eqn:gauss_mlr_prob}
    \max_{\zeta} \prod_{c=1}^{K} P(\hat{s}_c^{(i)} \geq 0)^{\mathbbm{I}[y_c \in \mathcal{Y}^{(i)}]} P(\hat{s}_c^{(i)} < 0)^{\mathbbm{I}[y_c \notin \mathcal{Y}^{(i)}]} \prod_{(y_u,y_v) \in \mathcal{B}^{(i)}} P(\hat{s}^{(i)}_u \geq \hat{s}^{(i)}_v).
\end{equation}

Letting $\hat{d}_{(u, v)}^{(i)} = \hat{s}^{(i)}_{u} - \hat{s}^{(i)}_{v}$, and using the above-mentioned definition: $\hat{d}_{(u, v)}^{(i)} \sim \mathcal{N}(\hat{\mu}^{(i)}_{u} - \hat{\mu}^{(i)}_{v}, \hat{\sigma}^{2(i)}_{u} + \hat{\sigma}^{2(i)}_{v})$, we can re-write Equation \ref{eqn:gauss_mlr_prob} as follows:
\begin{equation}
    \label{eqn:gauss_mlr_prob_with_diff}
    \max_{\zeta} \prod_{c=1}^{K} P(\hat{s}_c^{(i)} \geq 0)^{\mathbbm{I}[y_c \in \mathcal{Y}^{(i)}]} P(\hat{s}_c^{(i)} < 0)^{\mathbbm{I}[y_c \notin \mathcal{Y}^{(i)}]} \prod_{(y_u,y_v) \in \mathcal{B}^{(i)}} P(\hat{d}_{(u, v)}^{(i)} \geq 0).
\end{equation}

For a Gaussian random variable $z \sim \mathcal{N}(\mu, \sigma^2)$, $P(z > 0) = (1/2) [ 1 - \erf( -\mu / (\sigma \sqrt{2}) ) ]$ where $\erf(\cdot)$ is the Gaussian error function. Let $Q(\mu, \sigma) = (1/2)[ 1 -\erf\left( -\mu / (\sigma \sqrt{2}) \right) ]$, then we can re-write Equation \ref{eqn:gauss_mlr_prob_with_diff} as:
\begin{equation}
    \max_{\zeta} \prod_{c=1}^{K} Q(\hat{\mu}_c^{(i)}, \hat{\sigma}_c^{(i)})^{\beta_c^{(i)}}
    (1 - Q(\hat{\mu}_c^{(i)}, \hat{\sigma}_c^{(i)}))^{(1 - \beta_c^{(i)})}
     \prod_{(y_u, y_v) \in \mathcal{B}^{(i)}} Q(\hat{\mu}_{(u,v)}^{(i)}, \hat{\sigma}_{(u,v)}^{(i)}).
\end{equation}
Here $\beta_c^{(i)} = \mathbbm{I}[y_c \in \mathcal{Y}^{(i)}]$, $\hat{\mu}_{(u,v)} = \hat{\mu}^{(i)}_{u} - \hat{\mu}^{(i)}_{v}$ and $\hat{\sigma}_{(u,v)}^{(i)} = \sqrt{\hat{\sigma}^{2(i)}_{u} - \hat{\sigma}^{2(i)}_{v}}$. Applying negative log likelihood, we define our loss function in two parts as follows:

$L_c(\hat{\mu}^{(i)}, \hat{\sigma}^{(i)}, \mathcal{Y}^{(i)}) = \sum_{c=1}^{K} - \beta_c^{(i)}\log (Q(\hat{\mu}_c^{(i)}, \hat{\sigma}_c^{(i)})) -
(1 - \beta_c^{(i)})\log(1 - Q(\hat{\mu}_c^{(i)}, \hat{\sigma}_c^{(i)}))$,\\
$L_r(\hat{\mu}^{(i)}, \hat{\sigma}^{(i)}, \mathcal{B}^{(i)}) = \sum_{(y_u,y_v) \in \mathcal{B}^{(i)}} -\log(Q(\hat{\mu}_{(u,v)}^{(i)}, \hat{\sigma}_{(u,v)}^{(i)}))$.\\

Summing up, the objective function of the GaussianMLR is given by:
\begin{equation}
    \label{eqn:gaussian_mlr_loss}
    \min_{\zeta} \dfrac{1}{N} \sum_{i=1}^N L_c(\hat{\mu}^{(i)}, \hat{\sigma}^{(i)}, \mathcal{Y}^{(i)}) + L_r(\hat{\mu}^{(i)}, \hat{\sigma}^{(i)}, \mathcal{B}^{(i)}),
\end{equation}

\textbf{Training and inference.} GaussianMLR provides a differentiable loss function (\ref{eqn:gaussian_mlr_loss}), thus in the context of our work we choose to use our objective to train neural networks using stochastic gradient descent. After training a network with any dataset $\mathcal{D}$, at inference we use $\hat{\mu}^{(i)}$ as the predicted ranking score for a given $\vb*{x}^{(i)}$. Further details on the implementation can be found in Appendix \ref{sec:apx2}.

\textbf{Learning Implicit Class Significance.} For a large enough dataset, we claim that our predictions will be proportional to the real underlying significance values. This claim is theoretically supported in Appendix \ref{sec:apx4} and we further support our claim with empirical studies in Section \ref{sec:experiments}.


\section{Experiments}
\label{sec:experiments}

\subsection{Datasets}
\label{sec:datasets}

We conduct our experiments on three distinct datasets: natural scene images database\cite{InconsistentRankers}, architectural VDP dataset\cite{DEMIR2021103826} and Ranked MNIST datasets, which we introduce Section \ref{sec:ranked_mnist}. These datasets have the common trait that they not only bipartite the labels into negatives and positives, but also provide how relevant each of these positive classes are to the instances. All datasets we use follow the notation provided in Section \ref{sec:dataset_notation}, further details can be found in Appendix \ref{sec:apx1}. In the paper, we only provide Ranked MNIST Gray experiments, for the Ranked MNIST Color, the experiments can be found in Appendix \ref{sec:apx6}.

\subsection{Ranked MNIST}
\label{sec:ranked_mnist}
Ranked MNIST is a family of datasets with two main branches named as Ranked MNIST Gray and Ranked MNIST Color, where the first is in grayscale and the latter has varying random hue and saturation values for each digit. These datasets are generated by placing unique digits from the MNIST dataset \cite{deng2012mnist} on a 224x224 canvas, where the number of digits in a single image vary from 1 up to 10. For each branch, we have two different importance factors that change: scale and brightness. According to these factors, we rank each positive digit such that for scale: the larger digits have greater ranks and for brightness: the brighter digits have greater ranks. For both the Ranked MNIST Gray-S/B (scale/brightness) and Color-S/B datasets we have four different setups: changing scale, changing brightness, changing both and training on scales, changing both and training on brightness. Further explanation and examples can be seen in Appendix \ref{sec:apx5}.

\subsection{Baselines}

To evaluate our GaussianMLR (GMLR) method, we selected two pairwise baseline methods, namely: CRPC\cite{InconsistentRankers}, which is the calibrated\cite{calibrated_lr} version of RPC, and LSEP\cite{DBLP:journals/corr/LiSL17}. To our knowledge, GMLR is the first multi-label ranking method which utilizes the positive class ranks, thus we aim to provide fairness in competition of the baseline algorithms with our method. To that end, we introduce CRPC-Strong and LSEP-Strong, where we develop the existing baselines into \textit{Strong} versions that can process the positive class ranks by adding all $(y_u, y_v)$ pairs, where $y_u$ and $y_v$ are positive classes and $y_u \succ y_v$. Similarly, we call the methods which do not use the positive class relations as \textit{Weak} versions.

\begin{table}[ht]
\caption{Quantitative results on Ranked MNIST Gray datasets. Ranked MNIST S and B stands for changing the scale or brightness of the digits, while (Mix) means both of the features are changing, but the ground truth indicates only one of the features. Bold-marked results show the best scores in Strong(S) baselines, and underlined scores show the best scores in Weak(W) baselines.}
 \renewcommand{\arraystretch}{1.2}
\centering
\resizebox{1.0\textwidth}{!}{ 
\begin{tabular}{c|cccccc|cccccc|cccccc|cccccc}
\Xhline{3\arrayrulewidth}
 \multirow{2}{*}{Method} & \multicolumn{6}{c|}{Ranked MNIST Gray-S} & \multicolumn{6}{c|}{Ranked MNIST Gray-B} & \multicolumn{6}{c|}{Ranked MNIST Gray-S (Mix)} & \multicolumn{6}{c}{Ranked MNIST Gray-B (Mix)} \\ \cline{2-25}
 & \multicolumn{1}{c}{$\tau_b$ $\uparrow$} & \multicolumn{1}{c}{$ S \rho$ $\uparrow$} & \multicolumn{1}{c}{$\gamma$ $\uparrow$} & \multicolumn{1}{c}{HL $\downarrow$} & \multicolumn{1}{c}{M-1 $\downarrow$} & F1 $\uparrow$ & \multicolumn{1}{c}{$\tau_b$ $\uparrow$} & \multicolumn{1}{c}{$ S \rho$ $\uparrow$} & \multicolumn{1}{c}{$\gamma$ $\uparrow$} & \multicolumn{1}{c}{HL $\downarrow$} & \multicolumn{1}{c}{M-1 $\downarrow$ } & F1 $\uparrow$ & \multicolumn{1}{c}{$\tau_b$ $\uparrow$} & \multicolumn{1}{c}{$ S \rho$ $\uparrow$} & \multicolumn{1}{c}{$\gamma$ $\uparrow$} & \multicolumn{1}{c}{HL $\downarrow$} & \multicolumn{1}{c}{M-1 $\downarrow$} & F1 $\uparrow$& \multicolumn{1}{c}{$\tau_b$ $\uparrow$} & \multicolumn{1}{c}{$ S \rho$ $\uparrow$} & \multicolumn{1}{c}{$\gamma$ $\uparrow$} & \multicolumn{1}{c}{HL $\downarrow$} & \multicolumn{1}{c}{M-1 $\downarrow$} & F1 $\uparrow$ \\
 \Xhline{2\arrayrulewidth}
\multicolumn{1}{c|}{CRPC(W)} & 49.26 & 59.92 &
\multicolumn{1}{c}{59.80} & 17.13 & 0.20 & 86.45 & 51.36 & 61.45 & \multicolumn{1}{c}{59.27} & 12.85 & 0.44 & 89.44 & 51.73 & 61.71 & \multicolumn{1}{c}{58.83} & 13.29 & 0.43 & 89.12 & 52.35 & 62.50 & \multicolumn{1}{c}{59.29} & 12.89 & 0.47 & 89.42 \\
\multicolumn{1}{c|}{LSEP (W)} & 61.38 & 70.67 & \multicolumn{1}{c}{61.49} & \underline{0.49} & \underline{0.10} & \underline{99.56} & 59.45 & 68.67 & \multicolumn{1}{c}{59.77} & \underline{0.96} & 0.14 & \underline{99.12} & 60.45 & 69.47 & \multicolumn{1}{c}{60.68} & 0.97 & \underline{0.14} & 99.11 & \underline{60.04} & 69.20 & \multicolumn{1}{c}{\underline{60.31}} & 0.99 & \underline{0.11} & 99.10 \\
\multicolumn{1}{c|}{GMLR (W)} & \underline{62.52} & \underline{71.86} & \multicolumn{1}{c}{\underline{62.62}} & \underline{0.51} & \underline{0.10} & \underline{99.54} & \underline{60.18} & \underline{69.36} & \multicolumn{1}{c}{\underline{60.54}} & \underline{0.97} & \underline{0.10} & \underline{99.12} & \underline{60.56} & \underline{69.74} & \multicolumn{1}{c}{\underline{60.80}} & \underline{0.92} & \underline{0.15} & \underline{99.16} & 60.00 & \underline{69.23} & \multicolumn{1}{c}{60.24} & \underline{0.93} & \underline{0.13} & \underline{99.15} \\ \hline
\multicolumn{1}{c|}{CRPC(S)} & 64.09 & 75.56 & \multicolumn{1}{c}{75.20} & 18.69 & 0.18 & 85.37 & 61.71 & 73.62 & \multicolumn{1}{c}{74.70} & 24.15 & 0.37 & 81.73 & 62.58 & 74.03 & \multicolumn{1}{c}{73.80} & 18.89 & 0.32 & 85.17 & 63.32 & 74.85 & \multicolumn{1}{c}{74.44} & 18.58 & 0.34 & 85.37 \\
\multicolumn{1}{c|}{LSEP (S)} & 93.99 & 97.35 & \multicolumn{1}{c}{\textbf{94.50}} & 1.38 & 0.23 & 98.75 & \textbf{93.62} & \textbf{97.01} & \multicolumn{1}{c}{94.46} & \textbf{1.95} & \textbf{0.24} & \textbf{98.21} & \textbf{91.71} & \textbf{95.89} & \multicolumn{1}{c}{\textbf{92.54}} & 2.15 & \textbf{0.27} & 98.04 & \textbf{93.03} & \textbf{96.81} & \multicolumn{1}{c}{93.64} & 1.80 & \textbf{0.23} & 98.36 \\
\multicolumn{1}{c|}{GMLR (S)} & \textbf{94.23} & \textbf{97.41}  & \multicolumn{1}{c}{94.43} & \textbf{0.58} & \textbf{0.20} & \textbf{99.47} & 93.38 & 96.65 & \multicolumn{1}{c}{\textbf{94.49}} & 2.04 & 0.33 & 98.15 & 90.99 & 95.05 & \multicolumn{1}{c}{91.75} & \textbf{1.45} & 0.79 & \textbf{98.67} & 92.94 & 96.46 & \multicolumn{1}{c}{\textbf{93.73}} & \textbf{1.52} & 0.44 & \textbf{98.62} \\ \hline
\end{tabular}
}
\label{tbl:ranked_mnist}
\end{table}

\begin{wraptable}{r}{0.6\textwidth}
    \caption{Quantitative results on Natural Scene Images Database (NSID) and Architectural VDP Dataset (AVDP). The annotations of scores are the same with Table \ref{tbl:ranked_mnist}.}
     \renewcommand{\arraystretch}{1.2}
    \centering
    \resizebox{0.6\textwidth}{!}{
    
    \begin{tabular}{c|cccccc|cccccc} 
     \Xhline{3\arrayrulewidth}
     \multirow{2}{*}{Method} & \multicolumn{6}{c|}{NSID} & \multicolumn{6}{c}{AVDP}  \\
     \Xcline{2-13}{1\arrayrulewidth}
     & \multicolumn{1}{c}{$\tau_b \uparrow $} & \multicolumn{1}{c}{$ S \rho \uparrow$} & \multicolumn{1}{c}{$\gamma \uparrow$} & \multicolumn{1}{c}{HL $\downarrow$} & \multicolumn{1}{c}{M-1 $\downarrow$} & F1 $\uparrow$ & \multicolumn{1}{c}{$\tau_b \uparrow$} & \multicolumn{1}{c}{$ S \rho \uparrow$} & \multicolumn{1}{c}{$\gamma \uparrow$} & \multicolumn{1}{c}{HL $\downarrow$} & \multicolumn{1}{c}{M-1 $\downarrow$} & F1 $\uparrow$ \\
     \Xhline{2\arrayrulewidth}
    \multicolumn{1}{c|}{CRPC(W)} & 57.89 & 64.62 &
    \multicolumn{1}{c}{70.65} & 20.36 & 9.60 & 68.25 & 37.57 & 40.24 & \multicolumn{1}{c}{41.70} & 24.28 & 33.81 & 50.75\\
    \multicolumn{1}{c|}{LSEP (W)} & 71.85 & 75.80 & \multicolumn{1}{c}{79.13} & \underline{10.74} & 6.03 & \underline{79.92} & 39.79 & 41.93 & \multicolumn{1}{c}{\underline{44.54}} & \underline{19.23} & \underline{30.29} & \underline{54.66} \\
    \multicolumn{1}{c|}{GMLR (W)} & \underline{73.41} & \underline{77.77} & \multicolumn{1}{c}{\underline{80.81}} & 11.16 & \underline{5.58} & \underline{79.94} & \underline{40.29} & \underline{42.69} & \multicolumn{1}{c}{41.12} & 20.10 & 31.75 & 54.31 \\ \hline
    \multicolumn{1}{c|}{CRPC(S)} & 59.54 & 66.02 & \multicolumn{1}{c}{72.15} & 19.22 & 9.82 & 69.61 & 39.69 & 41.95 & \multicolumn{1}{c}{44.74} & 22.17 & 34.43 & 52.57 \\
    \multicolumn{1}{c|}{LSEP (S)} & 72.57 & 76.57 & \multicolumn{1}{c}{80.58} & \textbf{10.54} & \textbf{4.69} & \textbf{80.12} & 40.54 & 42.60 & \multicolumn{1}{c}{42.55} & \textbf{18.89} & 31.41 & \textbf{55.86} \\
    \multicolumn{1}{c|}{GMLR (S)} & \textbf{75.44} & \textbf{78.66} & \multicolumn{1}{c}{\textbf{82.87}} & 11.06 & 6.03 & 80.08 & \textbf{41.27} & \textbf{43.34} & \multicolumn{1}{c}{\textbf{45.27}} & 20.08 & \textbf{29.22} & 52.54 \\ \hline
    \end{tabular}
    }
    \label{tbl:real_datasets}
\end{wraptable}

\subsection{Quantitative Results}

We provide quantitative results on both the Ranked MNIST and real datasets using two set of metrics: ranking and classification. For ranking we use Kendall's Tau-b ($\tau_b$), Spearman's Rho ($S_\rho$) and Goodman and Kruskal's Gamma ($\gamma$), for classification we use Hamming Loss (HL), Max-1 (M-1) loss and F1 score. Max-1 loss yields the percentage of instances such that the label with the maximum predicted score is not in the ground truth positive set. The details for the metrics are described in Appendix \ref{sec:apx3}. Table \ref{tbl:ranked_mnist} shows the results for each method trained on each Ranked MNIST Gray dataset. Here, GMLR slightly outperforms LSEP, and CRPC performs the worst amongst the three on all metrics. Quantitative results of our experiments on real datasets are given in Table \ref{tbl:real_datasets}. GMLR visibly outperforms the baselines for the ranking metrics, and produces comparable results to LSEP for the classification. It should be noted that both of the real datasets comprise inherent noise due to the labeling process being subjective.

\subsection{Adjusting Significance Effects Experiment}
\label{55}

\begin{figure}[ht]
    \centering
    \begin{subfigure}[b]{1.0\linewidth}
        \hspace{0.10in}
        \includegraphics[width=0.48\linewidth]{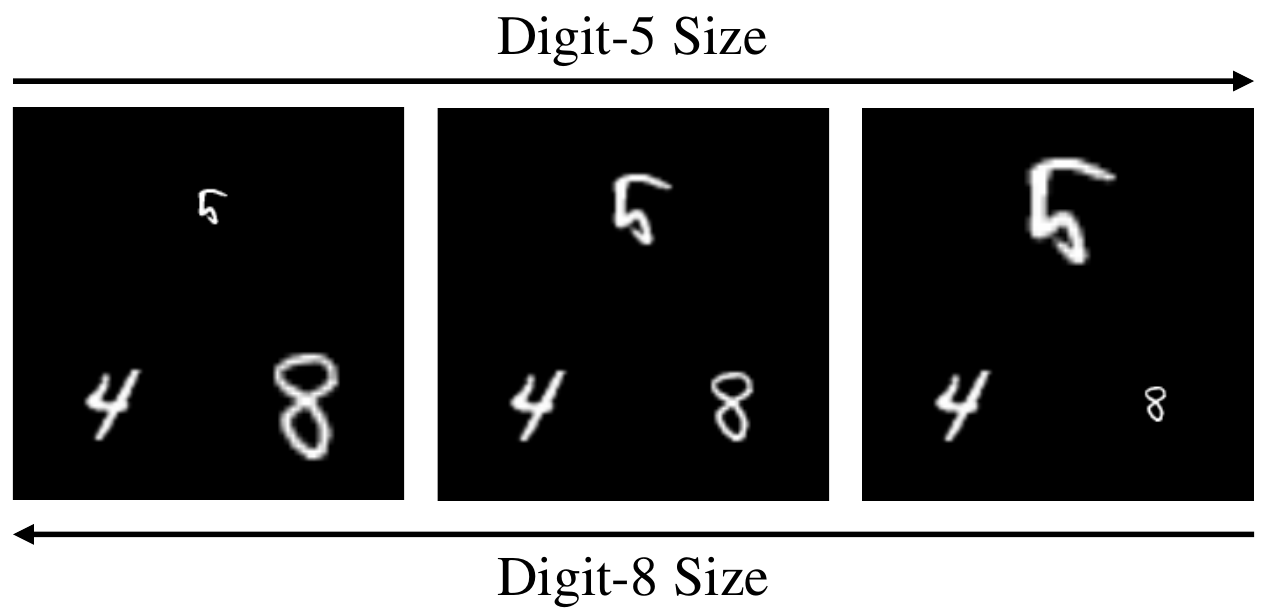}
        \includegraphics[width=0.48\linewidth]{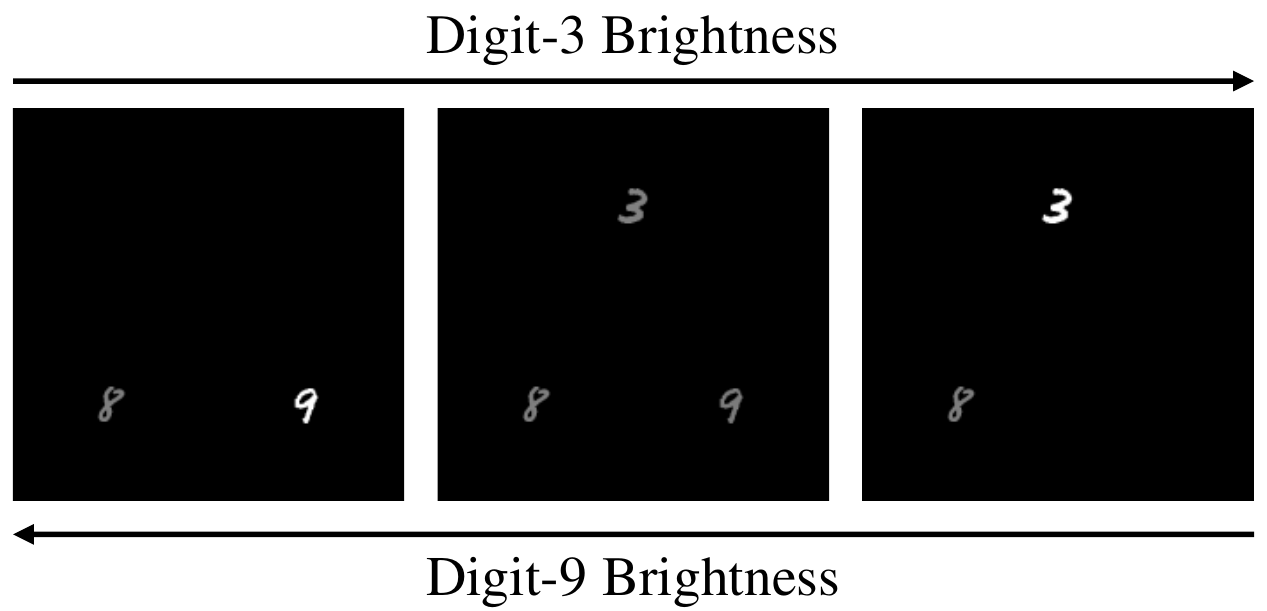}
    \label{fig:sequence}
    \end{subfigure}
    
    \begin{subfigure}[b]{1.0\linewidth}
        \centering
    
        \setlength\tabcolsep{0.2pt}
        \begin{tabular}[b]{cccccc}
    
            \includegraphics[width=0.16\linewidth]{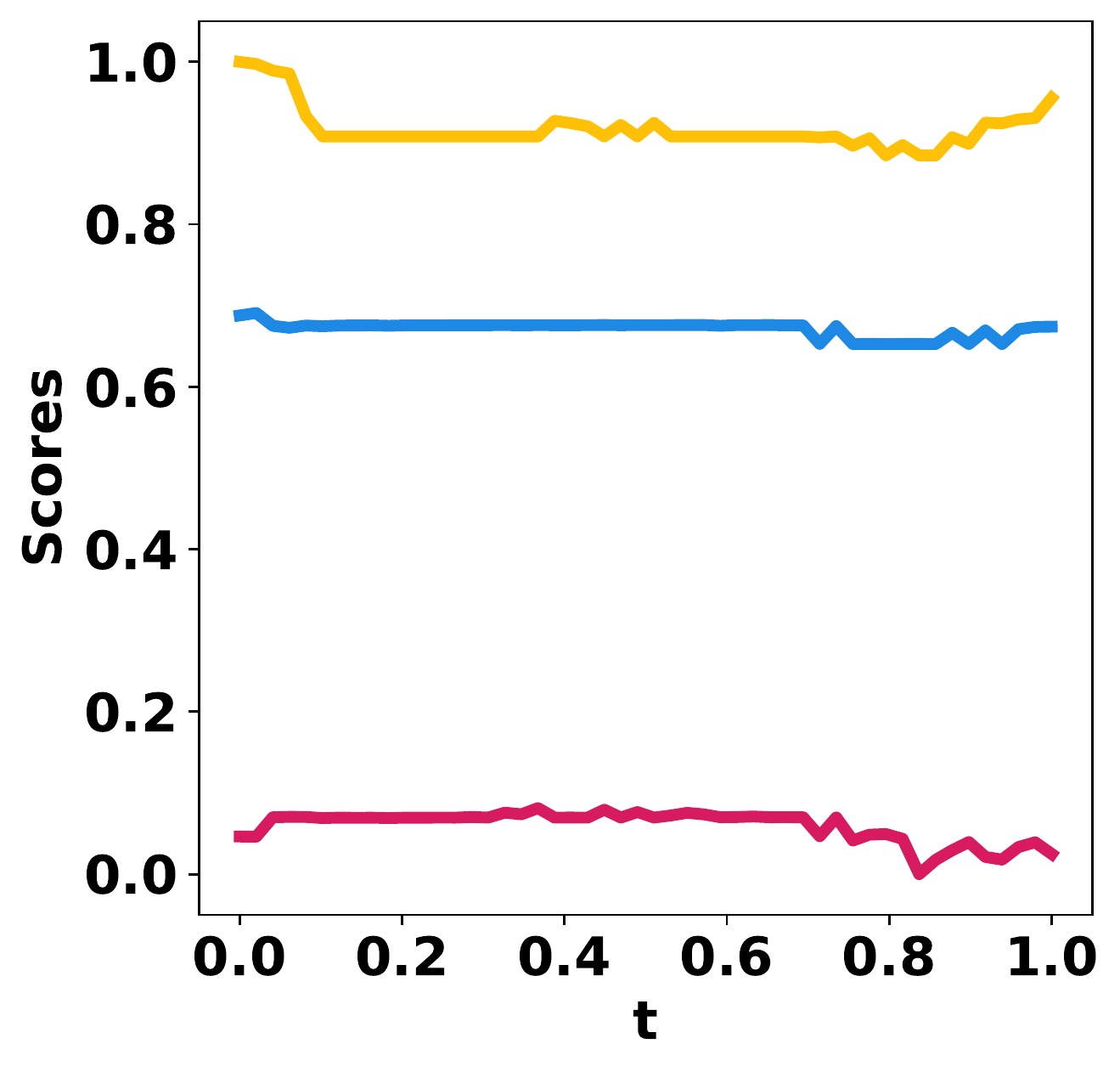} &
            \includegraphics[width=0.16\linewidth]{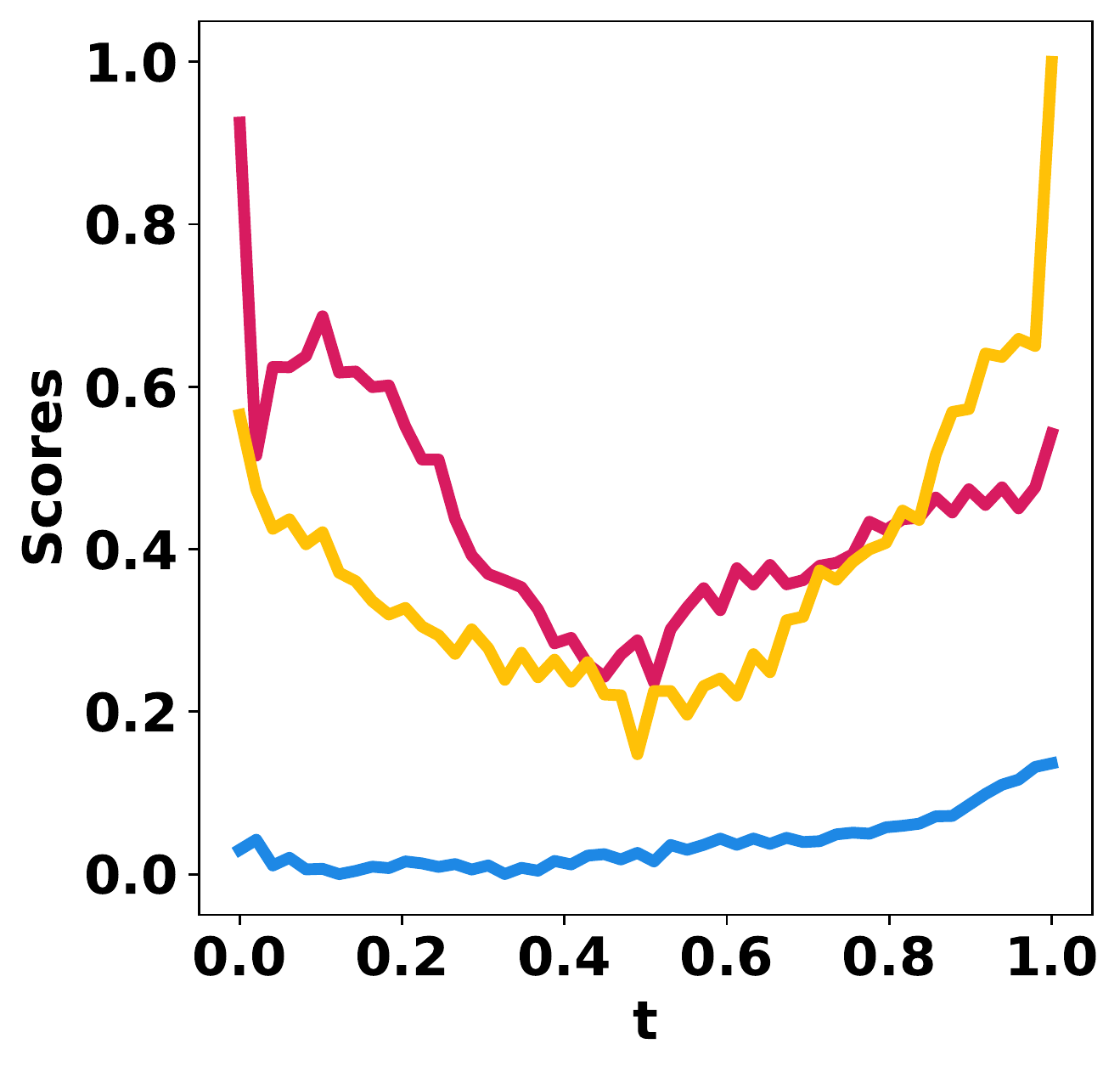} &
            \includegraphics[width=0.16\linewidth]{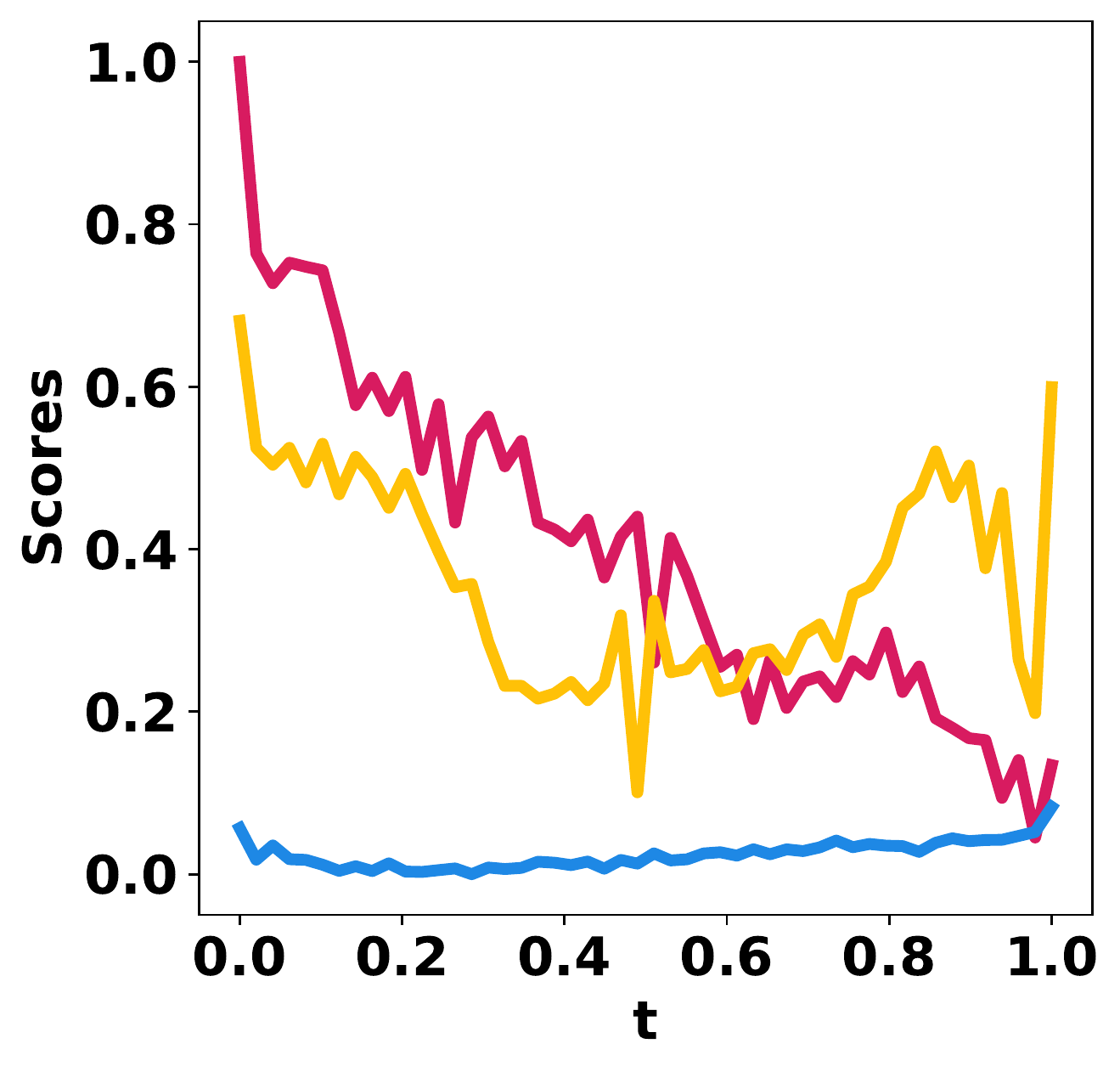} &
            \includegraphics[width=0.16\textwidth]{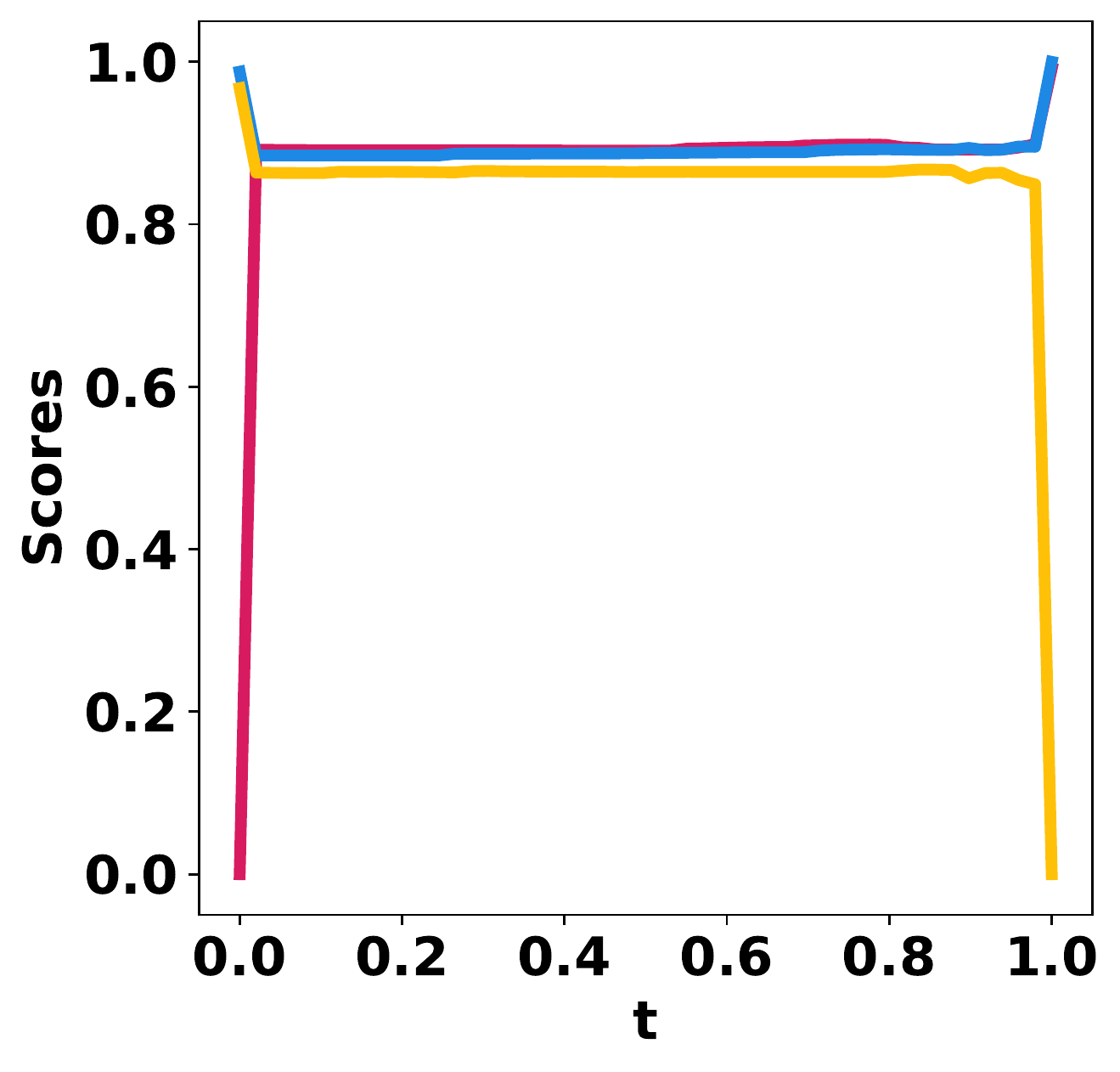} &
            \includegraphics[width=0.16\textwidth]{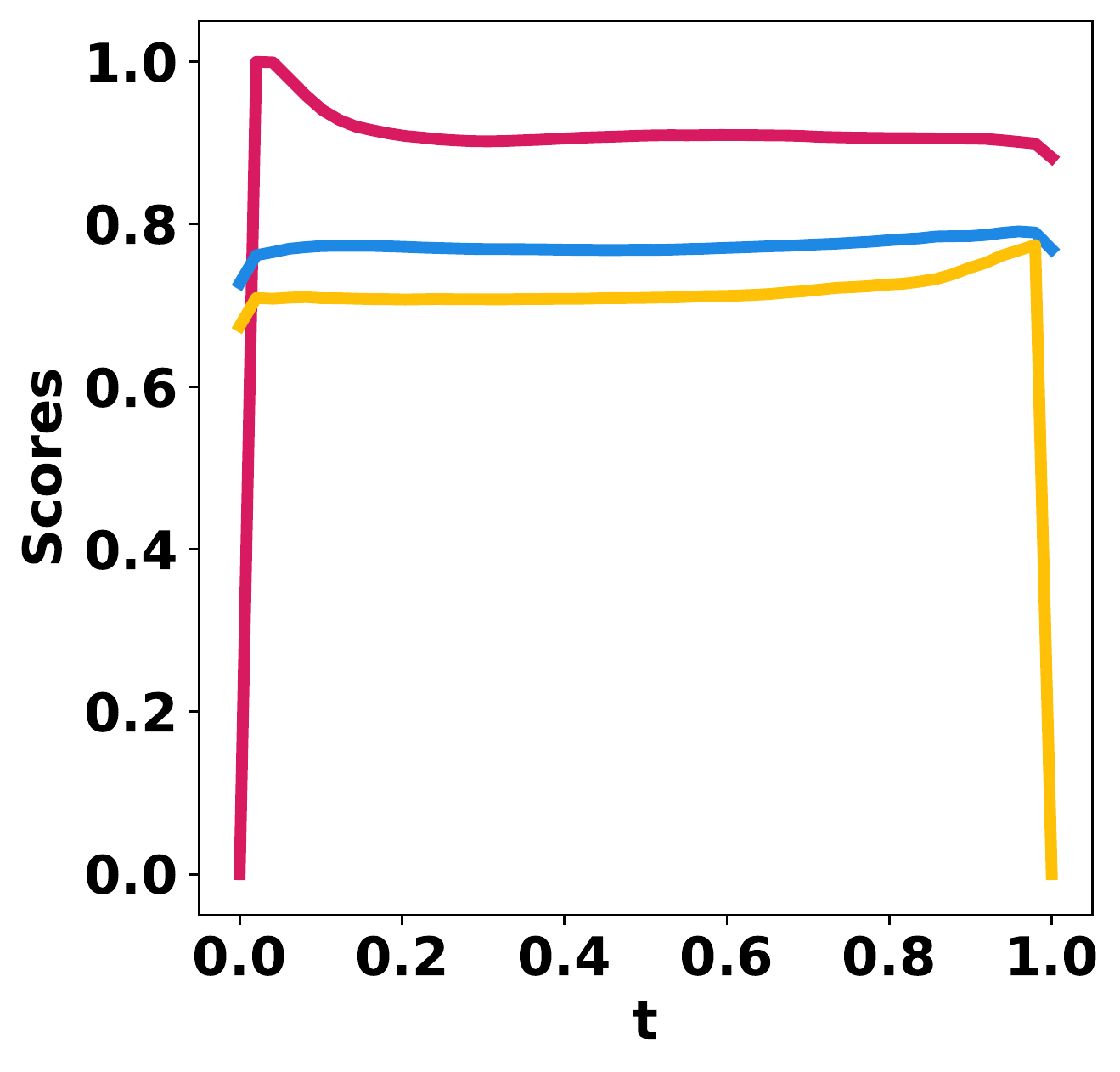} &
            \includegraphics[width=0.16\textwidth]{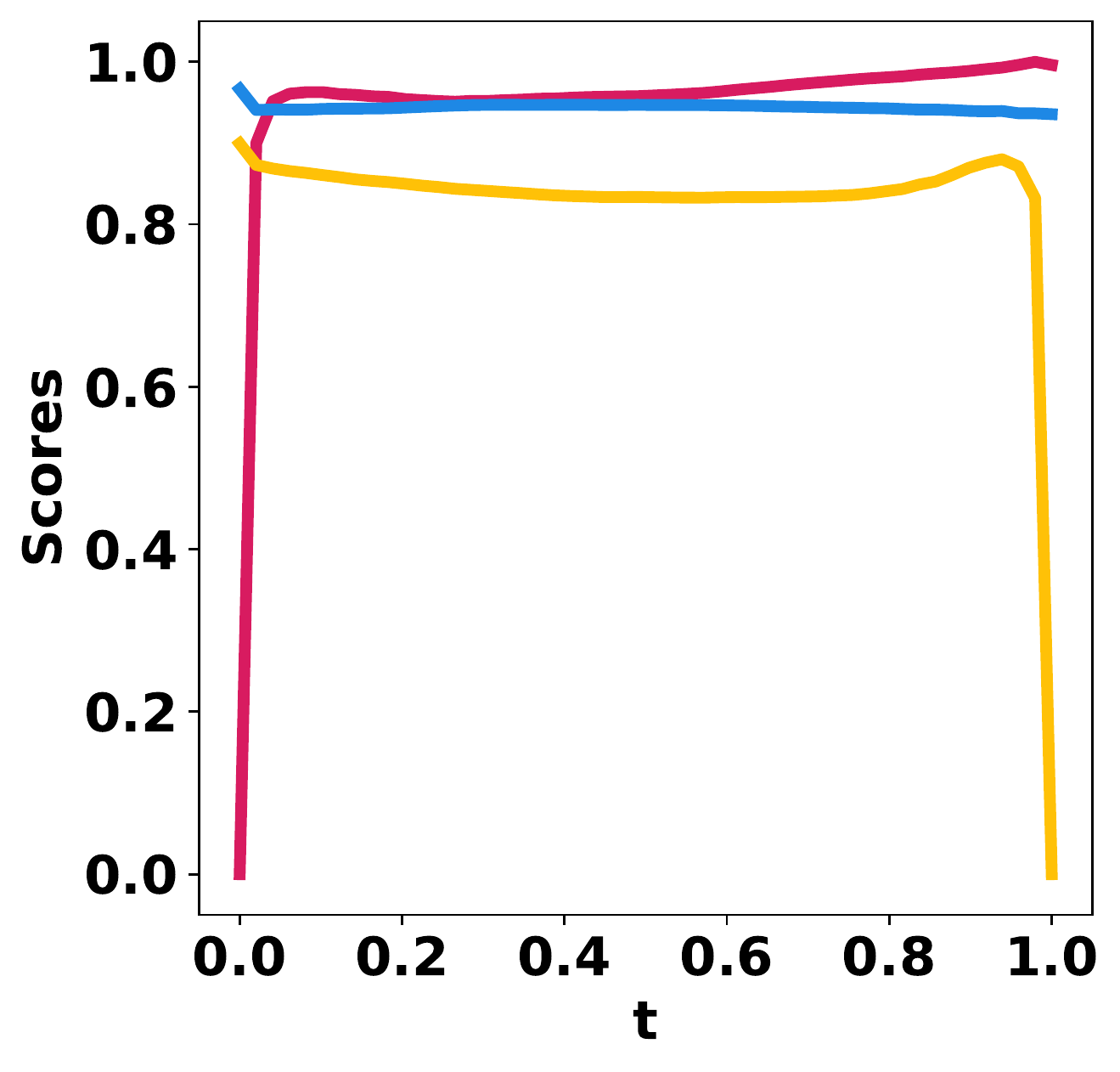} \\
            \hspace{0.12in} \small CRPC-Weak & \hspace{0.12in}  \small LSEP-Weak & \hspace{0.12in} \small GMLR-Weak &   \hspace{0.12in} \small CRPC-Weak & \hspace{0.12in} \small LSEP-Weak & \hspace{0.12in} \small GMLR-Weak
        
        \end{tabular}
    \end{subfigure}
    
    \begin{subfigure}[b]{1.0\linewidth}
        \centering
        
        \setlength\tabcolsep{0.2pt}
        \begin{tabular}[b]{cccccc}
            &&&&&\\
            \includegraphics[width=0.16\linewidth]{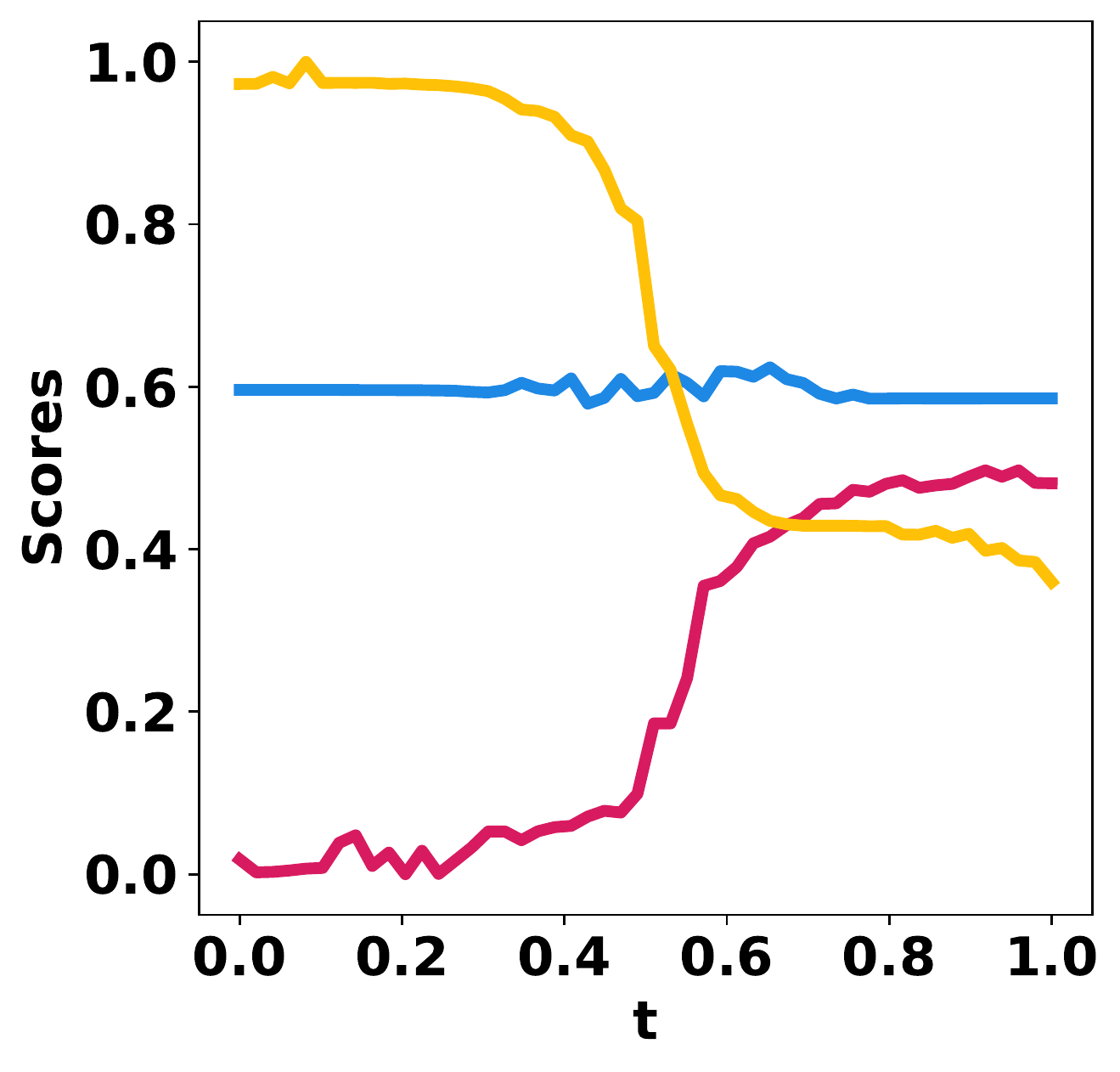} &
            \includegraphics[width=0.16\linewidth]{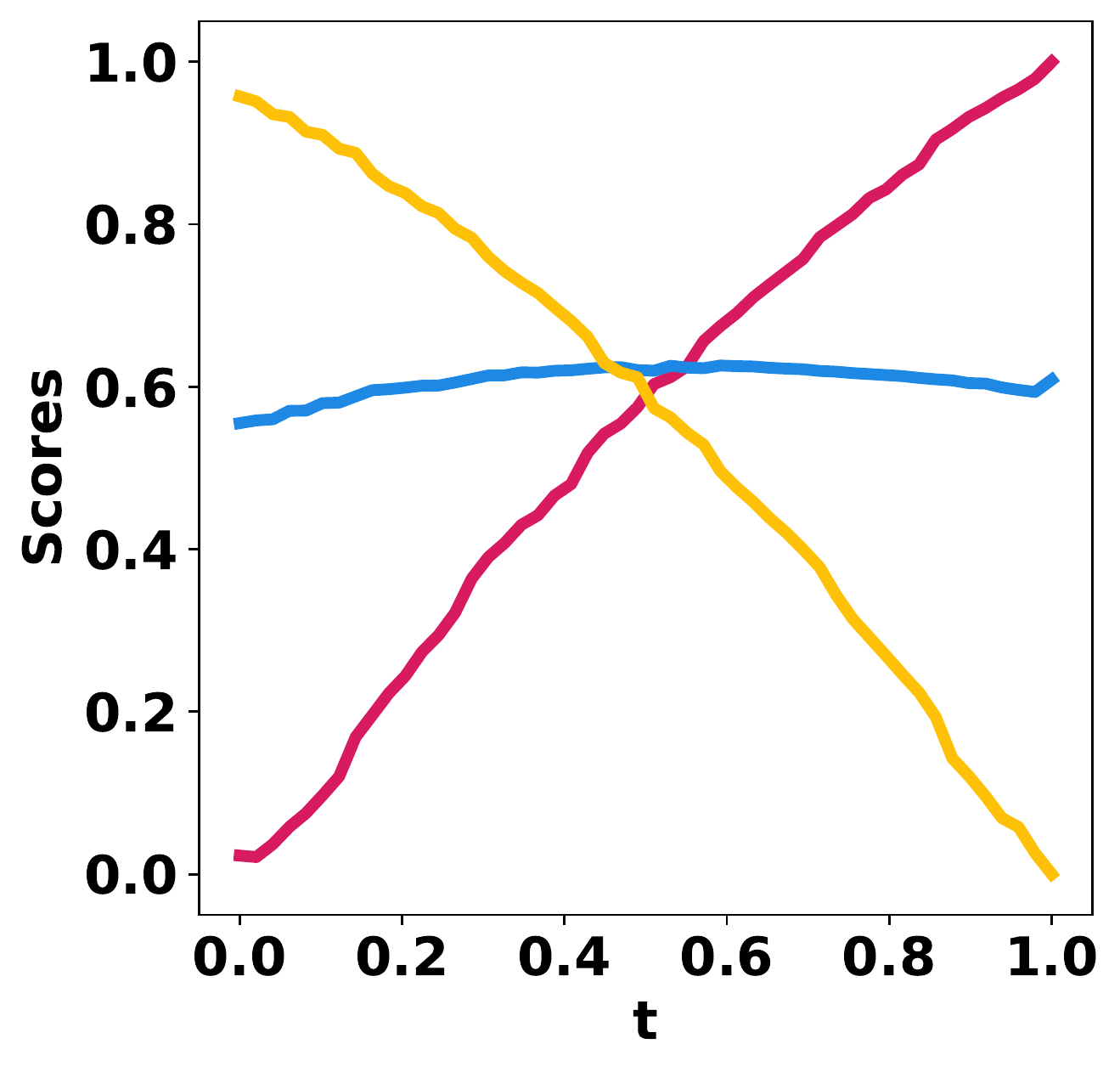} &
            \includegraphics[width=0.16\linewidth]{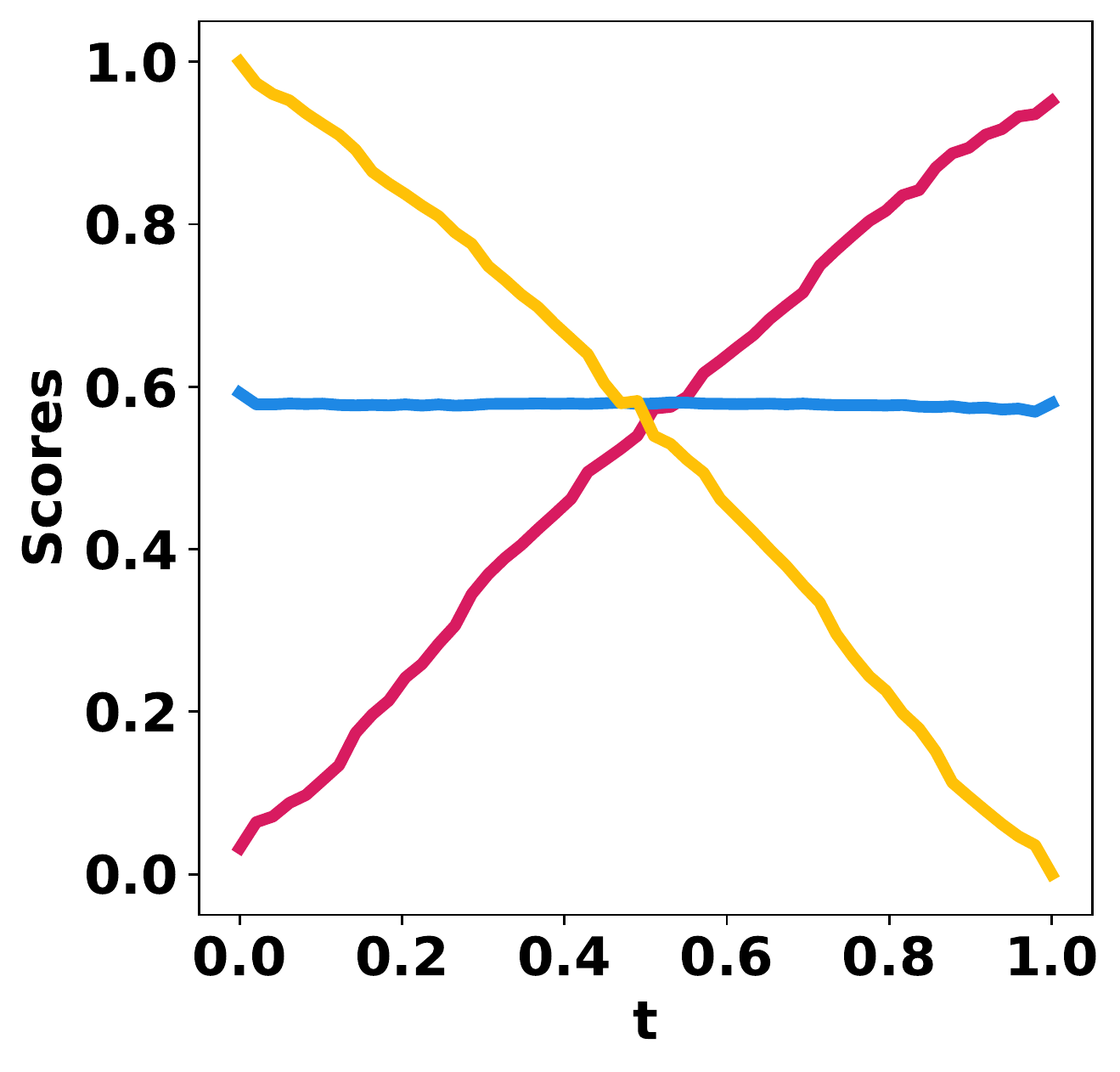} &
            \includegraphics[width=0.16\textwidth]{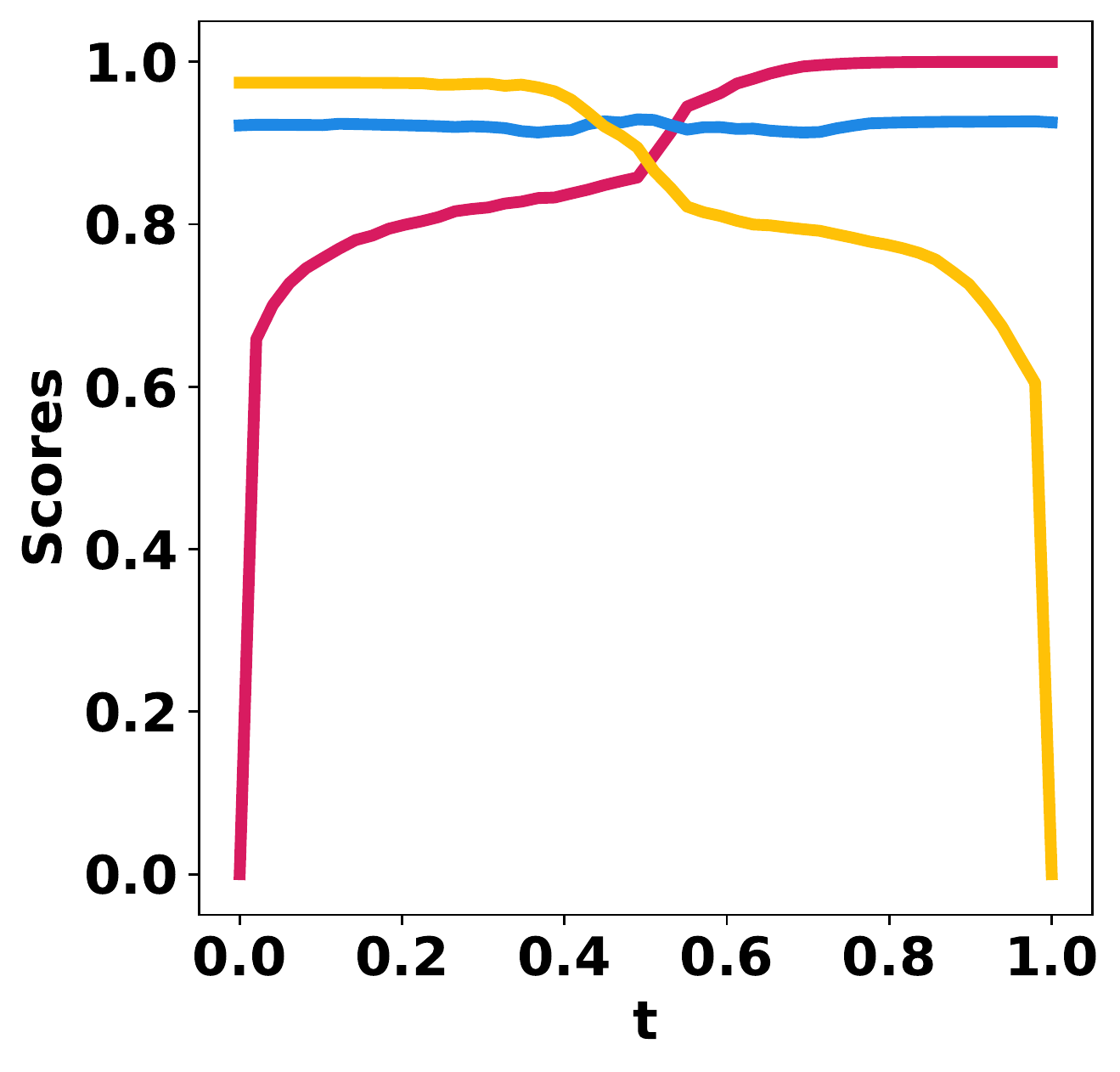} &
            \includegraphics[width=0.16\textwidth]{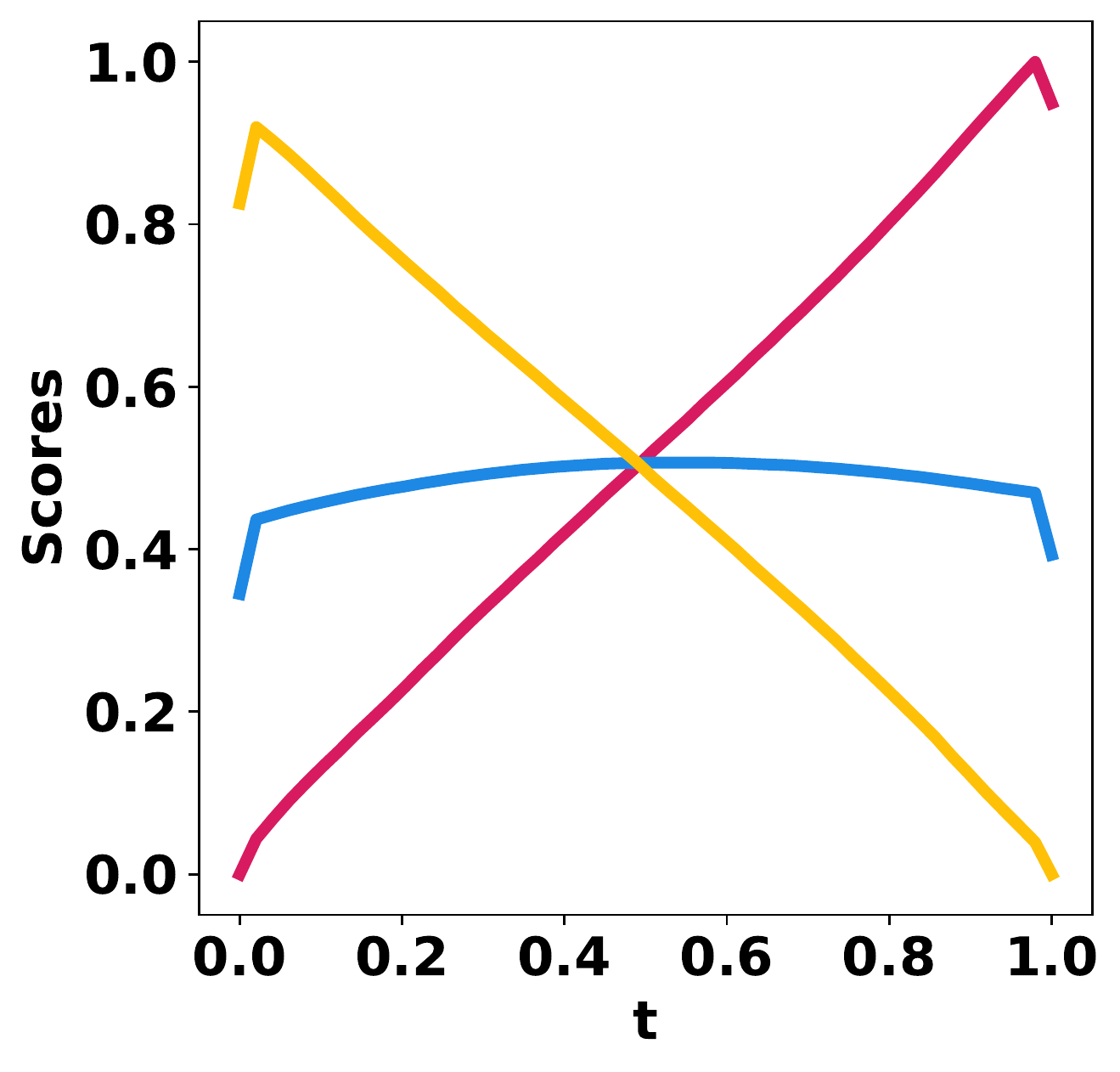} &
            \includegraphics[width=0.16\textwidth]{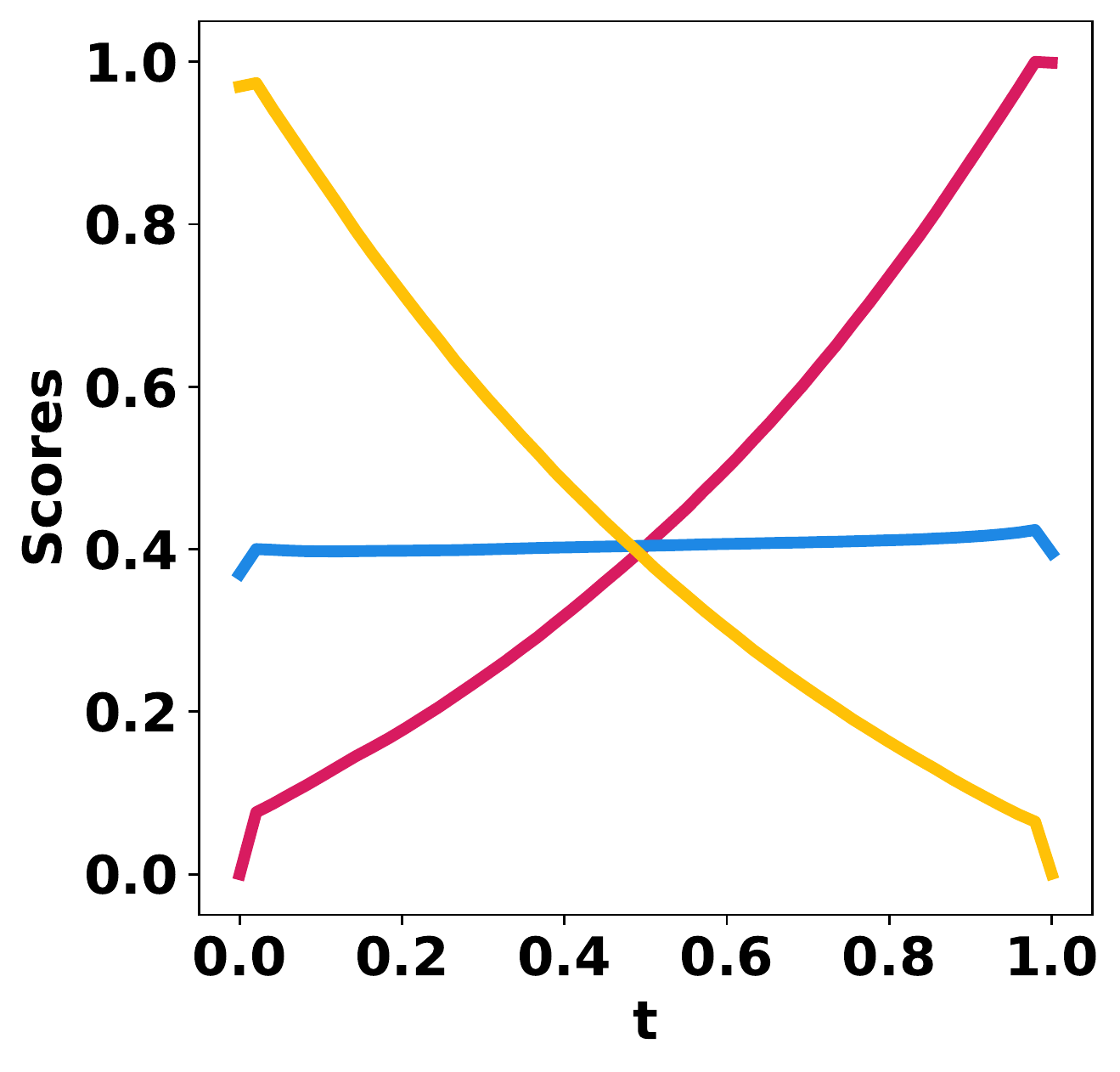} \\
             \hspace{0.12in} \small CRPC-Strong & \hspace{0.12in} \small LSEP-Strong & \hspace{0.12in} \small GMLR-Strong &   \hspace{0.12in} \small CRPC-Strong & \hspace{0.12in} \small LSEP-Strong & \hspace{0.12in} \small GMLR-Strong
        \end{tabular}
    \end{subfigure}
    \caption{Gradually changing significance effects in the sequences are shown at the top row, where the importance factor is the size of digits in top-left and brightness of digits in top-right. Lines demonstrate changes in scores of $\langle$\textbf{\textcolor{1st}{1st}}, \textbf{\textcolor{2nd}{2nd}}, \textbf{\textcolor{3rd}{3rd}}$\rangle$ digits, which are in the order of $\langle$5, 4, 8$\rangle$ in top-left and $\langle$3, 8, 9$\rangle$ in top-right. As GaussianMLR (GMLR) produces concurrently adjusting significance scores, it compares favorably over the baseline methods: CRPC and LSEP.}
    \label{fig:interpolation}
\end{figure}

To analyze the learned rank scores of GMLR and baseline methods, we first conduct an experiment where we gradually adjusted the selected effects. We have two setups in the experiment: changing scale and changing brightness. For each setup, we generate a set of sequences $\mathcal{D}_a = \{\mathcal{S}_1, ..., \mathcal{S}_{50}\}$, where each sequence consists of gradually changing images, i.e. $\mathcal{S}_i = \langle \vb*{x}_{i}^{(1)}, ..., \vb*{x}_{i}^{(50)} \rangle$. Each sequence $S_i$ consists of three random MNIST digits, let us name them $y_i^{low}$, $y_i^{middle}$ and $y_i^{high}$, the starting image of the sequence $\vb*{x}_i^{(1)}$ has the corresponding significance values $s^{low}$, $s^{middle}$ and $s^{high}$. Iterating over the images of any sequence $\mathcal{S}_i$, the significance value for $y_i^{low}$ linearly changes from $s^{low}$ to $s^{high}$, for $y_i^{high}$ changes from $s^{high}$ to $s^{low}$, and $y_i^{middle}$ remains constant. In the top row of Figure \ref{fig:interpolation} you can see how the images change for each setup. For each method and setup, we obtain the rank scores of the images in $\mathcal{D}_a$ using the network trained with the corresponding method on the corresponding dataset Ranked MNIST Gray-S/B. The average value of each position over all sequences $\mathcal{S}_i$ are calculated for $y_i^{low}$, $y_i^{middle}$ and $y_i^{high}$, and we show how the predicted rank scores change in Figure \ref{fig:interpolation} for both strong and weak baselines, and GMLR.

\begin{figure}[ht]
\centering
\vspace*{1cm}
    \begin{subfigure}[b]{1.0\linewidth}
        \centering
        \setlength\tabcolsep{0.2pt}
        \begin{tabular}[b]{ccccc}
            \includegraphics[width=0.14\linewidth]{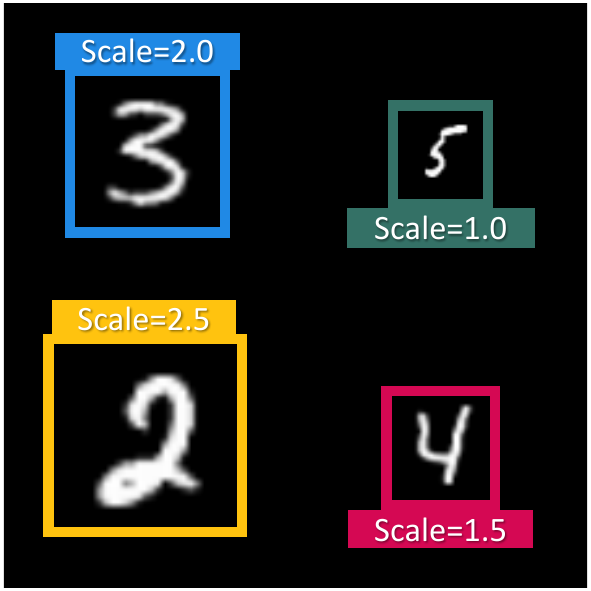} &
            \includegraphics[width=0.14\linewidth]{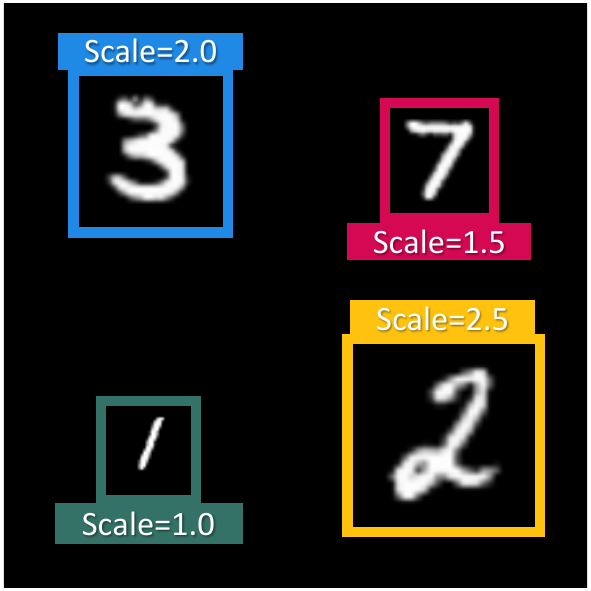} &
            \includegraphics[width=0.23\textwidth]{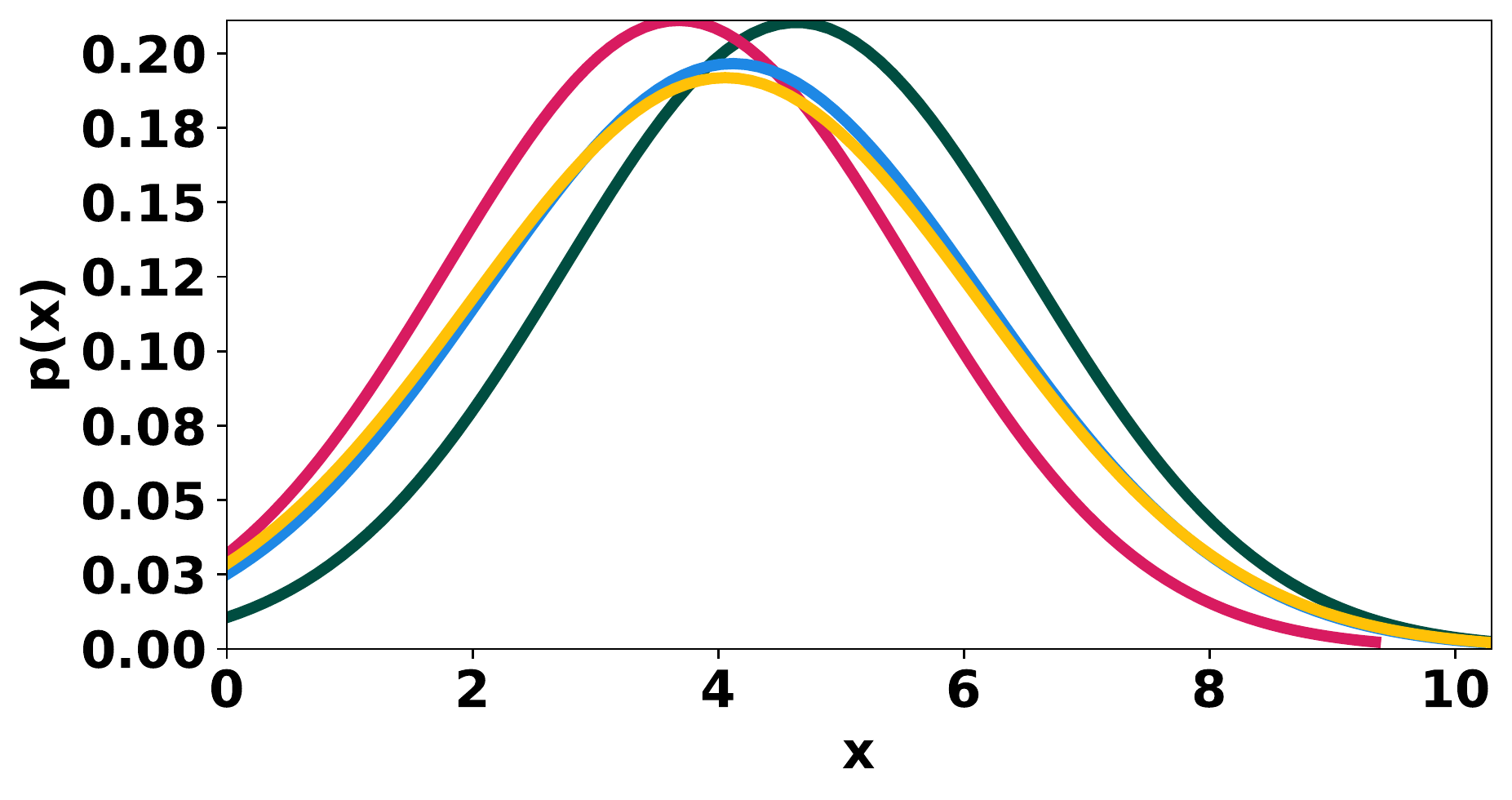} &
            \includegraphics[width=0.23\textwidth]{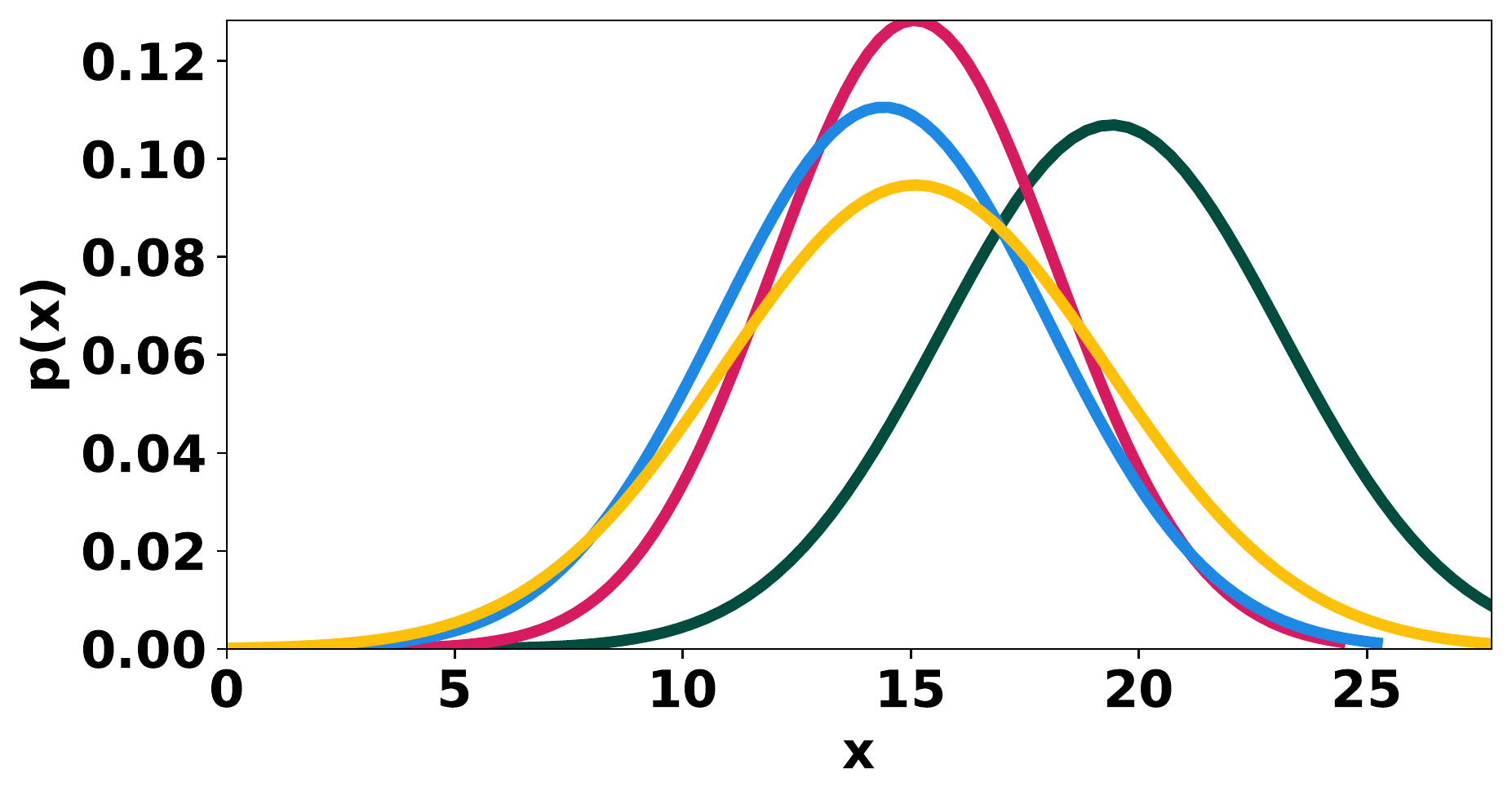} &
            \includegraphics[width=0.23\textwidth]{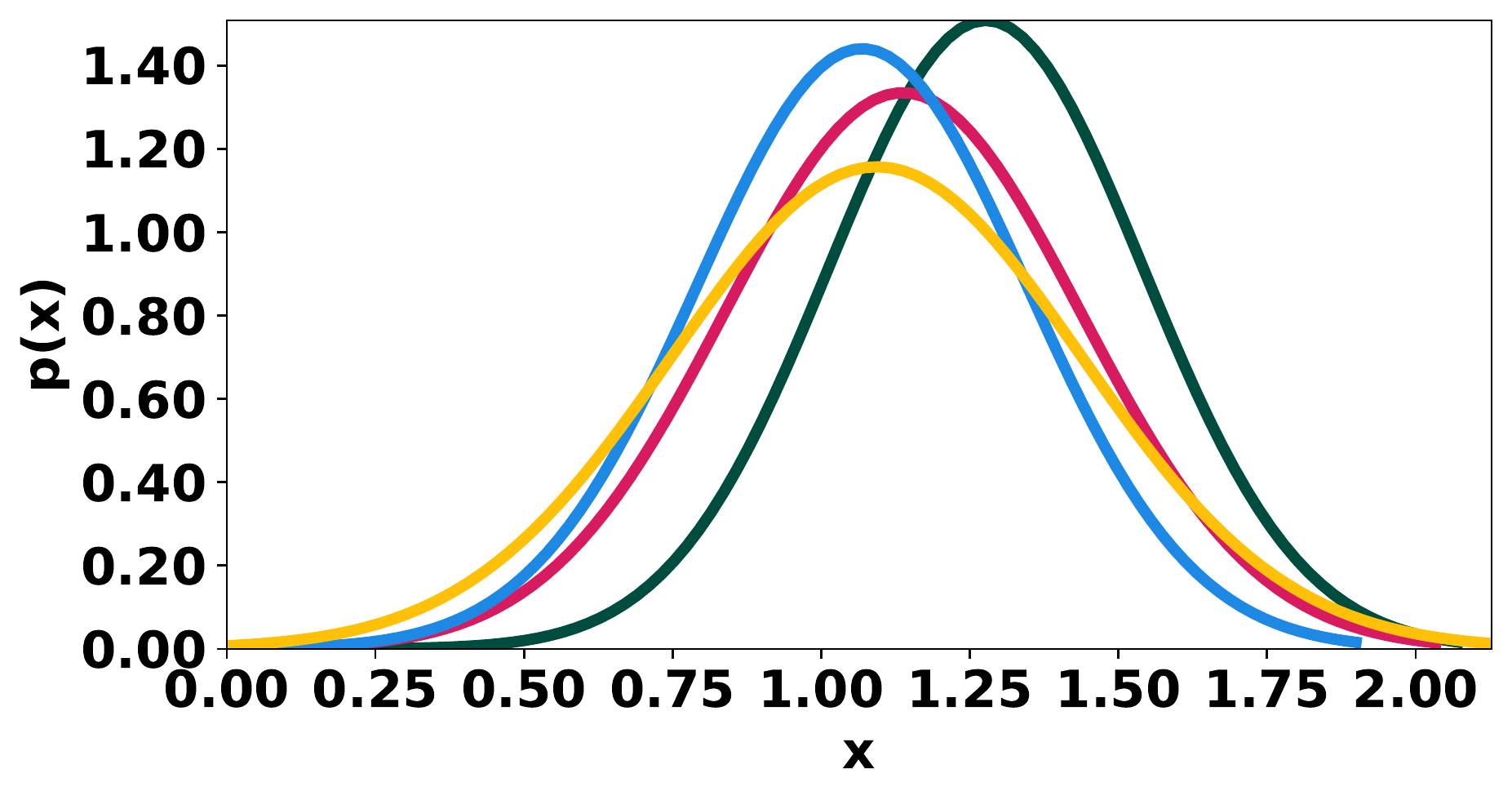}\\
            & &  \hspace{0.12in} CRPC-Weak & \hspace{0.12in}  LSEP-Weak & \hspace{0.12in}  GMLR-Weak \\
        \end{tabular}
    \end{subfigure}
    
    \begin{subfigure}[b]{1.0\linewidth}
        \centering
        \setlength\tabcolsep{0.2pt}
        \centering
        \begin{tabular}[b]{ccccc}
            \includegraphics[width=0.14\linewidth]{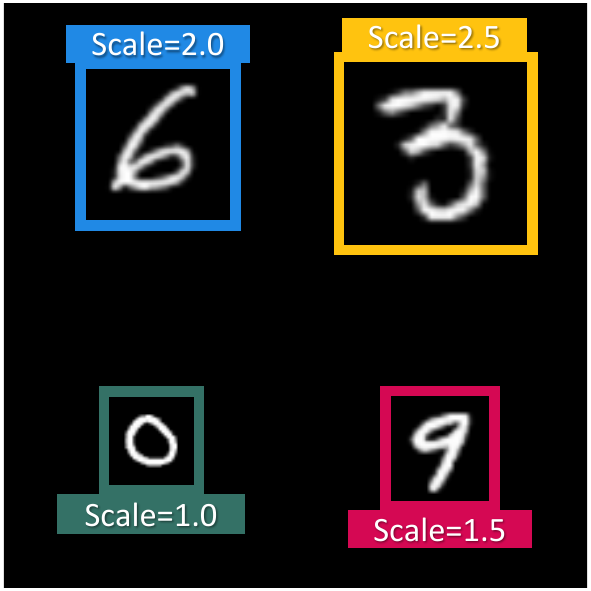} &
            \includegraphics[width=0.14\linewidth]{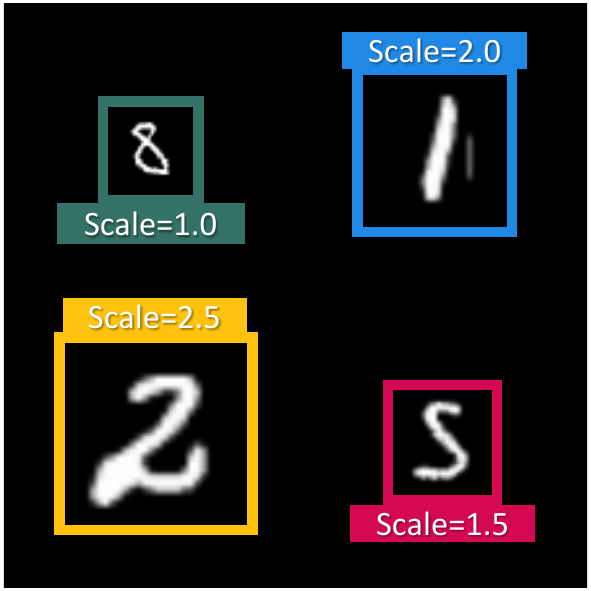} &
            \includegraphics[width=0.23\textwidth]{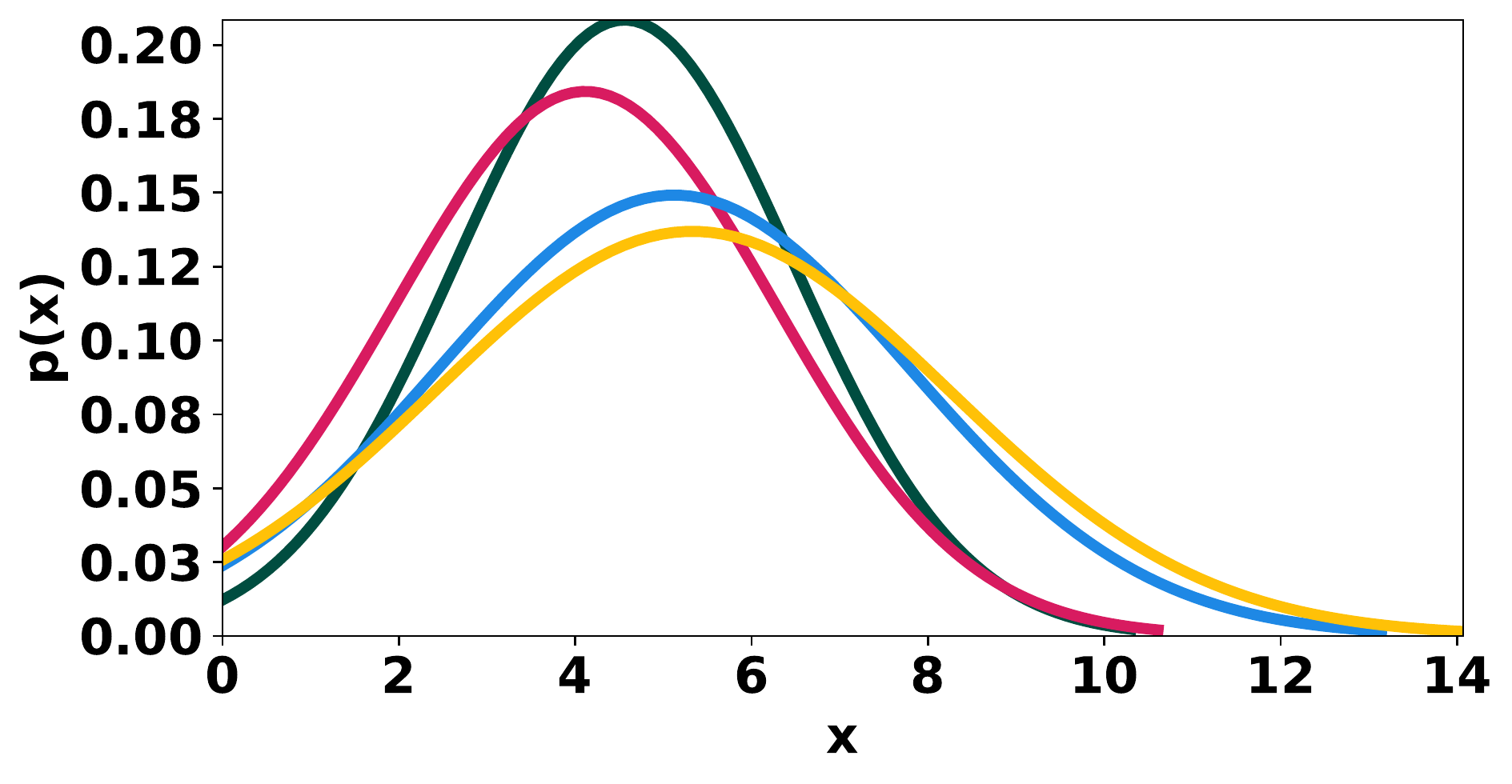} &
            \includegraphics[width=0.23\textwidth]{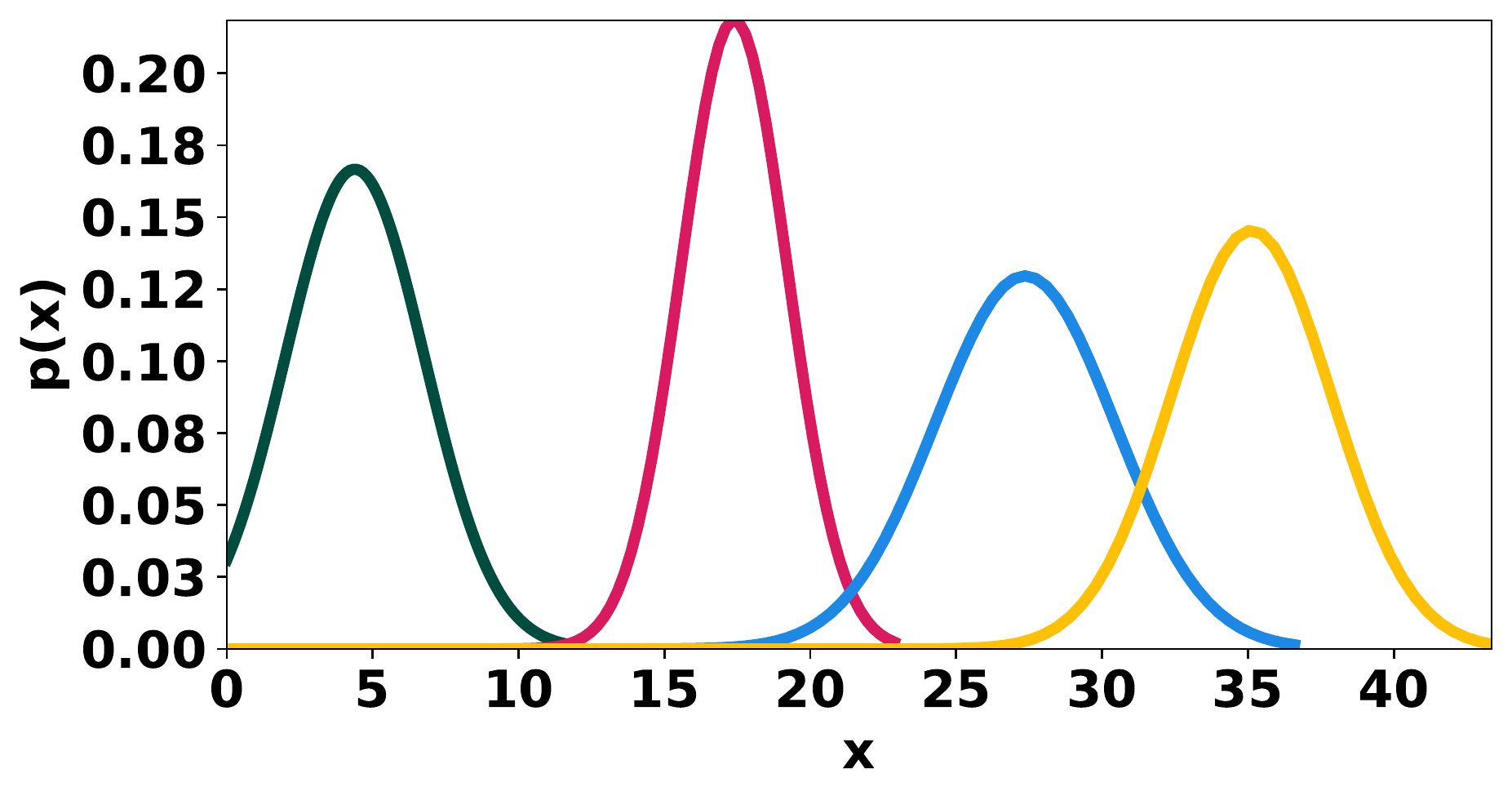} &
            \includegraphics[width=0.23\textwidth]{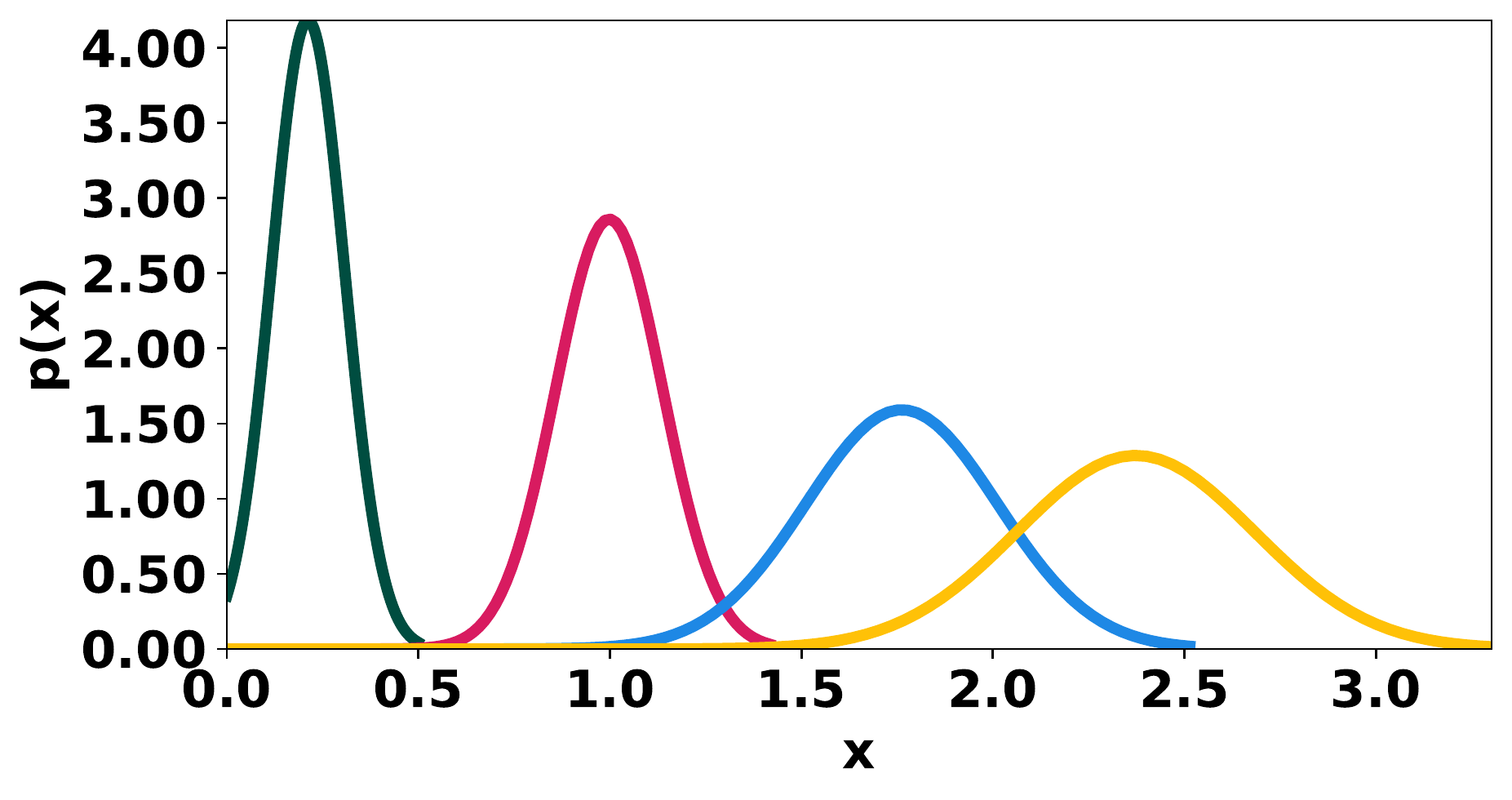}\\
            & & \hspace{0.12in} CRPC-Strong  & \hspace{0.12in} LSEP-Strong & \hspace{0.12in} GMLR-Strong \\
        \end{tabular}
    \end{subfigure}
\caption{Annotated samples of 4-digit Ranked MNIST Gray-S dataset are shown in the first two columns from the left, where the scale factors of each digit is demonstrated by the bounding boxes for \textbf{\textcolor{4th}{Scale=1.0}}, \textbf{\textcolor{1st}{Scale=1.5}}, \textbf{\textcolor{2nd}{Scale=2.0}}, \textbf{\textcolor{3rd}{Scale=2.5}}. Resulting plots for each method when the set of scores for each significance value (scale) is fit to a Gaussian distribution are given in the other three columns. GMLR captures proportional scores to significance values of each digit.} 
\label{fig:calibration}
\end{figure}

\subsection{Calibration Experiment}
\label{56}
The calibration experiments are conducted to see how the scores for each baseline are distributed for different instances of the same significance values. We start by generating an image set $\mathcal{D}_C = \{\vb*{x}^{(1)}, ..., \vb*{x}^{(50)}\}$, where each $\vb*{x}^{(i)}$ consists of MNIST digits and is associated with a label set $\mathcal{Y}^{(i)} \subseteq Y$ where $|\mathcal{Y}^{(i)}| = 4$. Each positive class $y_j \in \mathcal{Y}^{(i)}$ in an image $\vb*{x}^{(i)}$ has a one-to-one mapping to $\{1.0, 1.5, 2.0, 2.5\}$ which defines their significance value, in this case the scale value. For each method in our experimental setup, we train a network on Ranked MNIST Gray-S, then we feed each image $\vb*{x}^{(i)}$ to the network to produce score vectors $\hat{\vb{s}}^{(i)}$. From each score vector $\hat{\vb{s}}^{(i)}$ we select the scores associated with each value in $\{1.0, 1.5, 2.0, 2.5\}$ then create a set of scores for each underlying significance value $\mathcal{S}_{1.0}$, $\mathcal{S}_{1.5}$, $\mathcal{S}_{2.0}$, $\mathcal{S}_{2.5}$. For each set we fit a Gaussian distribution to the values, the resulting plots for each method are given in Figure \ref{fig:calibration} with a visual expressing the experiment. Both LSEP and GMLR extract an inherent calibration of significance scores, whereas in terms of their magnitude and variance, the order of distributions for each digit is best reflected by GMLR. Another observation from our experiments is that, which is also exemplified in Figure \ref{fig:calibration}, GMLR produces significance scores with larger variance for larger objects. This is due to naturally increased data variations owing to larger object extent. This positively distinct impact is not observed with other methods that entail intrinsic noise surpassing this effect, which is further explained in Appendix \ref{sec:apx9}.

\subsection{Extracted Significance Value Experiment}
\label{57}
\begin{figure}[H]
    \centering
    \includegraphics[width=0.8\textwidth]{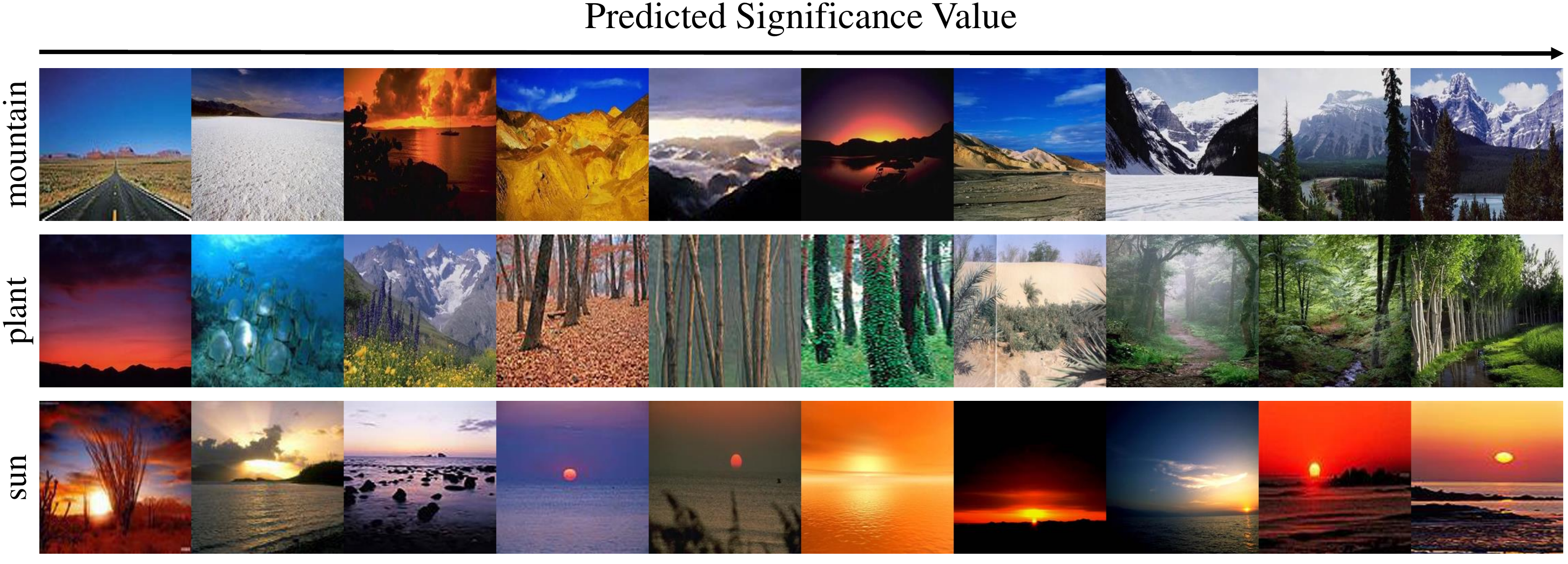} 
    \caption{Three sequences of images sampled from the test set of Natural Scene Images database, sorted in the order of predicted significance values for each class by GMLR-Strong. The ordering of images demonstrates that GaussianMLR extracts class significance values that determine their rank proportionally to the dominant class in the image.}
    \label{fig:int-gmlr}
\end{figure}

In this experiment, instead of creating gradually changing images to test if an MLR method produces consistent predictions with the underlying process, we do the opposite. Assuming the underlying process exists, we visualize which images in our test set would generate consistent scores with the process. We order the images according to the predicted significance value by our trained model for each class. Then, we select 10 images as checkpoints among the set of 400 test images, by choosing equidistant images in the interval. These image sequences sorted as in Figure \ref{fig:int-gmlr} demonstrate that as the predicted significance value for a class increases, the dominance of that class also increases. These results suggest that GaussianMLR extracts a proportional score to the underlying significance value for a class. 

\section{Conclusion}
\label{sec:conclusion}
While previously studied weak multi-label ranking methods learn almost no useful information about the underlying significance values of the positive classes, the strong multi-label ranking paradigm of GMLR yields remarkably calibrated significance values. GMLR compares favorably to the competing baselines as demonstrated by the experiments, where the concurrent gradual changes in the scores for the changing effects in images as well as the constant scores for the static effects indicate that the underlying appearance and geometric characteristics pertinent to ranking are learned by GMLR.

    
    
    

\paragraph{Broad Impact.}

GaussianMLR encourages new ideas by providing a fresh perspective into the field of MLR. It introduces a set of datasets (Ranked MNISTs), which construct a controllable experimental environment for the new MLR paradigm.  Not only GMLR shows the potential on learning more than a ranking between labels but also its output converges to a distribution that is proportional to the underlying significance value process of data characteristics. For a dataset where the labels of any instance are weighted by an unknown factor which controls their relevancy to the instance, GMLR provides a way to extract these unknown factors by only using the ordering between the labels.   These findings we believe have potential in numerous  applications where ordering of factors are of value for instance in generation, design, or in decision making where alternative choices are typically pairwise ranked.
  
\paragraph{Limitations and Ethical Concerns. } 
 In the absence of strong MLR paradigm, it is an open question whether the MLR can reach or surpass the results provided by GMLR.   The true potential of Strong MLR paradigm can be further appreciated with availability of more public synthetic and real-life datasets. 
 GMLR does not impact the explainability and fairness of the decisions made by the underlying design choices, such as the architecture of the used neural nets in the pipeline. 
 GMLR calls for further experimental and theoretical studies in learning calibrated significance values.

%
%

\clearpage

\appendix

\section{Datasets}
\label{sec:apx1}

\textbf{Natural Scene Images Database.} Natural scene images database\cite{InconsistentRankers} consists of 2000 images of natural scenes, for example: cloud, desert, mountain etc. The dataset has multiple labels per image, to convert them into single label, we applied mean rank ordering\cite{Brinker2007CaseBasedMR} as a ranking aggregation method described in their paper. We took a random split of images into sets with size 1600 and 400 to create our train and test sets accordingly. The dataset is public and can be found in \href{http://ldl.herokuapp.com/download}{http://ldl.herokuapp.com/download}.

\textbf{Architectural VDP Dataset.} Architectural Visual Design Principles (VDP)\cite{DEMIR2021103826} dataset consists of 3654 train and 407 test images where the labels are: asymmetric, color, crystallographic, flowing, isolation, progressive, regular, shape, symmetric. Images are associated with a maximum of 3 positive classes, and each positive class is ranked by their dominance over the other classes in representing the image. The authors of the paper can not provide the dataset publicly, and we obtained the dataset by asking from the authors. Some of the publishable images and their scores are provided in Figure \ref{fig:AVDP-bar}.

The datasets have no harmful or offensive content in them. In our work we only provide publishable material, further details of the datasets are held by the respective authors. 

\section{Implementation Details}
\label{sec:apx2}


For all experiments we use a neural network with ResNet18\cite{he2015deep} feature extractor with feature size $512$ and an additional $512\cross512$ fully connected layer.

For \textbf{CRPC} we add a fully connected layer with output size of $K(K-1)/2$, where K is the number of classes including the virtual label. The virtual label is used to determine \textit{positive} and \textit{negative} predictions such that if predicted score is higher than virtual label's score it is \textit{positive} and vice versa. Here each output value corresponds to the logits for the relation between a unique pair of items. Then for each logit sigmoid function is applied and used for a binary classification, where for a pair $(y_u, y_v)$ if $y_u \succ y_v$ then the ground truth is \textit{positive}, else it is \textit{negative}.

\begin{align*}
    L_{CRPC (weak)} = \sum_{i=1}^{N}\sum_{(y_u, y_v) \in \mathcal{S}} - \log \sigma(f_{(u,v)}(\vb*{x}^{(i)}))\beta_u -\log (1 - \sigma(f_{(u,v)}(\vb*{x}^{(i)})))\beta_v,
\end{align*}
where $\beta_u = \mathbb{I}[y_u \in \mathcal{Y}^{(i)} \wedge (y_v \notin \mathcal{Y}^{(i)} \lor y_v = v_0 )]$, $\beta_v = \mathbb{I}[(y_u \notin \mathcal{Y}^{(i)} \lor y_u = v_0) \wedge y_v \in \mathcal{Y}^{(i)}]$, $v_0$ is the virtual label, $\mathcal{Y}^{(i)}$ is the set of positive classes for the instance $\vb*{x}^{(i)}$, $\sigma(\cdot)$ is the sigmoid function, $\mathcal{S}$ is the set of unique pairs on the class set $Y$, where each unique pair corresponds to one output value $f_{(u,v)}$. Further details can be found in \cite{10.5555/1567016.1567123}.\\
For the strong version of CRPC we change the loss function into:
\begin{align*}
    L_{CRPC (strong)} = \sum_{i=1}^{N}\sum_{(y_u, y_v) \in \mathcal{B}^{(i)}} -\log \sigma(f_{(u,v)}(\vb*{x}^{(i)})),
\end{align*}
where $\mathcal{B}^{(i)}$ is the ground truth bucket order for instance $\vb*{x}^{(i)}$.

For \textbf{LSEP} we add two parallel fully connected layers with output size $K$ on top of the feature extractor for scores and thresholds. LSEP consists of two stages, first the score layer is trained with a ranking loss:

\begin{align*}
    L_{LSEP/R (weak)} = \sum_{i=1}^N \log \left( 1 + \sum_{(y_u, y_v) \in \mathcal{B}^{'(i)}} \exp ( f_v(\vb*{x}^{(i)}) - f_u(\vb*{x}^{(i)}) ) \right),
\end{align*}

where $\mathcal{B}^{'(i)}$ only consists of \textit{positive} and \textit{negative} pairs, such that $(y_u, y_v) \in \mathcal{B}^{'(i)} \iff y_u \in \mathcal{Y}^{(i)} \land y_v \notin \mathcal{Y}^{(i)}$. The strong version can be modeled similarly by replacing $\mathcal{B}^{'(i)}$ with the real bucket order for $\vb*{x}^{(i)}$, $\mathcal{B}^{(i)}$:

\begin{align*}
    L_{LSEP/R (strong)} = \sum_{i=1}^N \log \left( 1 + \sum_{(y_u, y_v) \in \mathcal{B}^{(i)}} \exp ( f_v(\vb*{x}^{(i)}) - f_u(\vb*{x}^{(i)}) ) \right).
\end{align*}

After training the score head, to be able the determine if a class is \textit{positive} or \textit{negative} all layers of the network except the threshold layer is frozen, then the network is optimized for classification:

\begin{align*}
    L_{LSEP/C} = -\sum_{i=1}^{N} \sum_{j=1}^{K} \mathbb{I}[y_j \in \mathcal{Y}^{(i)}]\log(\delta_j(\vb*{x}^{(i)})) + \mathbb{I}[y_j \notin \mathcal{Y}^{(i)}] \log(1 - \delta_j(\vb*{x}^{(i)})),
\end{align*}

where $\delta_k(\vb*{x}^{(i)}) = \sigma(f_k(\vb*{x}^{(i)}) - g_k(\vb*{x}^{(i)}))$, $f(\cdot)$ is the score head and $g(\cdot)$ is the threshold head. Further details for LSEP can be found in \cite{DBLP:journals/corr/LiSL17}. 

For \textbf{GaussianMLR} we add a fully connected layer with output size $2K$ for mean and log variance. The first $K$ values of the output vector are used as predicted mean $\hat{\mu}$ and the second $K$ values of the output vector are used as predicted logarithm variance $\log(\hat{\sigma}^{2})$ as commonly practiced in variational autoencoder frameworks. Our proposed loss function consists of two parts, to balance the magnitude of the two terms we use additional weights:

\begin{align*}
    \min_{\zeta} \dfrac{1}{N} \sum_{i=1}^N \lambda_1^{(i)} L_c(\hat{\mu}^{(i)}, \hat{\sigma}^{(i)}, \mathcal{Y}^{(i)}) + \lambda_2^{(i)} L_r(\hat{\mu}^{(i)}, \hat{\sigma}^{(i)}, \mathcal{B}^{(i)}),
\end{align*}

where $\lambda_1^{(i)} = 1 / K$ and $\lambda_2^{(i)} = 1 / |\mathcal{B}^{(i)}|$, $K$ is the number of classes and $\mathcal{B}^{(i)}$ is the bucket order for the instance $\vb*{x}^{(i)}$.

All networks are trained until convergence with Adam optimizer of learning rate $1.e-4$ and weight decay $1.e-5$. For real datasets we use the frozen ResNet18 feature extractor pretrained on ImageNet\cite{deng2009imagenet} and train for the rest of the layers. For Ranked MNIST the learning rate is decayed by 0.9 each epoch and for real datasets every 5 epochs. For real datasets we also use RandAugment\cite{DBLP:journals/corr/abs-1909-13719} and for RankedMNIST we apply no augmentation.

\section{Metrics}
\label{sec:apx3}

To evaluate our method quantitatively, we use both ranking and classification metrics. For ranking: Kendall's Tau-b ($\tau_b$), Spearman's rank correlation coefficient ($S\rho$), Goodman and Kruskal's gamma ($\gamma $), and for classification: Hamming Loss (HL), Max-1 error (M-1), and F1 Score are used. For a predicted and a ground truth ranking, the following are the notations used in formulations:

$N_c$ : the number of concordant pairs,\\
$N_d$ : the number of discordant pairs, \\
$N_0 = K(K-1)/2$ : the total number of pairs, \\
$N_1$ : total number of tied pairs in the prediction, \\
$N_2$ : total number of tied pairs in the labels, \\
$r_u$ : the rank of $y_u$ in the ground truth,\\
$\hat{r}_u$ : the predicted rank of $y_u$,\\
$\vb{y}$ :  the binary classification vector of the ground truth, where 1 indicates a positive class, \\
$\hat{\vb{y}}$ : the binary classification vector of the prediction, \\
TP, FP, FN : True Positive, False Positive and False Negative terms.

Formulations for each measure can be found in Table \ref{tab:metrics}. All metrics provided in the tables of quantitative results are the average of the individual metrics for the test sets. 

\begin{table}[H]
  \caption{Metrics used for quantitative evaluation. }
  \label{tab:metrics}
  \centering
  \begin{tabular}{p{0.16\linewidth}p{0.42\linewidth}p{0.30\linewidth}}
    \toprule
    \multicolumn{1}{l}{Algorithm} &Measure &  Formulation\\ 
    \midrule
    \multirow{3}{*}{ \vspace{-1.8cm} Ranking} & $\uparrow$ Kendall's Tau-b ($\tau_b$) & $\frac{(N_c - N_d)}{\sqrt{(N_0 - N_1)(N_0 - N_2)}}$ \\[4ex]
    
     & $\uparrow$ Spearman's $\rho$ ($ S \rho$)  & $1 - \dfrac{6\sum d_i^2}{K(K^2 - 1)}$\\
     &&\small (where $d_i = \hat{r}_i - r_i$) \\[2ex]
     & $\uparrow$ Goodman and Kruskal's gamma ($\gamma$)  & $\dfrac{N_c - N_d}{N_c + N_d}$ \\
    \midrule
     
    \multirow{3}{*}{ \vspace{-1.0cm} Classification} &$\downarrow$ Hamming Loss (HL)  & $\sum_{i=1}^K \mathbb{I}[\hat{\vb{y}}_i \neq \vb{y_i}] / K$ \\[3ex]
    
     & $\downarrow$ Max-1 Error (M-1)  &  $\mathbb{I}[\vb{y}_{\argmax_i r_i} \neq 1]$ \\[3ex]
     
     & $\uparrow$ F1 Score &   TP / (TP + 0.5(FP + FN))\\
    \bottomrule
  \end{tabular}
\end{table}

\section{Learning Implicit Class Significance}
\label{sec:apx4}

\begin{equation}
    \label{eqn:gaussian_mlr_loss2}
    \min_{\zeta} \dfrac{1}{N} \sum_{i=1}^N L_c(\hat{\mu}^{(i)}, \hat{\sigma}^{(i)}, \mathcal{Y}^{(i)}) + L_r(\hat{\mu}^{(i)}, \hat{\sigma}^{(i)}, \mathcal{B}^{(i)})
\end{equation}

Our dataset can be seen as a sample of a continuous data distribution $(\vb*{x}, \mathcal{Y}, \mathcal{B}) \sim \mathcal{D}$, with our objective function given in Equation \ref{eqn:gaussian_mlr_loss2}, we optimize the Monte Carlo estimation for the following intractable objective:
\begin{equation}
    \label{eqn:gaussian_mlr_expected}
    \min_{\zeta} \mathbb{E}_{(\vb*{x}, \mathcal{Y}, \mathcal{B}) \sim \mathcal{D}}\left[ L_c(\hat{\mu}, \hat{\sigma}, \mathcal{Y}) + L_r(\hat{\mu}, \hat{\sigma}, \mathcal{B})\right]
\end{equation}
where $f_{\mu}(\vb*{x}; \zeta) = \hat{\mu}$ and $f_{\sigma}(\vb*{x}; \zeta) = \hat{\sigma}$. For any input $\vb*{x}^{(i)}$ and for any two classes $y_u$ and $y_v$, let $s_u^{(i)}$ and $s_v^{(i)}$ be the underlying significance values of the classes, such that $s_u^{(i)} > s_v^{(i)} \iff y_u \succ y_v$ given $\vb*{x}^{(i)}$. Function $f(\vb*{x}; \zeta)$ is limited by $\zeta$ in terms of capacity, assuming $\zeta$ has finite dimensionality. Let us define $p = P(\hat{s}_u^{(i)} \geq \hat{s}_v^{(i)})$ where $\hat{s}_u^{(i)} \sim \mathcal{N}(\hat{\mu}_u^{(i)}, \hat{\sigma}_u^{2(i)})$ and $\hat{s}_v^{(i)} \sim \mathcal{N}(\hat{\mu}_v^{(i)}, \hat{\sigma}_v^{2(i)})$.

For the optimal parameter $\zeta^*$, the difference between significance values $\epsilon = s_u^{(i)} - s_v^{(i)}$, and the maximal difference $\Delta$, we expect as $\lim_{\epsilon \rightarrow \Delta} p = 1$, and $\lim_{\epsilon \rightarrow 0} p = 0.5$. This is the case since higher $\epsilon$ indicates higher perceivable difference in the instance $\vb*{x}^{(i)}$, thus the relation between $s_u^{(i)}$ and $s_v^{(i)}$ for the instance $\vb*{x}^{(i)}$ will be fairly obvious if $\epsilon$ is higher. Since $p \propto \hat{\mu}_u^{(i)} - \hat{\mu}_v^{(i)}$, for $\epsilon = s_u^{(i)} - s_v^{(i)}$, $\epsilon \propto \hat{\mu}_u^{(i)} - \hat{\mu}_v^{(i)}$ for optimal $\zeta^*$. Thus we can say the difference between our predicted means $\hat{\epsilon} = \hat{\mu}_u^{(i)} - \hat{\mu}_v^{(i)}$ is proportional to the real difference of underlying significance values $\epsilon = s_u^{(i)} - s_v^{(i)}$ for an instance $\vb*{x}^{(i)}$. Given the same soft constraint is applied to all pairs on our predicted vector, for a large enough dataset we claim that our predictions will be proportional to the real significance values, which we further supported with empirical studies in the paper Section \ref{55}, \ref{56} and \ref{57}.

\section{Ranked MNIST}
\label{sec:app-ranked-mnist}
\label{sec:apx5}

Ranked MNIST is a family of datasets with two main branches named as Ranked MNIST Gray and Ranked MNIST Color, where the first is in grayscale and the latter has varying random hue and saturation values for each digit. These datasets are generated by placing unique digits from the MNIST dataset \cite{deng2012mnist} on a 224x224 canvas, where the number of digits in a single image vary from 1 up to 10. Here we would like the further explain how we generate the datasets and how each of the 8 different datasets are defined.

For all the datasets we randomly pick the number of digits in the image from a discrete uniform distribution of numbers from 1 to 10, then we randomly place the digits on a 224x224 RGB canvas. For colored datasets we assign a random hue and saturation to color each digit.

Besides the coloring of the digits we have 3 different dataset setups to create the images: 1. Changing Scale, 2. Changing Brightness, 3. Changing Both. Yet we have 4 different datasets since for the scenario where we change both of the importance factors, the labels can be either for the changing scale or for the changing brightness, thus for each coloring we have 4 different in total 8 datasets, namely: Ranked MNIST Gray-S, Ranked MNIST Gray-B, Ranked MNIST Gray-S (Mix), Ranked MNIST Gray-B (Mix), Ranked MNIST Color-S, Ranked MNIST Color-B, Ranked MNIST Color-S (Mix), Ranked MNIST Color-B (Mix), where the letter in the end S and B stands for Scale and Brightness respectively.

For brightness change we use HSV color space and sample brightness values from uniform distribution $\mathcal{U}(0, 1)$ and for the scale change we sample coefficients from $\mathcal{U}(1, 3)$ then resize the digits such that the height and width of the digits are multiplied with the scale coefficients.

Each dataset consists of 60000 train, 10000 validation and 10000 test images. The test digits are sampled from the original MNIST test digits and the rest are sampled from the train digits.

\begin{figure}[H]
    \centering
    \includegraphics[width=1.0\linewidth]{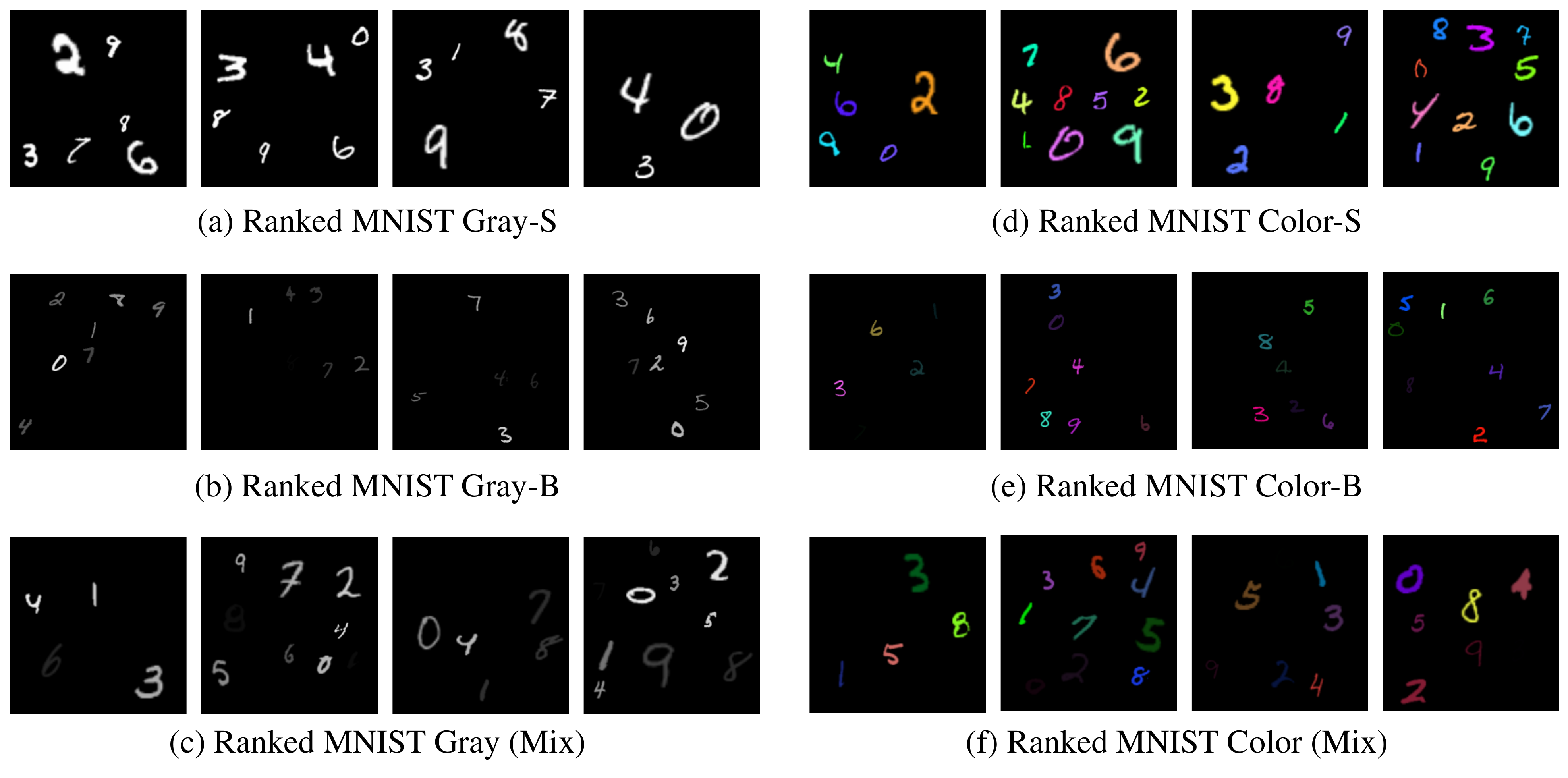}
    \caption{Sample images from the Ranked MNIST family.}
    \label{fig:ranked-mnist-family}
\end{figure}

The code to create the datasets will be provided in the supplementary material, and sample images can be seen in Figure \ref{fig:ranked-mnist-family}.

\section{Ranked MNIST Color Experiments}
\label{sec:apx6}

\begin{figure}[ht]
    \centering
    \begin{subfigure}[b]{1.0\linewidth}
        \hspace{0.10in}
        \includegraphics[width=0.48\linewidth]{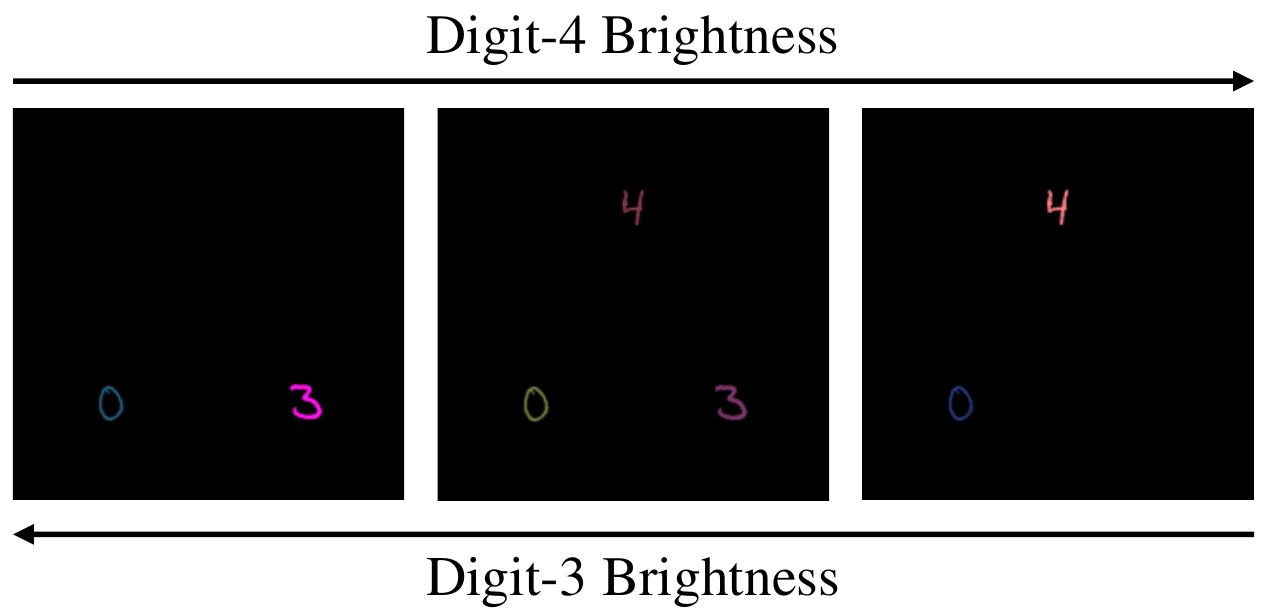}
        \includegraphics[width=0.48\linewidth]{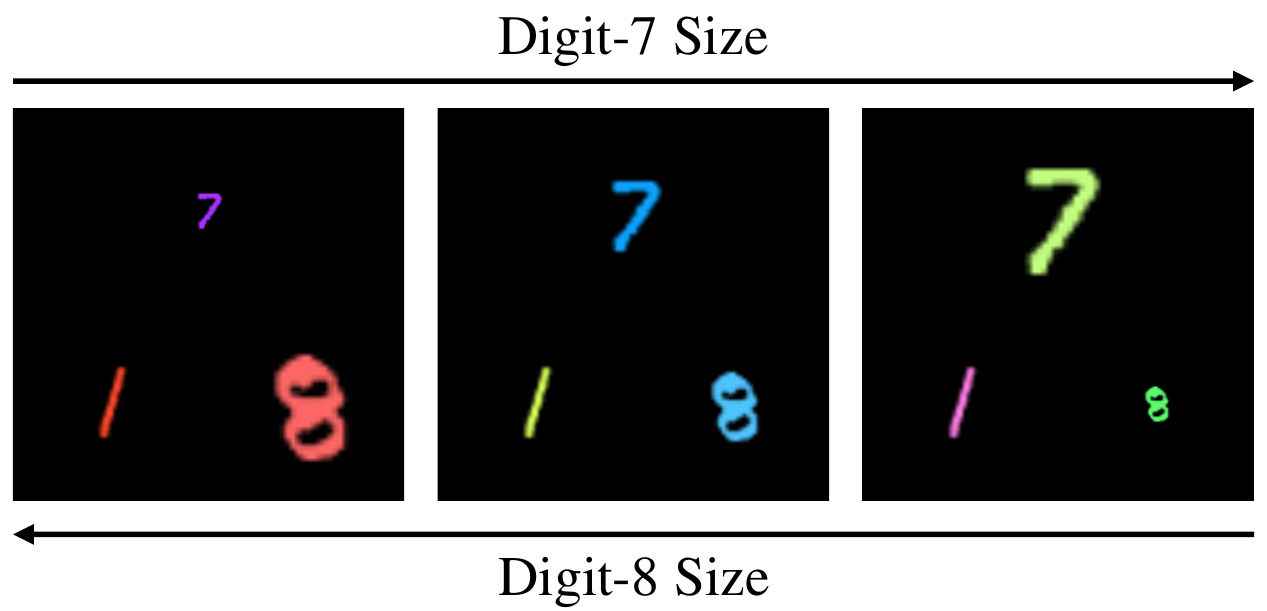}
    \label{fig:sequence1}
    \end{subfigure}
    \begin{subfigure}[b]{1.0\linewidth}
        \centering
        \setlength\tabcolsep{0.2pt}
        \begin{tabular}[b]{cccccc}
            \includegraphics[width=0.16\linewidth]{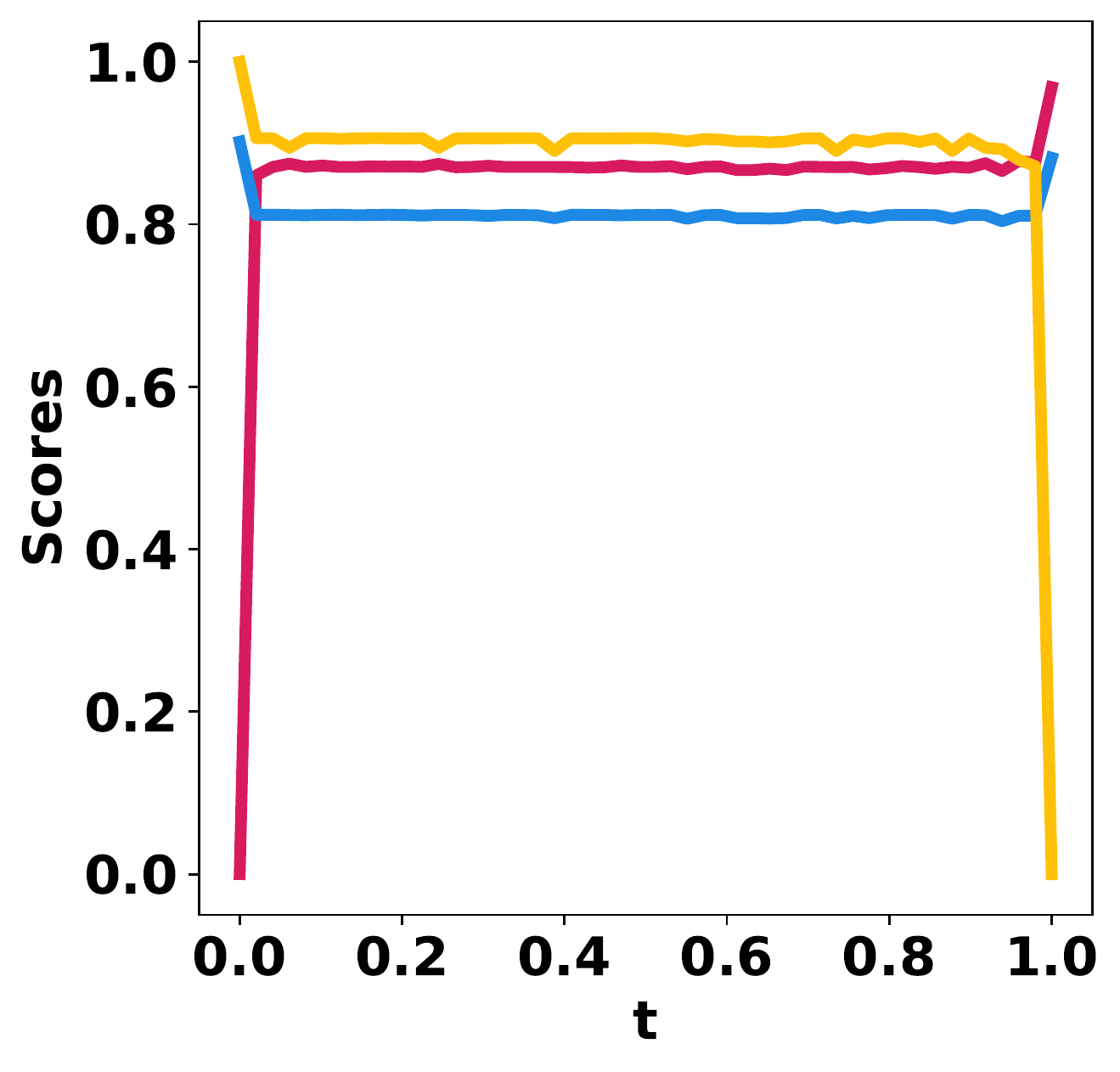} &
            \includegraphics[width=0.16\linewidth]{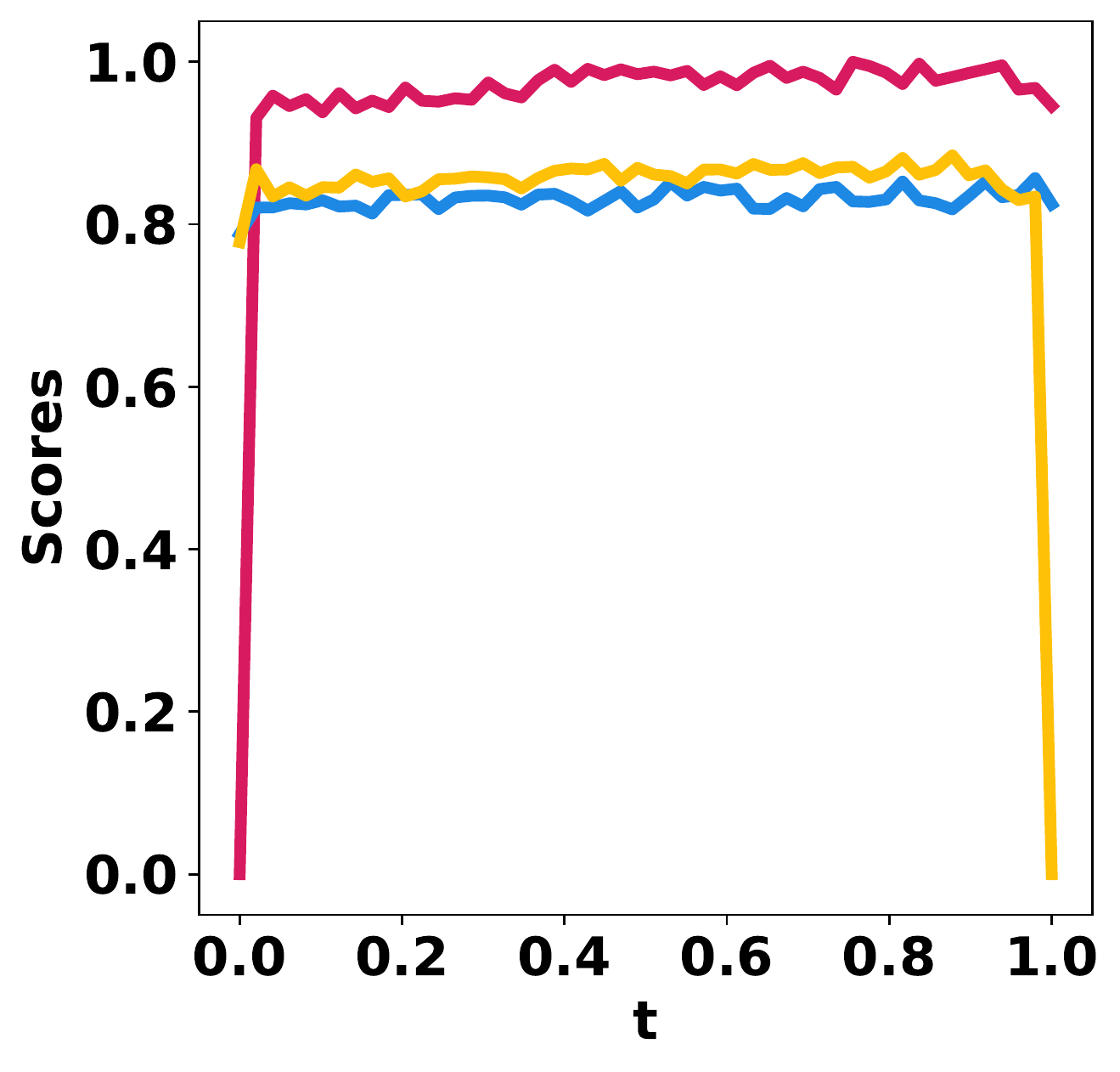} &
            \includegraphics[width=0.16\linewidth]{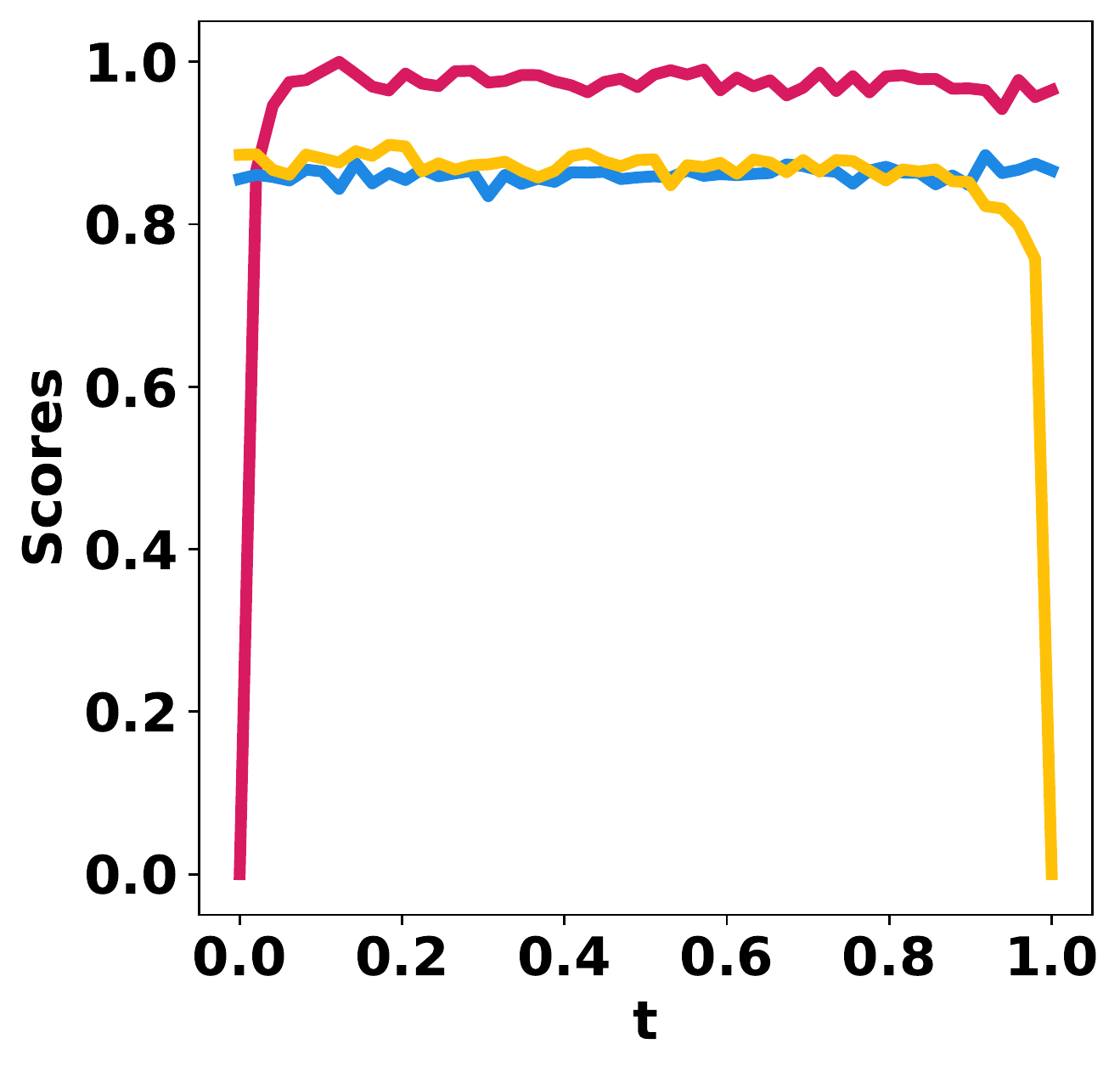} &
            \includegraphics[width=0.16\textwidth]{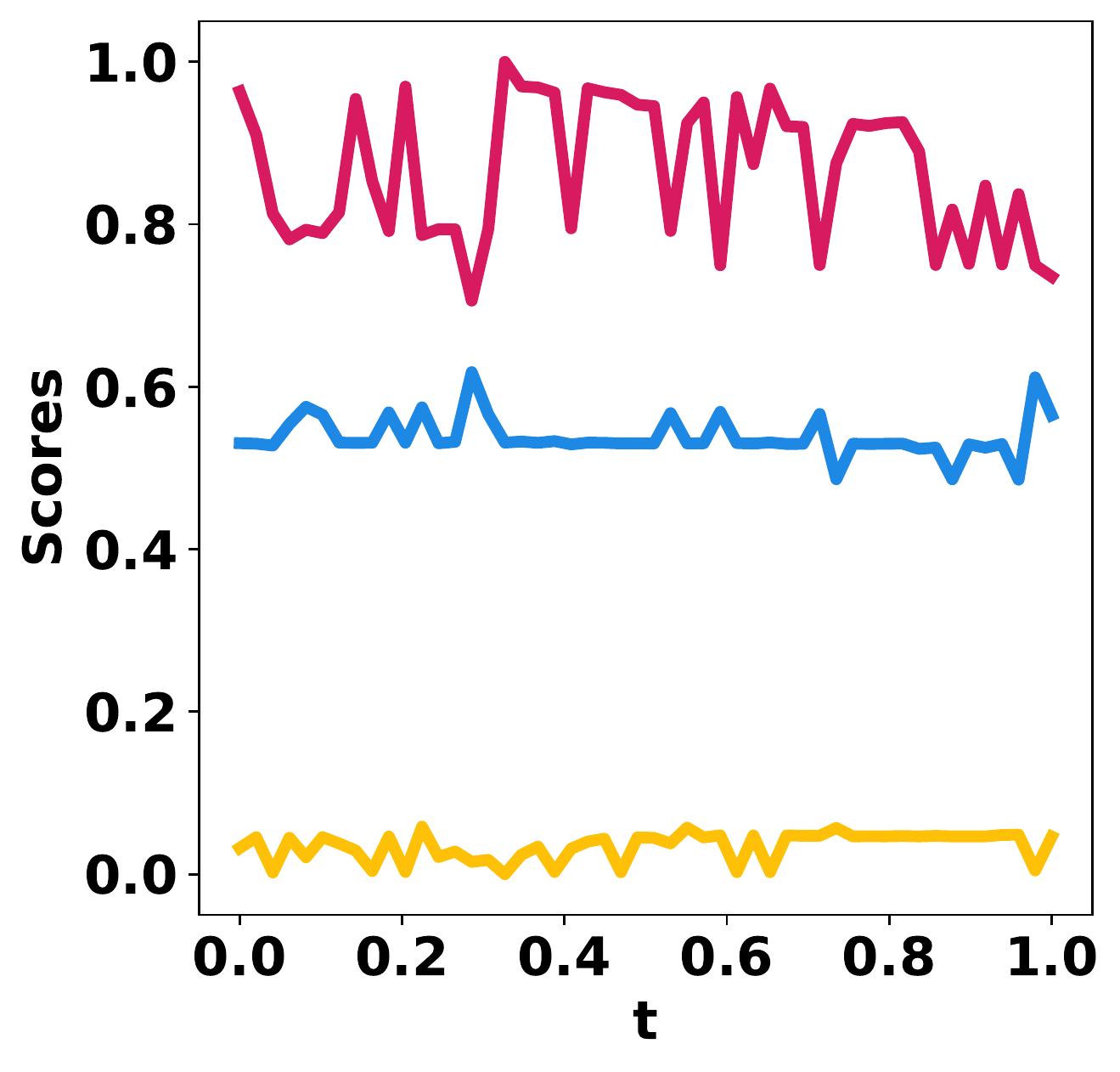} &
            \includegraphics[width=0.16\textwidth]{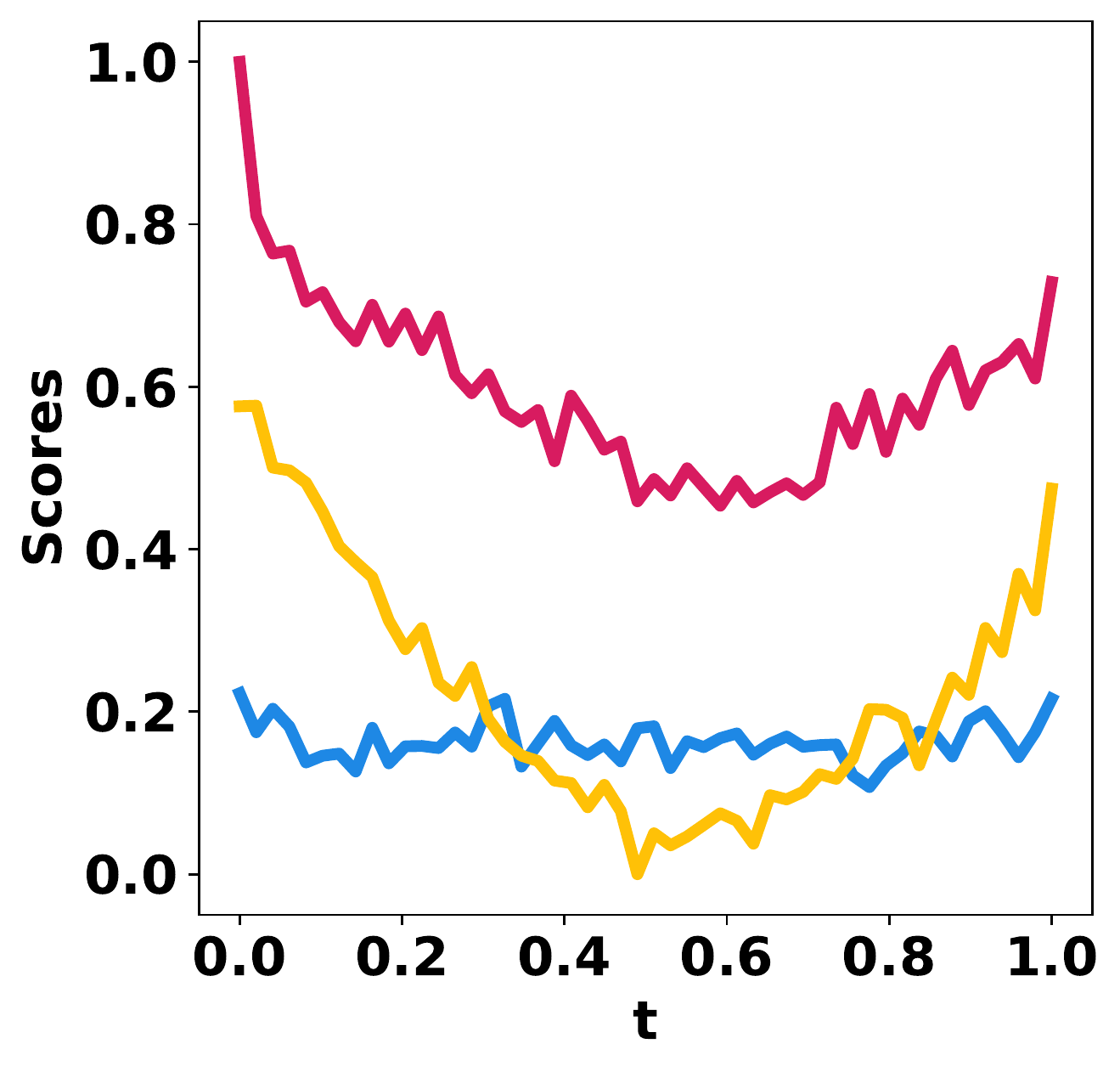} &
            \includegraphics[width=0.16\textwidth]{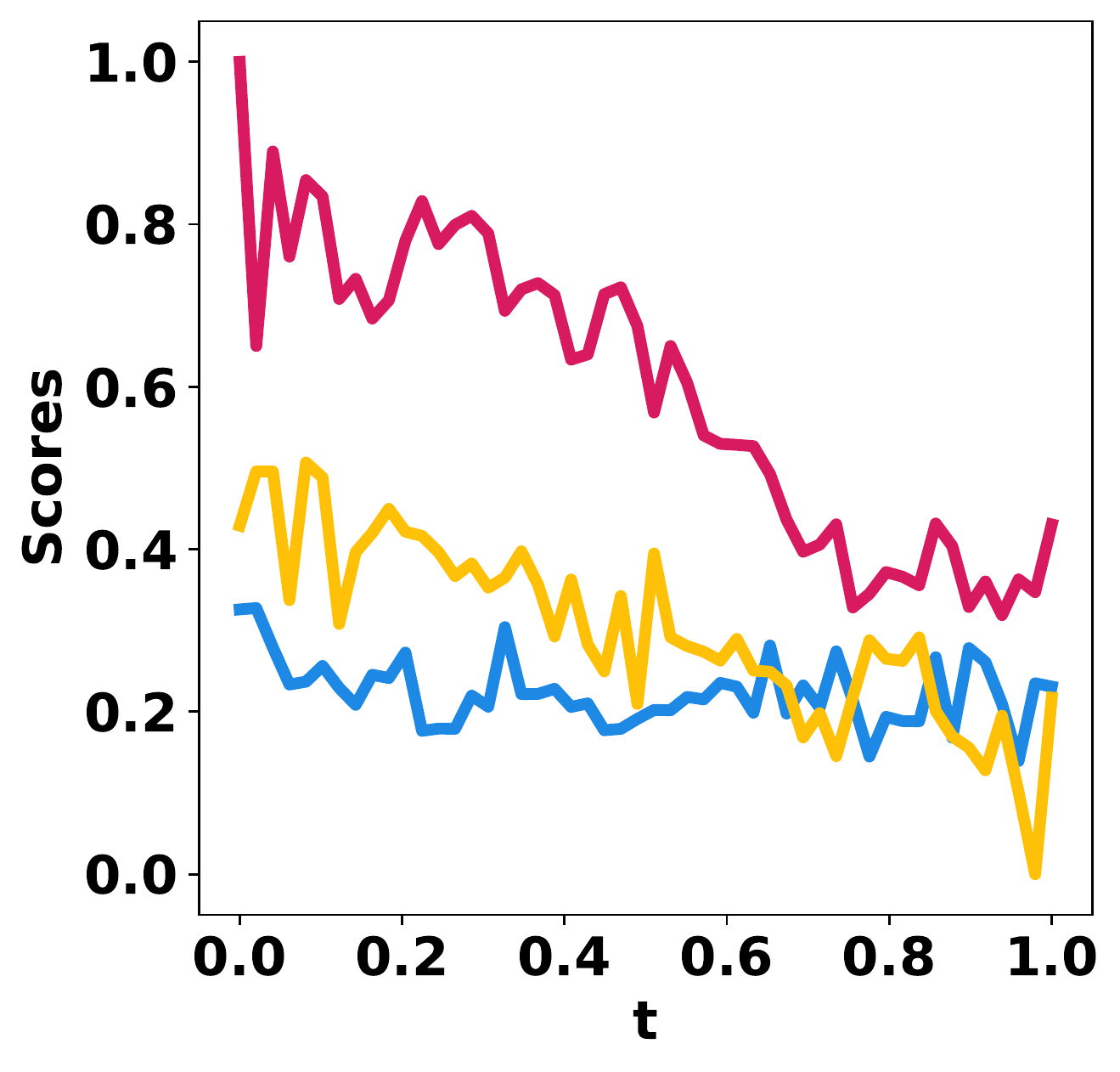} \\
            \hspace{0.12in} \small CRPC-Weak & \hspace{0.12in}  \small LSEP-Weak & \hspace{0.12in} \small GMLR-Weak &   \hspace{0.12in} \small CRPC-Weak & \hspace{0.12in} \small LSEP-Weak & \hspace{0.12in} \small GMLR-Weak
        \end{tabular}
    \end{subfigure}
    \begin{subfigure}[b]{1.0\linewidth}
        \centering
        \setlength\tabcolsep{0.2pt}
        \begin{tabular}[b]{cccccc}
            &&&&&\\
            \includegraphics[width=0.16\linewidth]{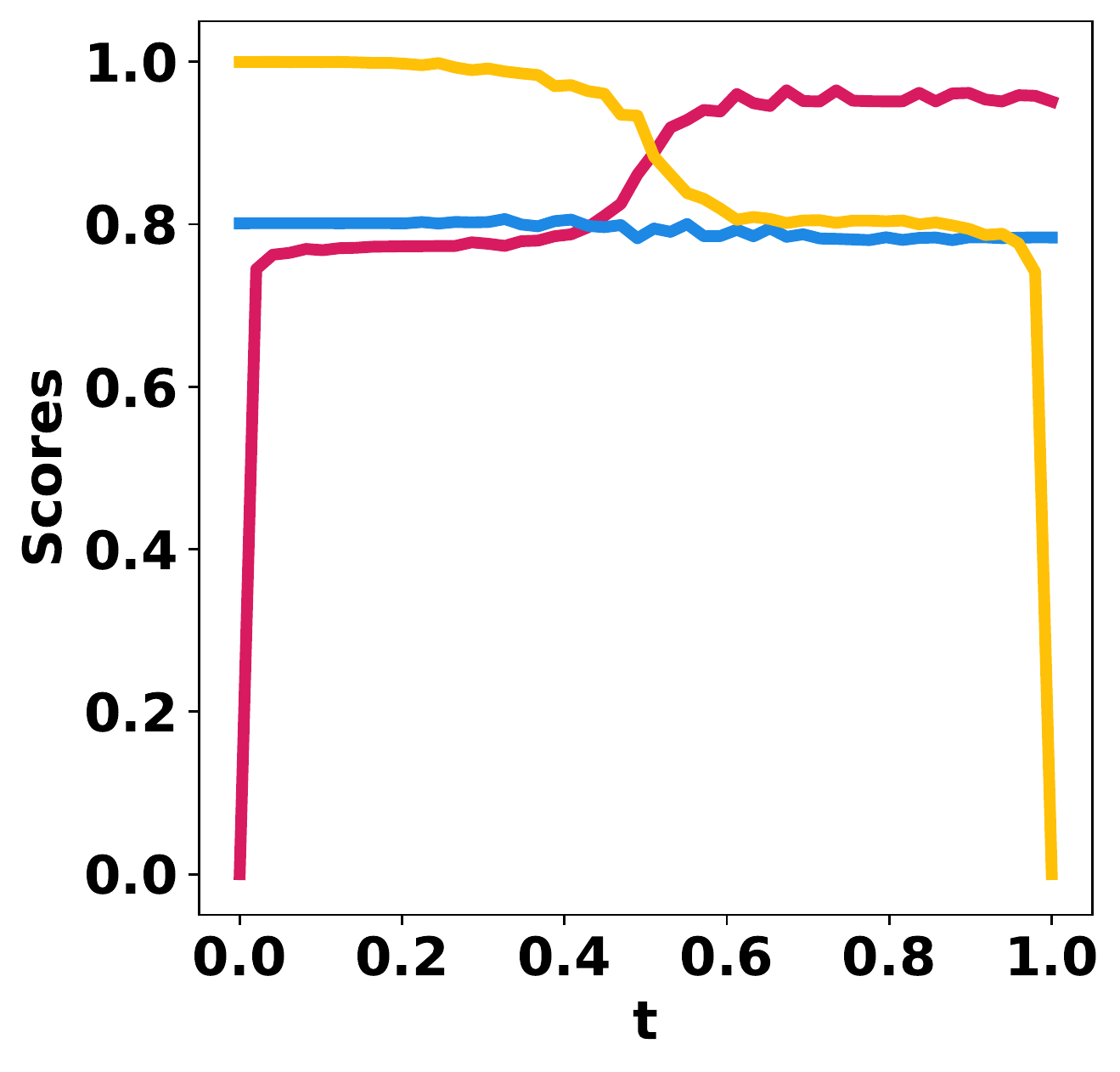} &
            \includegraphics[width=0.16\linewidth]{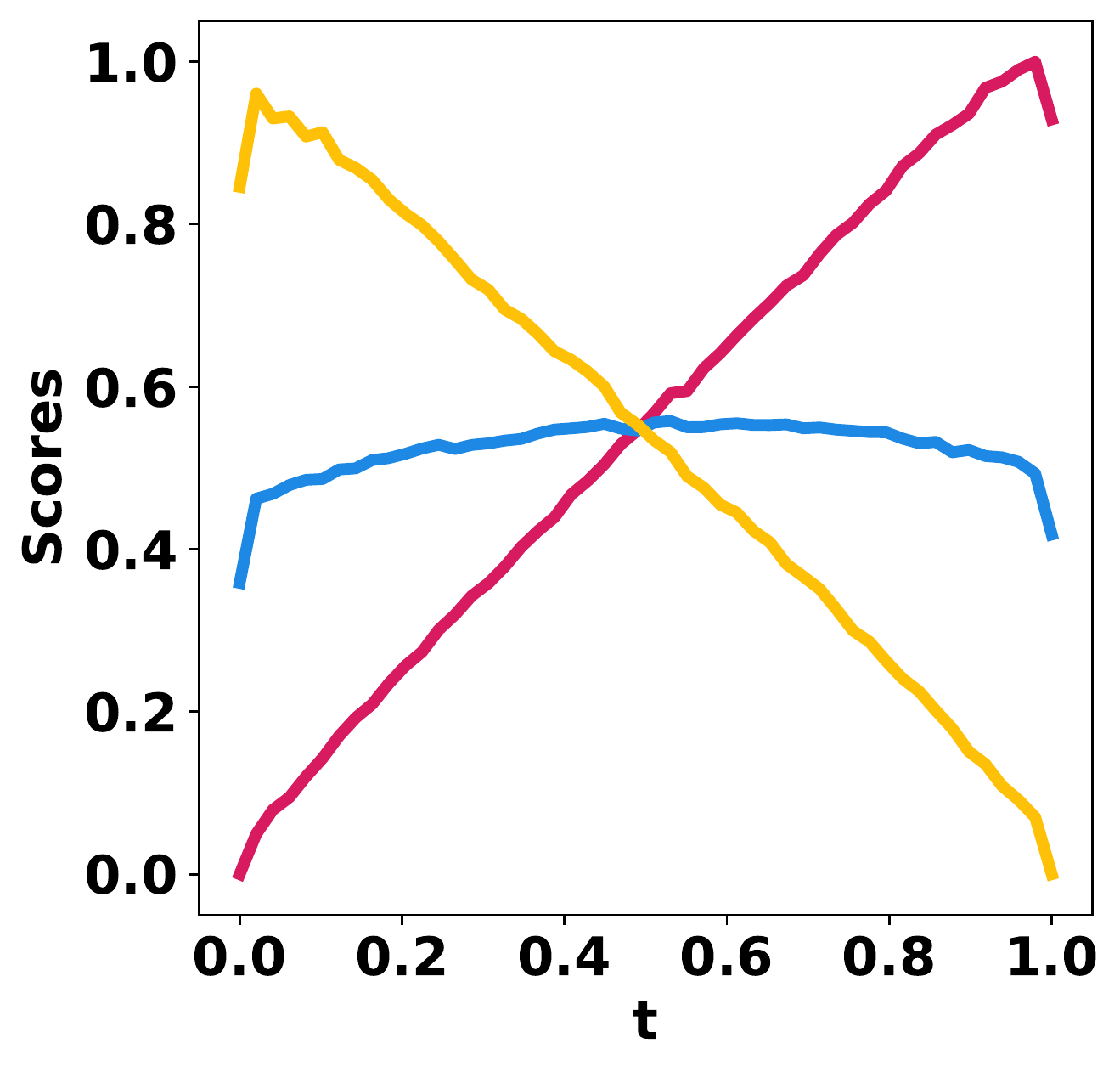} &
            \includegraphics[width=0.16\linewidth]{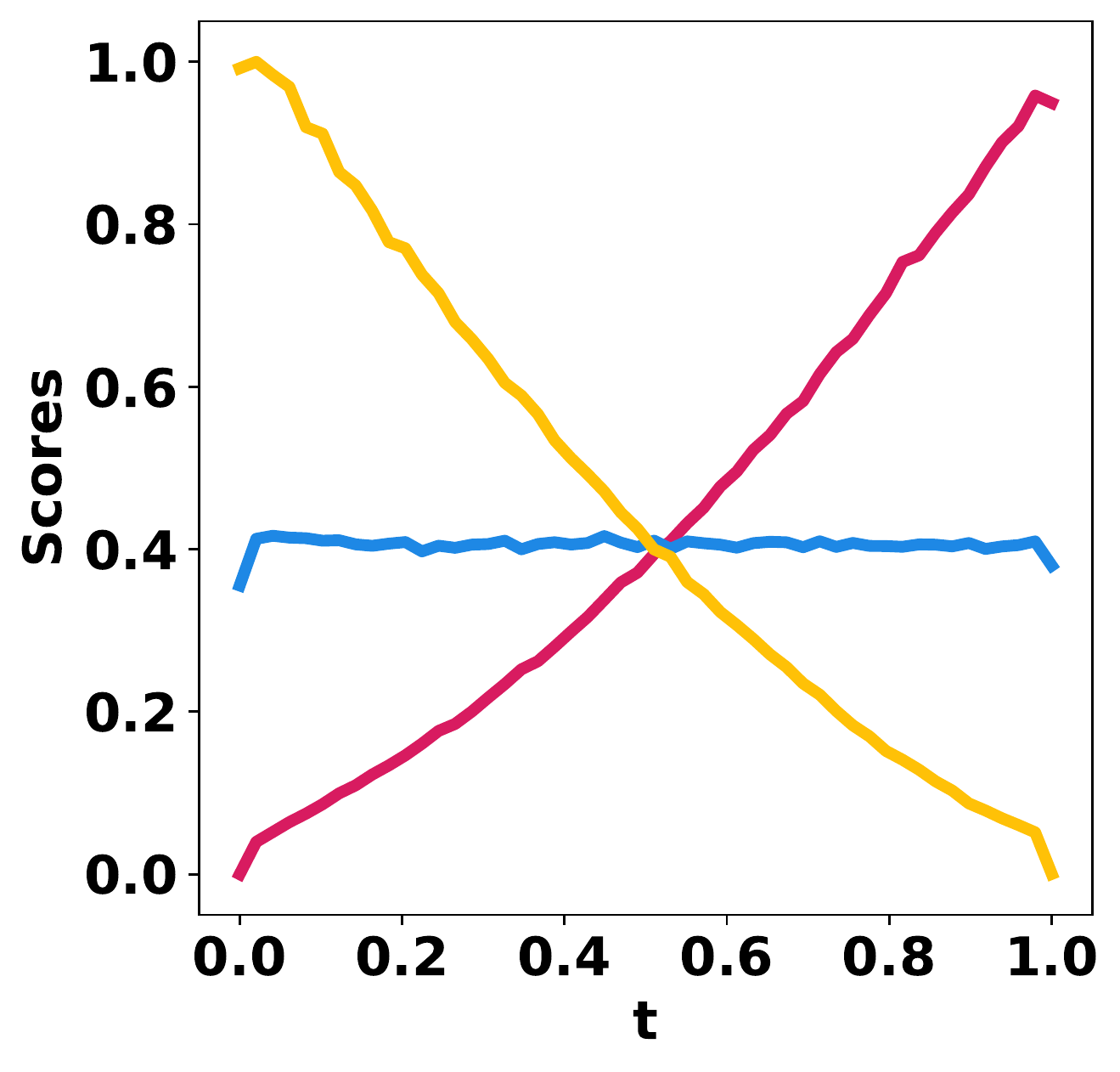} &
            \includegraphics[width=0.16\textwidth]{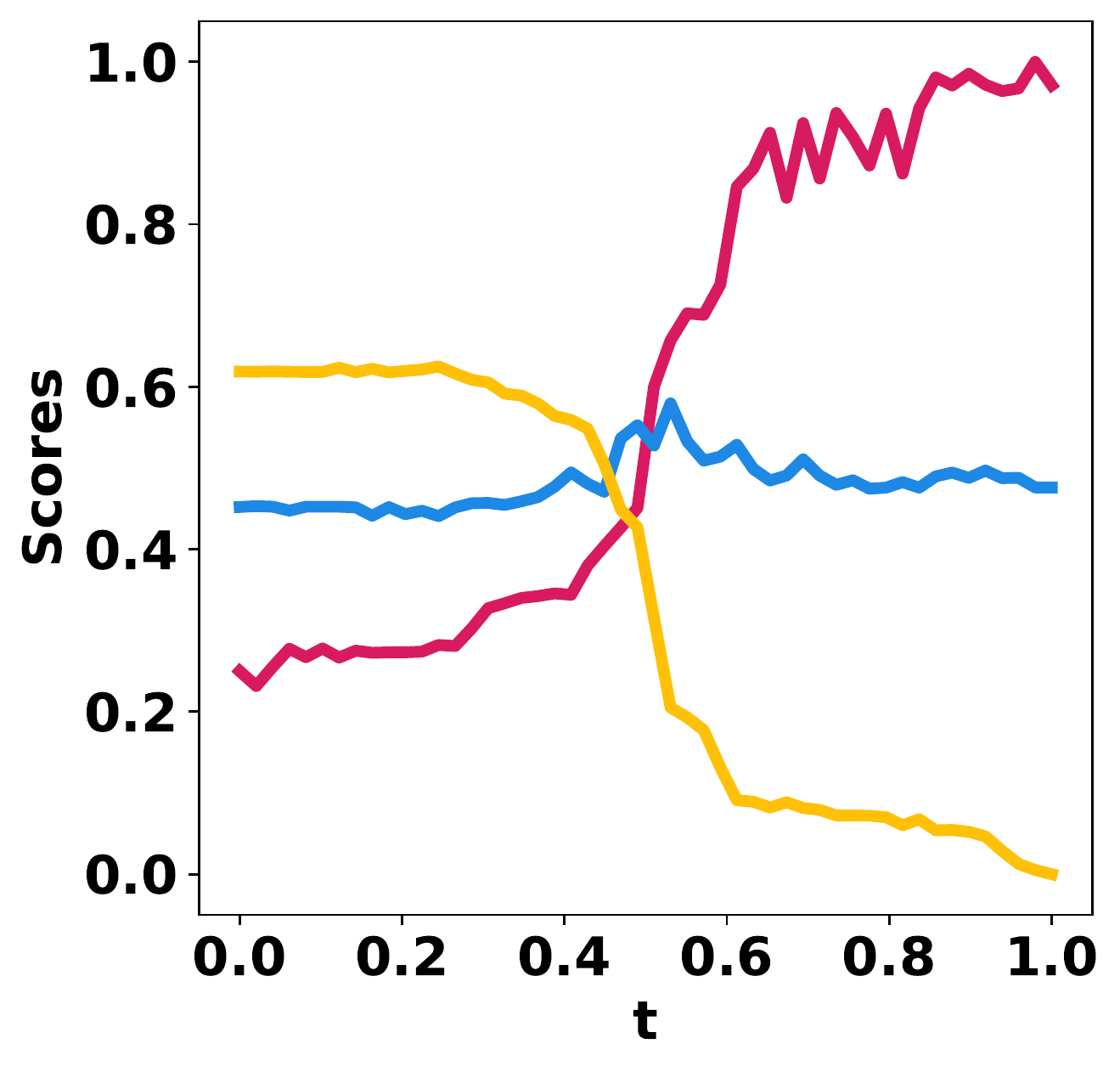} &
            \includegraphics[width=0.16\textwidth]{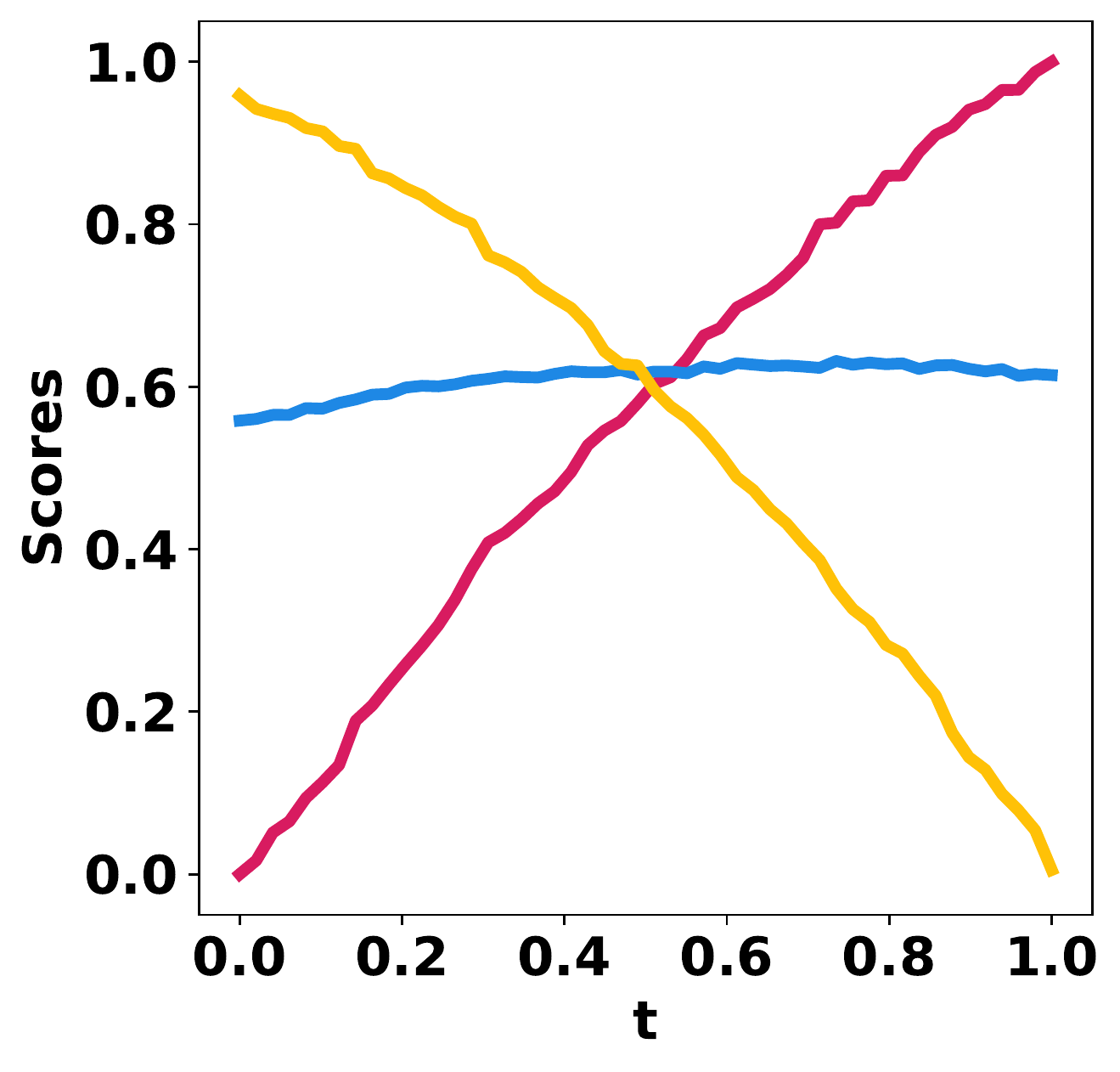} &
            \includegraphics[width=0.16\textwidth]{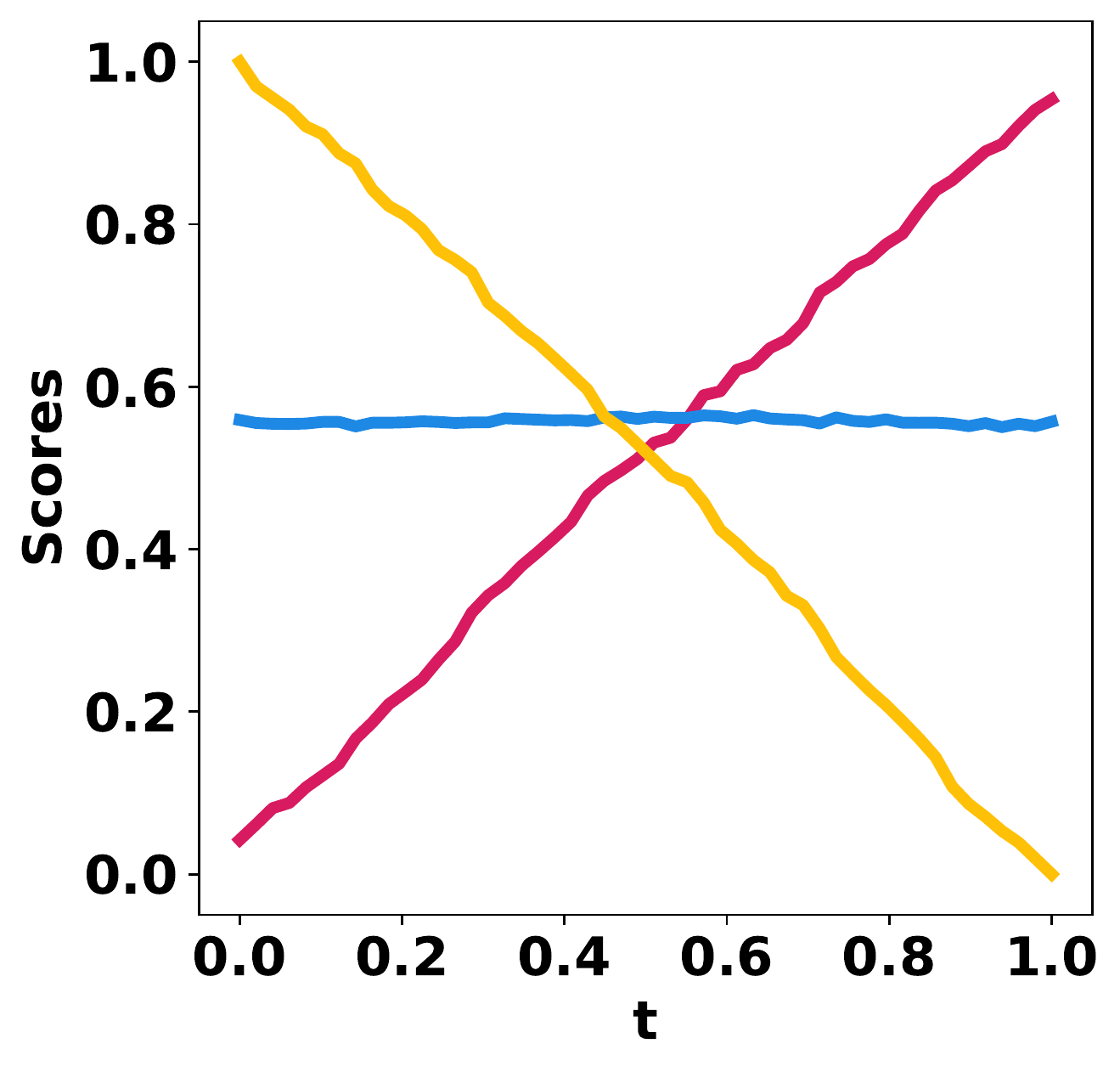} \\
             \hspace{0.12in} \small CRPC-Strong & \hspace{0.12in} \small LSEP-Strong & \hspace{0.12in} \small GMLR-Strong &   \hspace{0.12in} \small CRPC-Strong & \hspace{0.12in} \small LSEP-Strong & \hspace{0.12in} \small GMLR-Strong
        \end{tabular}
    \end{subfigure}
   
    \caption{Gradually changing significance effects in the sequences are shown at the top row, where the importance factor is the brightness of digits in top-left, Ranked MNIST Color-B, and size of digits in top-right, Ranked MNIST Color-S. Lines demonstrate changes in scores of $\langle$\textbf{\textcolor{1st}{1st}}, \textbf{\textcolor{2nd}{2nd}}, \textbf{\textcolor{3rd}{3rd}}$\rangle$ digits, which are in the order of $\langle$4, 0, 3$\rangle$ in top-left and $\langle$7, 1, 8$\rangle$ in top-right.}
    \label{fig:interpolation1}
\end{figure}

\begin{figure}[ht]
    \centering 
    \begin{subfigure}[b]{1.0\linewidth}
        \hspace{0.10in}
        \includegraphics[width=0.48\linewidth]{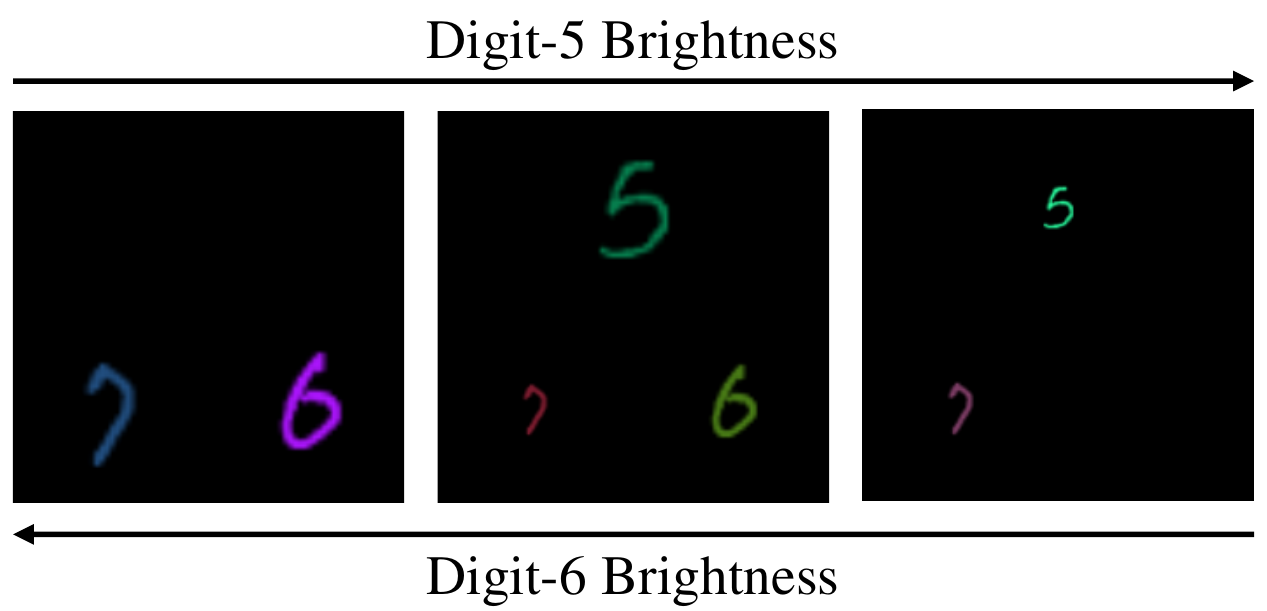}
        \includegraphics[width=0.48\linewidth]{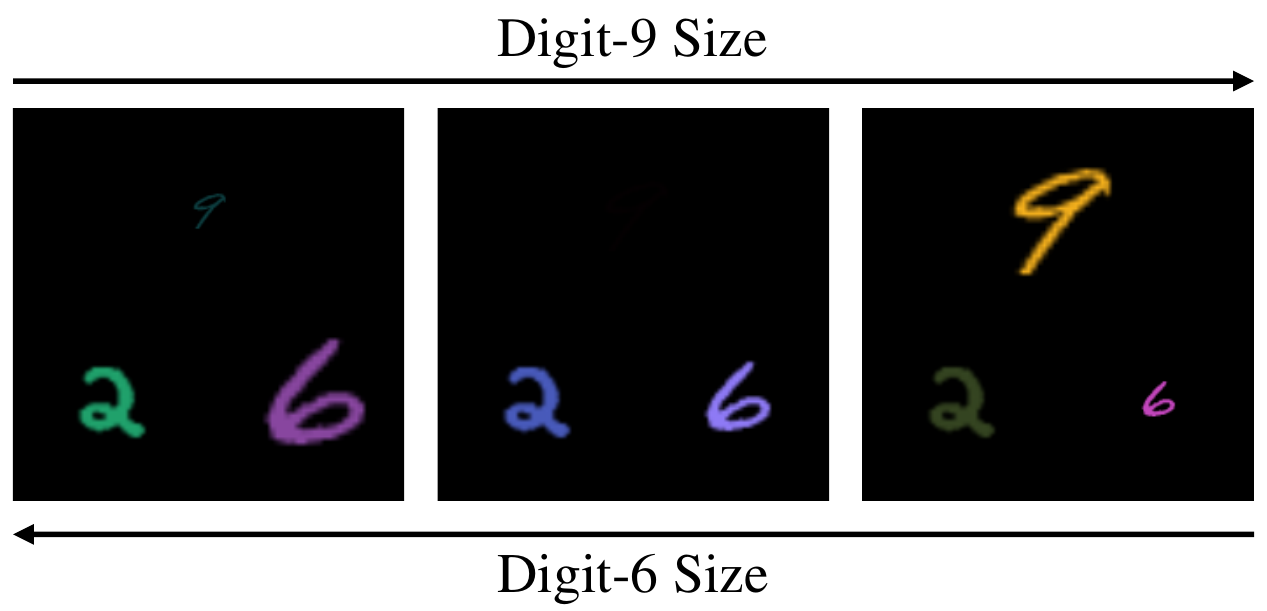}
    \label{fig:sequence2}
    \end{subfigure}
    \begin{subfigure}[b]{1.0\linewidth}
        \centering
        \setlength\tabcolsep{0.2pt}
        \begin{tabular}[b]{cccccc}
            \includegraphics[width=0.16\linewidth]{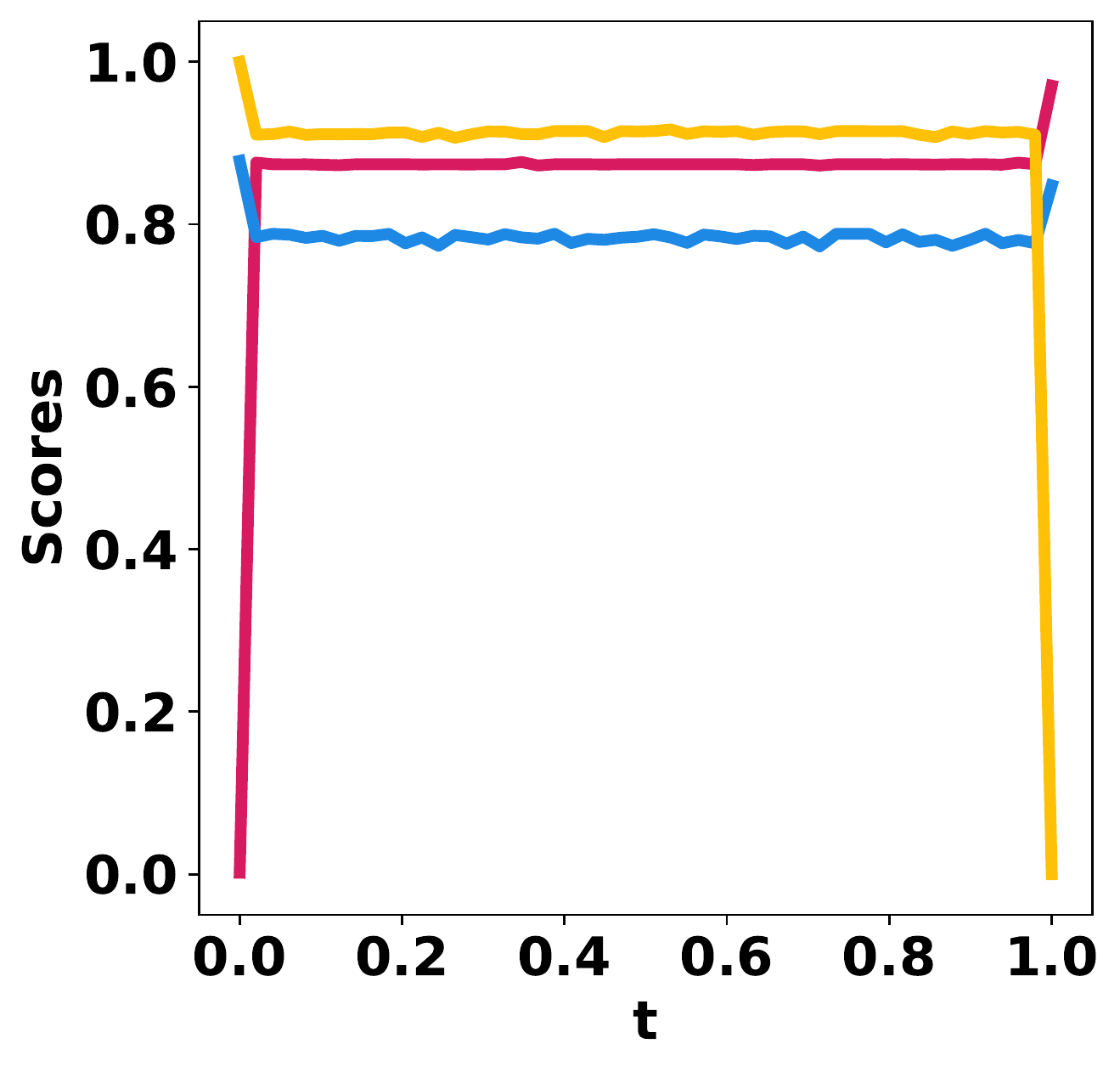} &
            \includegraphics[width=0.16\linewidth]{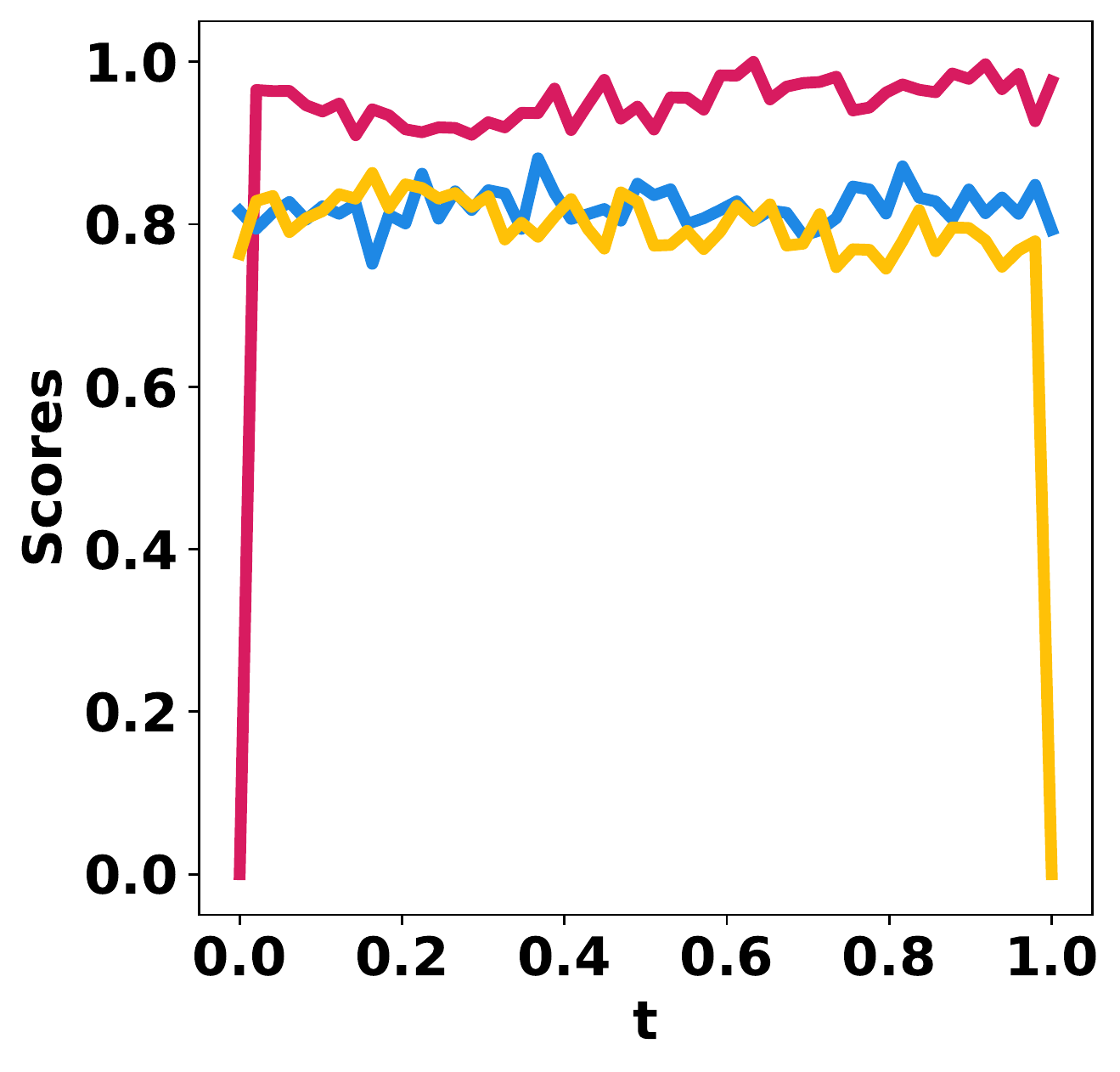} &
            \includegraphics[width=0.16\linewidth]{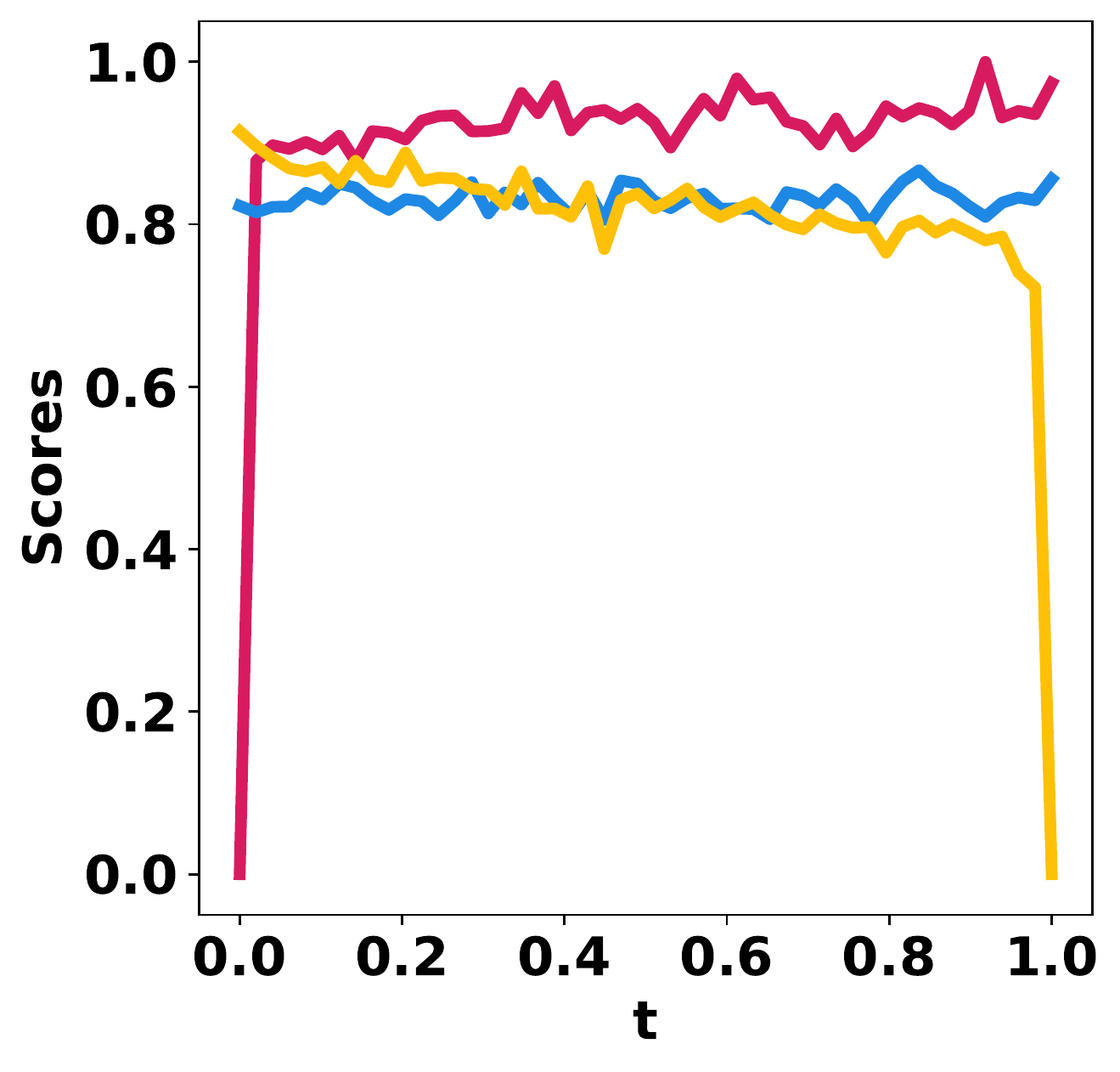} &
            \includegraphics[width=0.16\textwidth]{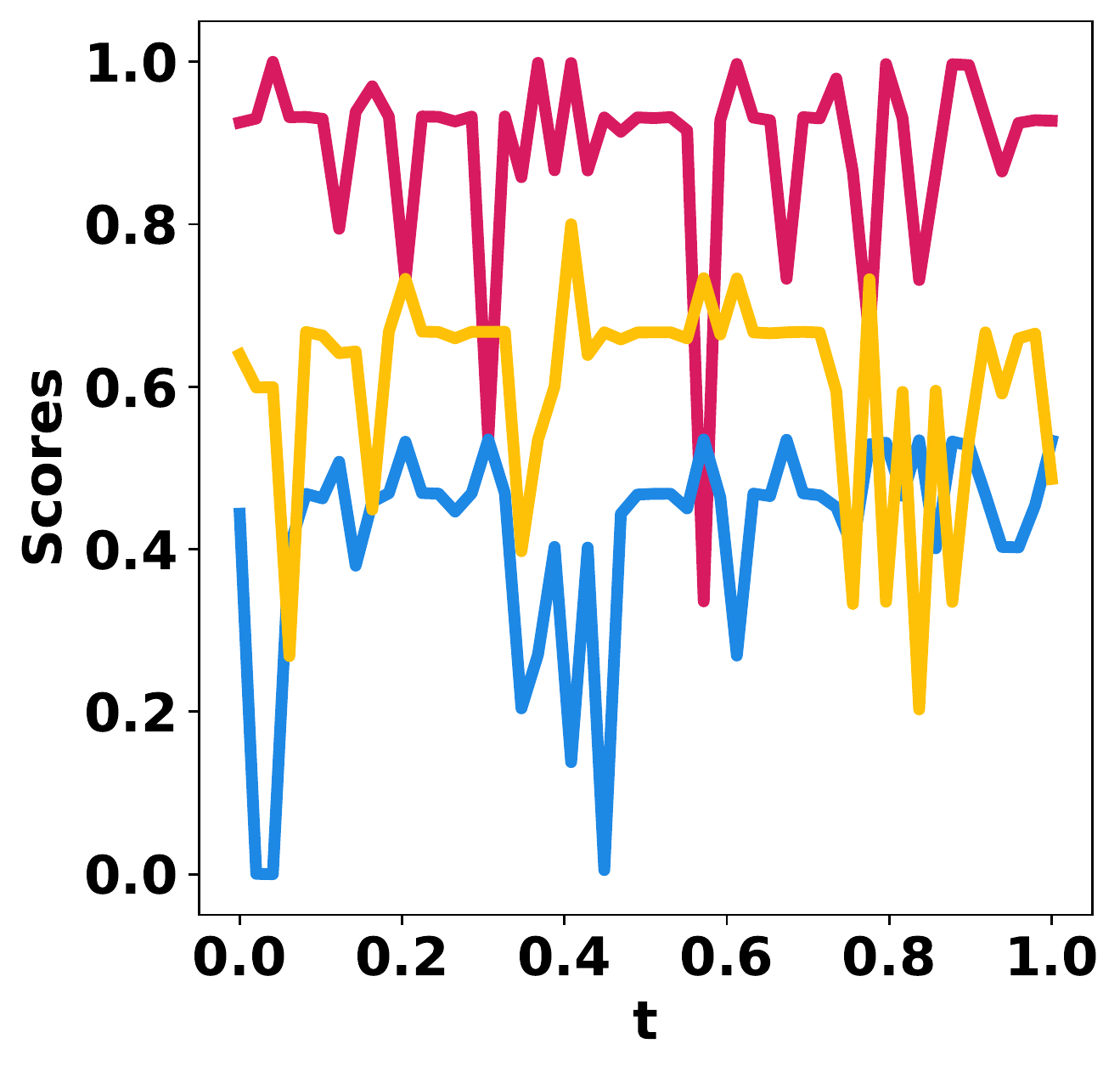} &
            \includegraphics[width=0.16\textwidth]{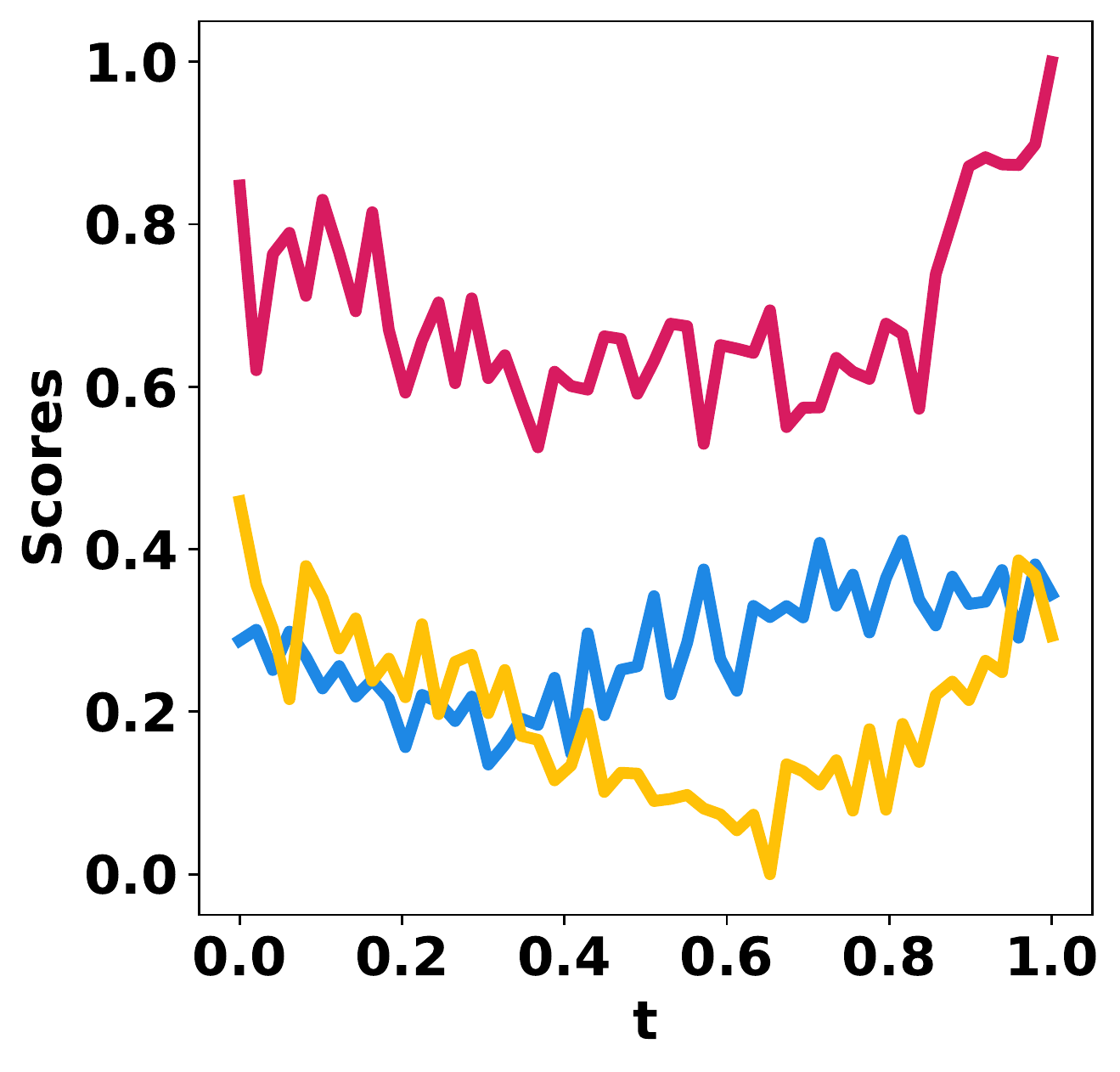} &
            \includegraphics[width=0.16\textwidth]{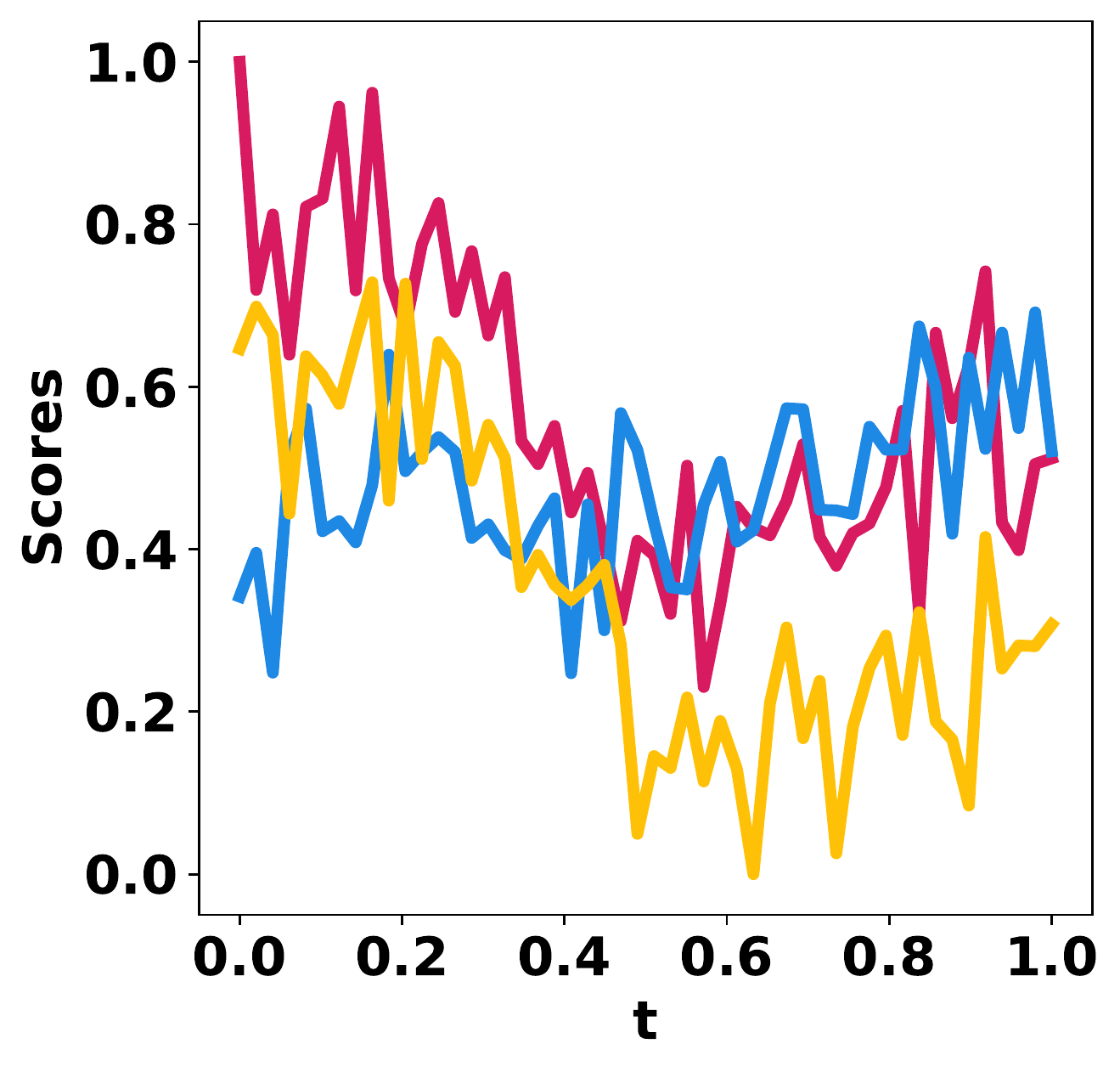} \\
            \hspace{0.12in} \small CRPC-Weak & \hspace{0.12in}  \small LSEP-Weak & \hspace{0.12in} \small GMLR-Weak &   \hspace{0.12in} \small CRPC-Weak & \hspace{0.12in} \small LSEP-Weak & \hspace{0.12in} \small GMLR-Weak
        \end{tabular}
    \end{subfigure}
    \begin{subfigure}[b]{1.0\linewidth}
        \centering
        \setlength\tabcolsep{0.2pt}
        \begin{tabular}[b]{cccccc}
            &&&&&\\
            \includegraphics[width=0.16\linewidth]{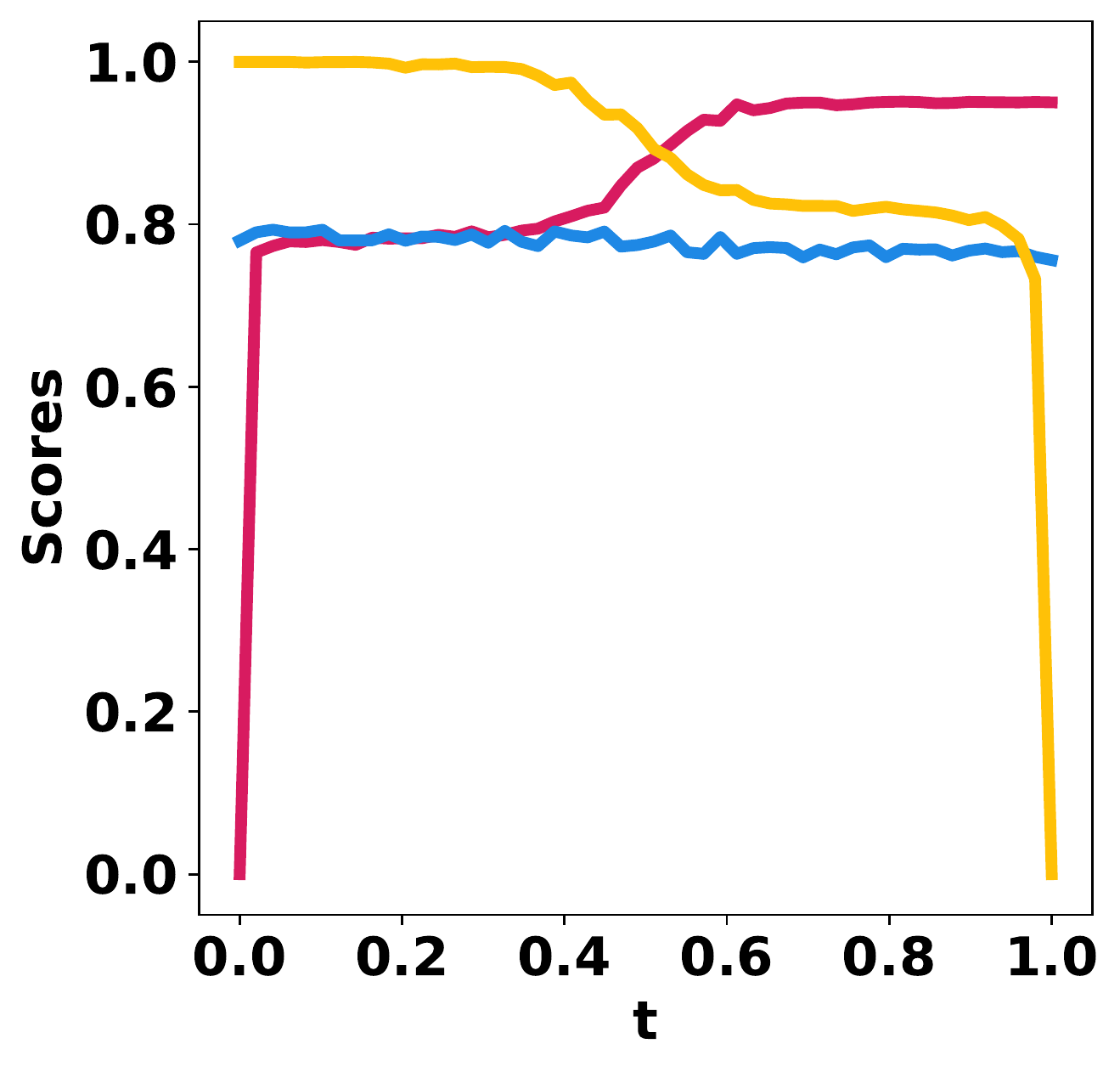} &
            \includegraphics[width=0.16\linewidth]{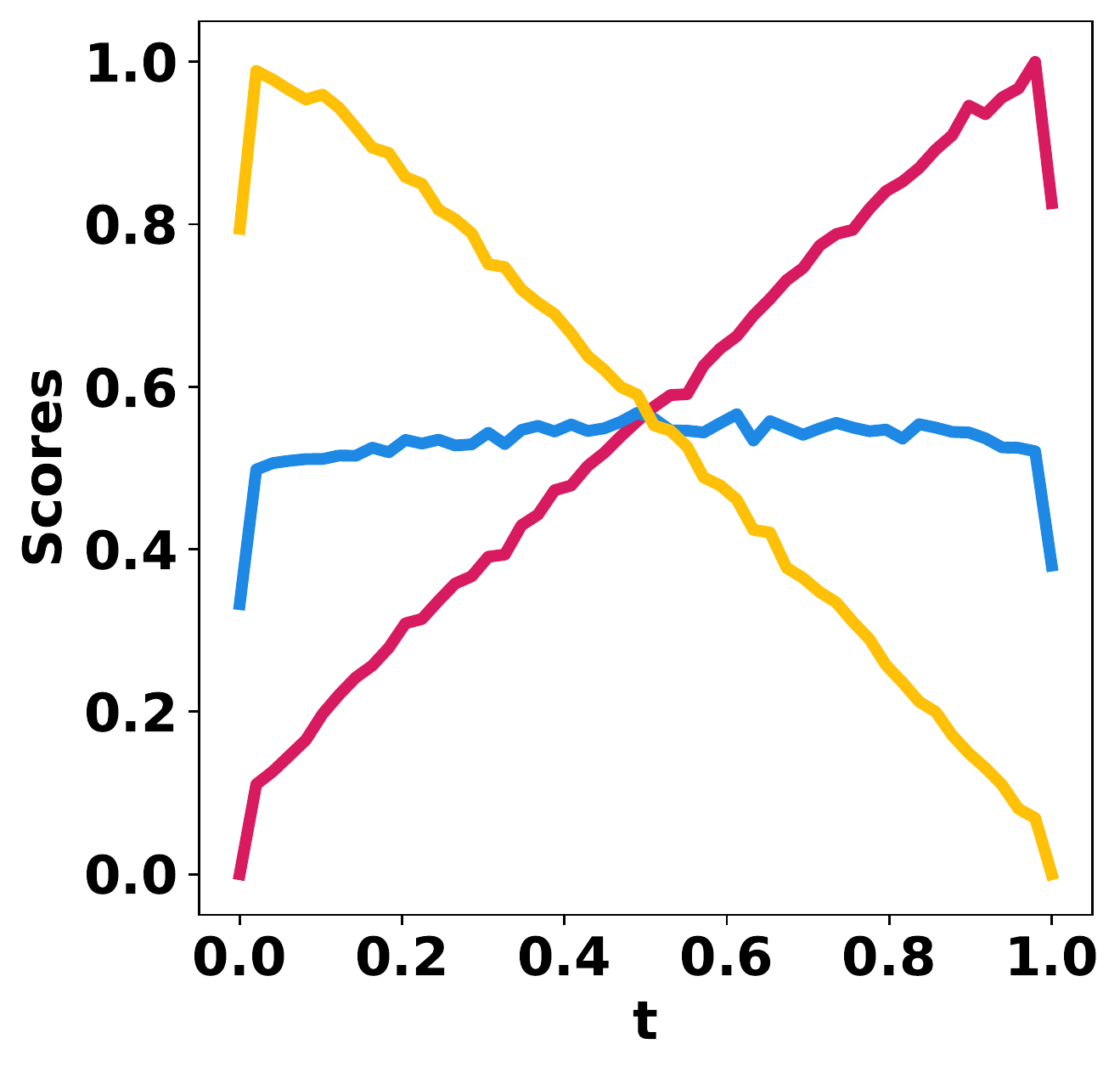} &
            \includegraphics[width=0.16\linewidth]{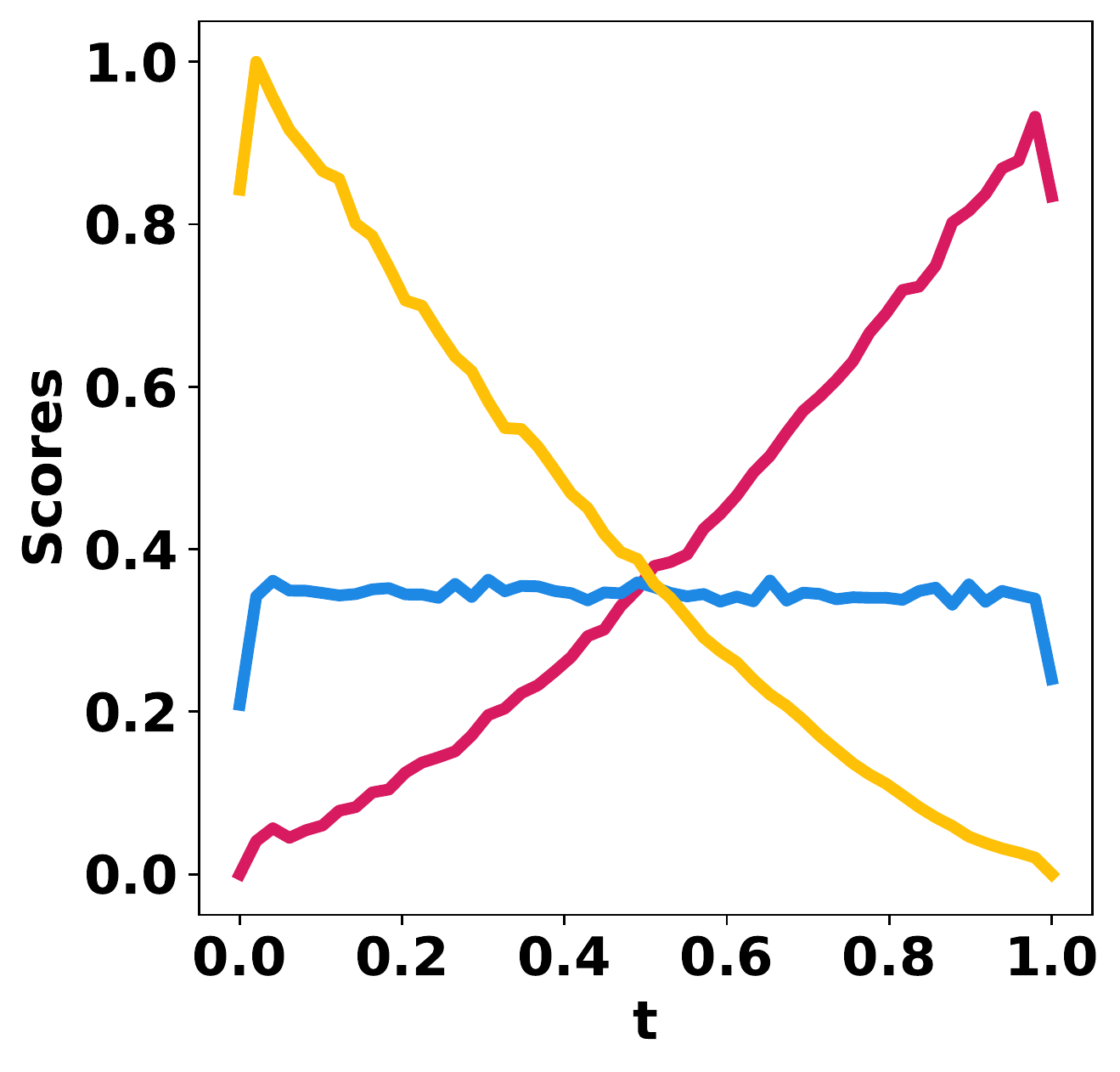} &
            \includegraphics[width=0.16\textwidth]{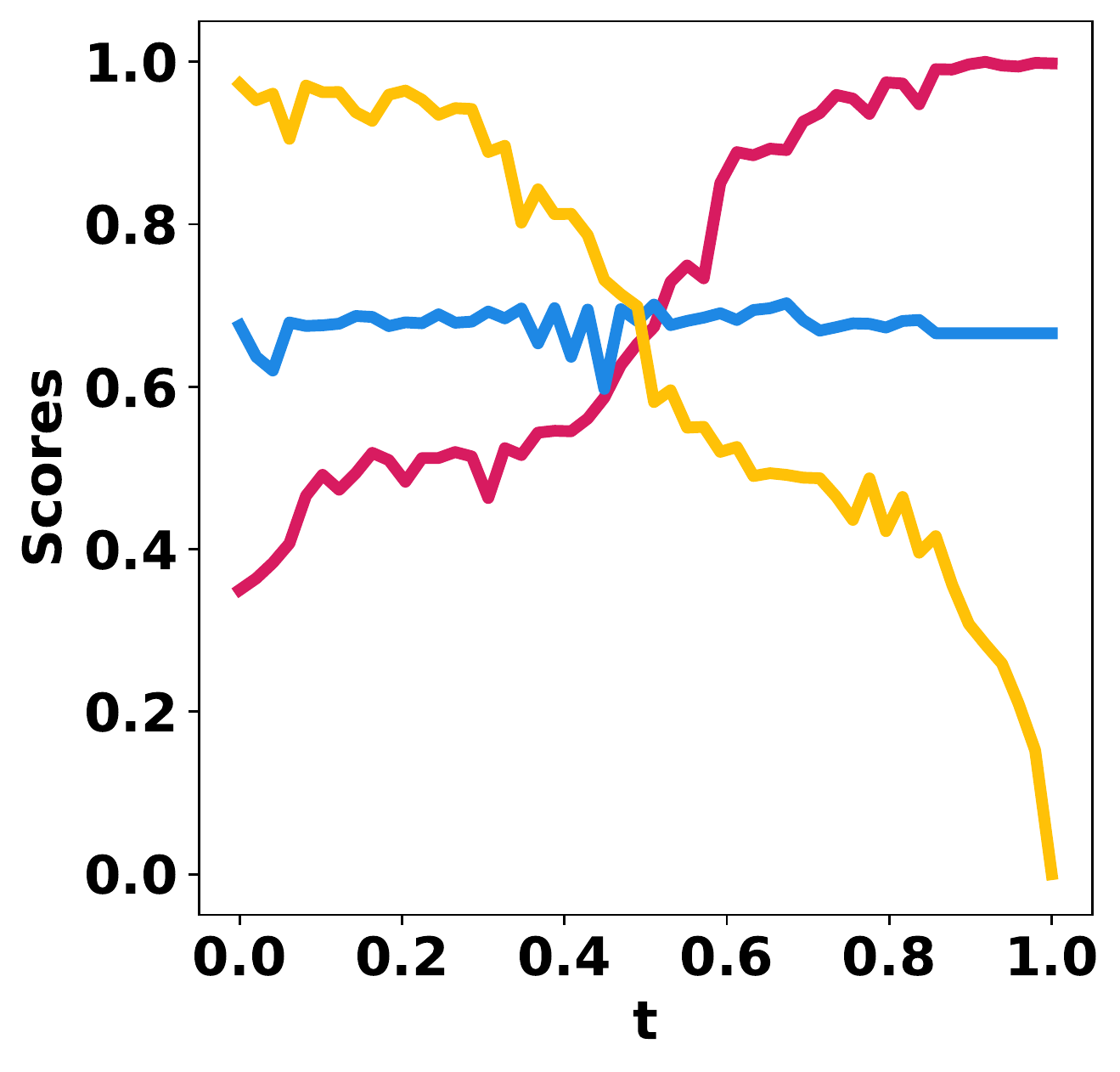} &
            \includegraphics[width=0.16\textwidth]{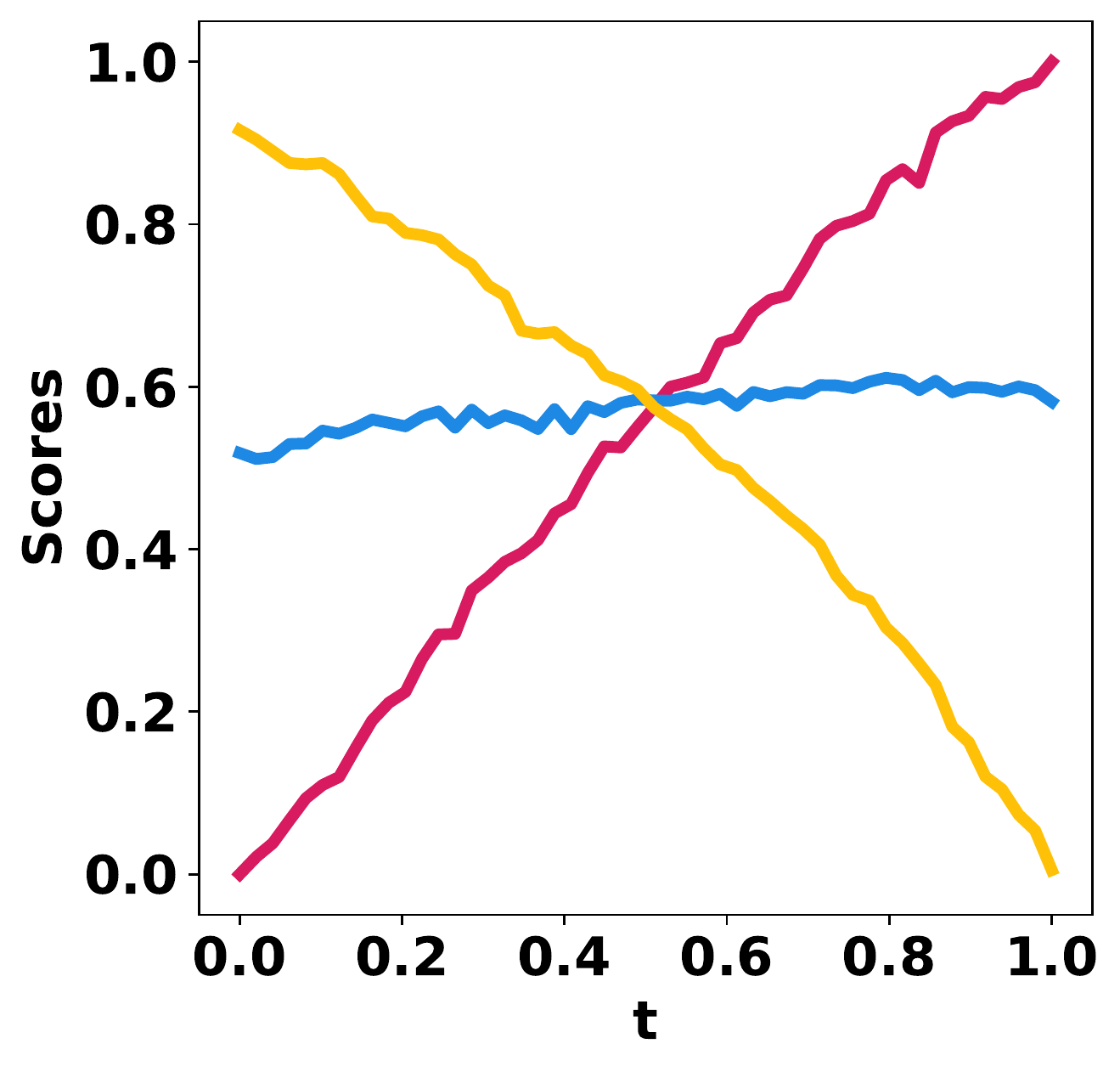} &
            \includegraphics[width=0.16\textwidth]{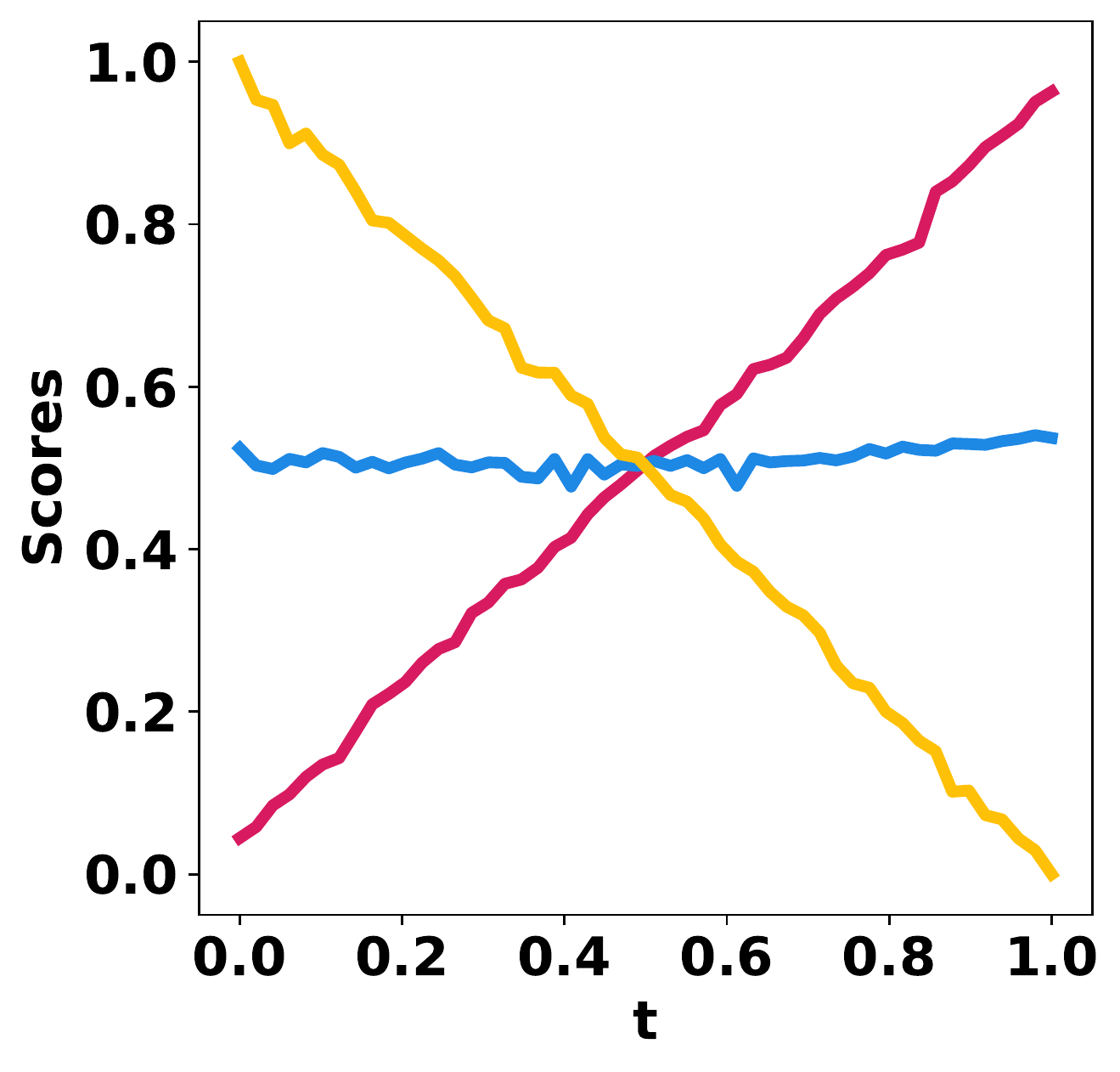} \\
             \hspace{0.12in} \small CRPC-Strong & \hspace{0.12in} \small LSEP-Strong & \hspace{0.12in} \small GMLR-Strong &   \hspace{0.12in} \small CRPC-Strong & \hspace{0.12in} \small LSEP-Strong & \hspace{0.12in} \small GMLR-Strong
        \end{tabular}
    \end{subfigure}
    \caption{Gradually changing significance effects in the sequences are shown at the top row, where the importance factor is the brightness of digits in top-left, Ranked MNIST Color-B (Mix), and size of digits in top-right, Ranked MNIST Color-S (Mix). Size of the digits for Ranked MNIST Color-B (Mix) and brightness of the digits for Ranked MNIST Color-S (Mix) change randomly as explained in Section \ref{sec:app-ranked-mnist}. Lines demonstrate changes in scores of $\langle$\textbf{\textcolor{1st}{1st}}, \textbf{\textcolor{2nd}{2nd}}, \textbf{\textcolor{3rd}{3rd}}$\rangle$ digits, which are in the order of $\langle$5, 7, 6$\rangle$ in top-left and $\langle$9, 2, 6$\rangle$ in top-right.}
    \label{fig:interpolation2}
\end{figure}

The Adjusting Significance Effects Experiment explained in Section 5.5 of the main paper is performed on the Ranked MNIST Color dataset, for all four different setups. Corresponding results on Ranked MNIST Color-S and Color-B are given in Figure \ref{fig:interpolation1},and for the Mix datasets Color-S (Mix) and Color-B (Mix), results are given in Figure \ref{fig:interpolation2}. Quantitative results are provided in Table \ref{tab:ranked_mnist_color}. It can be observed that developing the baselines into \enquote{\textit{stronger}} versions that enhance the models' ability to adjust to the changing importance factors when compared to the \enquote{\textit{weaker}} versions which are the original versions that do not benefit from positive class relations, in terms of the predicted scores. When compared with the Strong versions, our method GMLR performs better in predicting consistent scores both for the increasing or decreasing gradual changes and for the digits with unchanged importance factors, while LSEP produces inconstant scores for the latter. Another interpretation is that the predictions of GMLR are rather independent of the correlation between the digits and it evaluates each digit separately, while LSEP bases the predictions on the relation between digits, causing the convex score line for the 2nd digit in the experiments. These results are in compliance with the experiments of the same setup on Ranked MNIST Gray on the main paper.

\begin{figure}[H]
    \centering
    \begin{subfigure}[b]{1.0\linewidth}
        \hspace{0.10in}
        \includegraphics[width=0.48\linewidth]{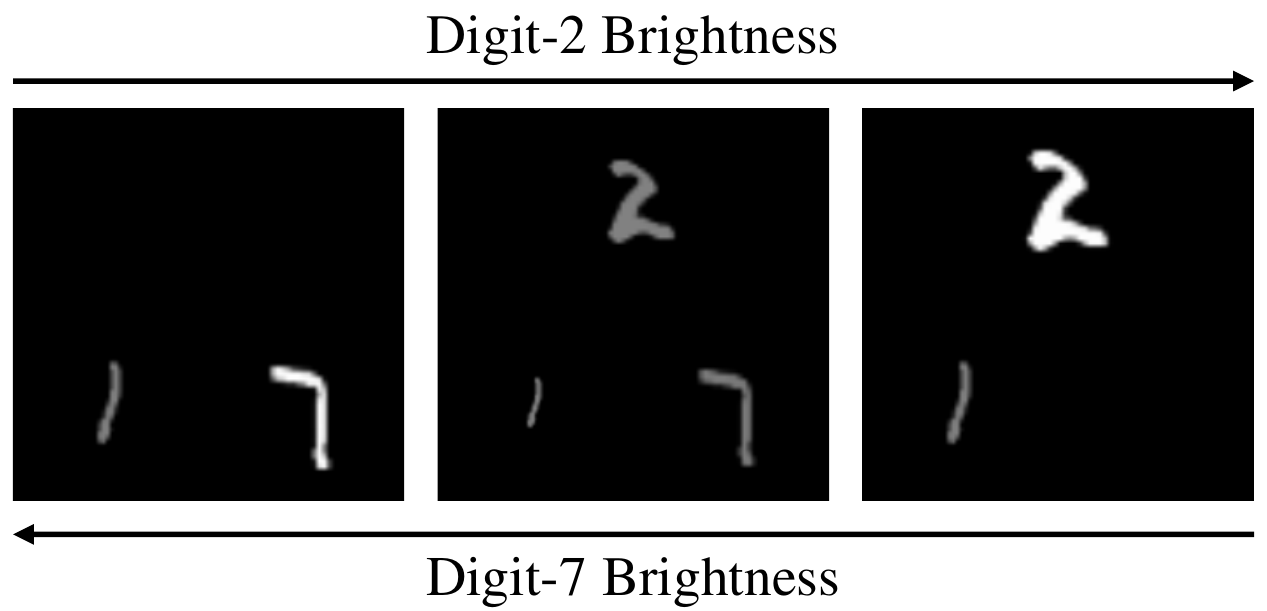}
        \includegraphics[width=0.48\linewidth]{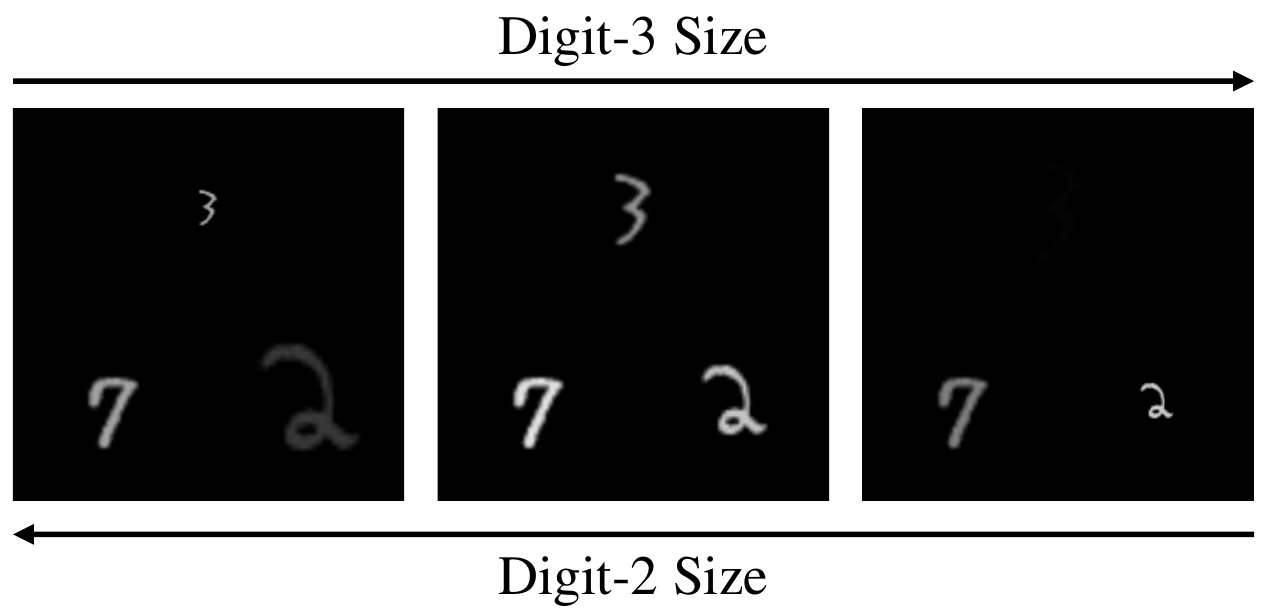}
    \label{fig:sequence3}
    \end{subfigure}
    \begin{subfigure}[b]{1.0\linewidth}
        \centering
        \setlength\tabcolsep{0.2pt}
        \begin{tabular}[b]{cccccc}
            \includegraphics[width=0.16\linewidth]{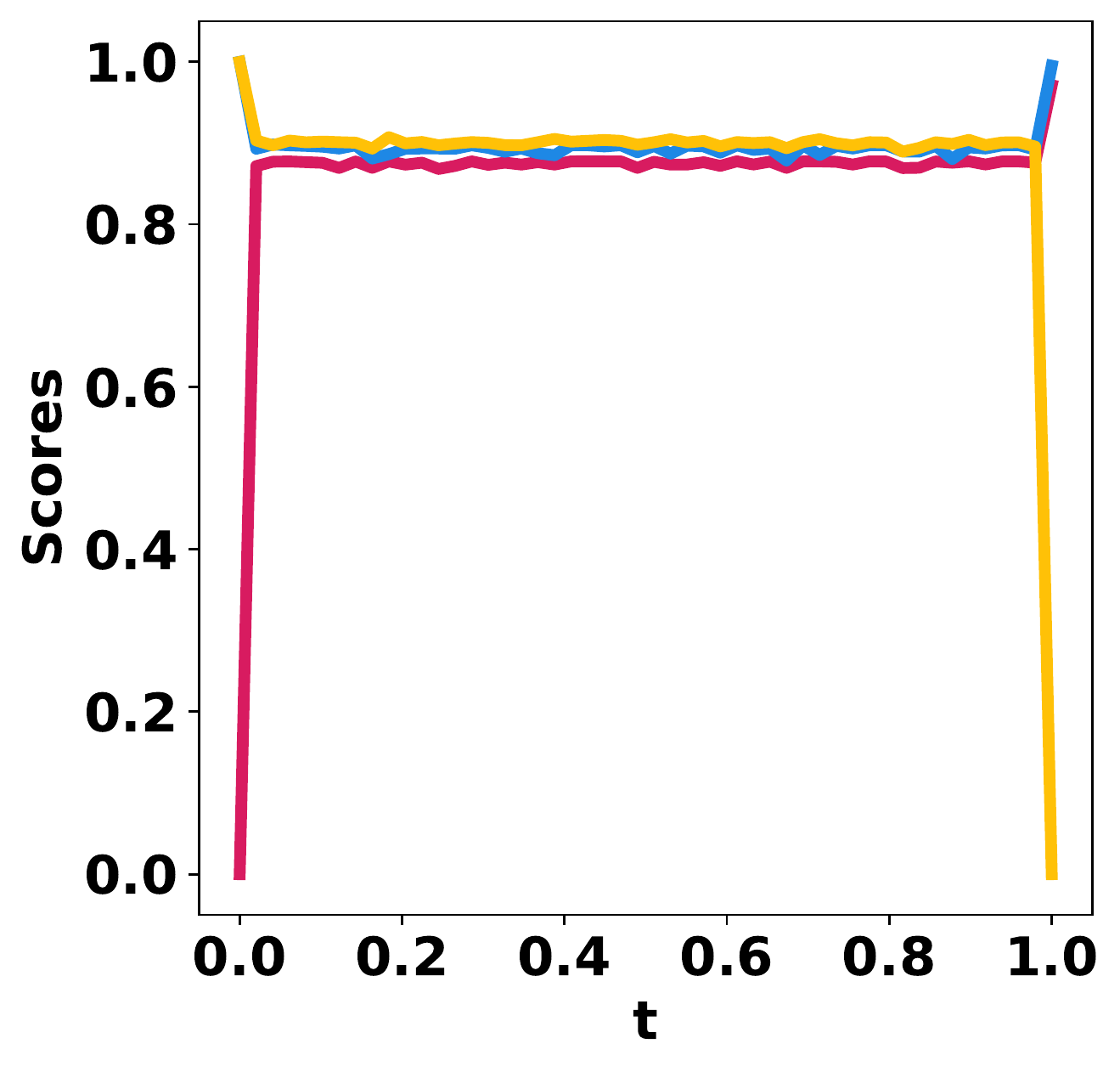} &
            \includegraphics[width=0.16\linewidth]{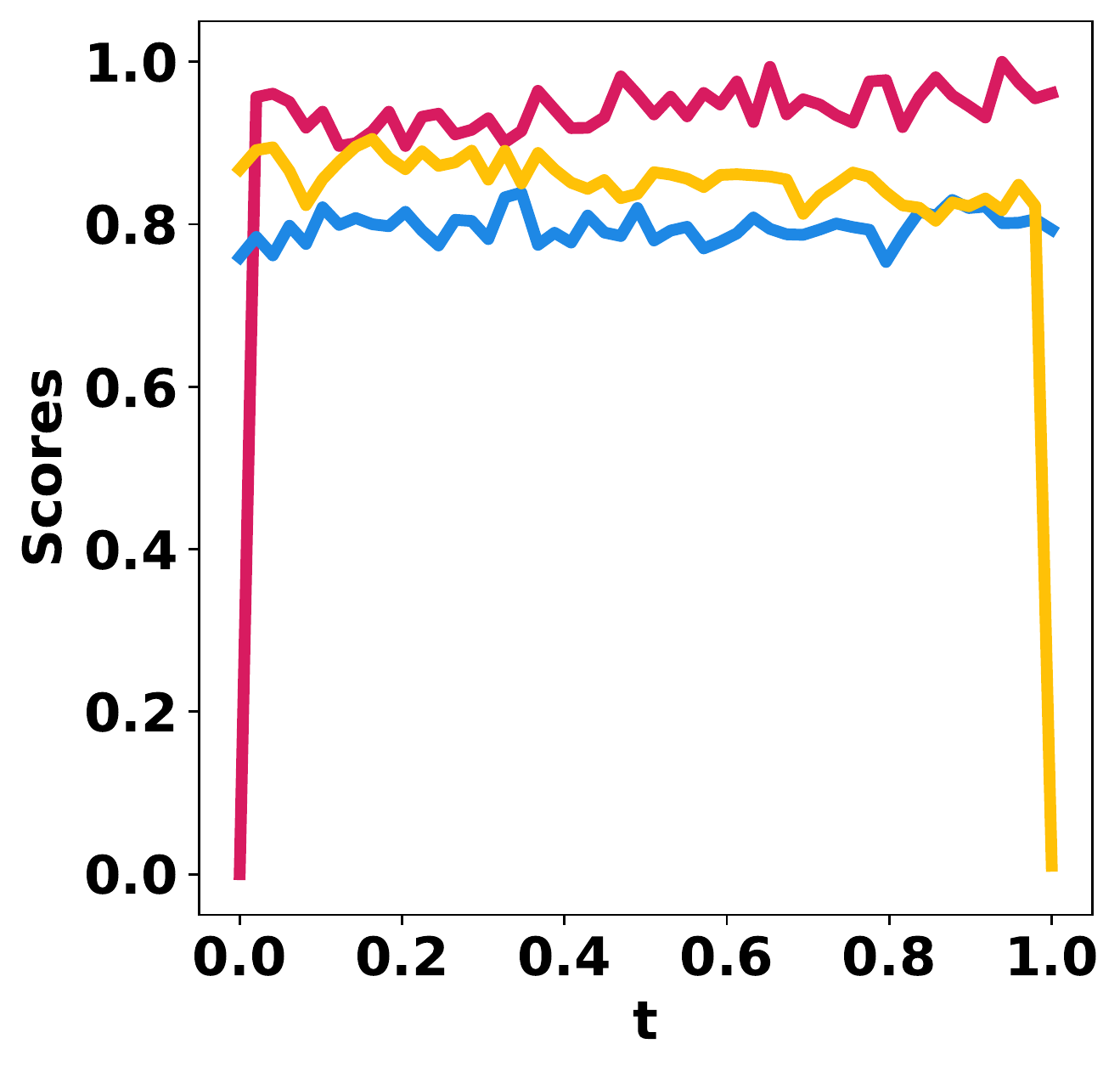} &
            \includegraphics[width=0.16\linewidth]{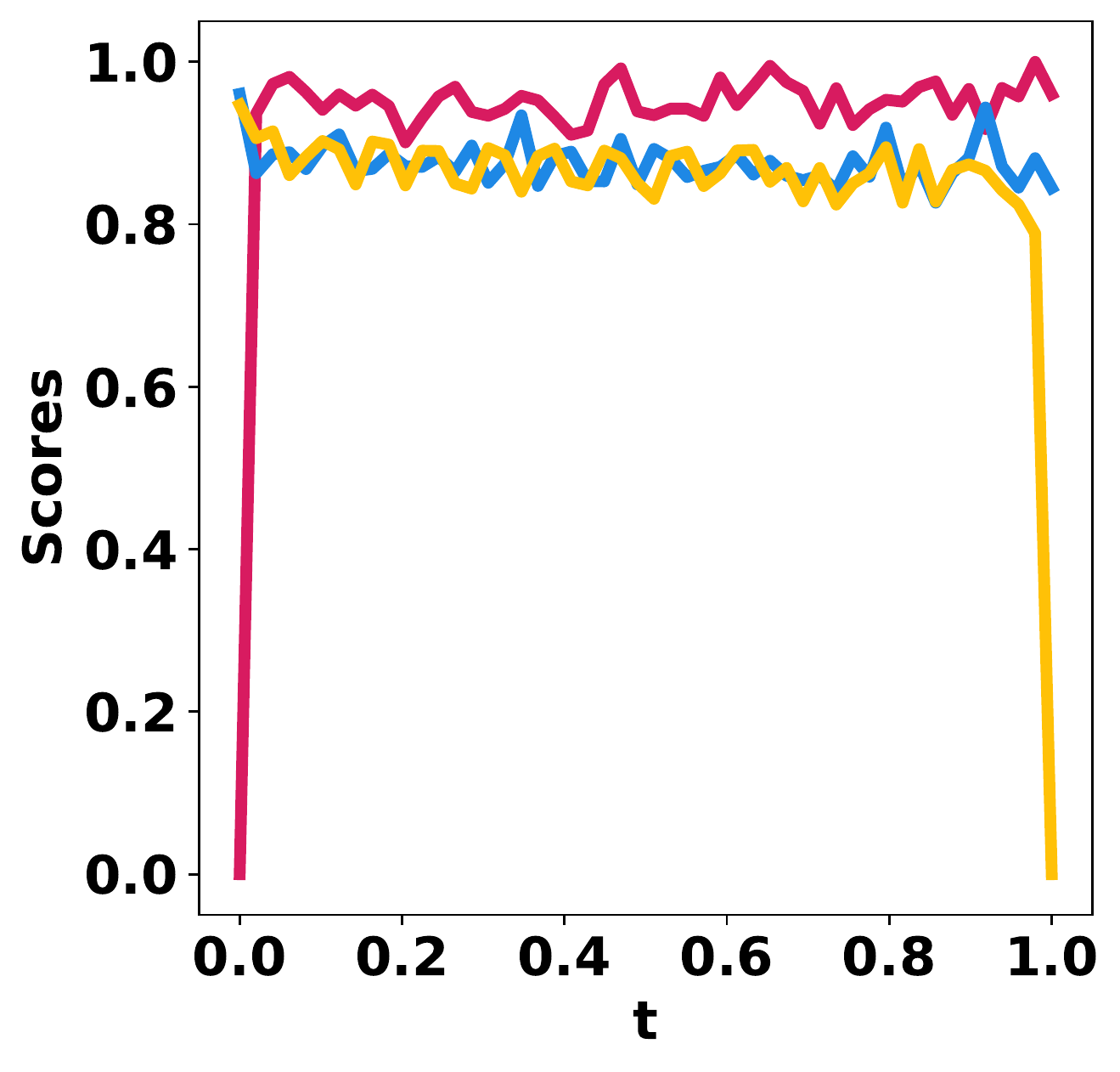} &
            \includegraphics[width=0.16\textwidth]{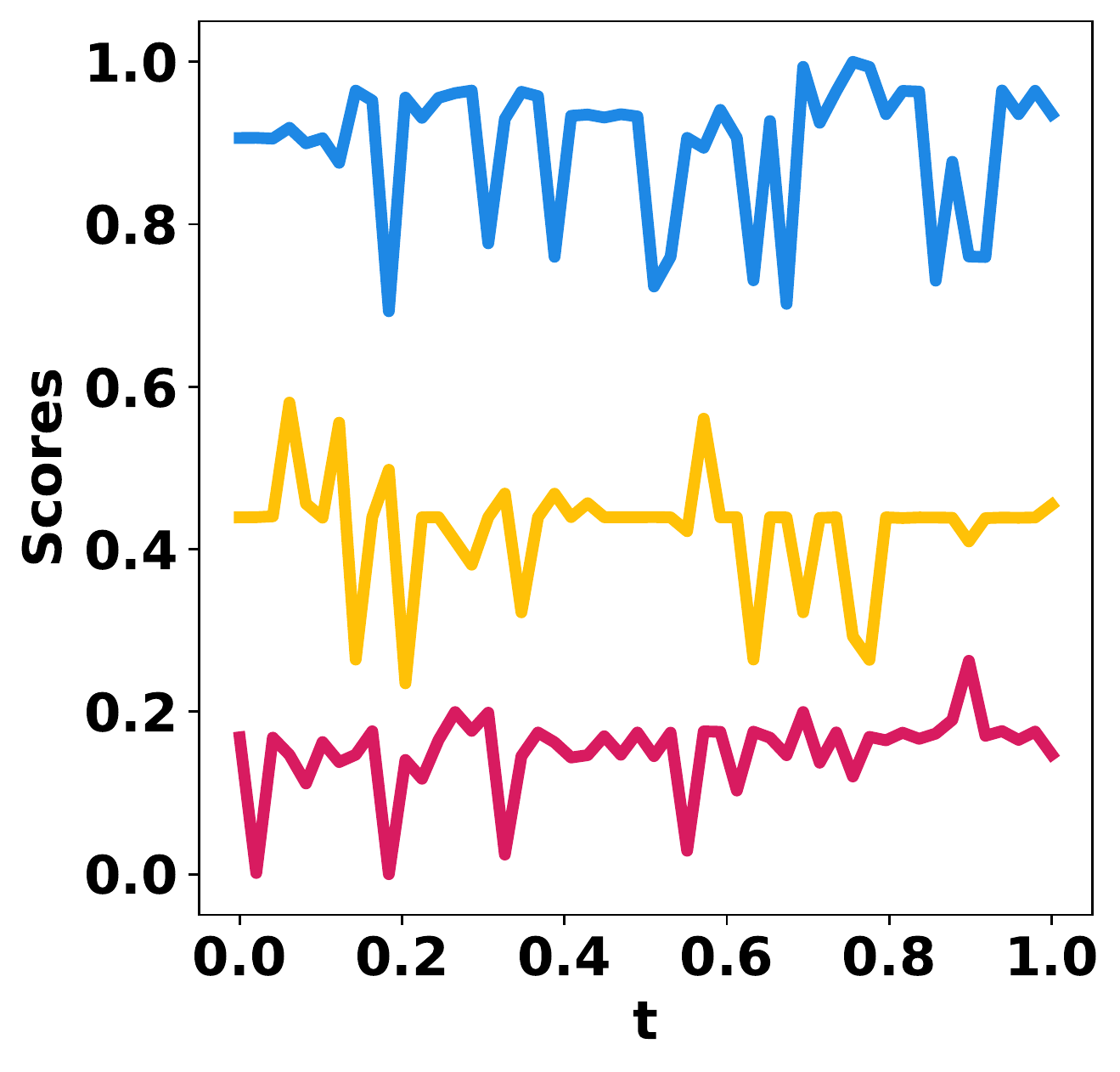} &
            \includegraphics[width=0.16\textwidth]{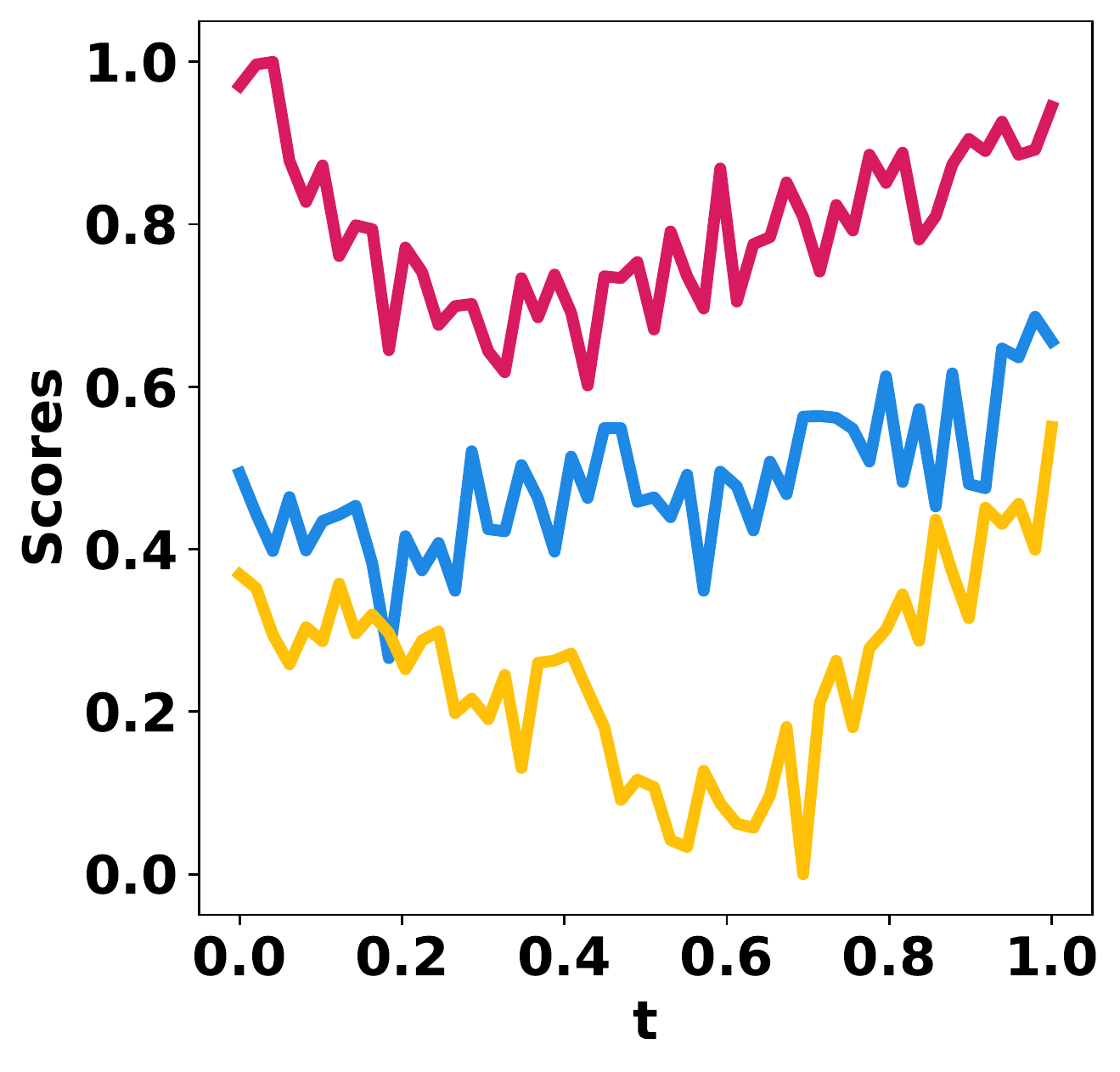} &
            \includegraphics[width=0.16\textwidth]{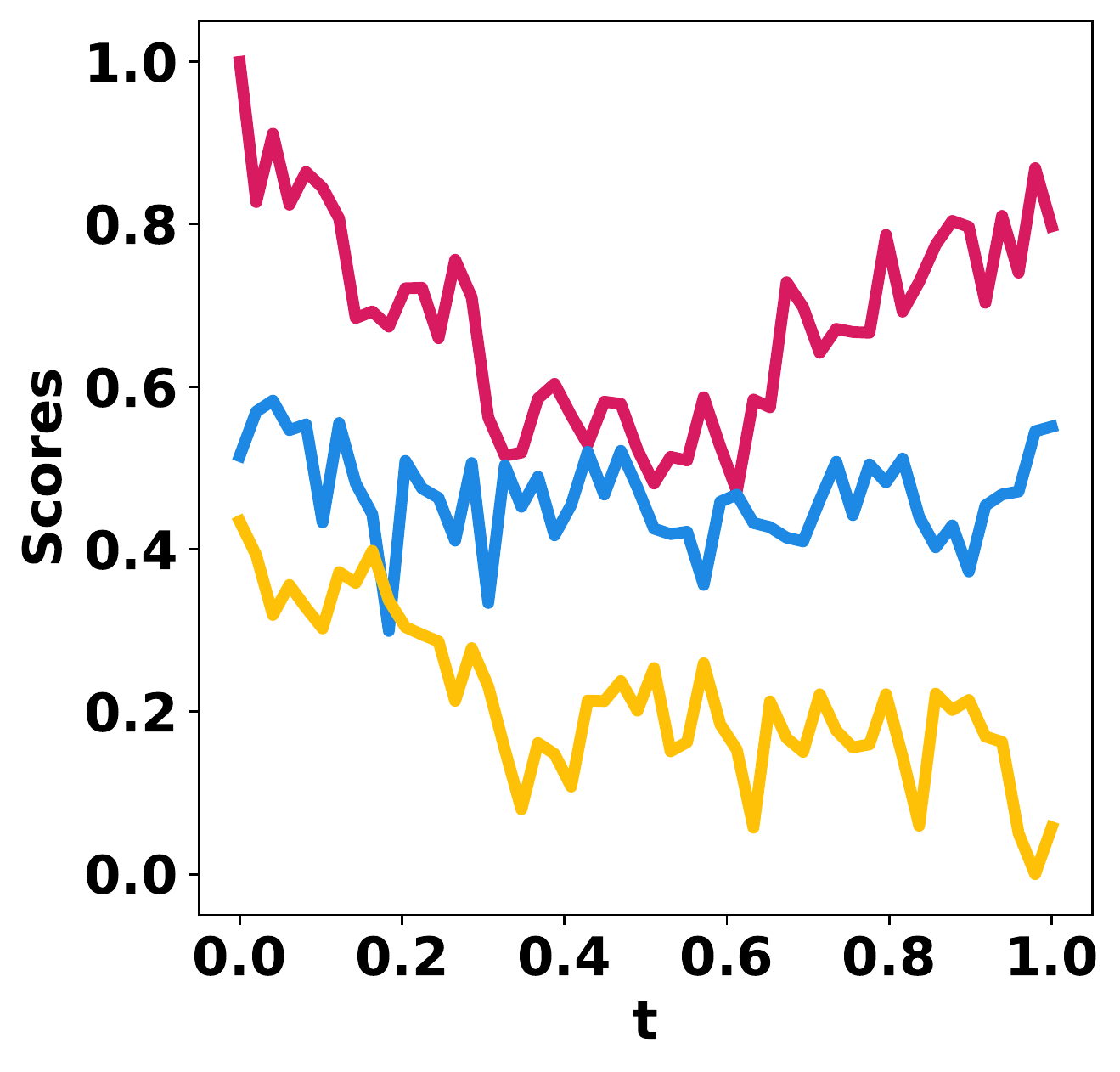} \\
            \hspace{0.12in} \small CRPC-Weak & \hspace{0.12in}  \small LSEP-Weak & \hspace{0.12in} \small GMLR-Weak &   \hspace{0.12in} \small CRPC-Weak & \hspace{0.12in} \small LSEP-Weak & \hspace{0.12in} \small GMLR-Weak
        \end{tabular}
    \end{subfigure}
    \begin{subfigure}[b]{1.0\linewidth}
        \centering
        \setlength\tabcolsep{0.2pt}
        \begin{tabular}[b]{cccccc}
            &&&&&\\
            \includegraphics[width=0.16\linewidth]{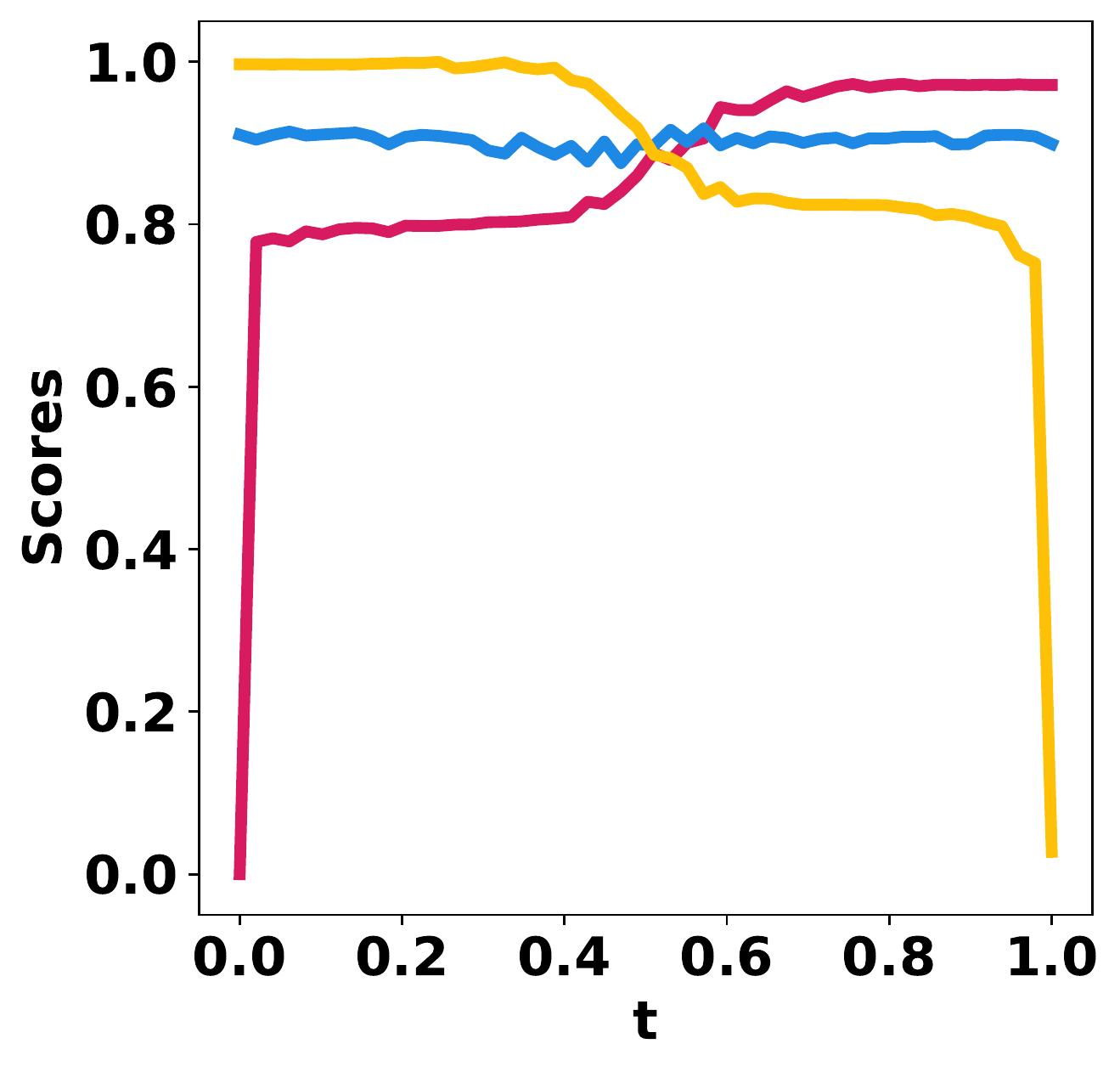} &
            \includegraphics[width=0.16\linewidth]{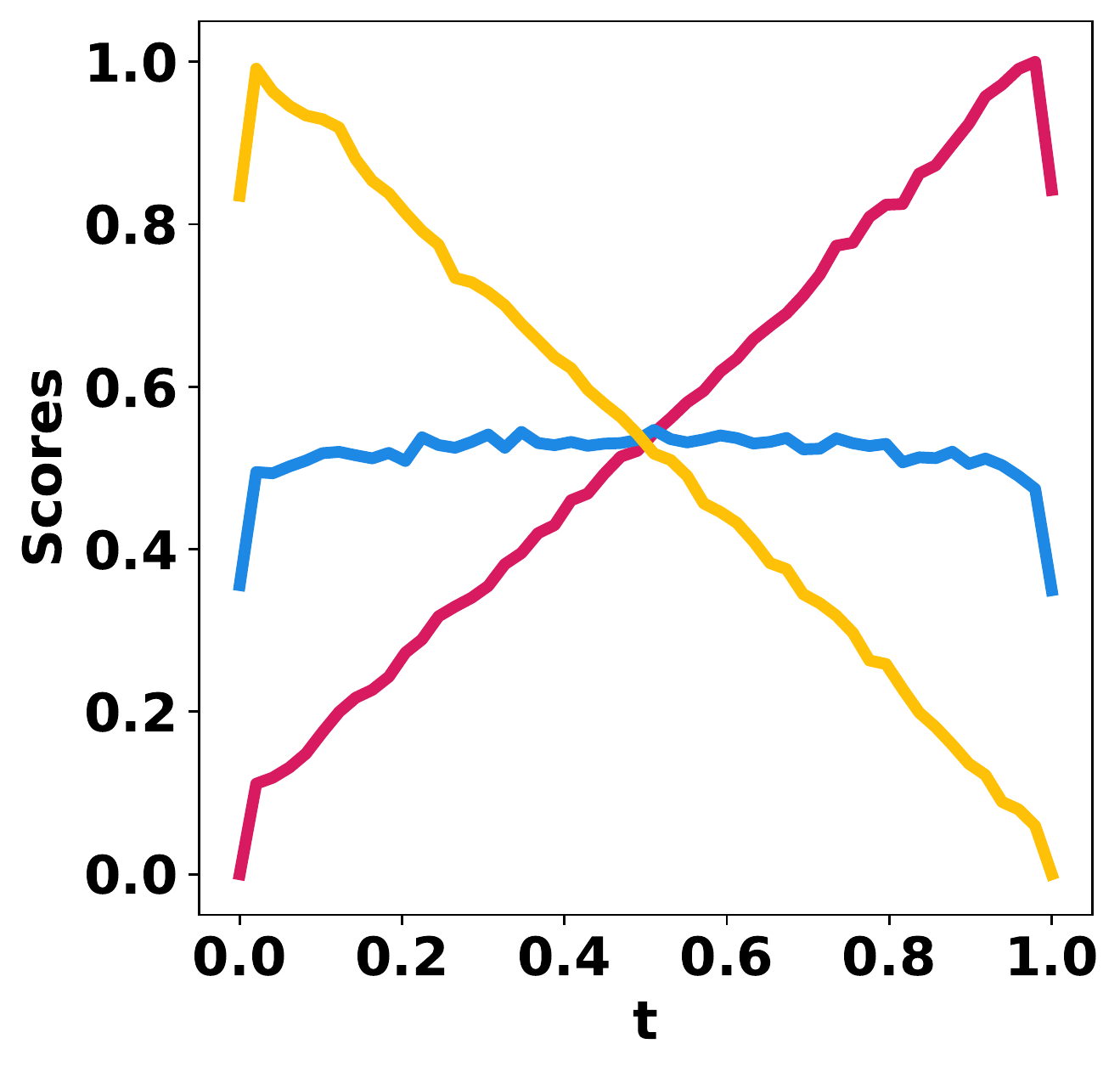} &
            \includegraphics[width=0.16\linewidth]{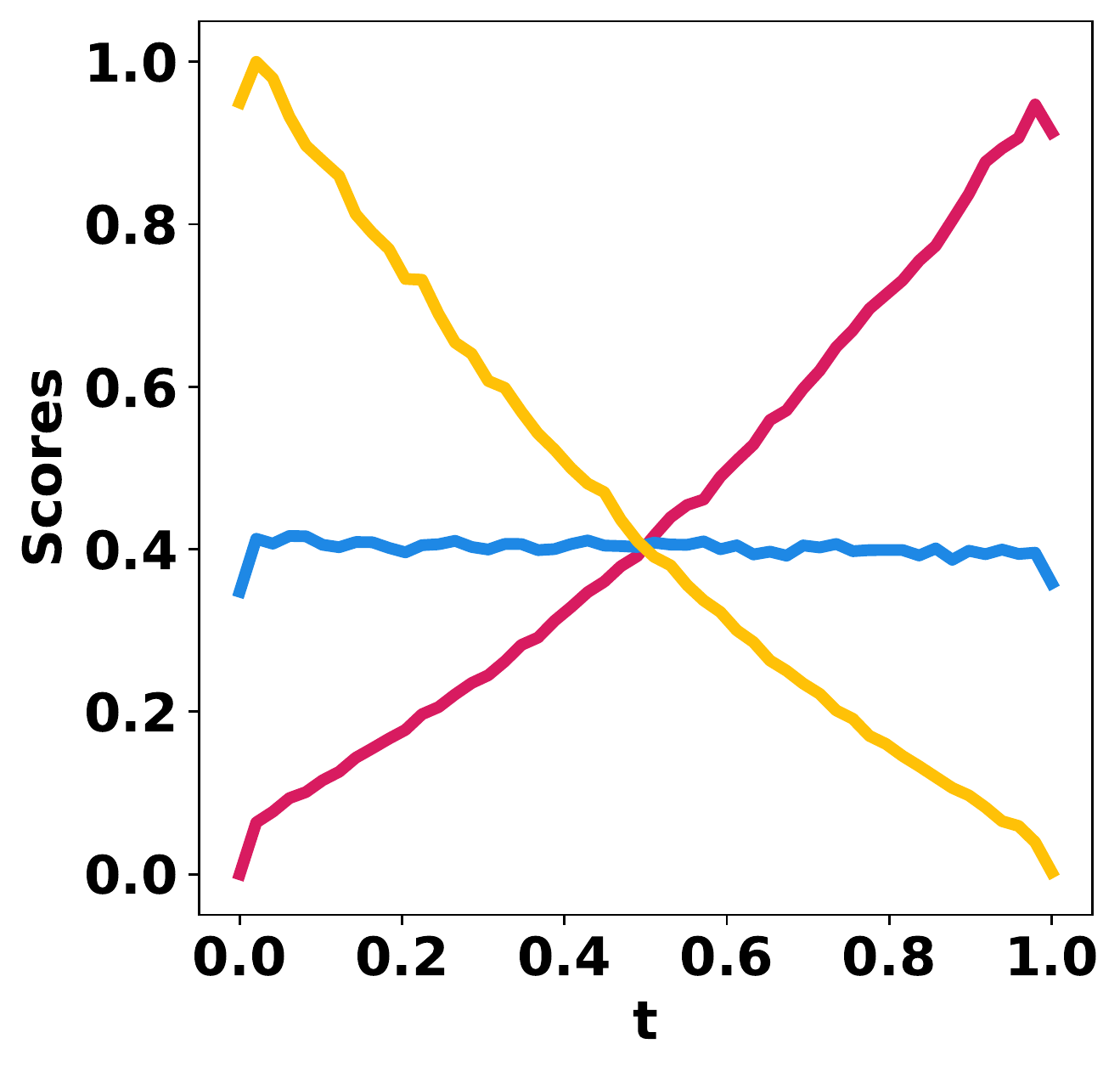} &
            \includegraphics[width=0.16\textwidth]{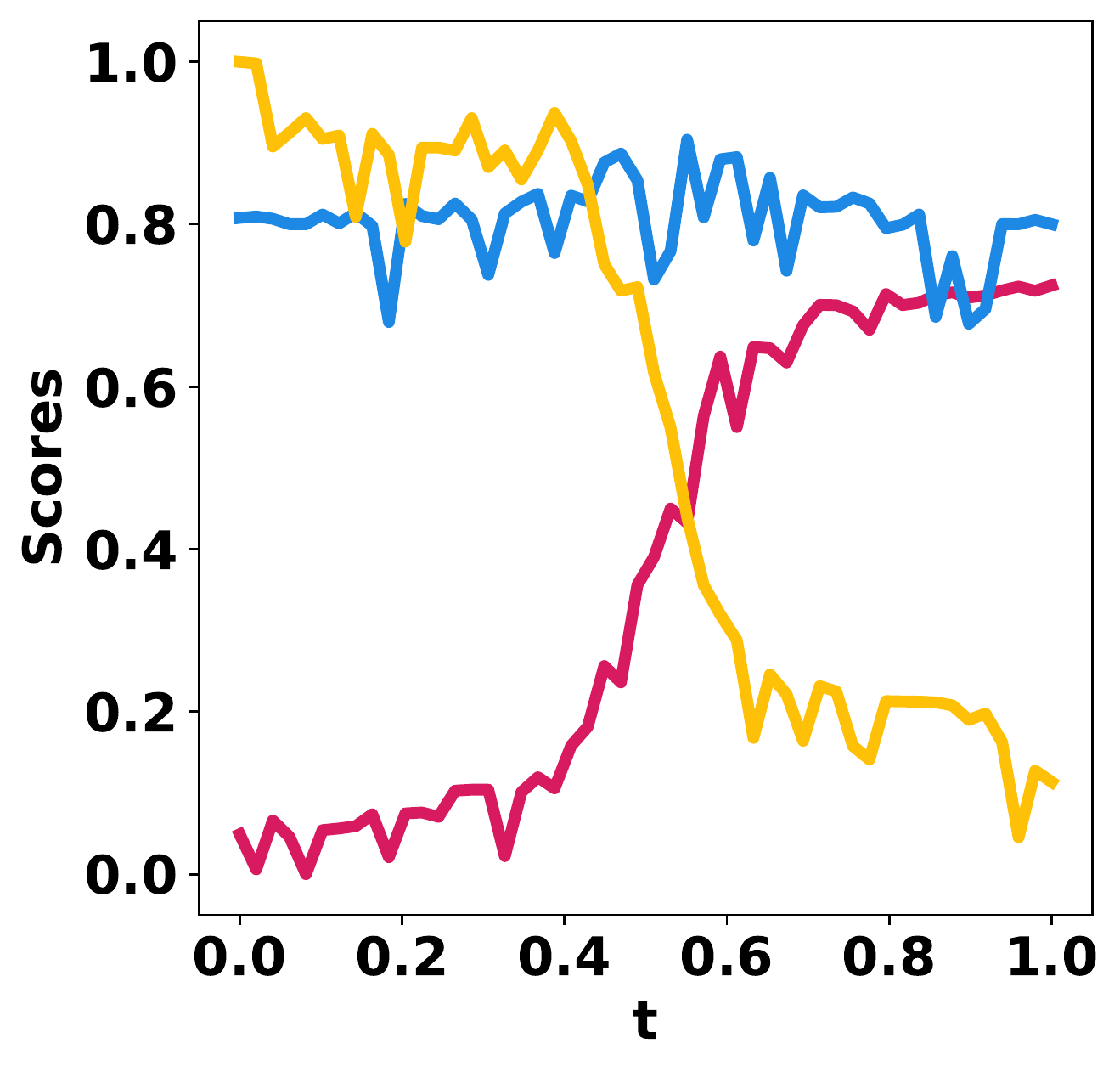} &
            \includegraphics[width=0.16\textwidth]{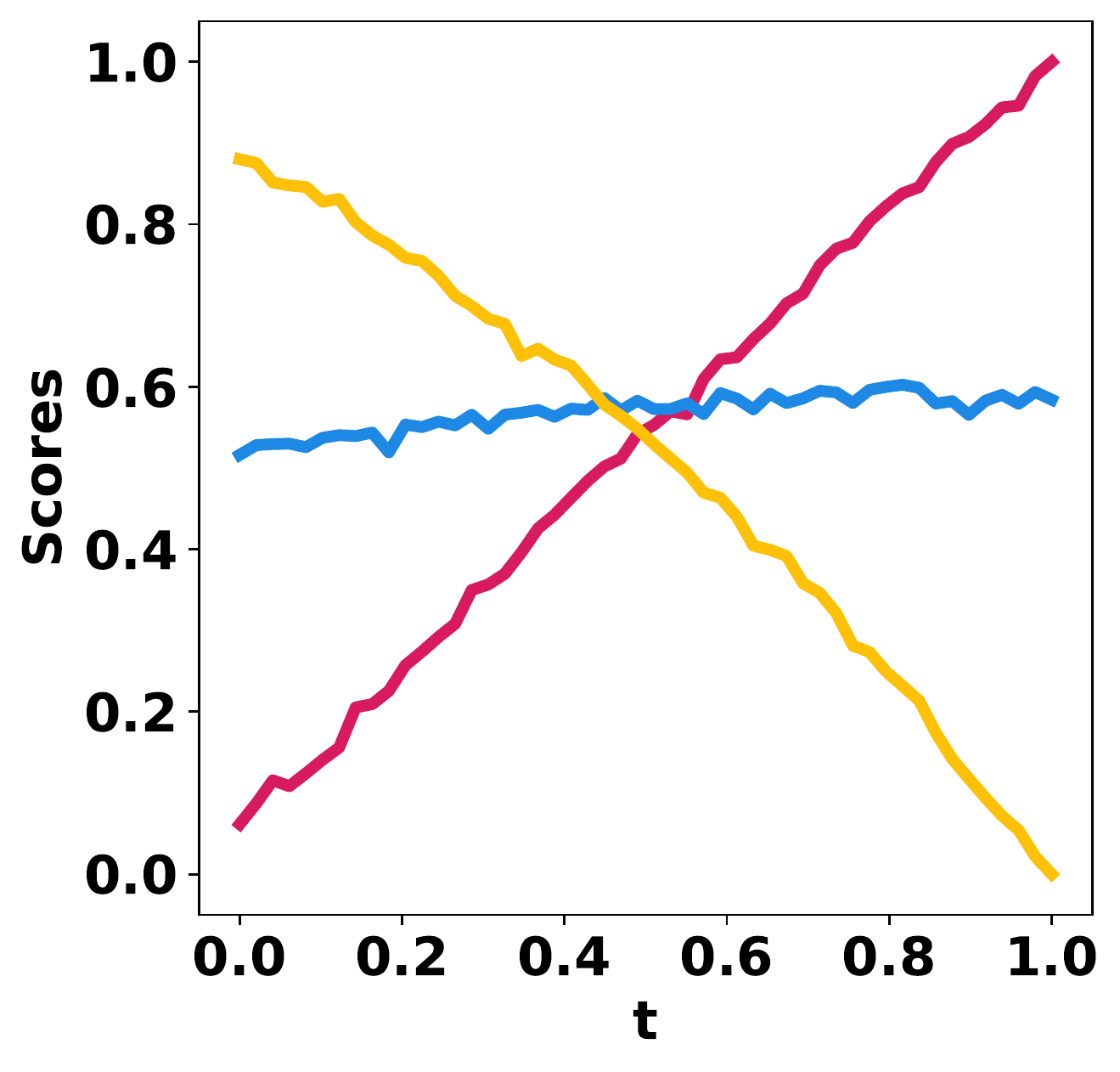} &
            \includegraphics[width=0.16\textwidth]{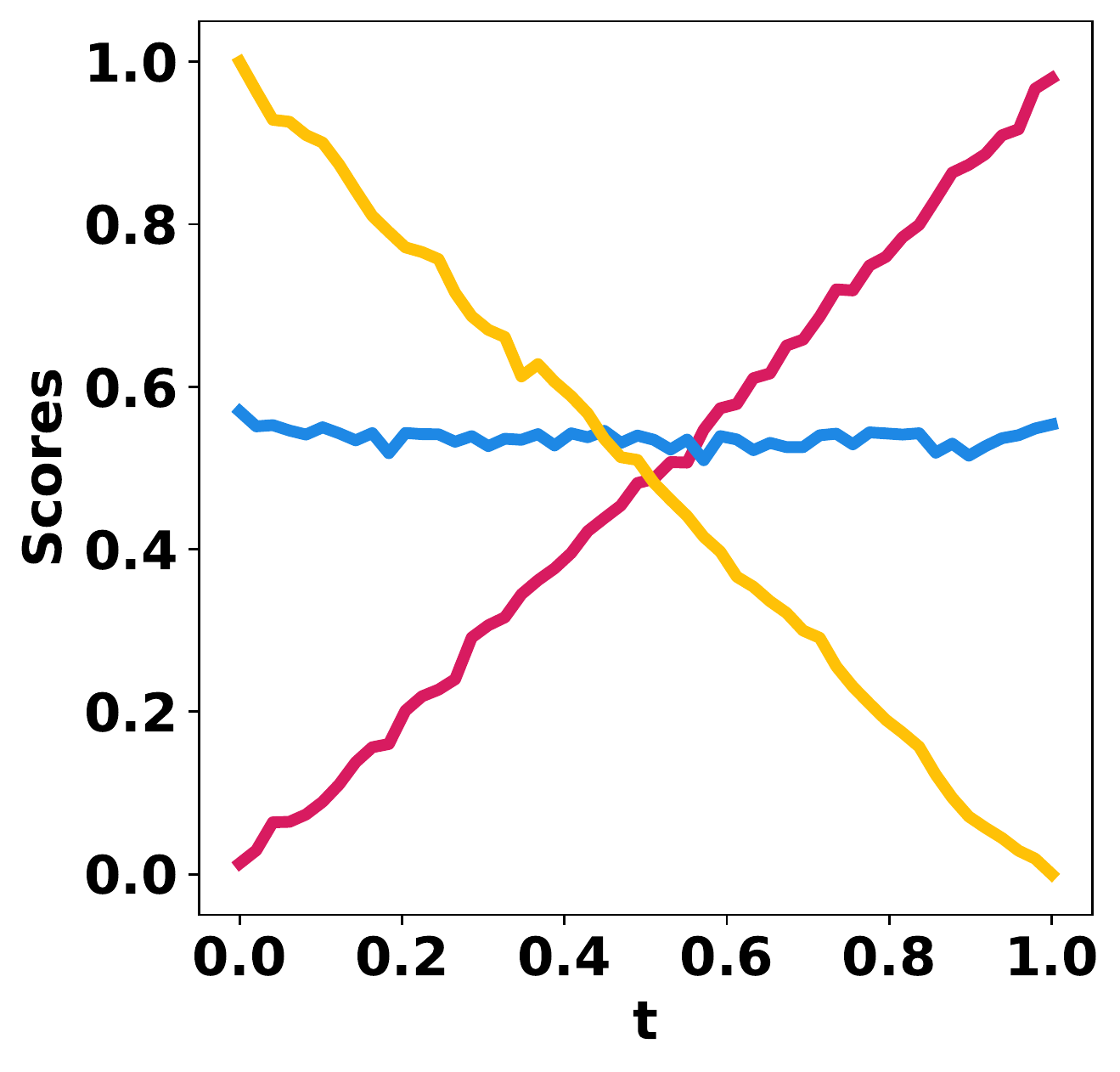} \\
             \hspace{0.12in} \small CRPC-Strong & \hspace{0.12in} \small LSEP-Strong & \hspace{0.12in} \small GMLR-Strong &   \hspace{0.12in} \small CRPC-Strong & \hspace{0.12in} \small LSEP-Strong & \hspace{0.12in} \small GMLR-Strong
        \end{tabular}
    \end{subfigure}
    \caption{Gradually changing significance effects in the sequences are shown at the top row, where the importance factor is the brightness of digits in top-left, Ranked MNIST Gray-B (Mix), and size of digits in top-right, Ranked MNIST Gray-S (Mix). Size of the digits for Ranked MNIST Gray-B (Mix) and brightness of the digits for Ranked MNIST Gray-S (Mix) change randomly as explained in Section \ref{sec:app-ranked-mnist}. Lines demonstrate changes in scores of $\langle$\textbf{\textcolor{1st}{1st}}, \textbf{\textcolor{2nd}{2nd}}, \textbf{\textcolor{3rd}{3rd}}$\rangle$ digits, which are in the order of $\langle$2, 1, 7$\rangle$ in top-left and $\langle$3, 7, 2$\rangle$ in top-right.}
    \label{fig:interpolation3}
\end{figure}

The additional Adjusting Significance Effects Experiment on Ranked MNIST Gray (Mix) that we could not provide in the main paper due to space limitations can also be seen in Figure \ref{fig:interpolation3}.
\newpage

\begin{table}[H]
\caption{Quantitative results on Ranked MNIST Color datasets. Ranked MNIST S and B stands for changing the scale or brightness of the digits, while (Mix) means both of the features are changing, but the ground truth indicates only one of the features. Bold-marked results show the best scores in Strong(S) baselines, and underlined scores show the best scores in Weak(W) baselines.}
 \renewcommand{\arraystretch}{1.2}
\centering
\resizebox{1.0\textwidth}{!}{ 
\begin{tabular}{c|cccccc|cccccc|cccccc|cccccc}
\Xhline{3\arrayrulewidth}
 \multirow{2}{*}{Method} & \multicolumn{6}{c|}{Ranked MNIST Color-S} & \multicolumn{6}{c|}{Ranked MNIST Color-B} & \multicolumn{6}{c|}{Ranked MNIST Color-S (Mix)} & \multicolumn{6}{c}{Ranked MNIST Color-B (Mix)} \\ \cline{2-25}
 & \multicolumn{1}{c}{$\tau_b$ $\uparrow$} & \multicolumn{1}{c}{$ S \rho$ $\uparrow$} & \multicolumn{1}{c}{$\gamma$ $\uparrow$} & \multicolumn{1}{c}{HL $\downarrow$} & \multicolumn{1}{c}{M-1 $\downarrow$} & F1 $\uparrow$ & \multicolumn{1}{c}{$\tau_b$ $\uparrow$} & \multicolumn{1}{c}{$ S \rho$ $\uparrow$} & \multicolumn{1}{c}{$\gamma$ $\uparrow$} & \multicolumn{1}{c}{HL $\downarrow$} & \multicolumn{1}{c}{M-1 $\downarrow$ } & F1 $\uparrow$ & \multicolumn{1}{c}{$\tau_b$ $\uparrow$} & \multicolumn{1}{c}{$ S \rho$ $\uparrow$} & \multicolumn{1}{c}{$\gamma$ $\uparrow$} & \multicolumn{1}{c}{HL $\downarrow$} & \multicolumn{1}{c}{M-1 $\downarrow$} & F1 $\uparrow$& \multicolumn{1}{c}{$\tau_b$ $\uparrow$} & \multicolumn{1}{c}{$ S \rho$ $\uparrow$} & \multicolumn{1}{c}{$\gamma$ $\uparrow$} & \multicolumn{1}{c}{HL $\downarrow$} & \multicolumn{1}{c}{M-1 $\downarrow$} & F1 $\uparrow$ \\
 \Xhline{2\arrayrulewidth}
\multicolumn{1}{c|}{CRPC (W)} & 49.80 & 60.40 & \multicolumn{1}{c}{60.11} & 17.07 & 0.32 & 86.45 & 36.06 & 62.67 & \multicolumn{1}{c}{59.29} & 11.69 & 0.41 & 90.28 & 52.07  & 62.40 & \multicolumn{1}{c}{60.15} & 14.77 & 0.35 & 87.93 & 50.76 & 61.36  & \multicolumn{1}{c}{60.23} & 15.99 & 0.38  & 87.05 \\

\multicolumn{1}{c|}{LSEP (W)} & 61.85 & 71.06 & \multicolumn{1}{c}{61.98} & \underline{0.51} & \underline{0.10} & \underline{99.53} & 59.70 & 69.03 &  \multicolumn{1}{c}{60.03} & 1.07 & \underline{0.17} &  \underline{99.02} & 61.58 & 70.75 & \multicolumn{1}{c}{61.85} & 1.01 & \underline{0.12} & 99.07 & 61.12 & 70.44 & \multicolumn{1}{c}{61.39} & \underline{1.00} & 0.16 & 99.08 \\

\multicolumn{1}{c|}{GMLR (W)} & \underline{63.60} & \underline{73.05} & \multicolumn{1}{c}{\underline{63.77}} & \underline{0.49} & \underline{0.08} & \underline{99.55} & \underline{60.0} & \underline{69.25} & \multicolumn{1}{c}{\underline{60.33}} & \underline{1.04}  & \underline{0.19} & \underline{99.04}  & \underline{62.43} & \underline{71.60} & \multicolumn{1}{c}{\underline{62.73}} & \underline{0.94} & \underline{0.11} & \underline{99.13} & \underline{61.68} & \underline{71.10} & \multicolumn{1}{c}{\underline{61.98}} & 1.06  & \underline{0.11} & \underline{99.03} \\ \hline

\multicolumn{1}{c|}{CRPC (S)} &  63.91 & 75.42 & \multicolumn{1}{c}{74.88} & 19.06 & 0.23 & 85.08 & 62.62 & 74.41 & \multicolumn{1}{c}{74.78} & 21.02 & 0.42 & 83.69 & 61.56 & 73.41 & \multicolumn{1}{c}{74.13} & 22.63 & 0.32 & 82.53 & 64.77 & 76.07 & \multicolumn{1}{c}{75.0} & 16.33 & 0.29  & 86.78  \\

\multicolumn{1}{c|}{LSEP (S)} & 93.76 & 97.32 & \multicolumn{1}{c}{94.34} & 1.42 & \textbf{0.09} & 98.70 & \textbf{92.91} & \textbf{96.58} & \multicolumn{1}{c}{\textbf{93.72}} & \textbf{2.02} & \textbf{0.31} & \textbf{98.14} & 91.74 & 95.93 & \multicolumn{1}{c}{92.61} & 2.29 & 0.20 & 97.90 & \textbf{92.48} & \textbf{96.51} & \multicolumn{1}{c}{\textbf{93.24}} & \textbf{2.07} & \textbf{0.21} & \textbf{98.09} \\

\multicolumn{1}{c|}{GMLR (S)} & \textbf{94.18} & \textbf{97.52}  & \multicolumn{1}{c}{\textbf{94.44}} & \textbf{0.66} & \textbf{0.07}  & \textbf{99.40} & 89.43 & 93.44  & \multicolumn{1}{c}{92.69} & 3.15 & 0.88 & 97.15 & \textbf{92.18} & \textbf{96.27} & \multicolumn{1}{c}{\textbf{92.67}} & \textbf{1.27} & \textbf{0.15} & \textbf{98.83} & 88.80 & 92.77 & \multicolumn{1}{c}{90.41} & 2.40 & 3.03 & 97.74 \\ \hline
\end{tabular}
}
\label{tab:ranked_mnist_color}
\end{table}

\section{Bar Graphs}
\label{sec:apx7}

\begin{figure}[ht]
    \centering
    \resizebox{1.0\textwidth}{!}{
    \begin{tabularx}{\textwidth}{cccc}
    AVDP  &\hspace{0.14in} CRPC & \hspace{0.14in}LSEP & \hspace{0.14in}GMLR \\
    \includegraphics[width=0.22\linewidth]{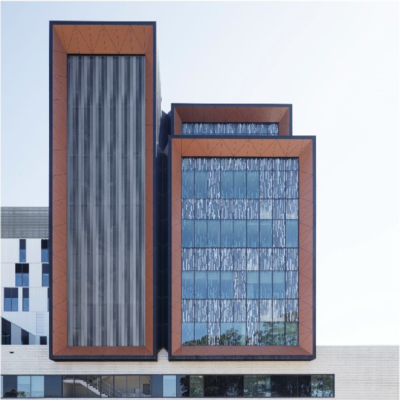}  &
    \includegraphics[width=0.22\linewidth]{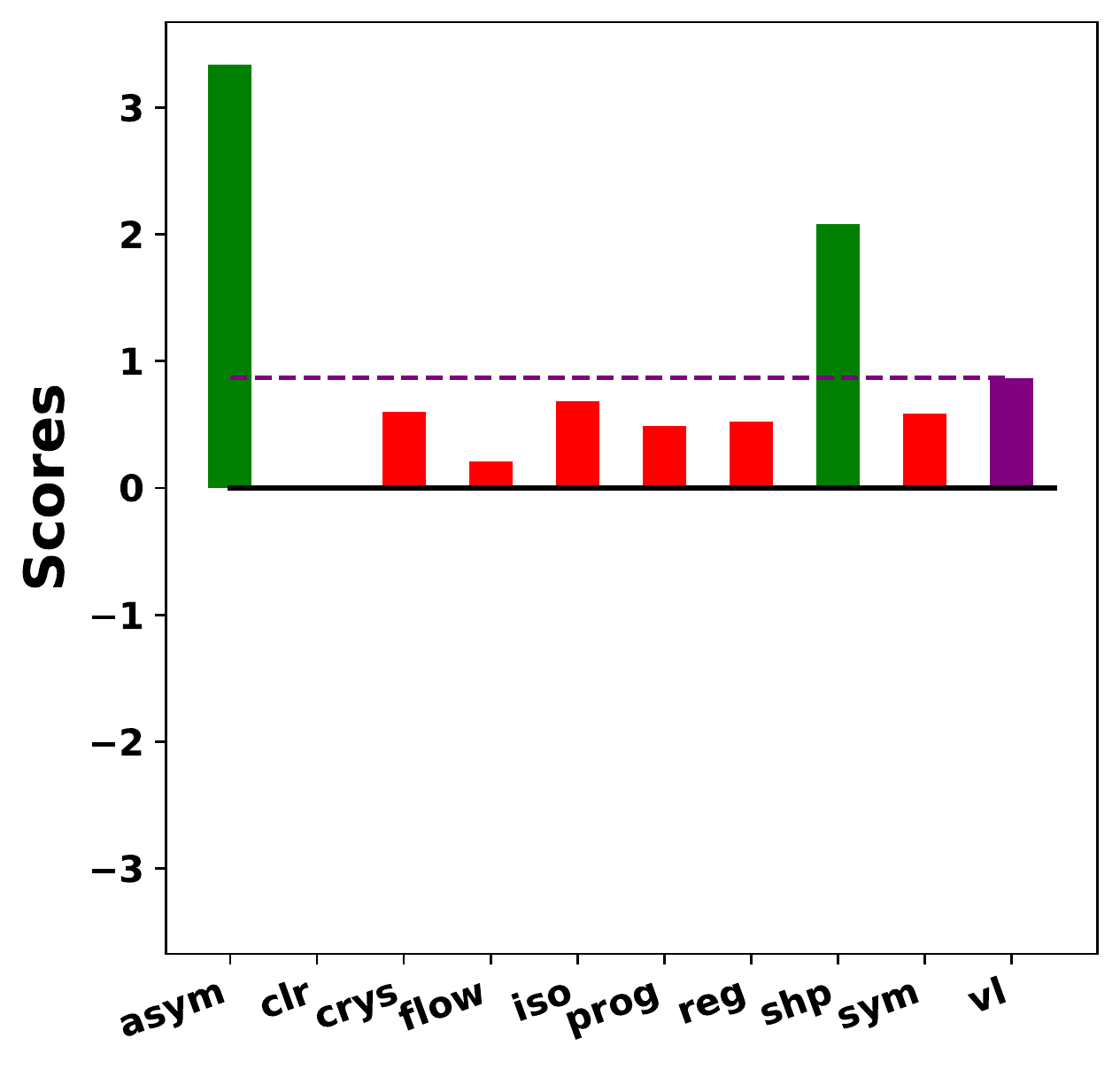} &
    \includegraphics[width=0.22\linewidth]{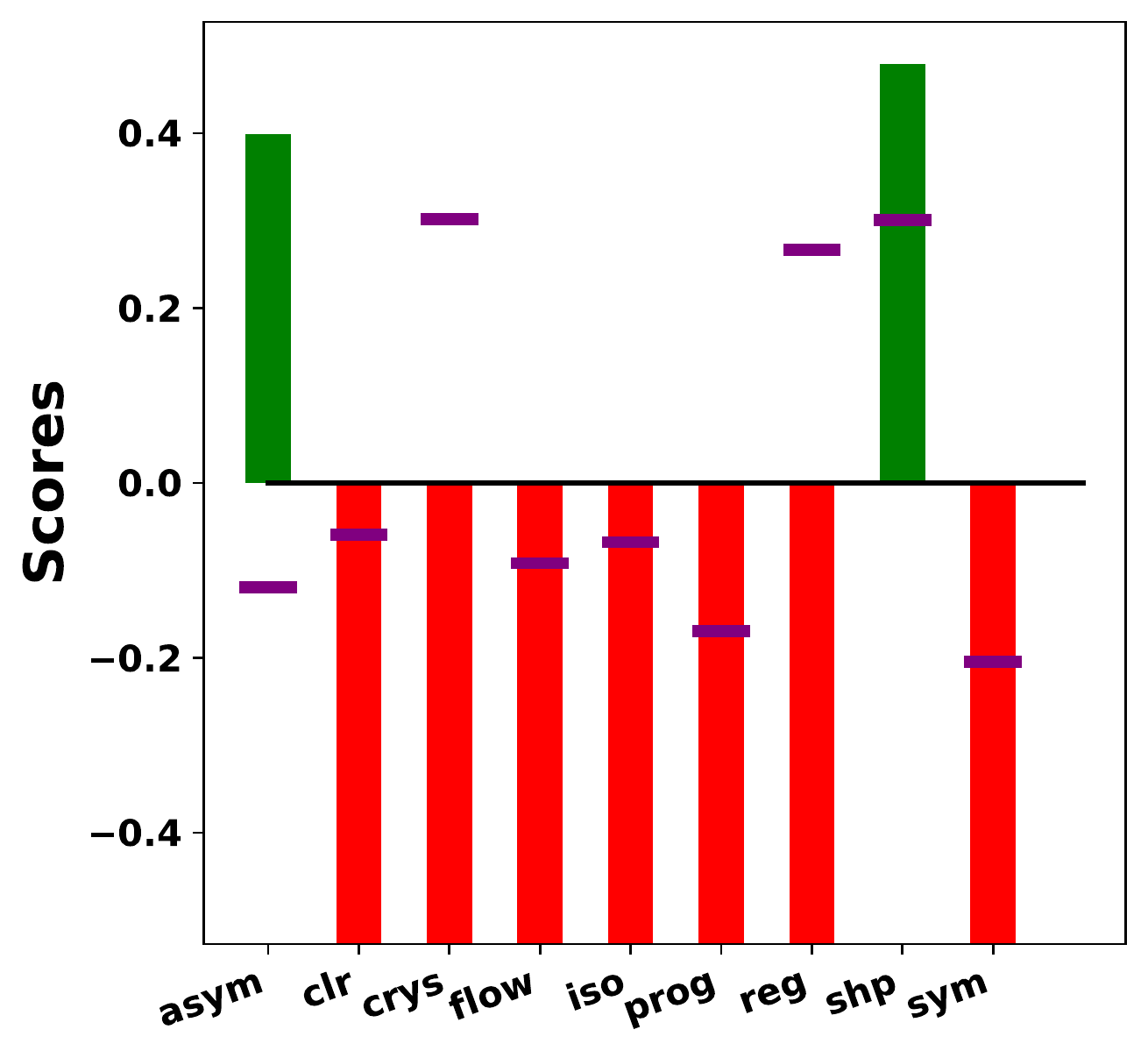} &
    \includegraphics[width=0.22\linewidth]{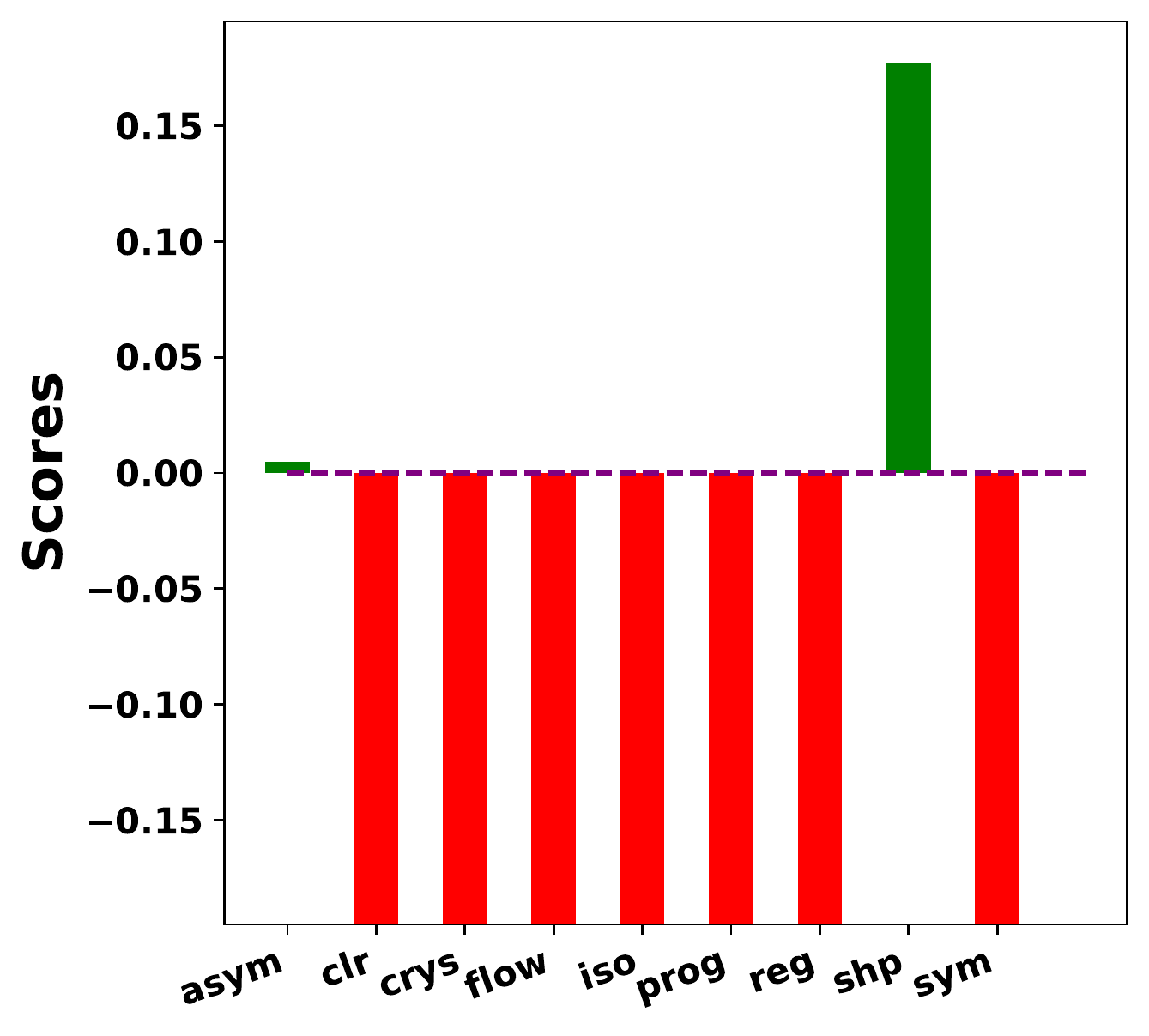} \\
    GT = \{asym $\succ$ shp $\succ$ reg\}& \hspace{0.14in} \{asym $\succ$ shp\}  & \hspace{0.14in} \{asym $\succ$ shp\} &  \hspace{0.14in} \{asym $\succ$ shp\} \\
    &&& \\
    \includegraphics[width=0.22\linewidth]{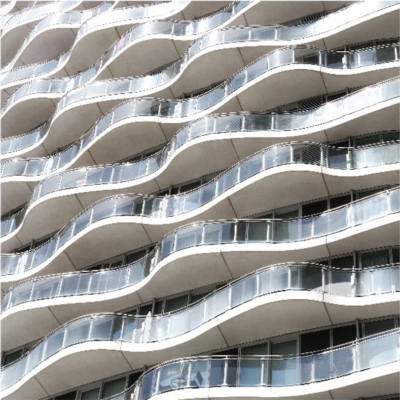}  &
    \includegraphics[width=0.22\linewidth]{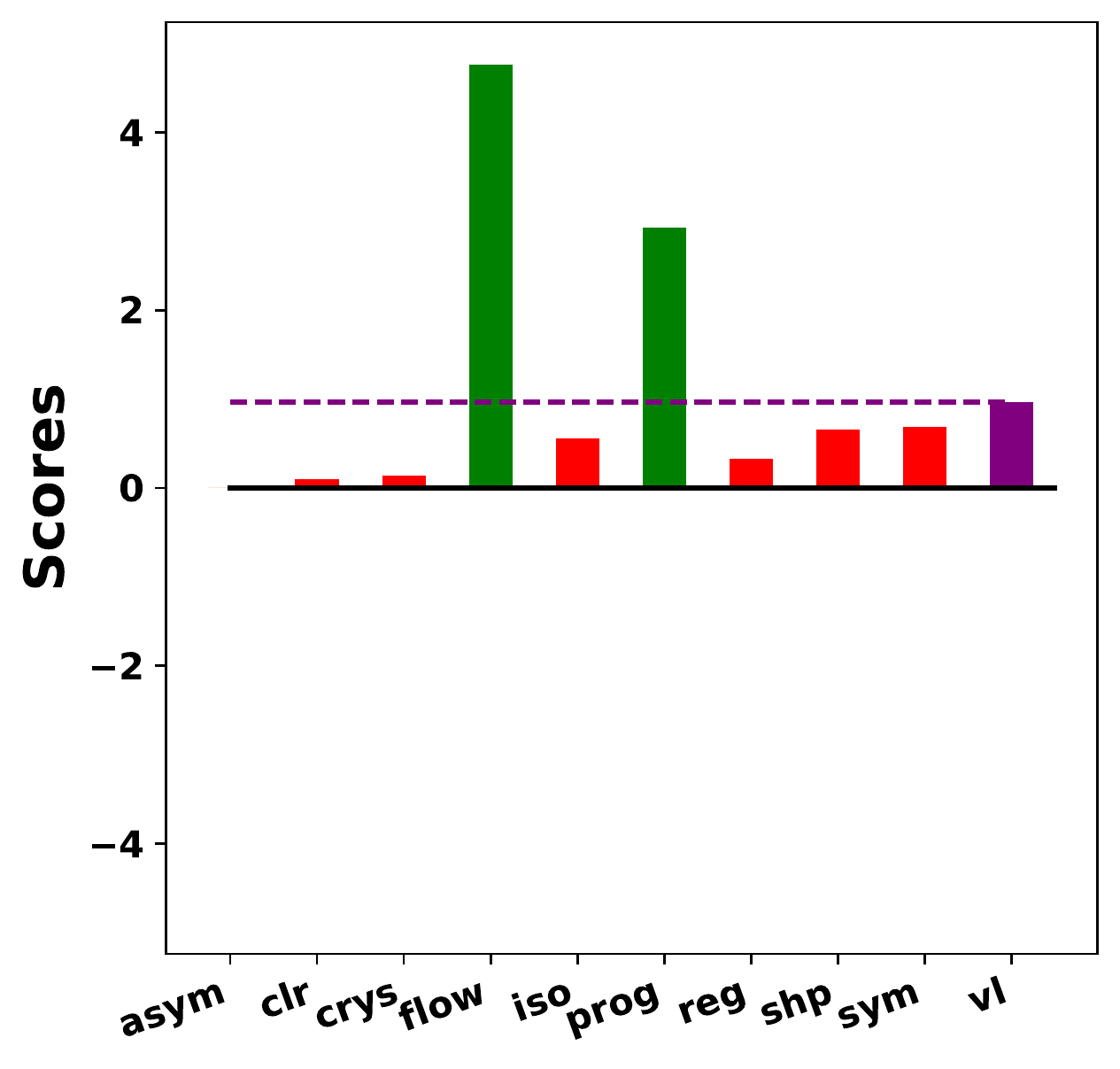} &
    \includegraphics[width=0.22\linewidth]{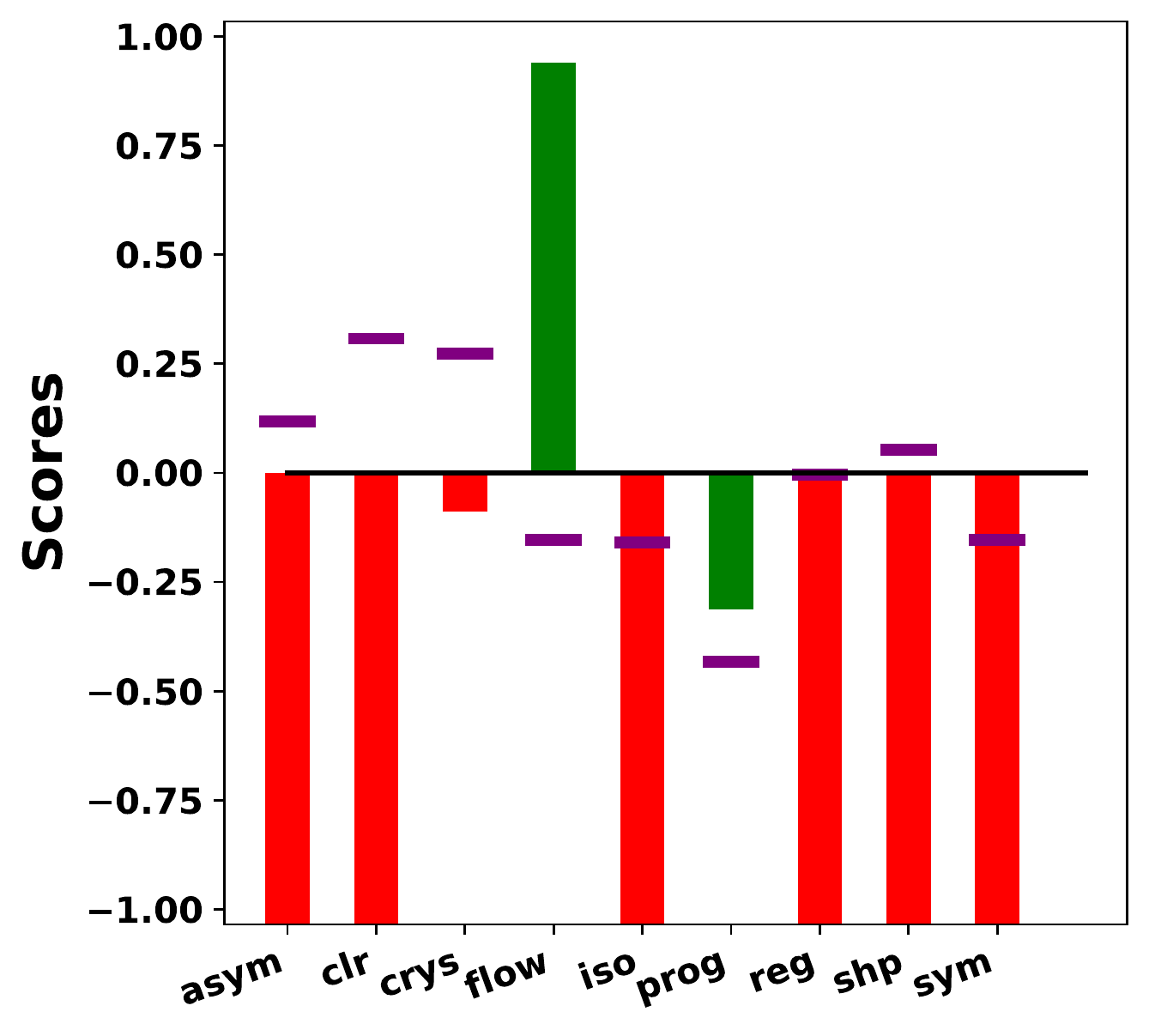} &
    \includegraphics[width=0.22\linewidth]{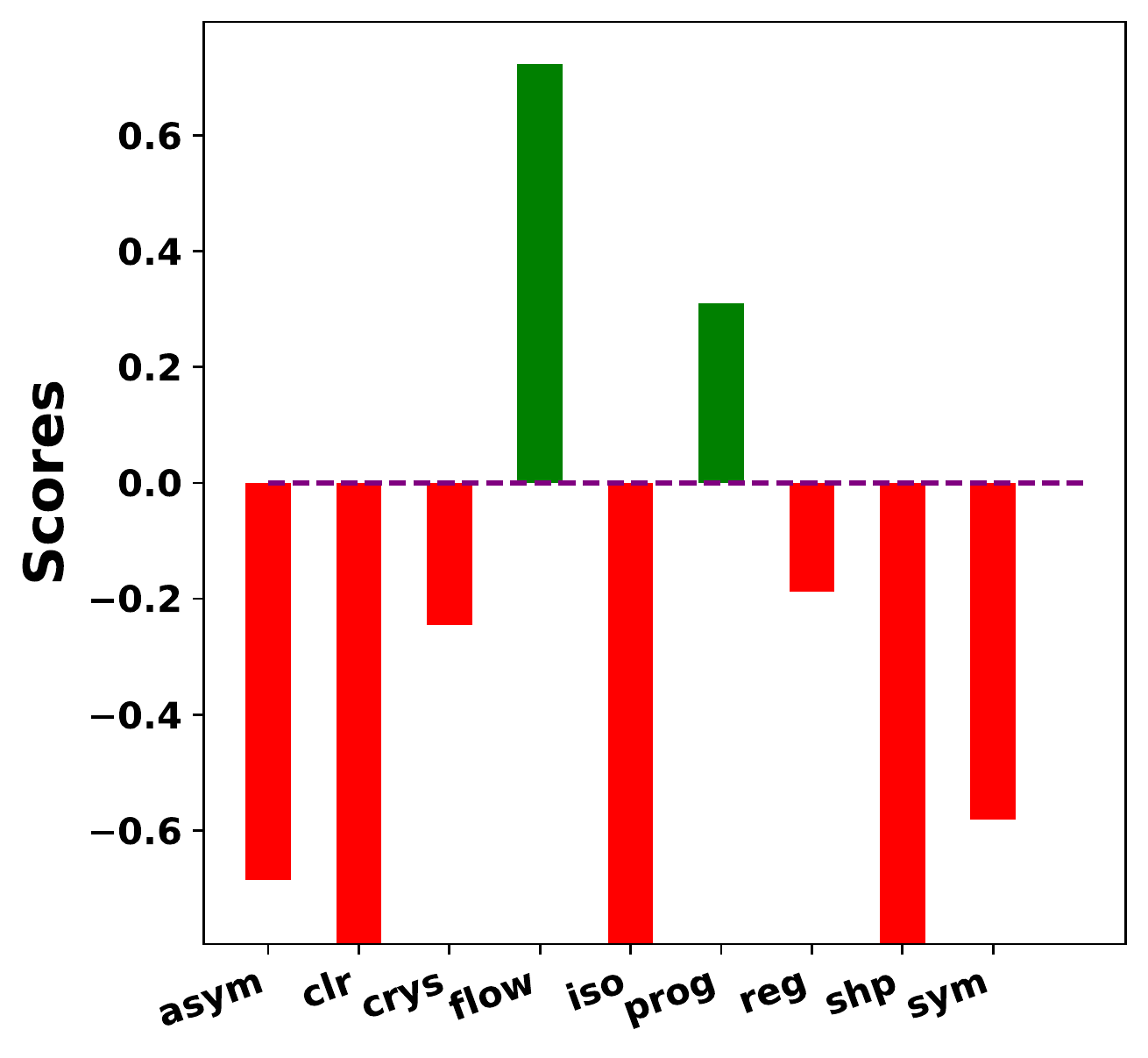} \\
    GT = \{flow\}&  \hspace{0.14in}\{flow $\succ$ prog\}& \hspace{0.14in}\{flow $\succ$ prog\} & \hspace{0.14in} \{flow $\succ$ prog\}\\
    &&& \\
    \includegraphics[width=0.22\linewidth]{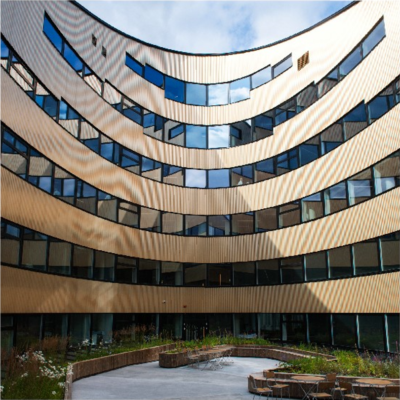}  &
    \includegraphics[width=0.22\linewidth]{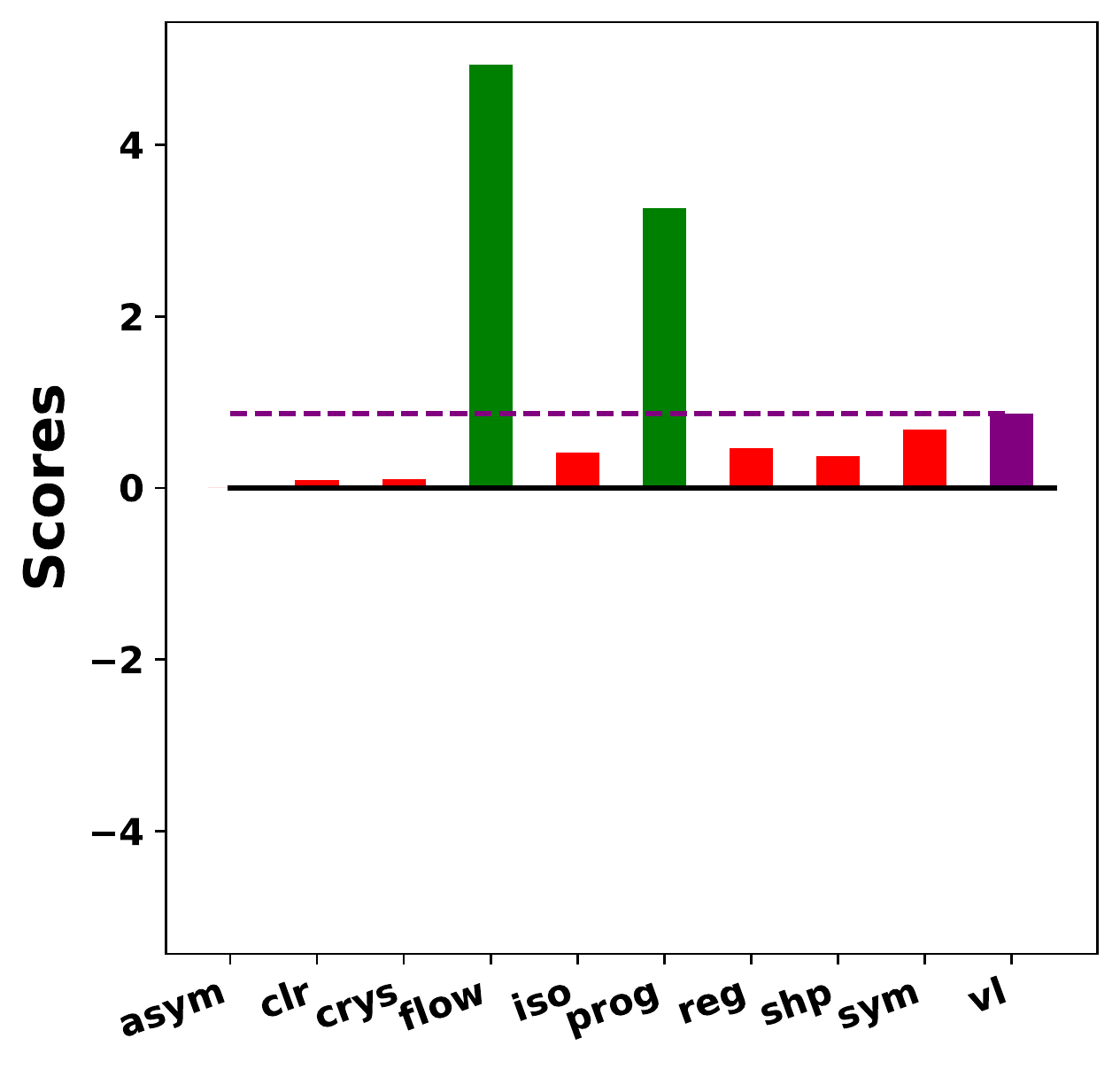} &
    \includegraphics[width=0.22\linewidth]{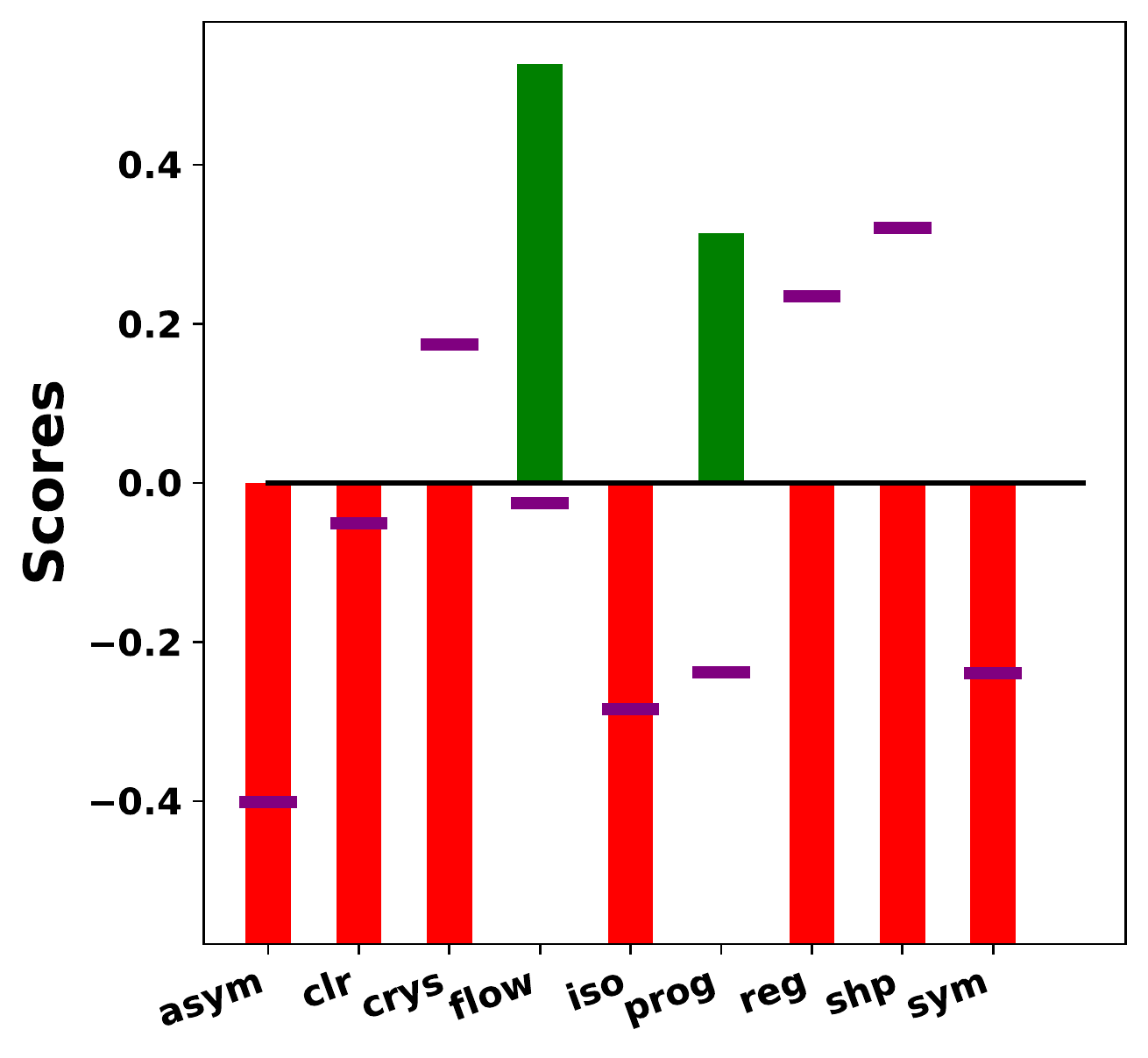} &
    \includegraphics[width=0.22\linewidth]{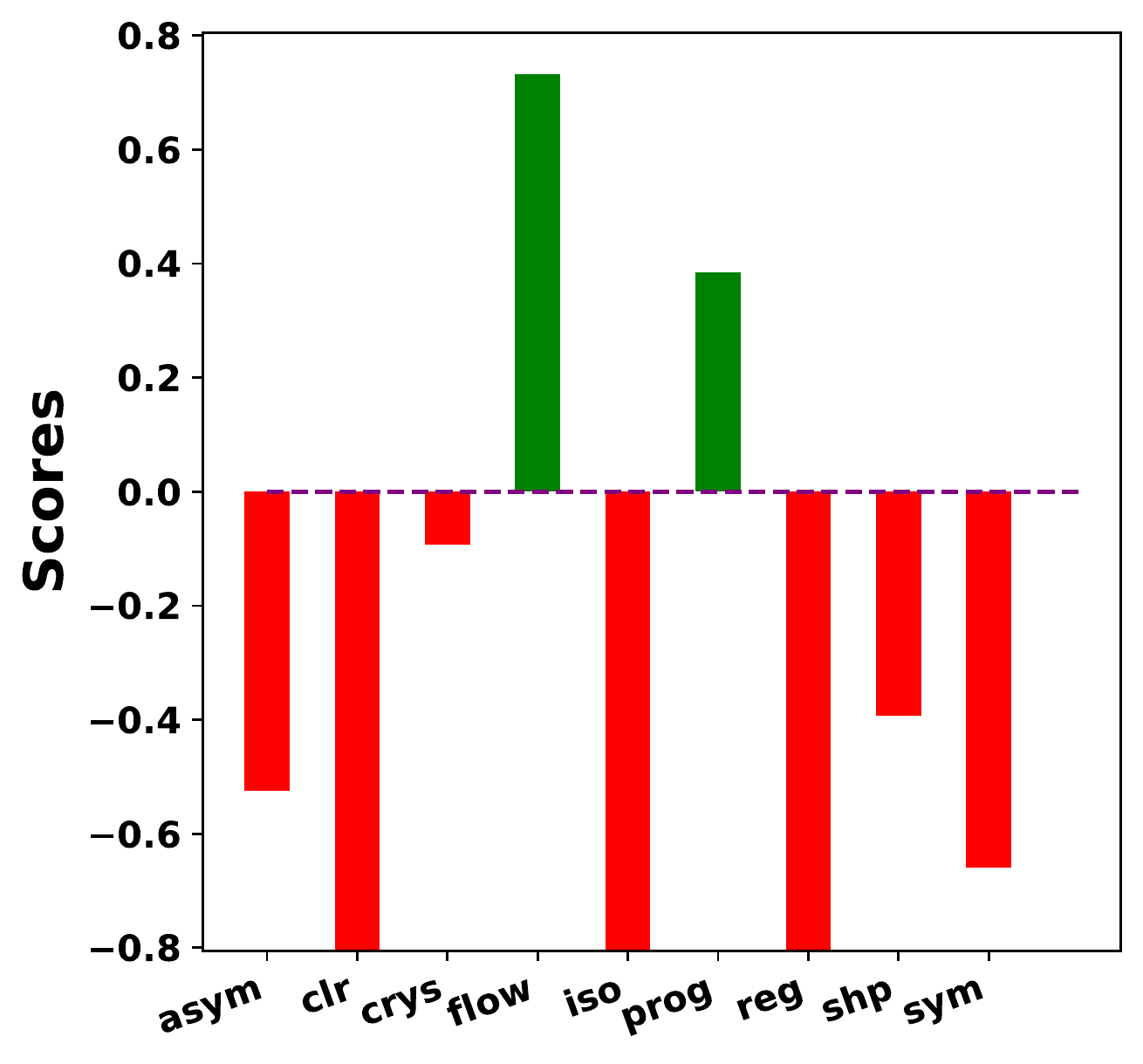} \\
    GT = \{flow $\succ$ prog\} & \hspace{0.14in}\{flow $\succ$ prog\} & \hspace{0.14in}\{flow $\succ$ prog\} & \hspace{0.14in}\{flow $\succ$ prog\}\\
    &&& \\
    \end{tabularx}
    }
    \caption{Samples from the test set of the AVDP dataset are given in the first column from left, with the corresponding ground truth labels denoted as GT. Each bar plot represents the predicted scores of baselines CRPC, LSEP and our method GMLR, in their respective columns. }
    \label{fig:AVDP-bar}
\end{figure}

\begin{figure}[H]
    \centering
    \resizebox{1.0\textwidth}{!}{
    \begin{tabularx}{\textwidth}{cccc}
    NSID & \hspace{0.14in} CRPC &\hspace{0.14in} LSEP & \hspace{0.14in}GMLR \\
    \includegraphics[width=3cm, height=3cm]{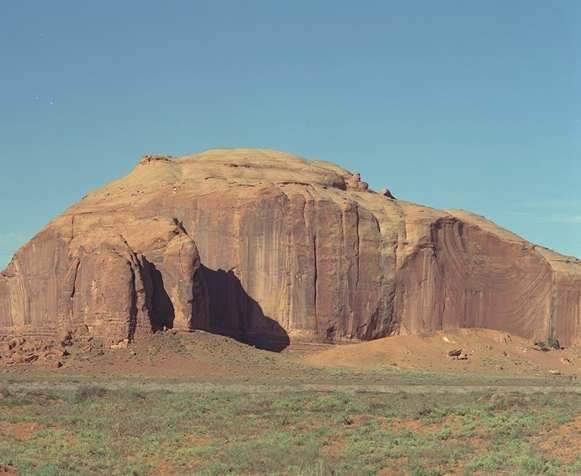} &
    \includegraphics[width=0.22\linewidth]{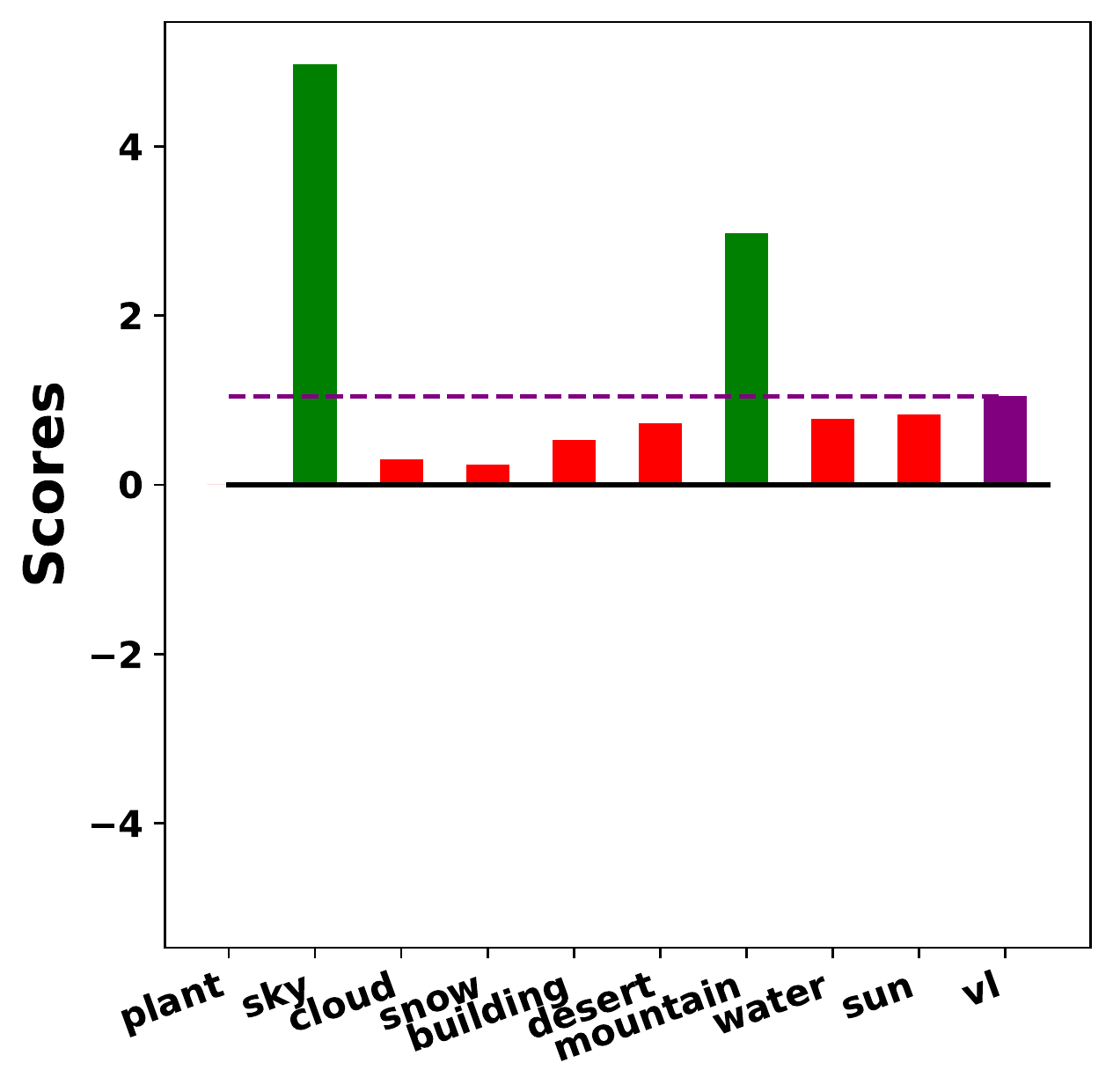} &
    \includegraphics[width=0.22\linewidth]{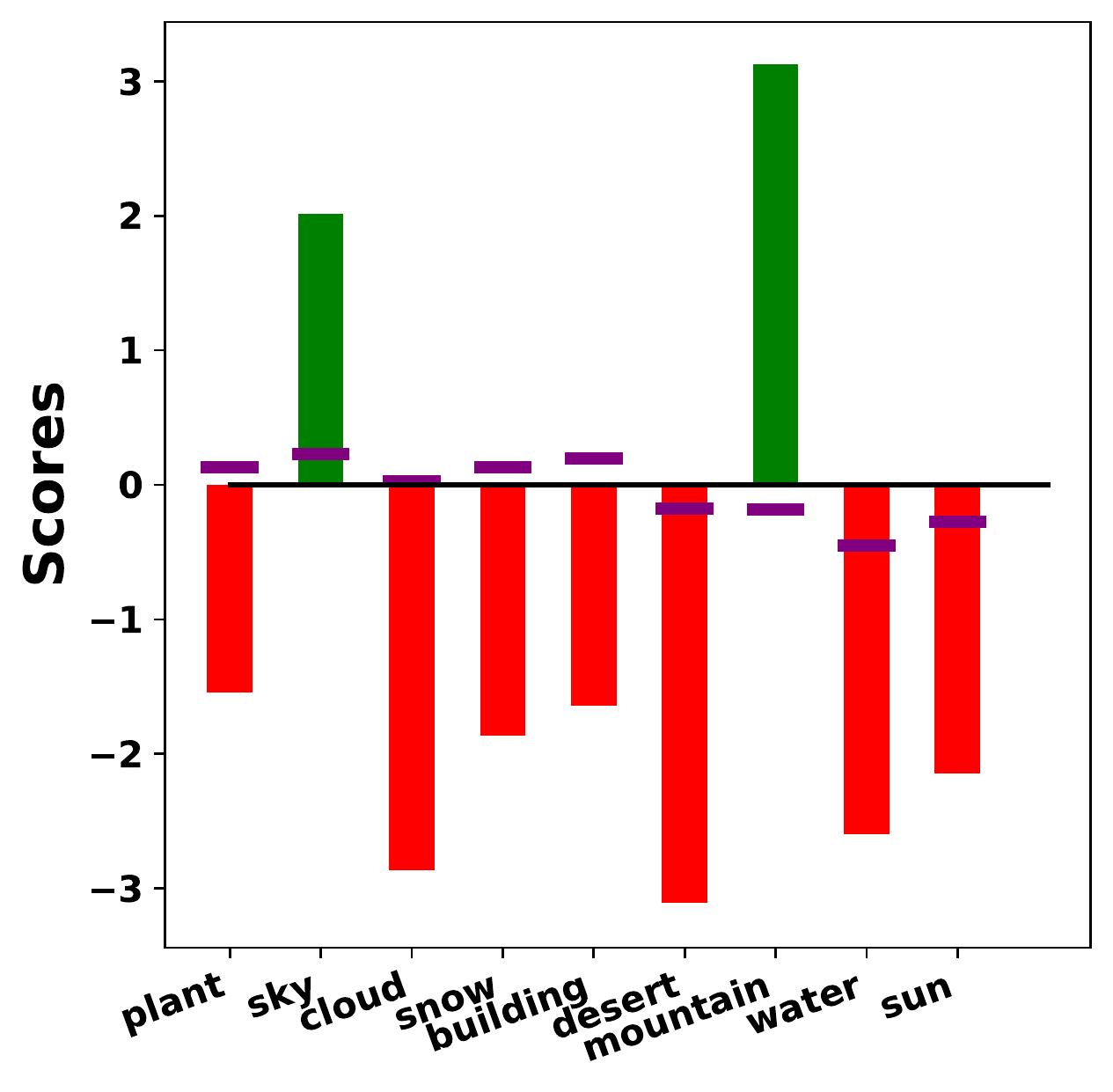} &
    \includegraphics[width=0.22\linewidth]{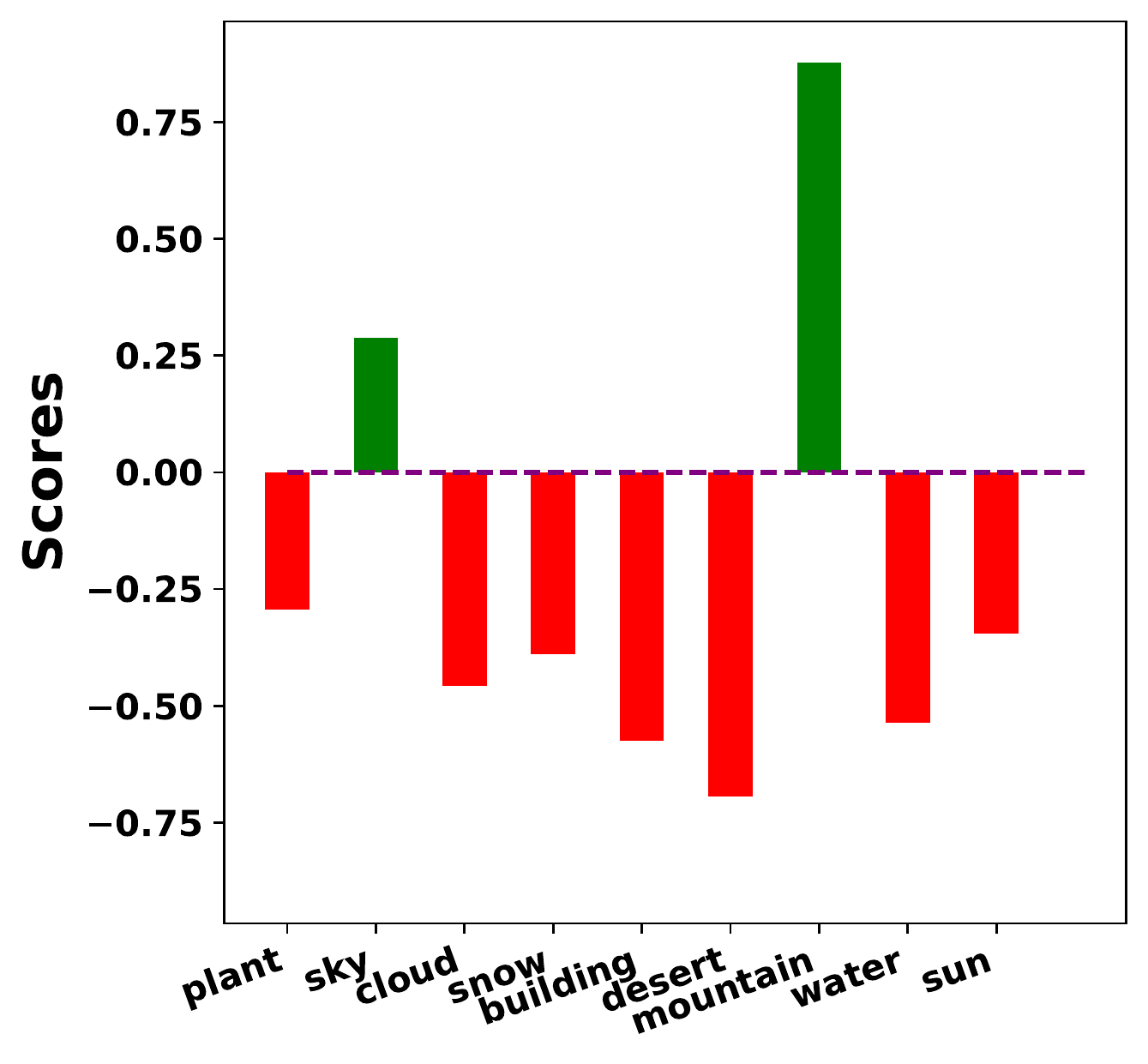} \\
    MRO = \{ mountain $\succ$
    & \{sky $\succ$ mountain\}  &  \{mountain $\succ$ sky\} &  \{mountain $\succ$ sky\} \\
    desert $\succ$ plant $\succ$ sky\} &&& \\
    &&& \\
    \includegraphics[width=3cm, height=3cm]{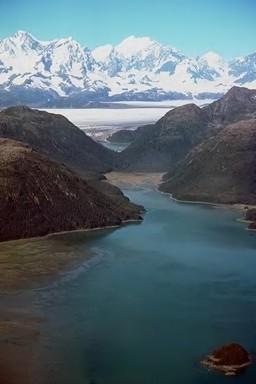} &
    \includegraphics[width=0.22\linewidth]{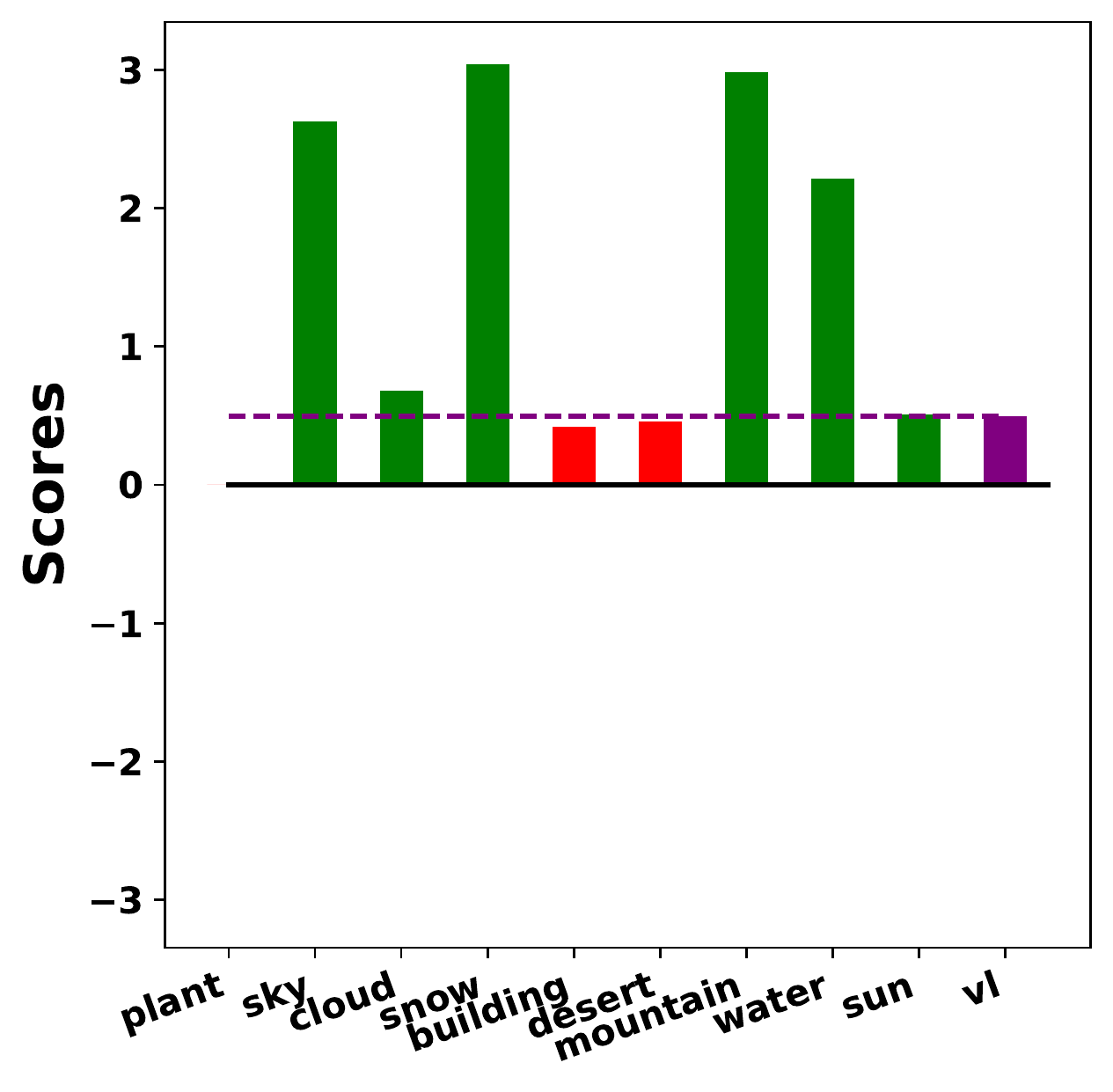} &
    \includegraphics[width=0.22\linewidth]{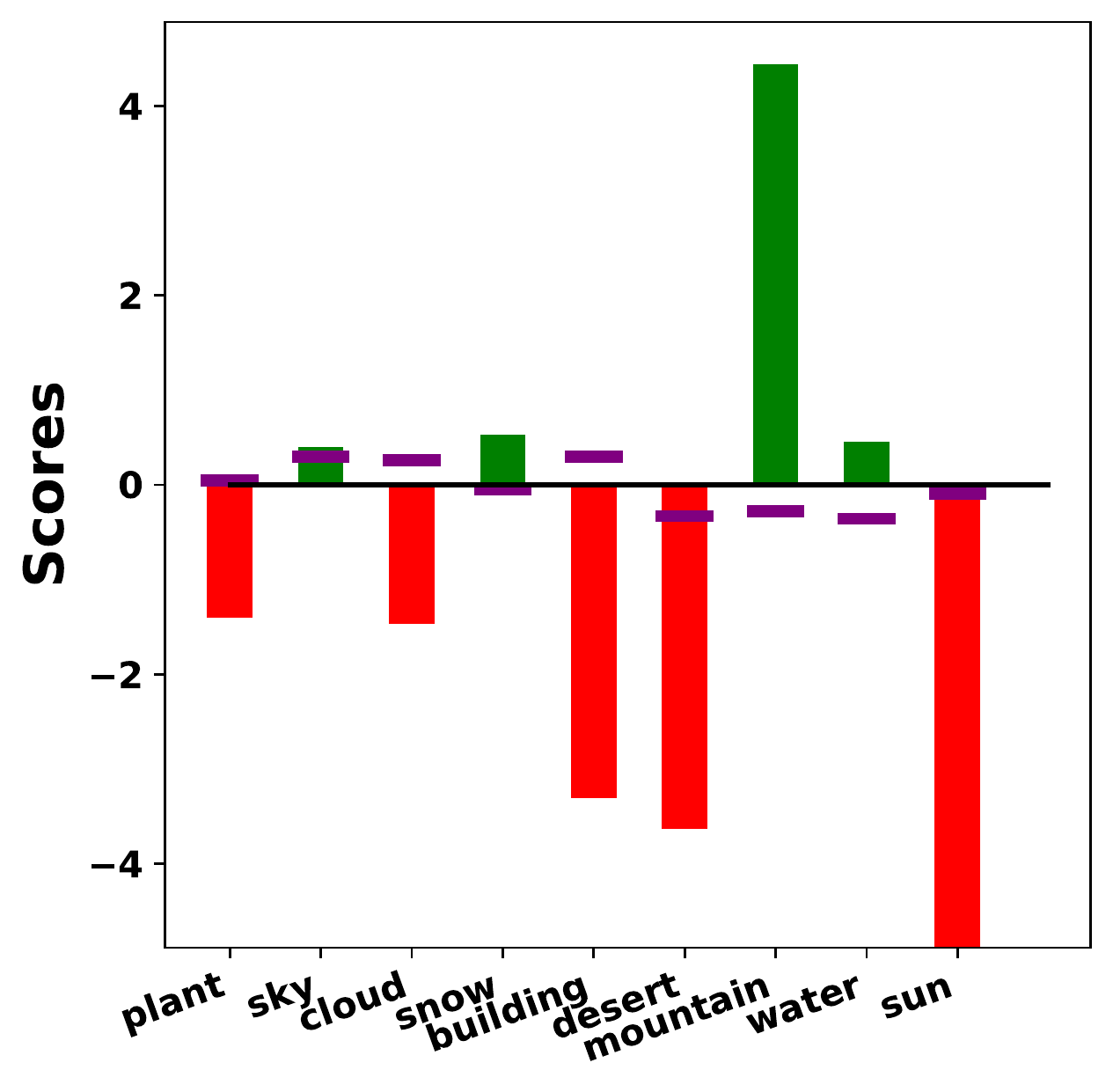} &
    \includegraphics[width=0.22\linewidth]{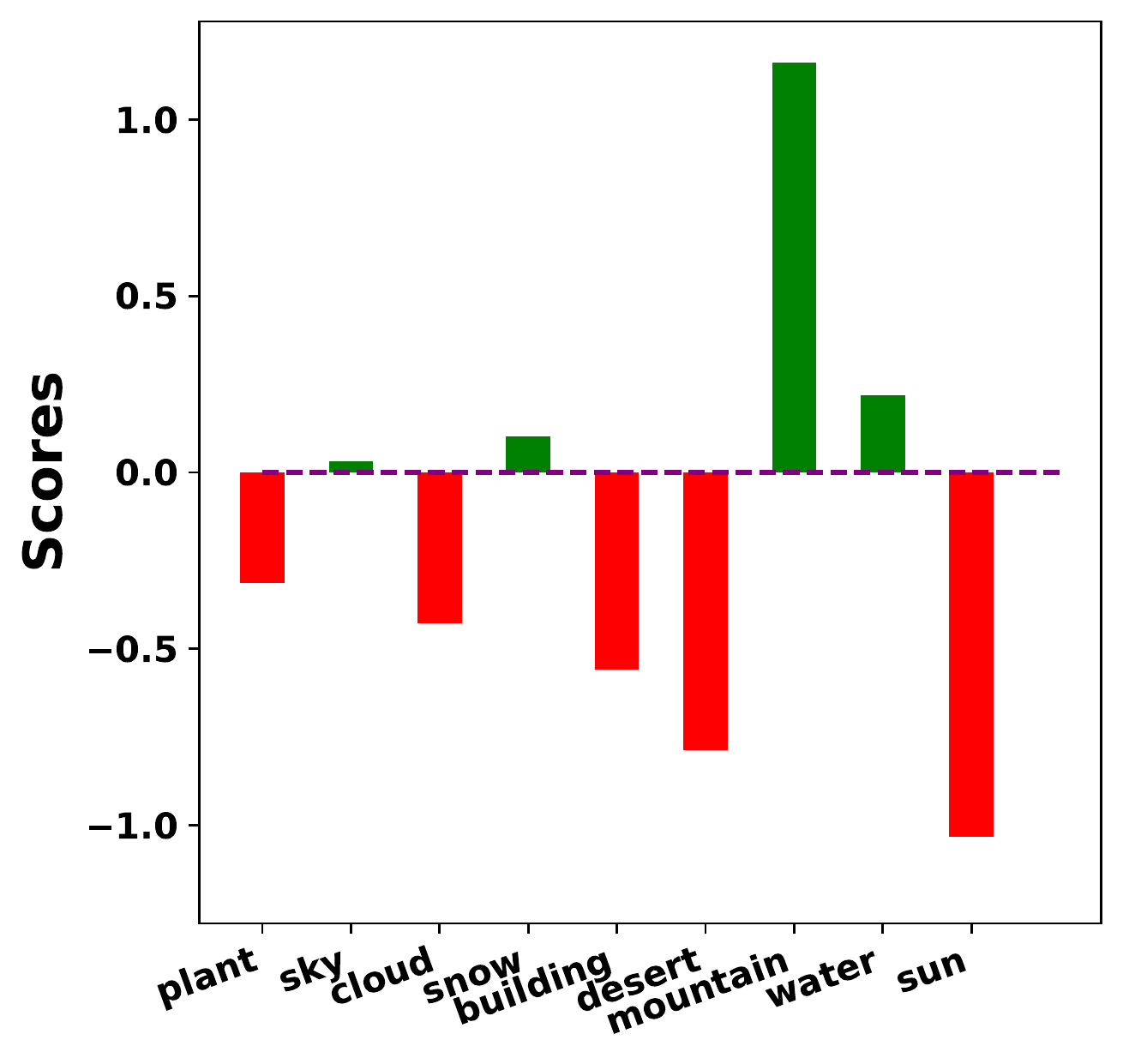} \\
    MRO = \{mountain $\succ$ &  \{snow $\succ$ mountain $\succ$ sky $\succ$ & \{mountain $\succ$ snow $\succ$ &  \{mountain $\succ$ water $\succ$ \\
     water $\succ$ snow\} & water $\succ$ cloud $\succ$ sun\} & water $\succ$ sky\} & snow $\succ$ sky\} \\
    &&& \\
    \includegraphics[width=3cm, height=3cm]{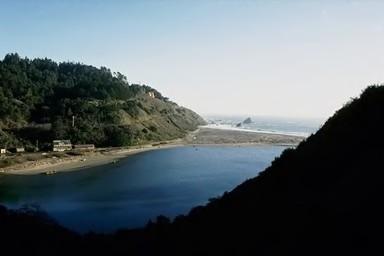}  &
    \includegraphics[width=0.22\linewidth]{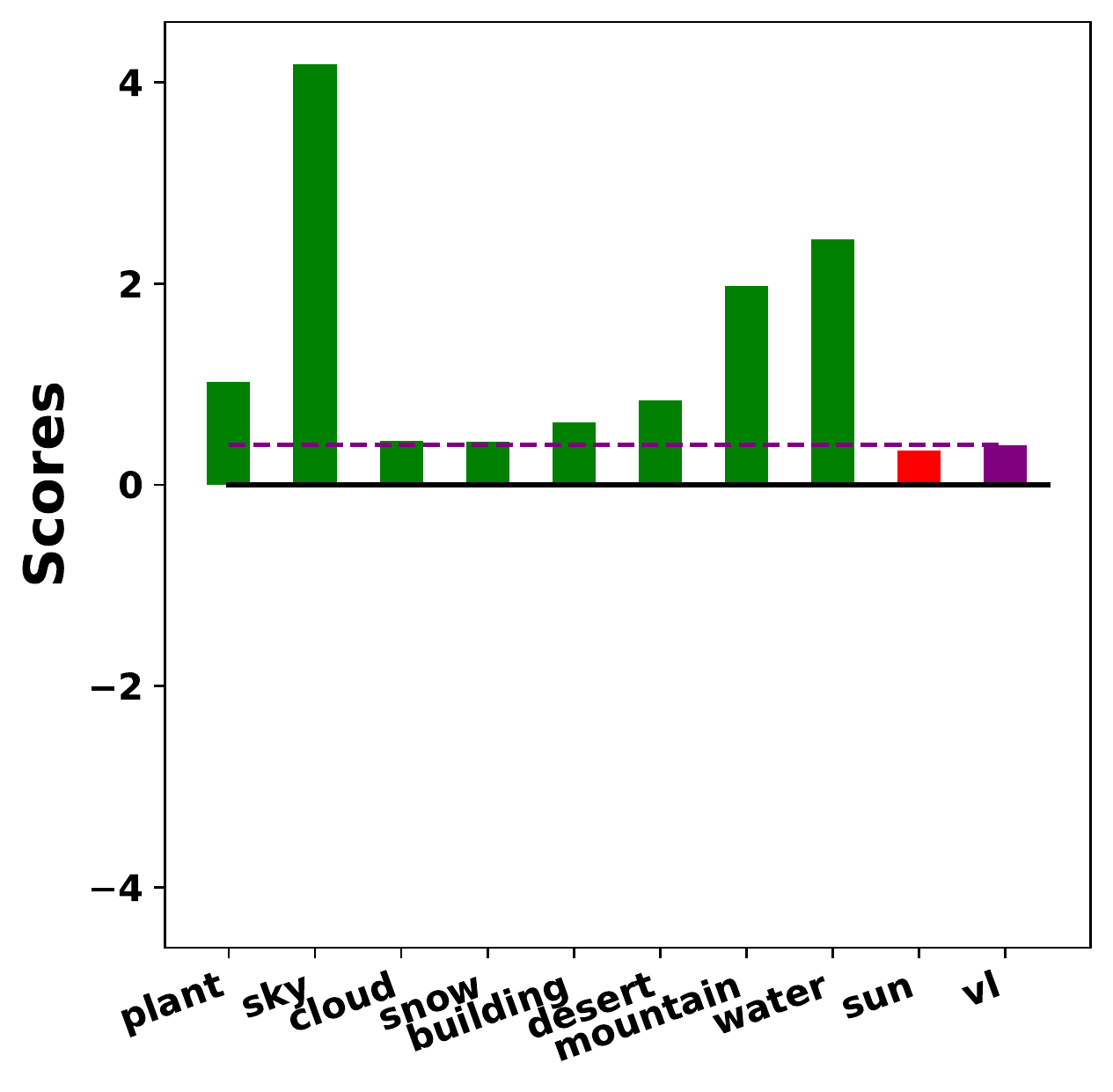} &
    \includegraphics[width=0.22\linewidth]{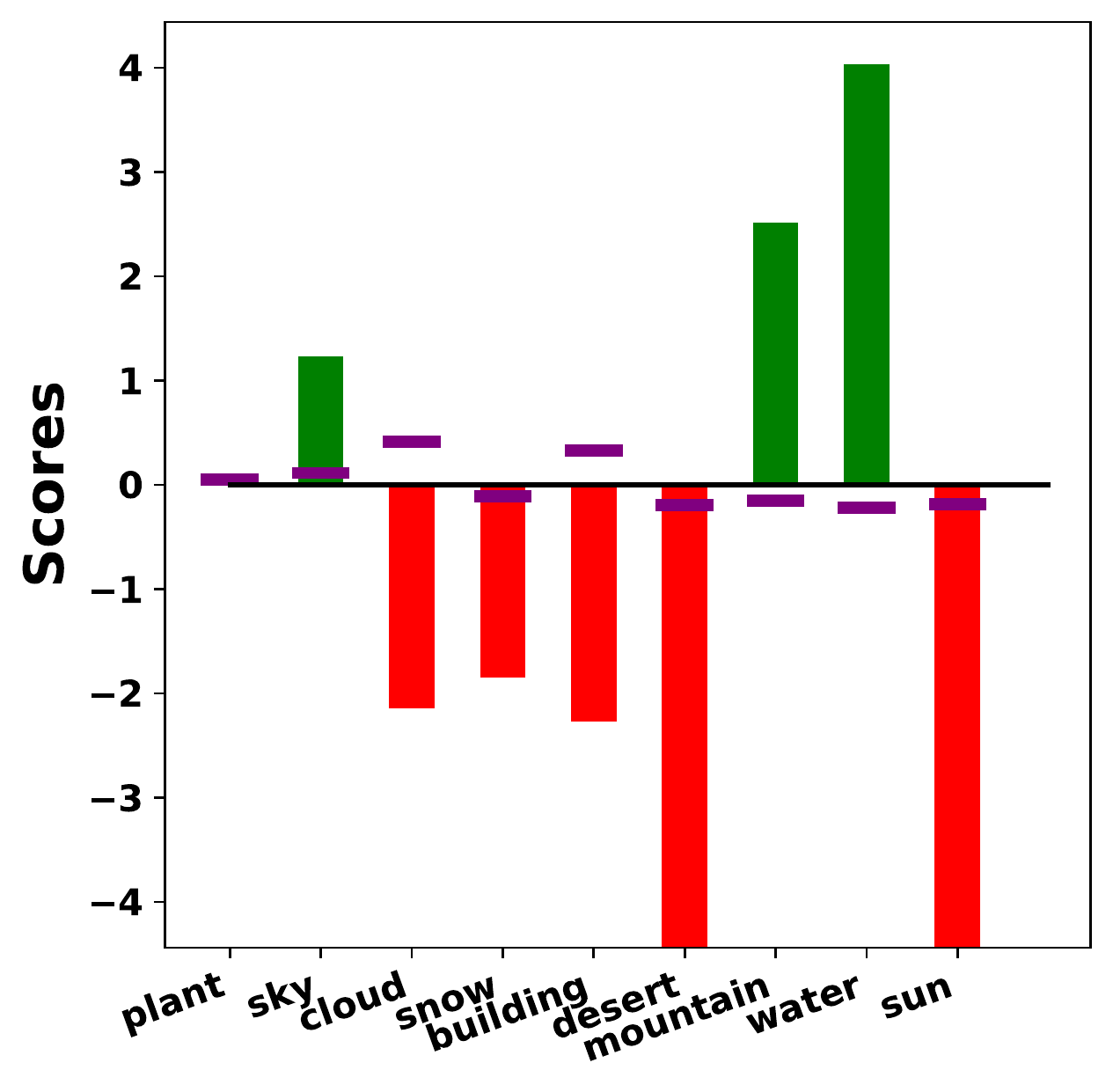} &
    \includegraphics[width=0.22\linewidth]{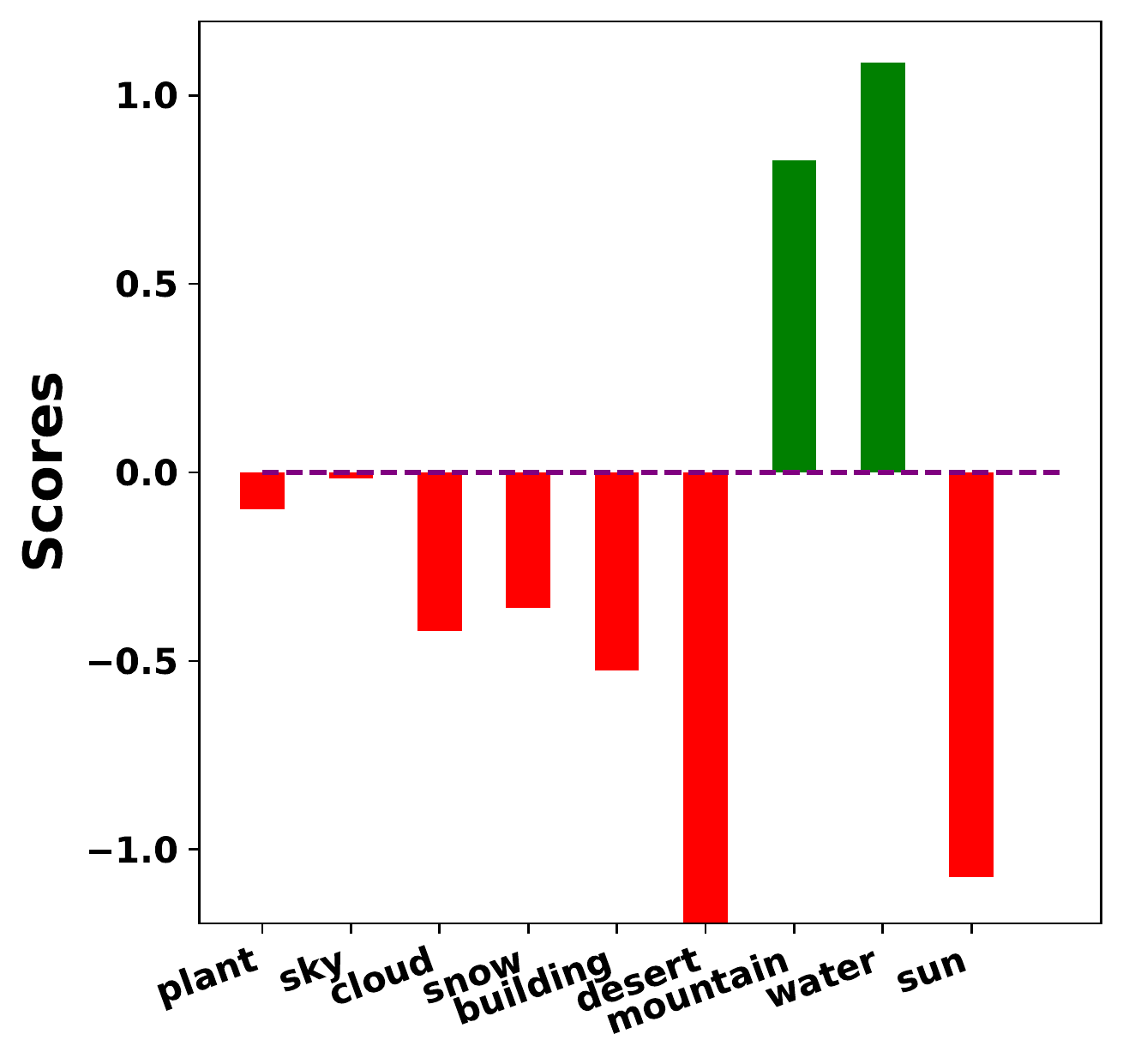} \\
    MRO = \{ mountain $\succ$ & \{sky $\succ$ water $\succ$ mountain $\succ$ & \{water $\succ$ mountain $\succ$ sky\} & \{water $\succ$ mountain \}\\
    water $\succ$ building\} & plant $\succ$ desert $\succ$ building $\succ$ && \\
    & cloud $\succ$ snow \} && \\
    \includegraphics[width=3cm, height=3cm]{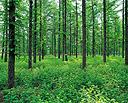}  &
    \includegraphics[width=0.22\linewidth]{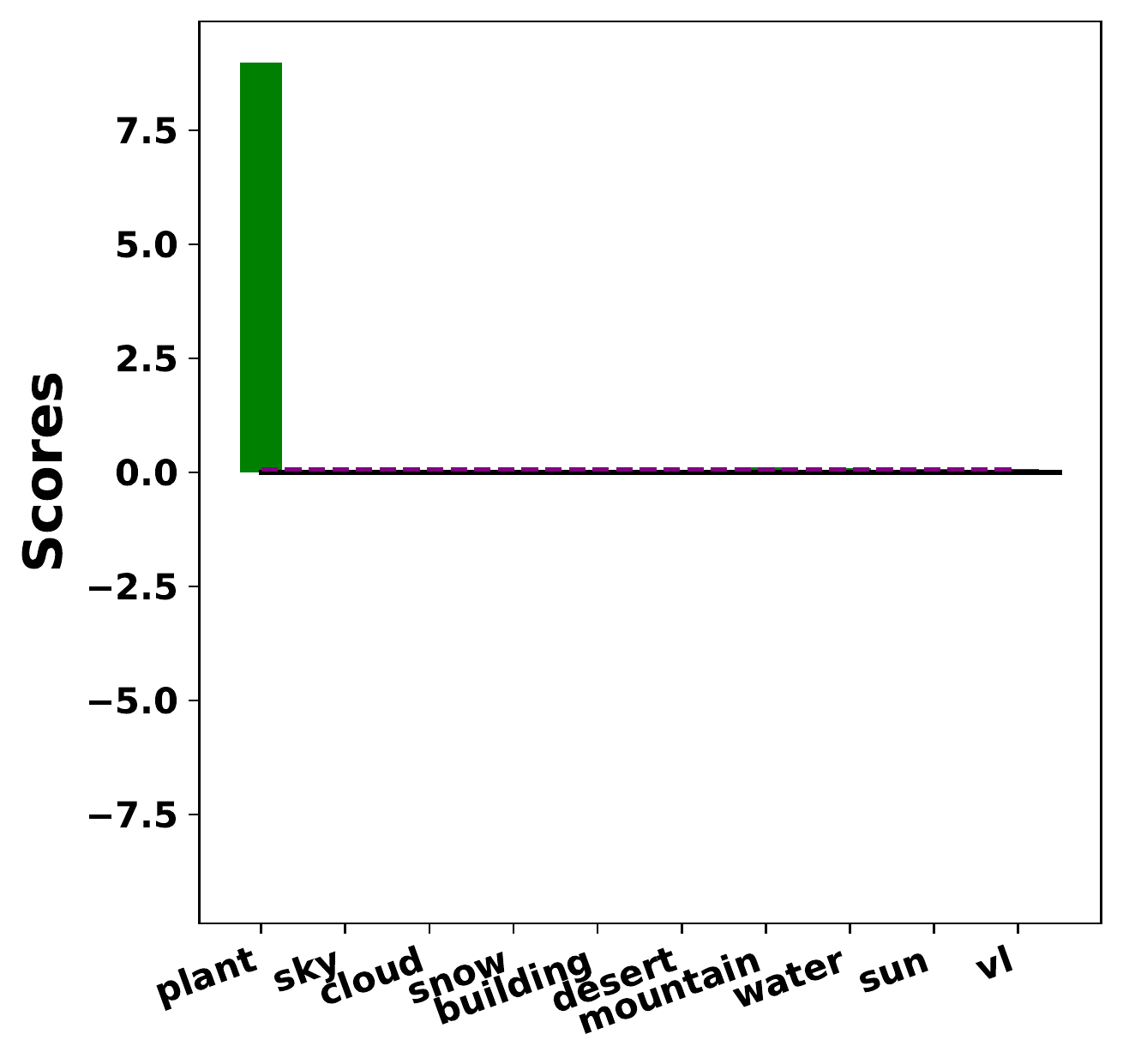} &
    \includegraphics[width=0.22\linewidth]{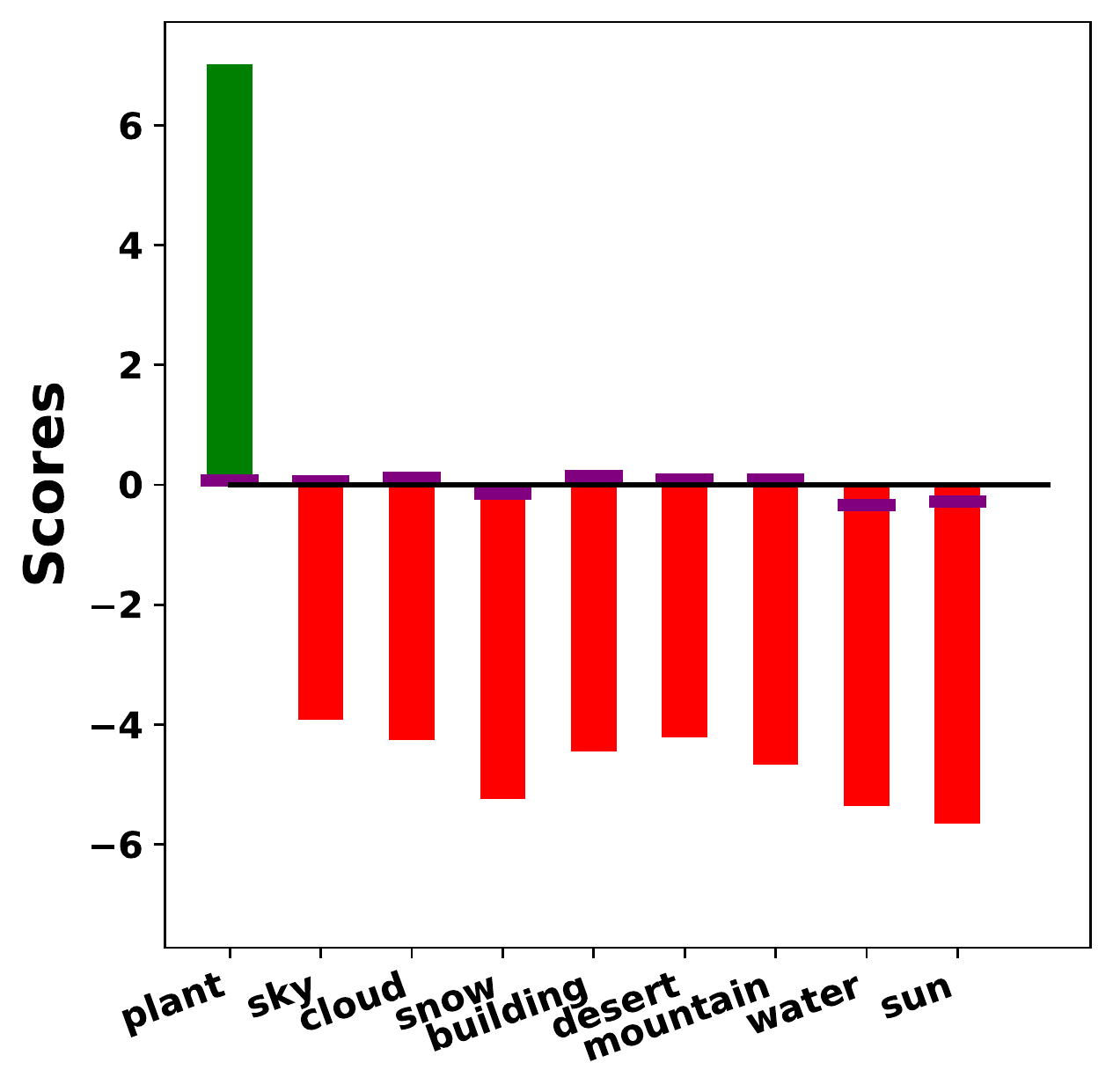} &
    \includegraphics[width=0.22\linewidth]{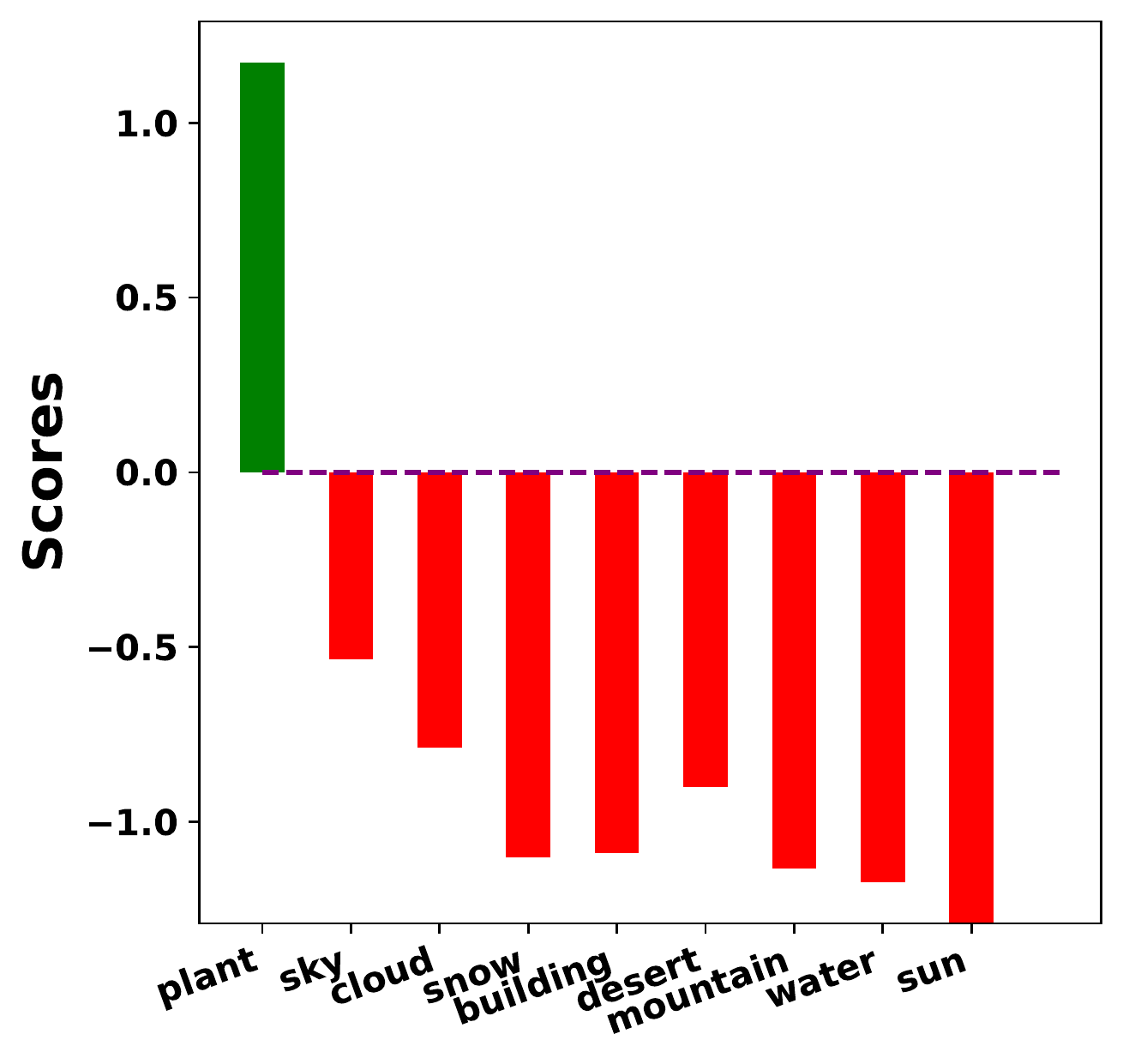} \\
    MRO = \{plant\} & \hspace{0.14in} \{plant\} & \hspace{0.14in}\{plant\} & \hspace{0.14in}\{plant\}\\
    \end{tabularx}
    }
    \caption{Samples from the test set of the NSID are given in the first column from left, with the corresponding aggregated rankings from ten rankers by the mean rank ordering method denoted as MRO. Each bar plot represents the predicted scores of baselines CRPC, LSEP and our method GMLR, in their respective columns.}
    \label{fig:NSID-bar}
\end{figure}

\begin{figure}[H]
    \centering
    \begin{tabularx}{\textwidth}{cccc}
    Ranked MNIST Gray-S  & \hspace{0.14in}CRPC & \hspace{0.14in}LSEP & \hspace{0.14in}GMLR \\
    \includegraphics[width=3cm, height=3cm]{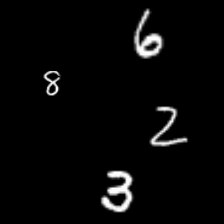}  &
    \includegraphics[width=0.22\linewidth]{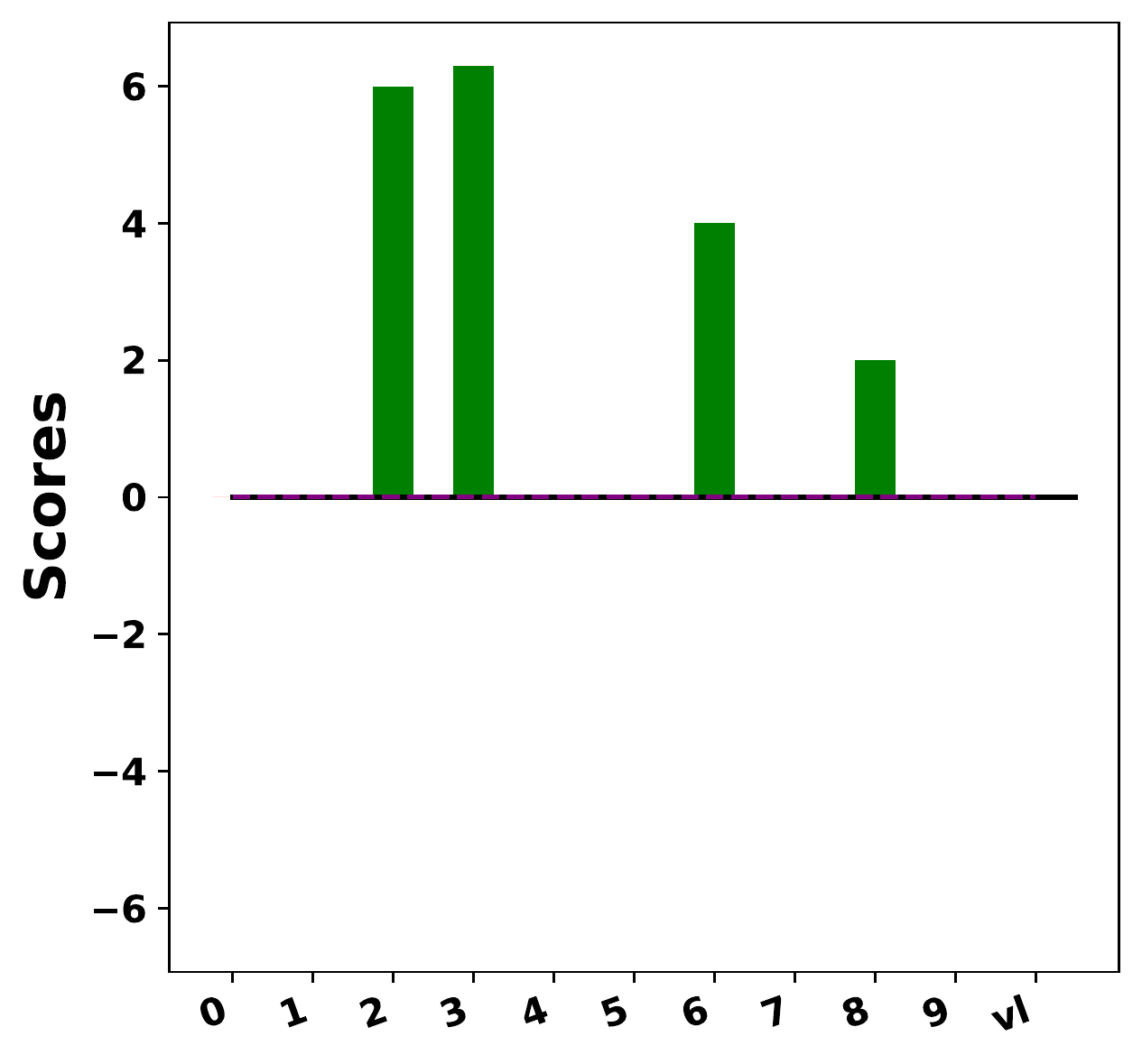} &
    \includegraphics[width=0.22\linewidth]{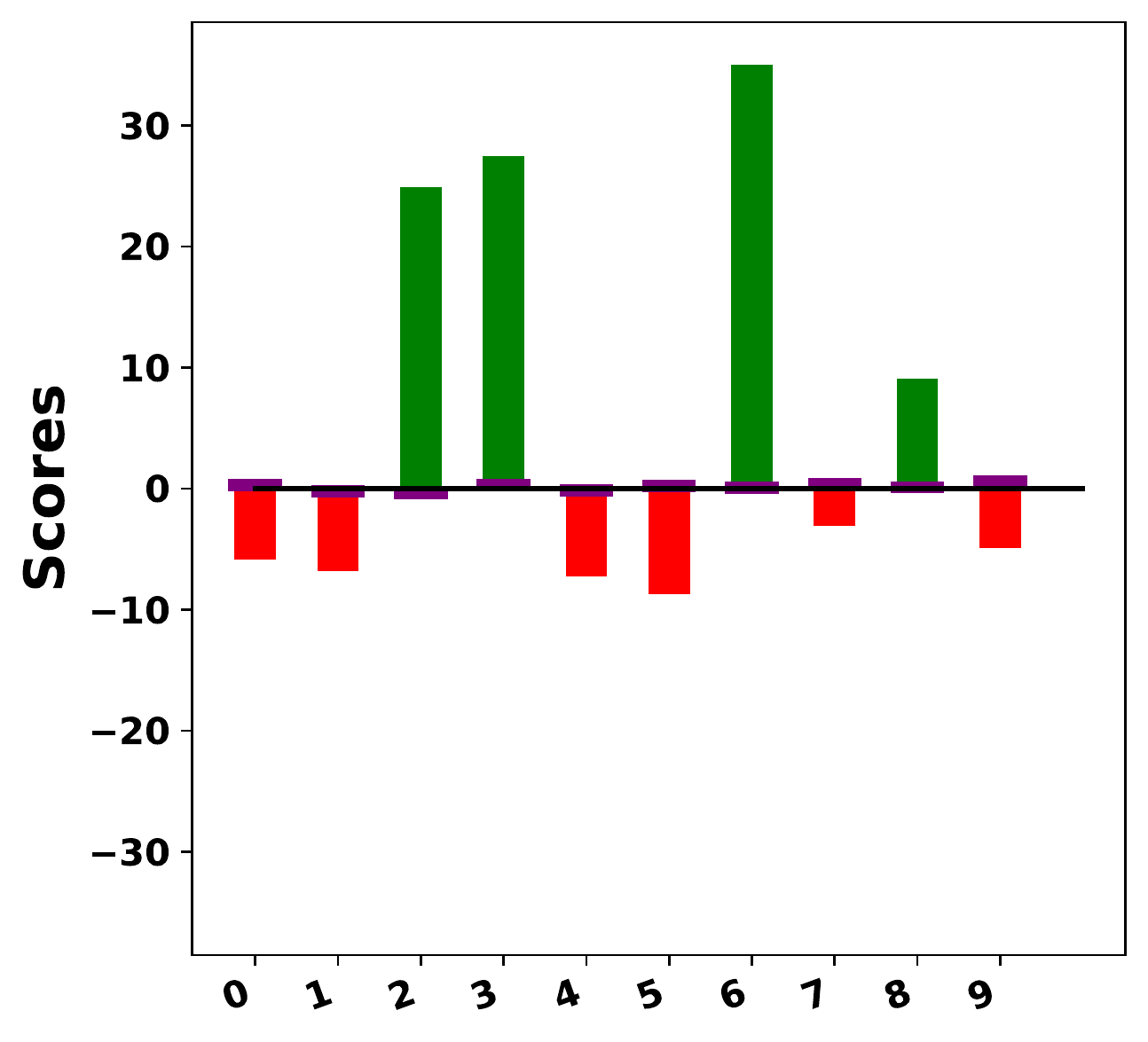} &
    \includegraphics[width=0.22\linewidth]{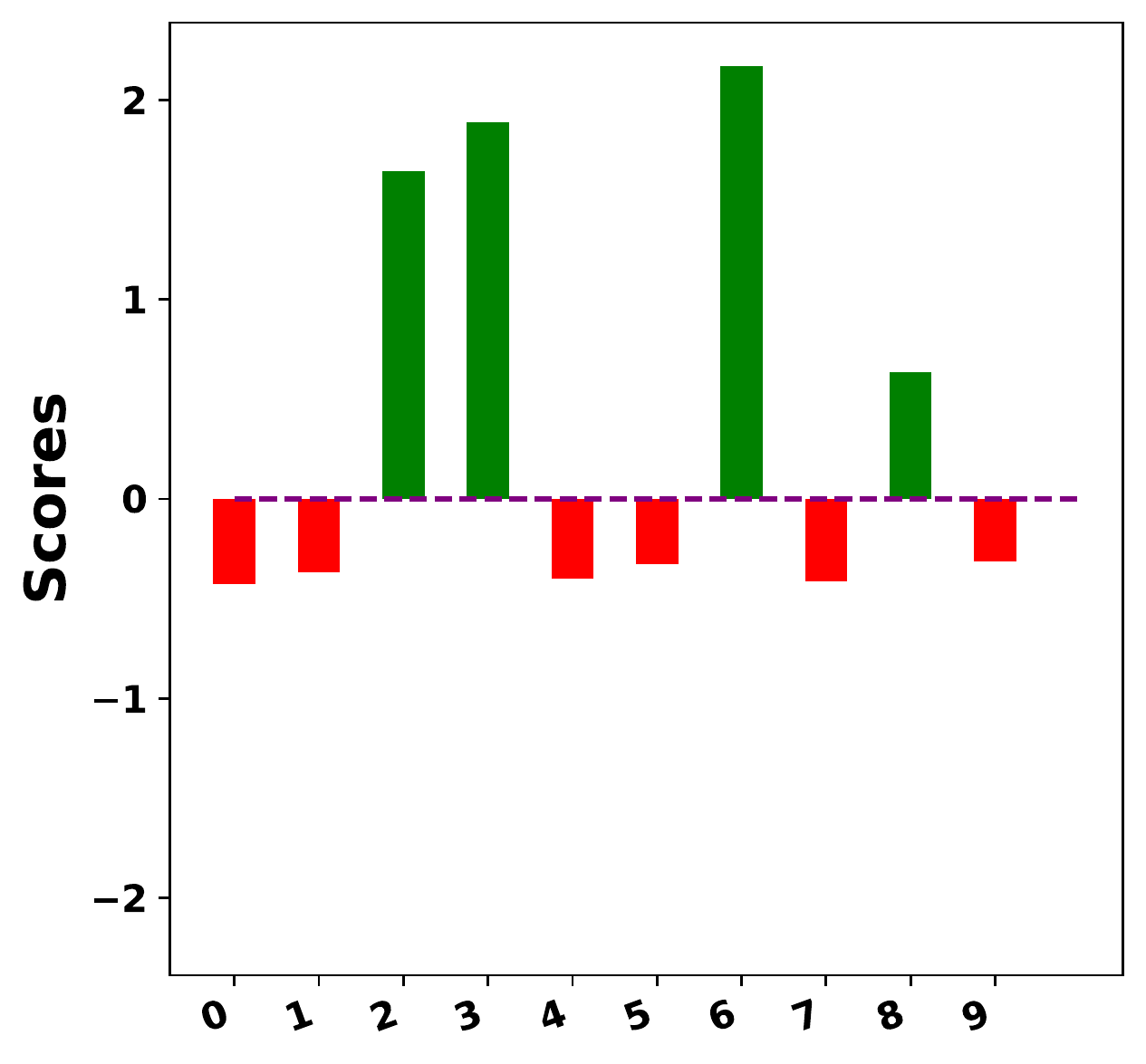} \\
    GT = \{ 6 $\succ$ 3 $\succ$ 2 $\succ$ 8\}& \hspace{0.12in} \{3 $\succ$ 2 $\succ$ 6 $\succ$ 8\} & \hspace{0.12in} \{6 $\succ$ 3 $\succ$ 2 $\succ$ 8\}  & \hspace{0.12in}  \{6 $\succ$ 3 $\succ$ 2 $\succ$ 8 \}  \\
    &&& \\
    \includegraphics[width=3cm, height=3cm]{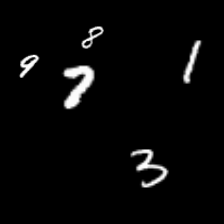}  &
    \includegraphics[width=0.22\linewidth]{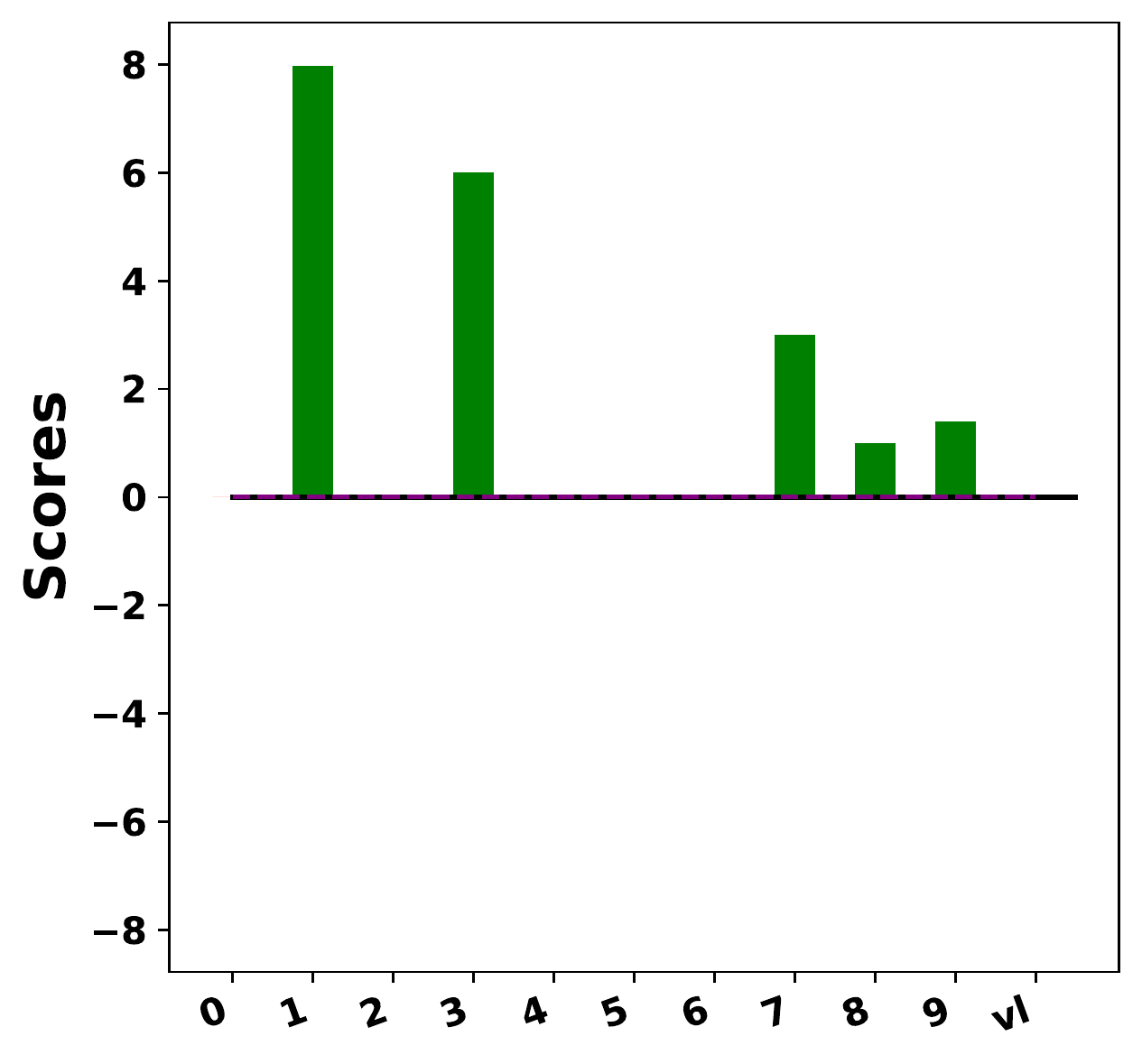} &
    \includegraphics[width=0.22\linewidth]{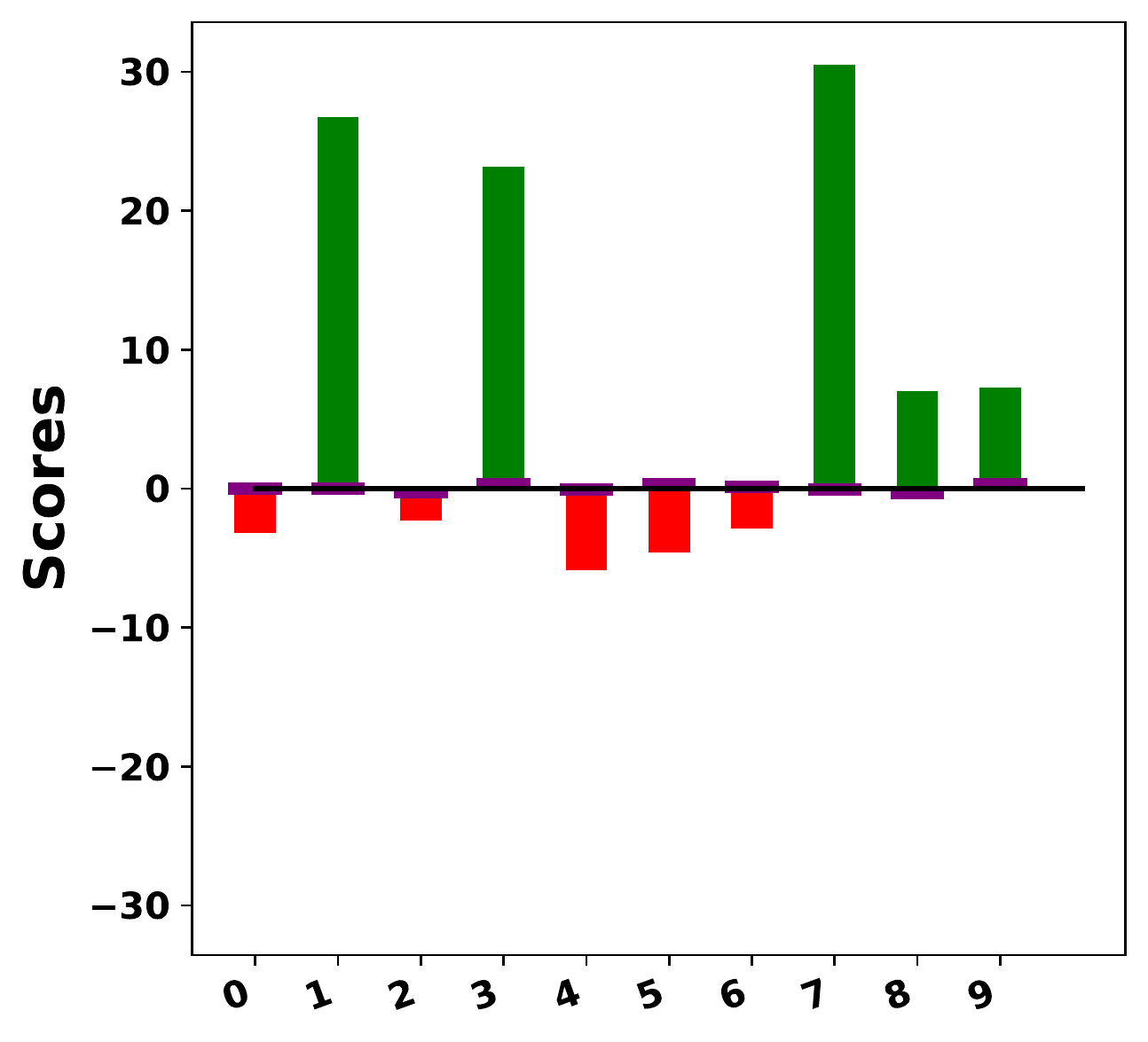} &
    \includegraphics[width=0.22\linewidth]{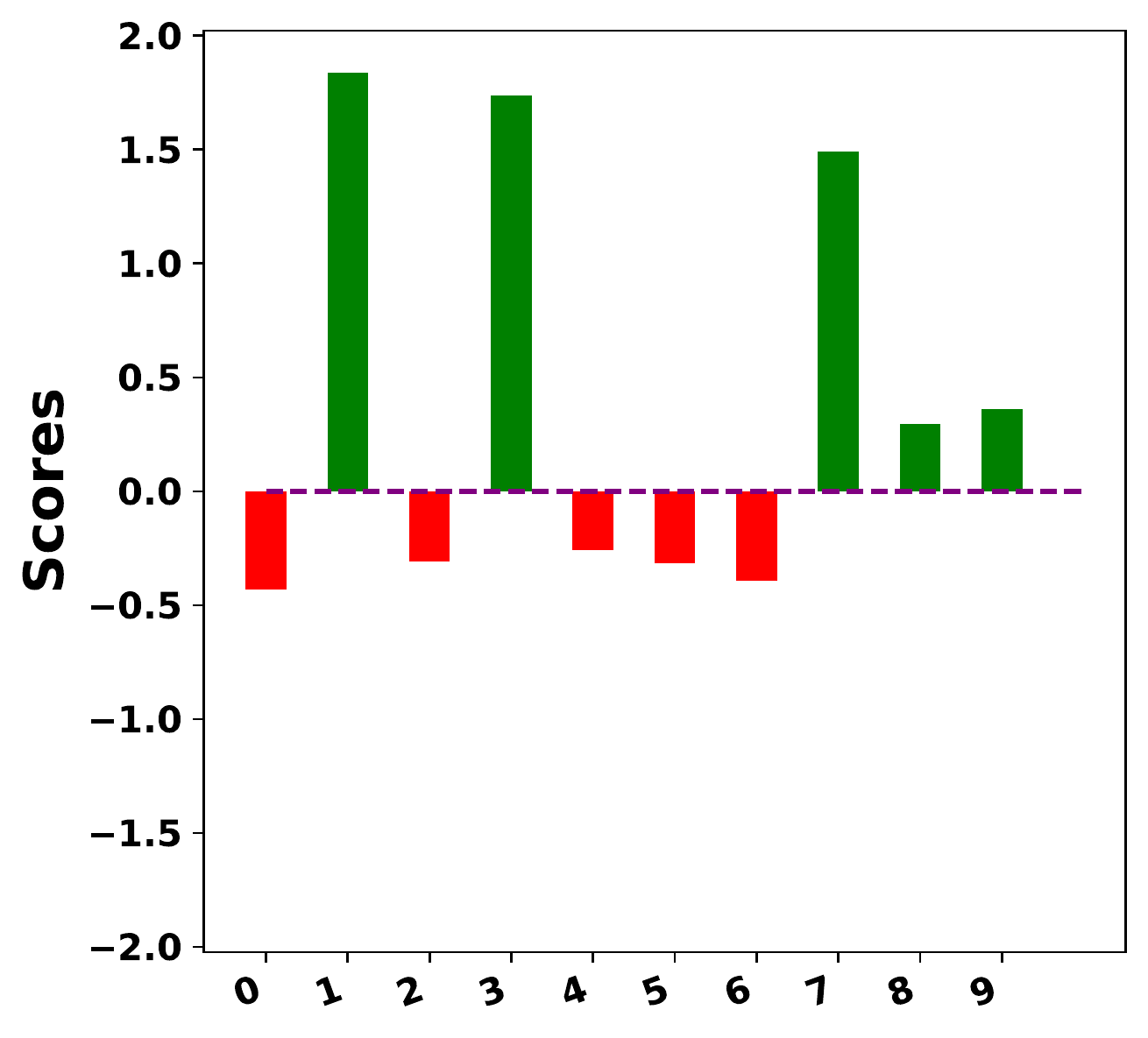} \\
    GT = \{ 7 $\succ$ 1 $\succ$ 3 $\succ$ 8 \}&  \hspace{0.12in}\{1 $\succ$ 3 $\succ$ 7 $\succ$ 9 $\succ$ 8 \} & \hspace{0.12in} \{7 $\succ$ 1 $\succ$ 3 $\succ$ 9 $\succ$ 8\} &  \hspace{0.12in}\{1 $\succ$ 3 $\succ$ 7 $\succ$ 9 $\succ$ 8 \}\\
     &   && \\
    &&& \\
    \includegraphics[width=3cm, height=3cm]{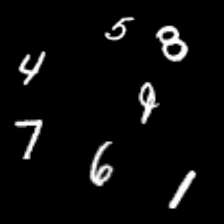} &
    \includegraphics[width=0.22\linewidth]{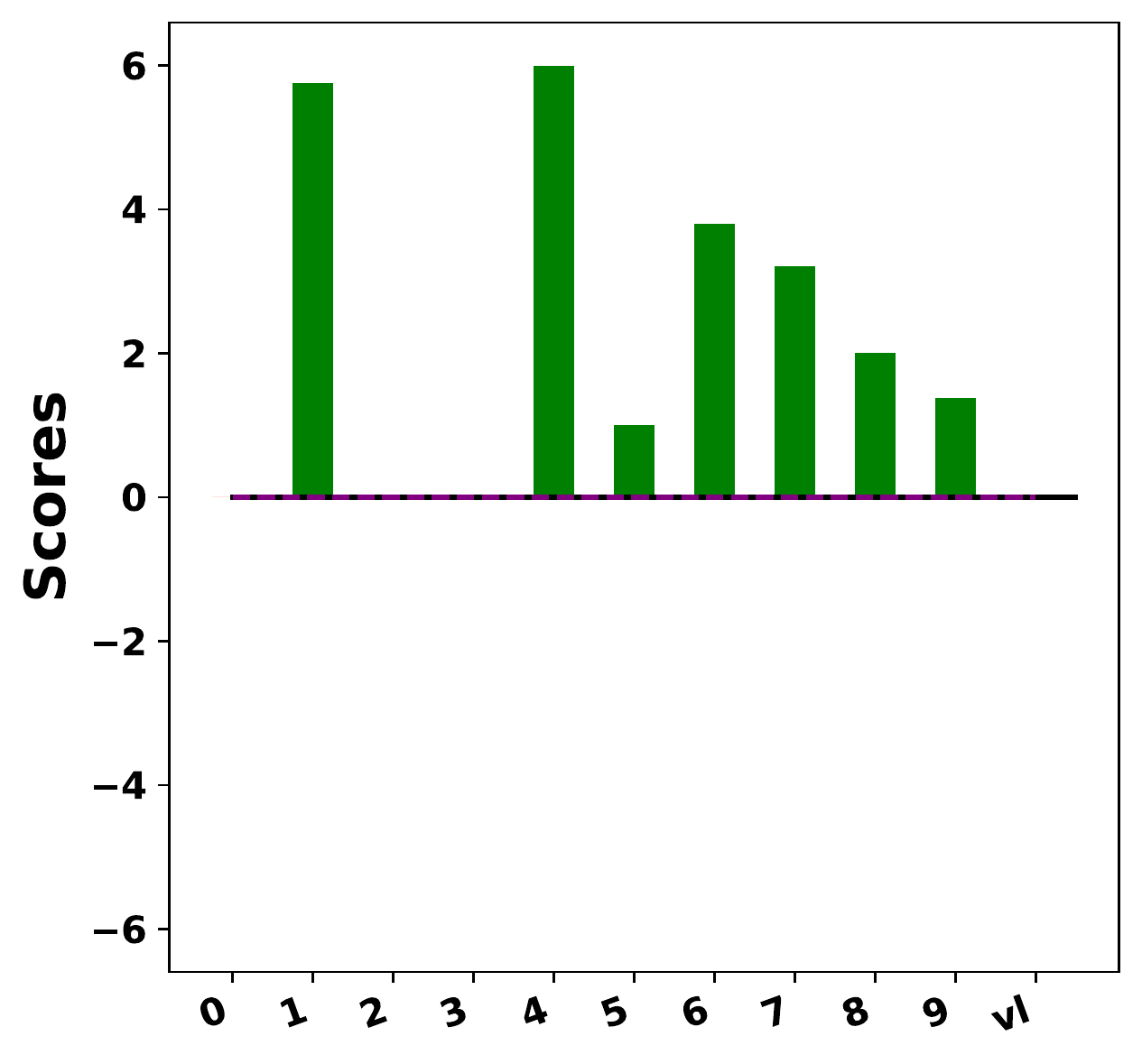} &
    \includegraphics[width=0.22\linewidth]{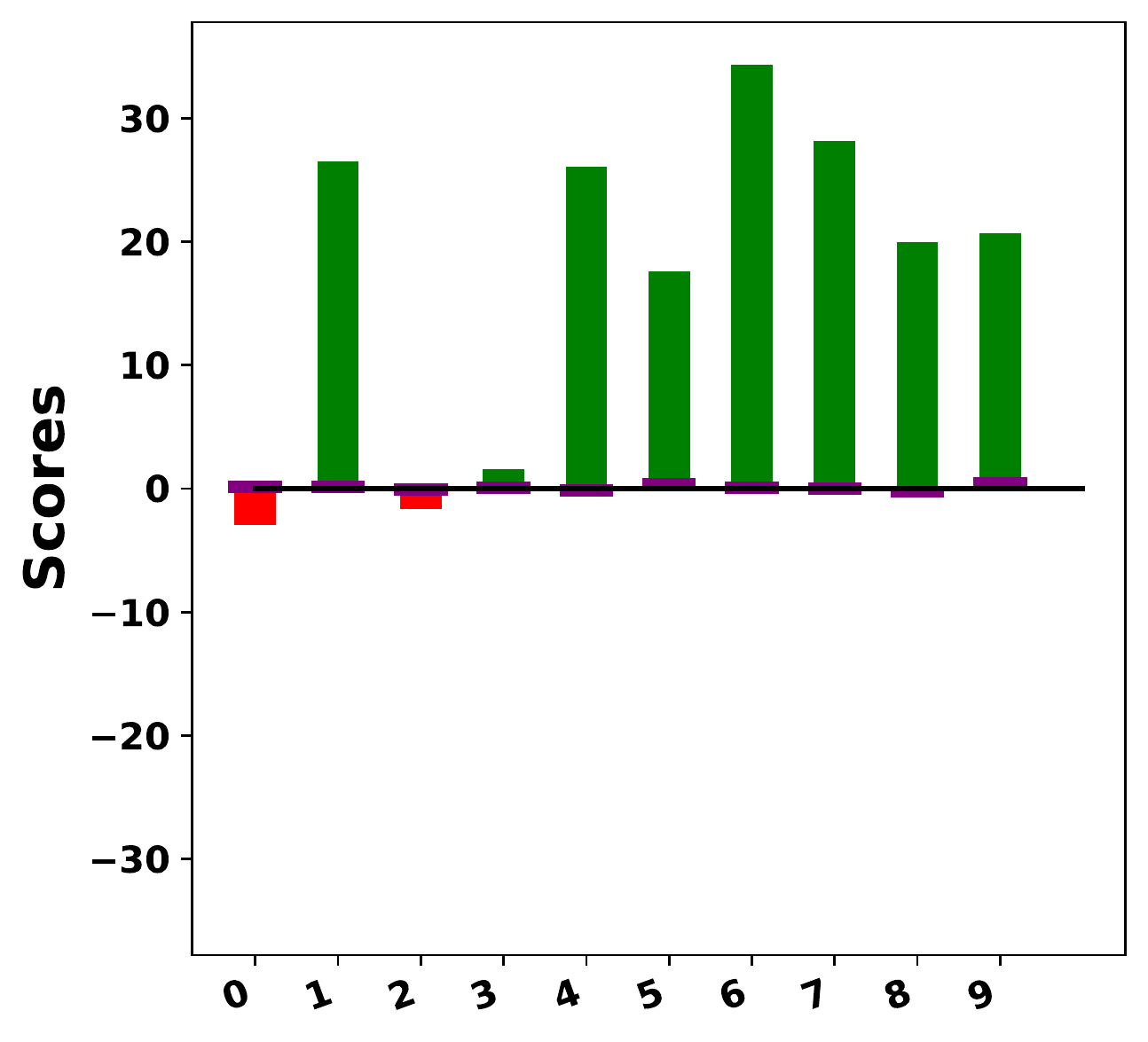} &
    \includegraphics[width=0.22\linewidth]{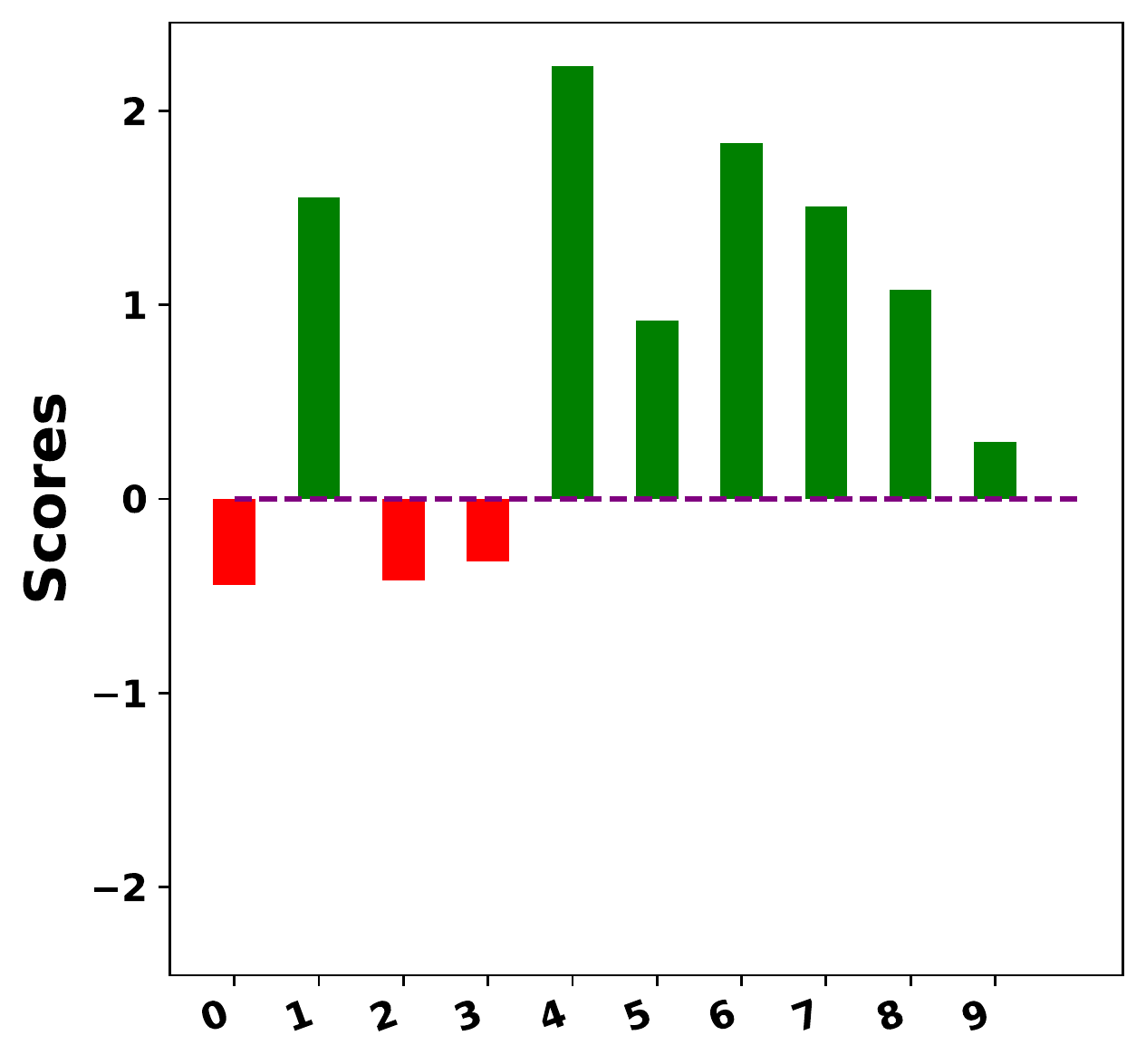} \\
    GT = \{ 6 $\succ$ 9 $\succ$ 7 $\succ$  1 $\succ$ & \hspace{0.12in} \{4 $\succ$ 1 $\succ$ 6 $\succ$ 7 $\succ$& \hspace{0.12in} \{ 6 $\succ$ 7 $\succ$ 1 $\succ$ 4 $\succ$ & \hspace{0.12in} \{ 4 $\succ$ 6 $\succ$ 7 $\succ$ 1 $\succ$ \\
    8 $\succ$ 4 $\succ$ 5 \}&  8 $\succ$ 9 $\succ$ 5 \}  &  9 $\succ$ 8 $\succ$ 5 \}& 8 $\succ$ 5 $\succ$ 9 \}\\
    &&& \\
    \end{tabularx}
    \caption{Samples from the test set of the Ranked MNIST Gray dataset are given in the first column from left, with the corresponding ground truth labels denoted as GT. Each bar plot represents the predicted scores of baselines CRPC, LSEP and our method GMLR, in their respective columns.}
    \label{fig:gray-bar}
\end{figure}

\begin{figure}[H]
    \centering
    \begin{tabularx}{\textwidth}{cccc}
    Ranked MNIST Color-S  & CRPC & LSEP & GMLR \\
    \includegraphics[width=3cm, height=3cm]{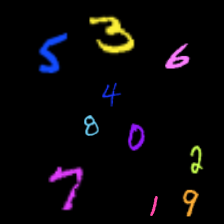}  &
    \includegraphics[width=0.22\linewidth]{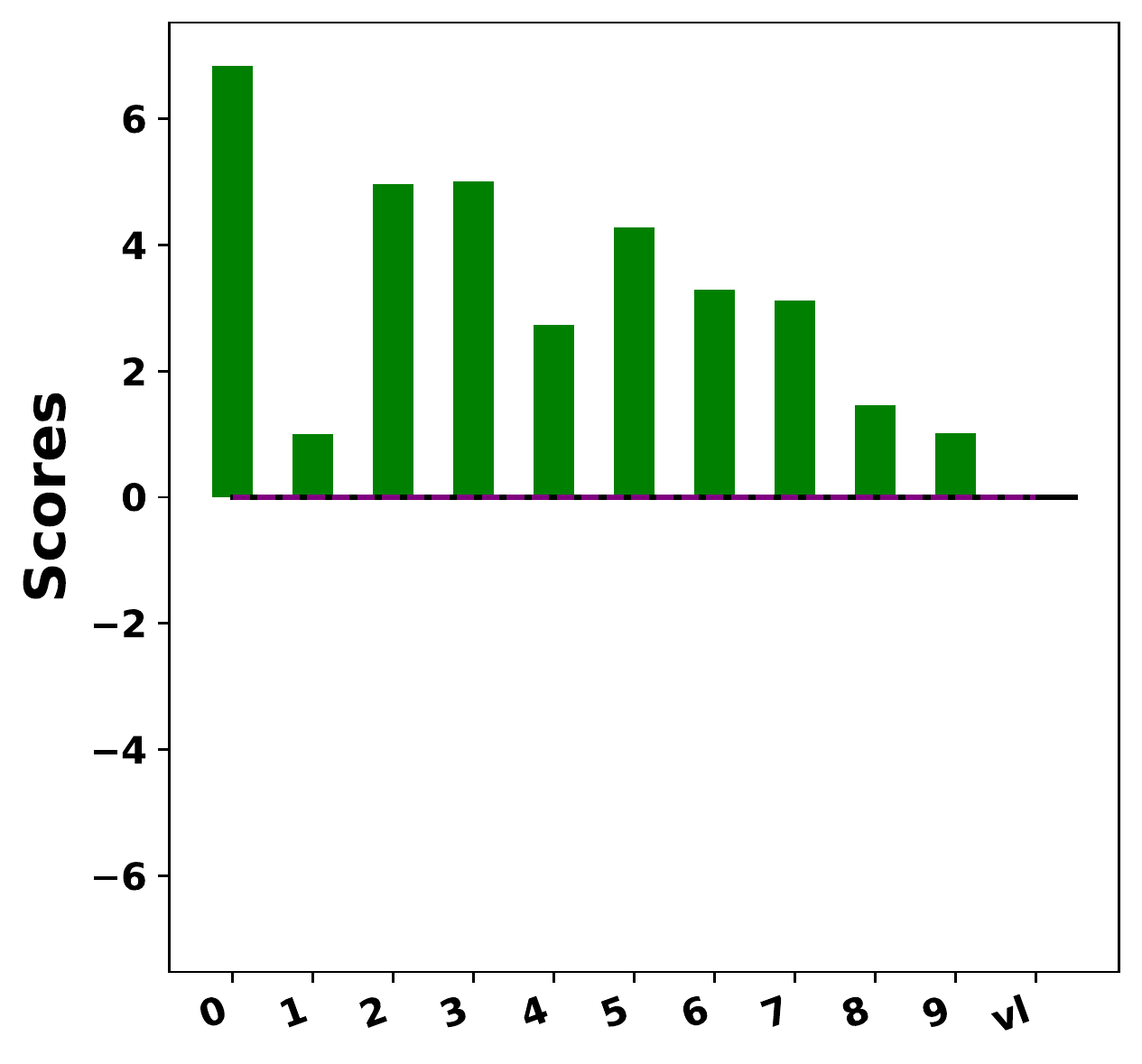} &
    \includegraphics[width=0.22\linewidth]{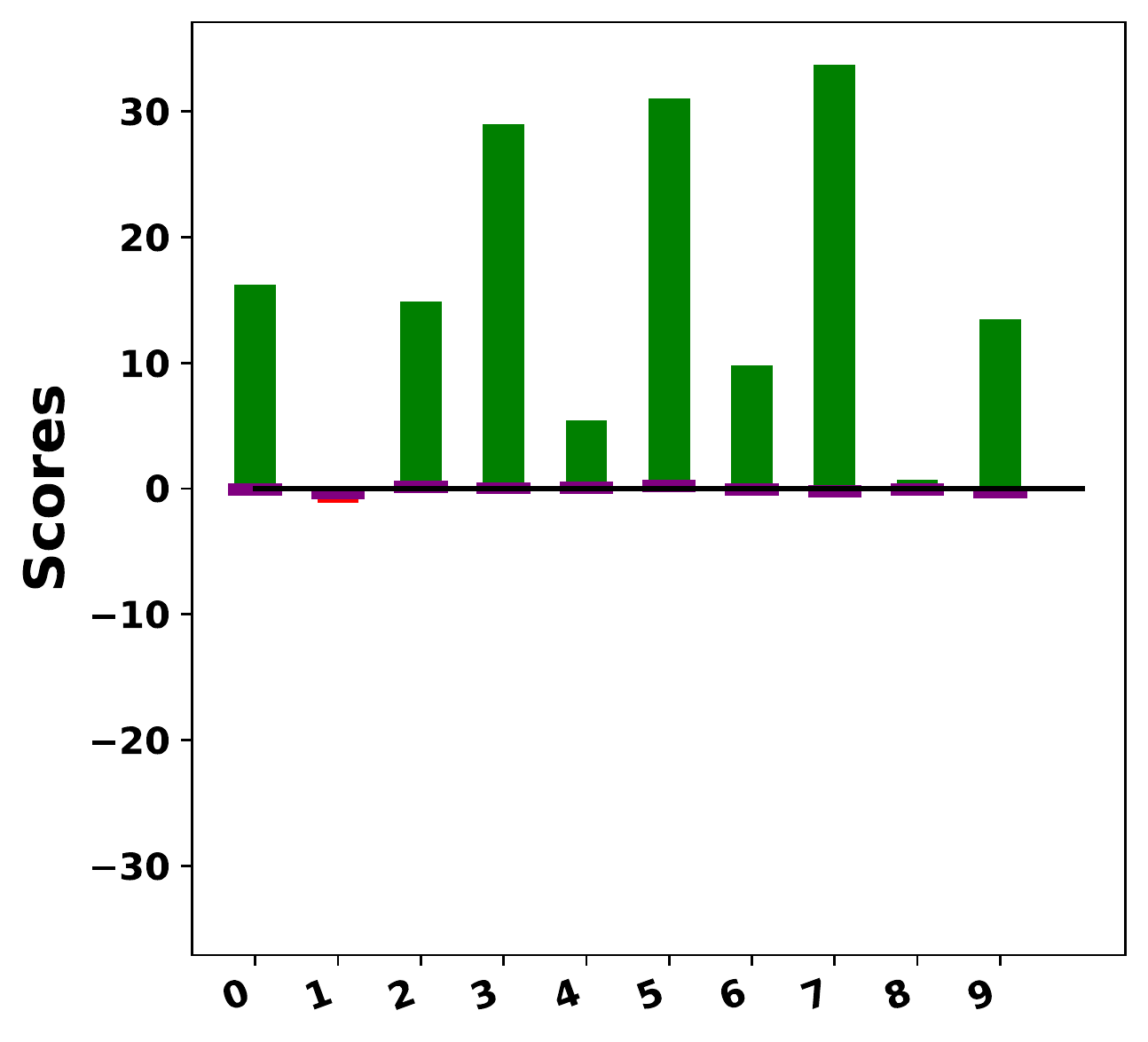} &
    \includegraphics[width=0.22\linewidth]{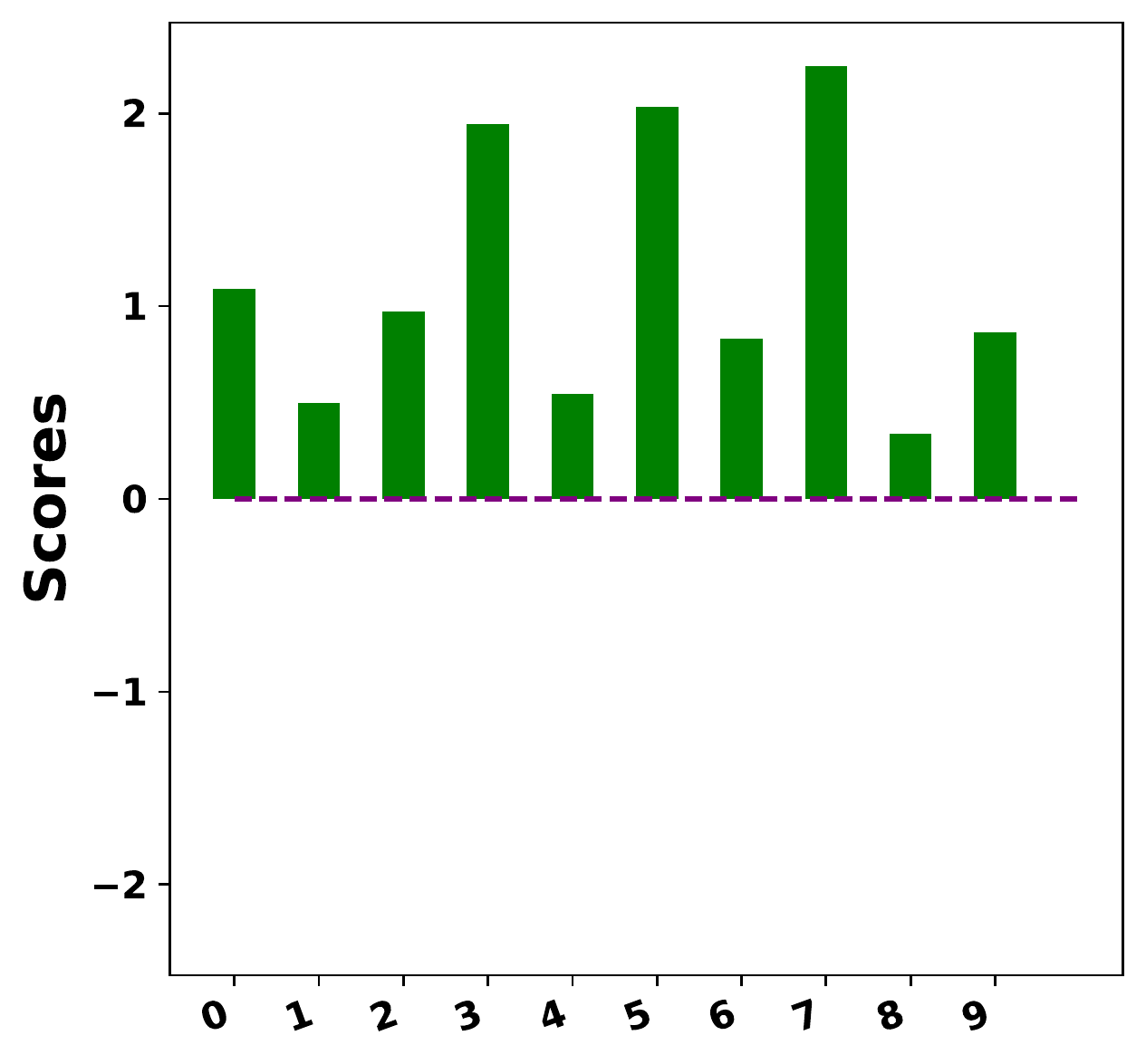} \\
    GT = \{ 3 $\succ$ 7 $\succ$ 5 $\succ$ 2 $\succ$ 0 $\succ$ & \hspace{0.14in}\{ 0 $\succ$ 3 $\succ$ 2 $\succ$ 5 $\succ$ 6 $\succ$ & \{ 7 $\succ$ 5 $\succ$ 3 $\succ$ 0 $\succ$ &  
    \{ 7 $\succ$ 5 $\succ$ 3 $\succ$ 0 $\succ$ 2 $\succ$ \\
    
    9 $\succ$ 6 $\succ$ 4 $\succ$ 8 $\succ$ 1 \} & 7 $\succ$ 4 $\succ$ 8 $\succ$ 9 $\succ$ 1 \} & 2 $\succ$ 9 $\succ$ 6 $\succ$ 4 \} & 9 $\succ$ 6 $\succ$ 4 $\succ$ 1 $\succ$ 8 \}\\
    &&& \\
    \includegraphics[width=3cm, height=3cm]{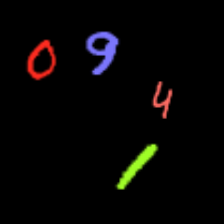}  &
    \includegraphics[width=0.22\linewidth]{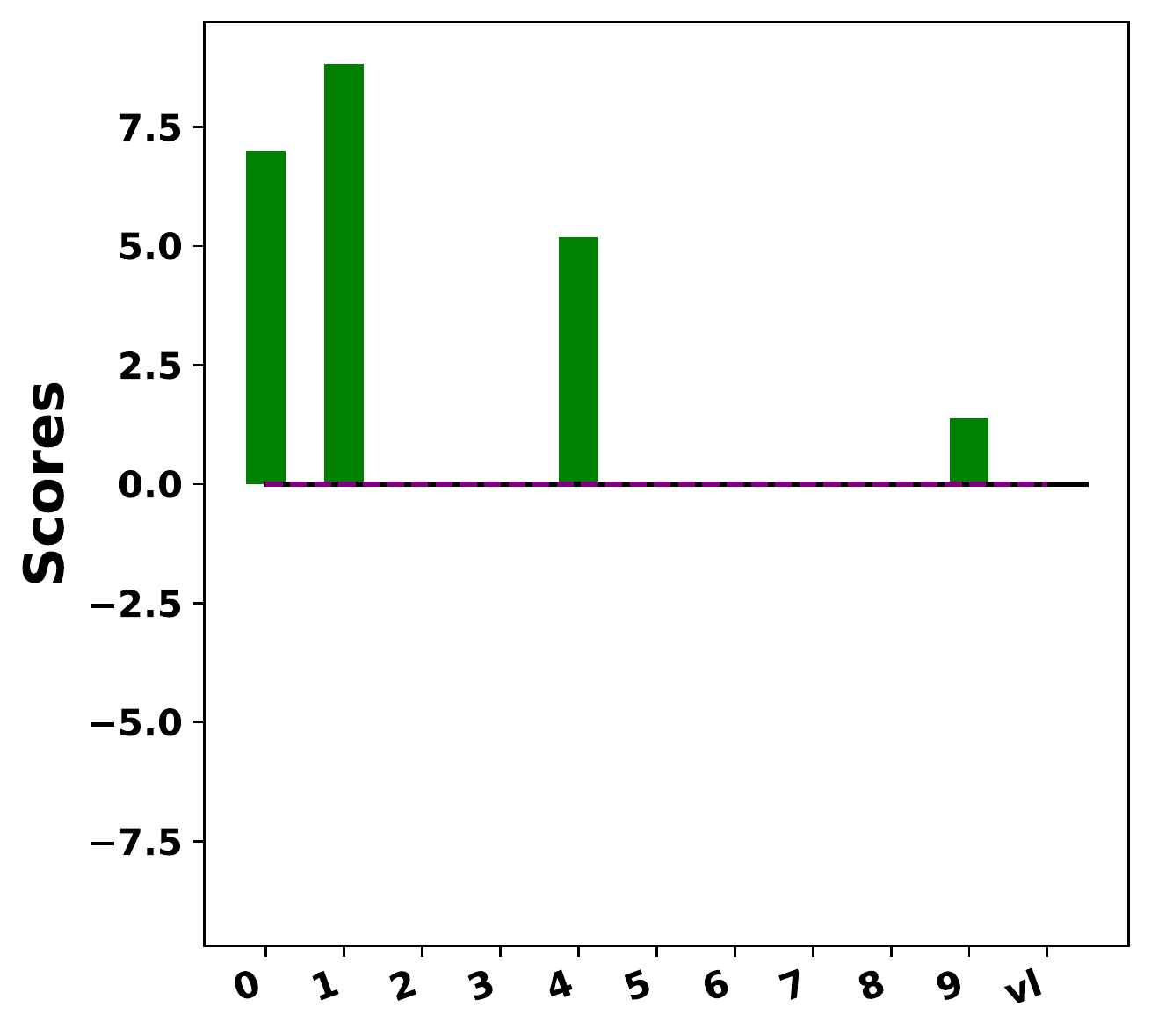} &
    \includegraphics[width=0.22\linewidth]{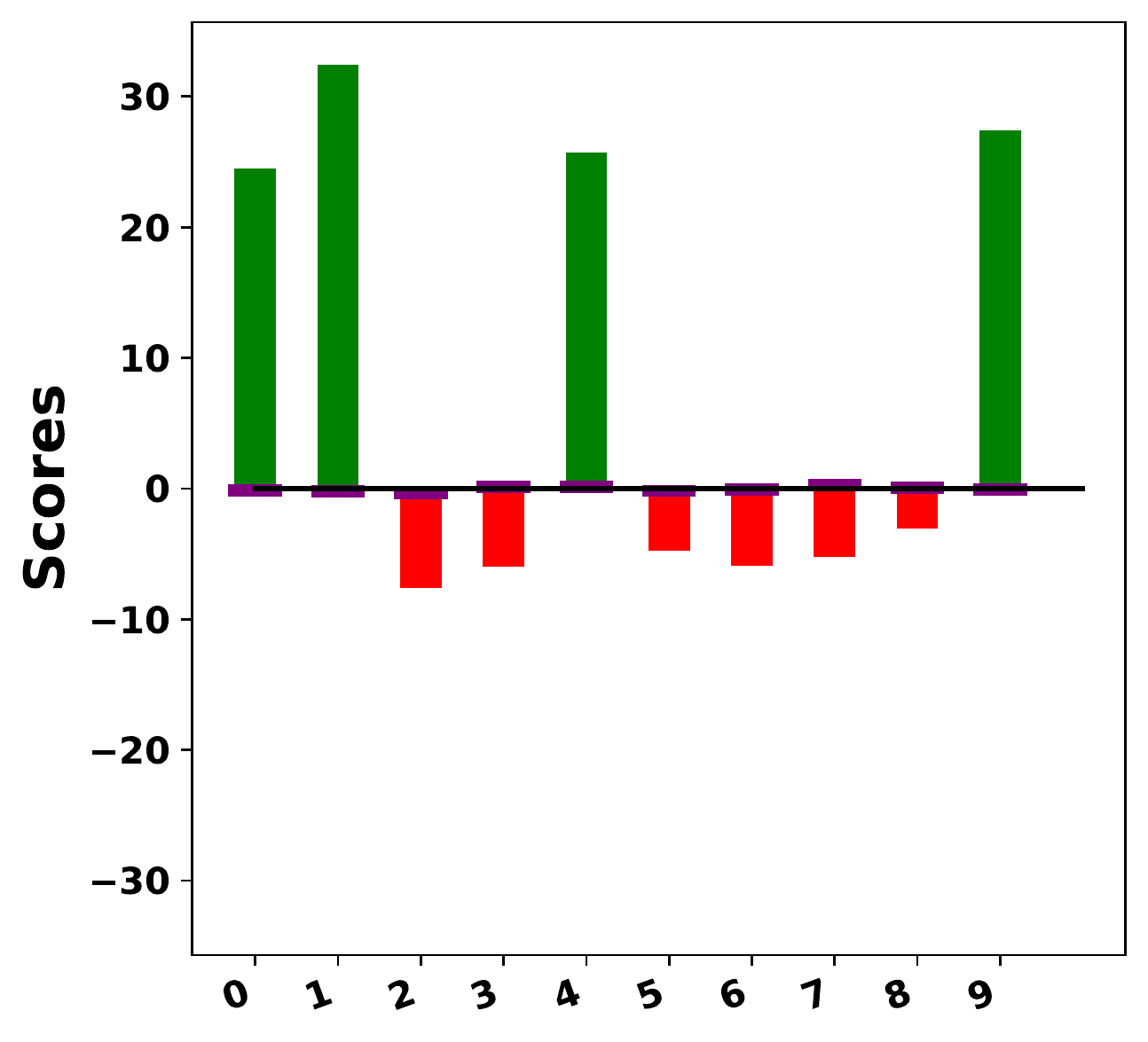} &
    \includegraphics[width=0.22\linewidth]{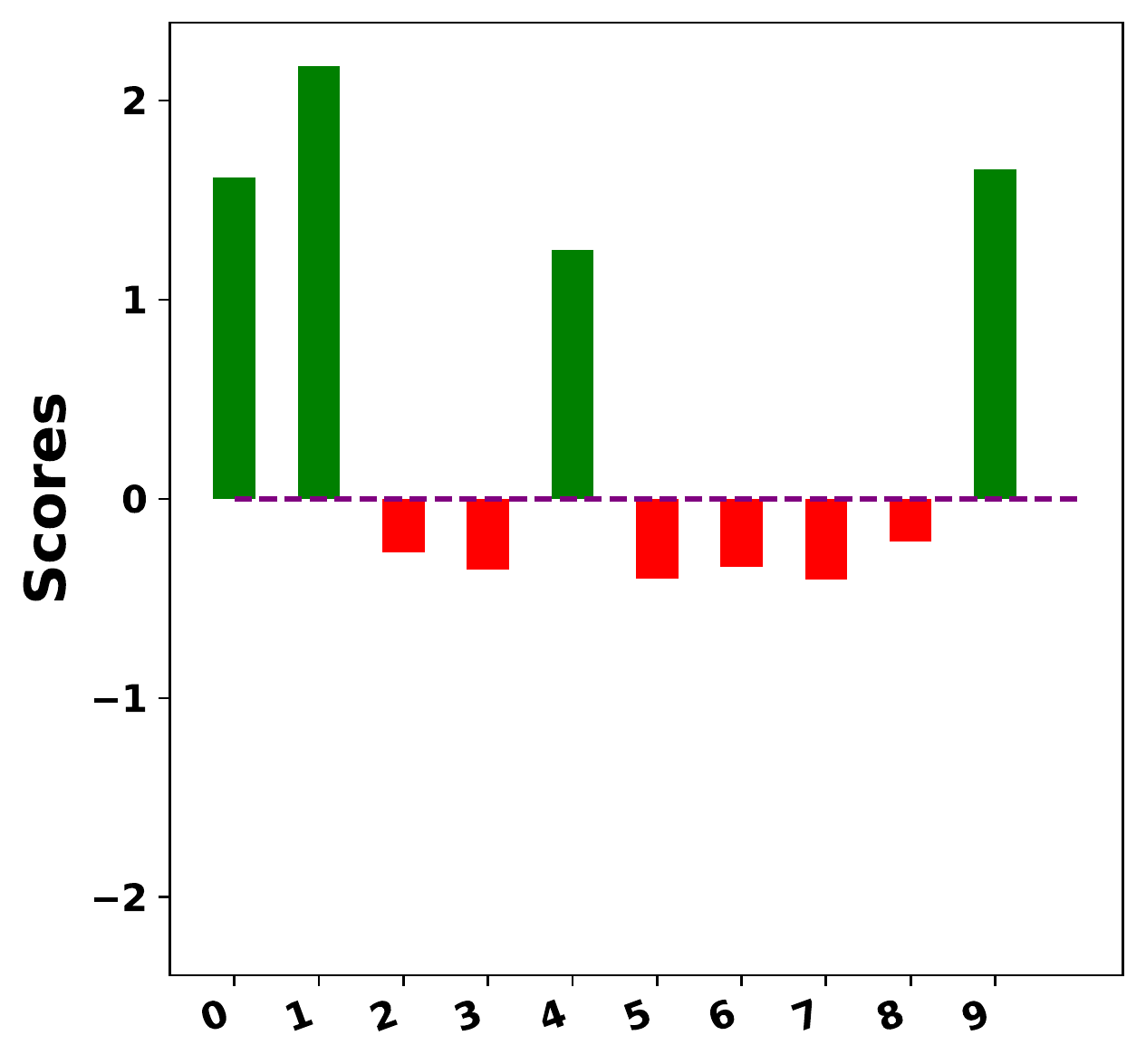} \\
    GT = \{ 1 $\succ$ 9 $\succ$ 0 $\succ$ 4 \}& \hspace{0.14in} \{ 1 $\succ$ 0 $\succ$ 4 $\succ$ 9 \}& \hspace{0.14in} \{ 1 $\succ$ 9 $\succ$ 4 $\succ$ 0 \} &  \hspace{0.14in} \{ 1 $\succ$ 9 $\succ$ 0 $\succ$ 4 \}\\
    &&& \\
    \includegraphics[width=3cm, height=3cm]{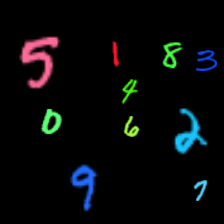}  &
    \includegraphics[width=0.22\linewidth]{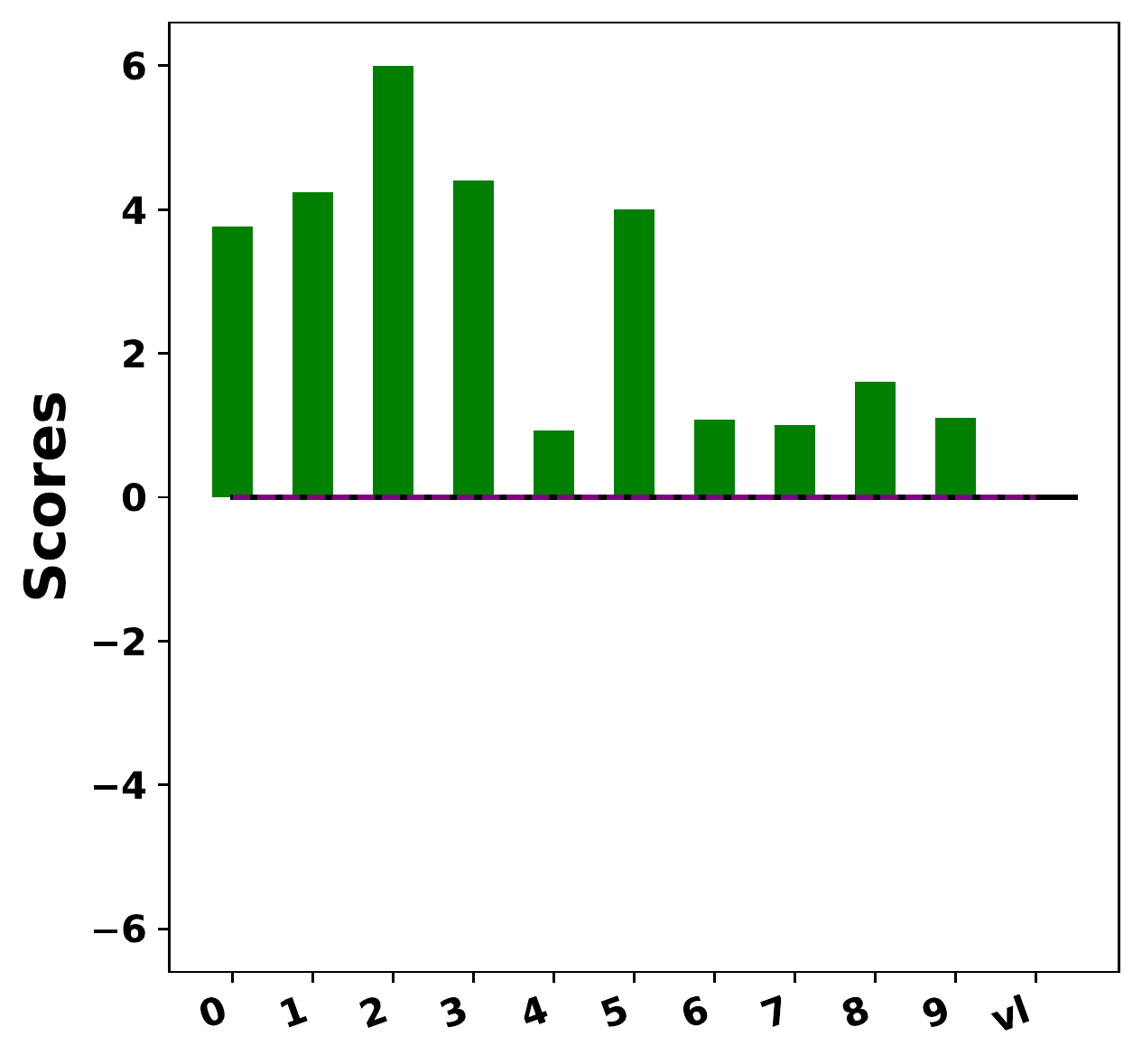} &
    \includegraphics[width=0.22\linewidth]{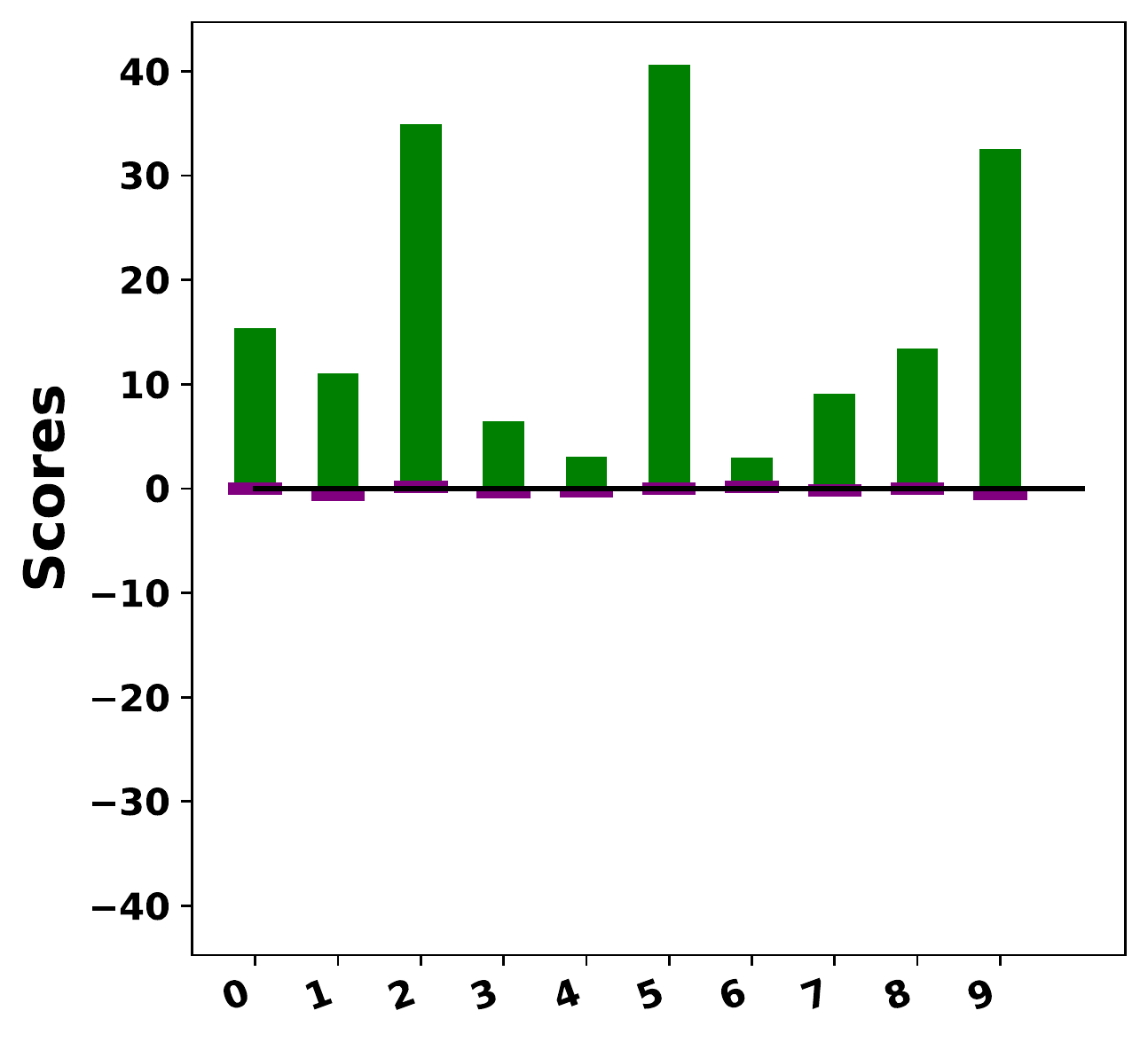} &
    \includegraphics[width=0.22\linewidth]{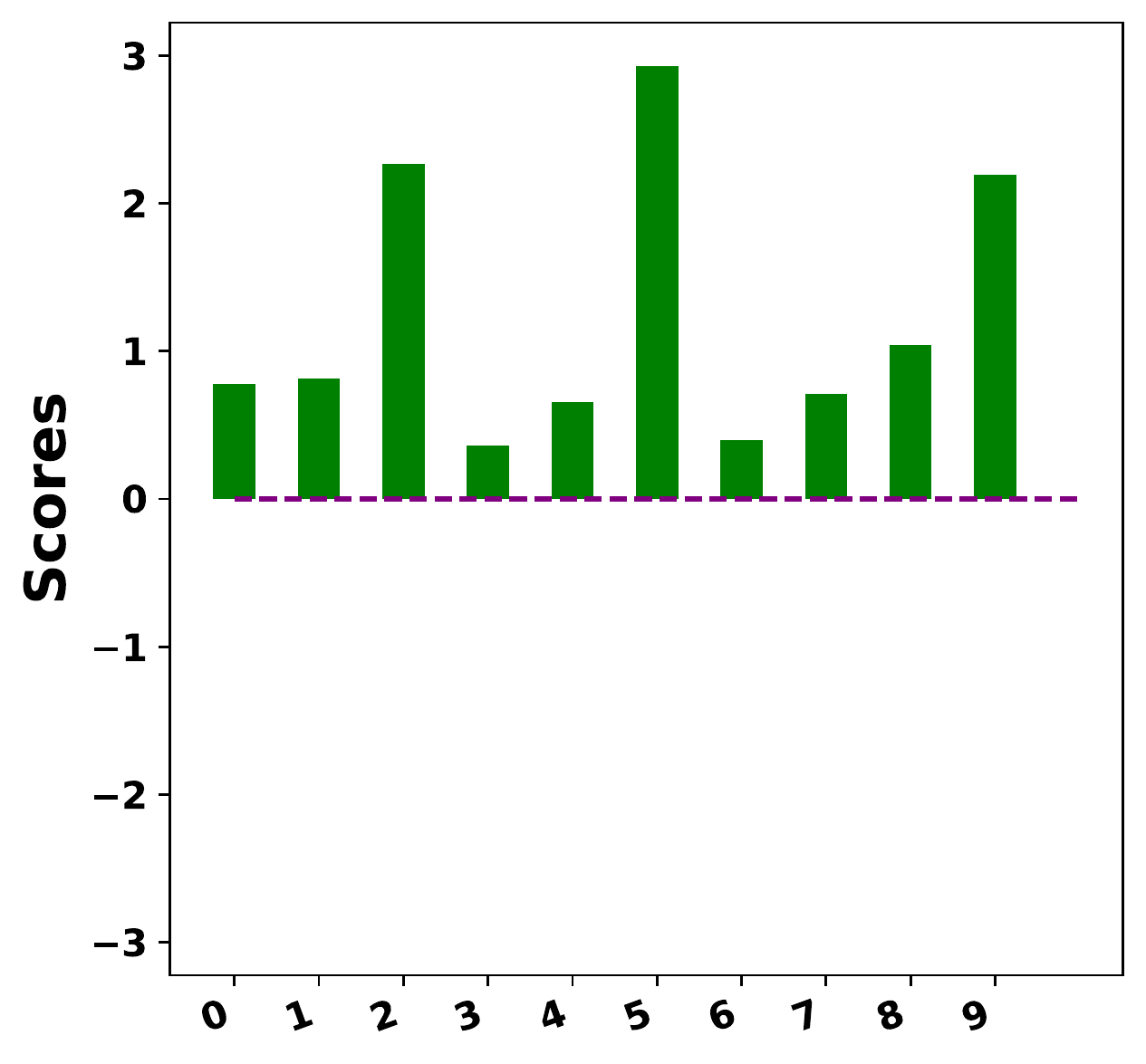} \\
    GT = \{ 5 $\succ$ 9 $\succ$ 2 $\succ$ 8 $\succ$ 1 $\succ$ &
    \{ 2 $\succ$ 3 $\succ$ 1 $\succ$ 5 $\succ$ 0 $\succ$ & 
    \{ 5 $\succ$ 2 $\succ$ 9 $\succ$ 0 $\succ$ 8 $\succ$ & \{ 5 $\succ$ 2 $\succ$ 9 $\succ$ 8 $\succ$ 1 $\succ$ \\
    
    0 $\succ$ 4 $\succ$ 3 $\succ$ 6 $\succ$ 7 \} &
    8 $\succ$ 9 $\succ$ 6 $\succ$ 7 $\succ$ 4 \}&
    1 $\succ$ 7 $\succ$ 3 $\succ$ 4 $\succ$ 6 \}& 
    0 $\succ$ 7 $\succ$ 4 $\succ$ 6 $\succ$ 3 \}\\
    &&& \\
    \end{tabularx}
    \caption{Samples from the test set of the Ranked MNIST Color dataset are given in the first column from left, with the corresponding ground truth labels denoted as GT. Each bar plot represents the predicted scores of baselines CRPC, LSEP and our method GMLR, in their respective columns.}
    \label{fig:color-bar}
\end{figure}

Bar graphs of the predictions of strong versions of CRPC, LSEP and GMLR are given for the four datasets: AVDP Dataset on Figure \ref{fig:AVDP-bar}, NSID on Figure \ref{fig:NSID-bar}, Ranked MNIST Gray-S on Figure \ref{fig:gray-bar} and Ranked MNIST Color-S on Figure \ref{fig:color-bar}. The graphs visualize the working principles of each baseline, where there is an additional virtual label for CRPC, learnable thresholds for each class for LSEP and the zero-point as the inherent threshold for GMLR, to perform binary classification where the positive predicted classes are denoted by green bars and negatives are in red, and the thresholding method is given in purple. The ranks of positive predicted classes are determined by the sorted scores and written below each graph with the $\succ$ operator denoting precedence.
\newpage

\section{Compute Resources}
\label{sec:apx8}

All the trainings, evaluations and experiments are done with an Intel i9-9900K CPU, 64 GB of RAM and two GPUs, NVIDIA TITAN RTX and NVIDIA RTX 2080-ti on an Ubuntu 18.04 machine. 

\section{Error Bars}
\label{sec:apx9}

The error bar table for real datasets is given in Table \ref{tab:error-bar-real}. It can be seen that Strong methods are consistently better compared to the weak methods. It should be noted that both of the real datasets consists of noisy and subjective labels which can affect the classification scores, while pairwise ranking explains the performance more accurately. Here GMLR manages to outperform the rest of the methods on most of the ranking metrics and yields comparable scores for classification, due to the noisy nature of the datasets we can say the difference between GMLR and LSEP for the classification does not strongly indicate that one is better while the other is not.

We ran each of the training setup given in Table \ref{tab:error-bar-real} for 5 times with random seeds, the scores on the table are the mean and standard deviation of the metrics for each random run. We are not providing error bars for Ranked MNIST datasets, due to resource limitations, since they are considerably larger compared to the real datasets.

\begin{table}[H]
    \caption{Error bar for real datasets NSID and AVDP. Mean scores and standard deviations of each baseline after 5 runs are reported. The annotations of scores are the same with Table \ref{tab:ranked_mnist_color}.}
     \renewcommand{\arraystretch}{1.2}
    \centering
    \resizebox{1.0\textwidth}{!}{ 
    \begin{tabular}{c|cccccc|cccccc} 
    \Xhline{3\arrayrulewidth}
    \multirow{2}{*}{Method} & \multicolumn{6}{c|}{NSID} & \multicolumn{6}{c}{AVDP}  \\
    \Xcline{2-13}{1\arrayrulewidth}
    & \multicolumn{1}{c}{$\tau_b \uparrow $} & \multicolumn{1}{c}{$ S \rho \uparrow$} & \multicolumn{1}{c}{$\gamma \uparrow$} & \multicolumn{1}{c}{HL $\downarrow$} & \multicolumn{1}{c}{M-1 $\downarrow$} & F1 $\uparrow$ & \multicolumn{1}{c}{$\tau_b \uparrow$} & \multicolumn{1}{c}{$ S \rho \uparrow$} & \multicolumn{1}{c}{$\gamma \uparrow$} & \multicolumn{1}{c}{HL $\downarrow$} & \multicolumn{1}{c}{M-1 $\downarrow$} & F1 $\uparrow$ \\
    \Xhline{2\arrayrulewidth}
    \multicolumn{1}{c|}{CRPC(W)}  & 57.68 $\pm$ 1.0 & 64.39 $\pm$ 1.0 & \multicolumn{1}{c}{70.43 $\pm$ 0.6 } & 20.73 $\pm$ 0.9 & 10.67 $\pm$ 0.6 & 67.97 $\pm$ 1.0 & 38.71 $\pm$ 1.0& 41.29 $\pm$ 1.0 & \multicolumn{1}{c}{42.83 $\pm$ 1.9} & 23.86 $\pm$ 0.4 & 33.43 $\pm$ 0.8 & 51.8 $\pm$ 0.8\\
    
    \multicolumn{1}{c|}{LSEP (W)} & 72.8 $\pm$ 0.6 & 77.0 $\pm$ 0.6 & \multicolumn{1}{c}{80.3 $\pm$ 0.5} & \underline{10.51 $\pm$ 0.2} & \underline{5.31 $\pm$ 0.6} & 80.35 $\pm$ 0.3 & 39.21 $\pm$ 2.8 & 41.3 $\pm$ 3.0 & \multicolumn{1}{c}{41.93 $\pm$ 2.7} &  \underline{19.21 $\pm$ 0.2} & 30.66 $\pm$ 1.4 & \underline{54.94 $\pm$ 1.2} \\
    
    \multicolumn{1}{c|}{GMLR (W)} & \underline{73.49 $\pm$ 0.9} & \underline{77.84 $\pm$ 1.0} & \multicolumn{1}{c}{\underline{80.96 $\pm$ 1.0}} & 10.64 $\pm$ 0.3 & 5.67 $\pm$ 0.5 & \underline{80.63 $\pm$ 0.6} & \underline{41.13 $\pm$ 0.4} & \underline{43.41 $\pm$ 0.4} &  \multicolumn{1}{c}{\underline{43.48 $\pm$ 1.8}} & 20.03 $\pm$ 0.2 & \underline{30.21 $\pm$ 1.0} & 54.31 $\pm$ 0.5\\ \hline
    
    \multicolumn{1}{c|}{CRPC(S)}  & 59.34 $\pm$ 0.5 & 65.87 $\pm$ 0.4 & \multicolumn{1}{c}{72.47 $\pm$ 0.5} & 19.36 $\pm$ 0.3 & 9.51 $\pm$ 0.4 & 69.44 $\pm$ 0.4 & 39.91 $\pm$ 0.6 & 42.43 $\pm$ 0.6 & \multicolumn{1}{c}{44.71 $\pm$ 1.2} & 23.11 $\pm$ 0.4 & 33.59 $\pm$ 0.7 & 52.66 $\pm$ 0.3\\
    
    \multicolumn{1}{c|}{LSEP (S)} & 71.95 $\pm$ 1.8 & 75.94 $\pm$ 1.9 & \multicolumn{1}{c}{79.25 $\pm$ 1.8} & \textbf{10.51 $\pm$ 0.2} & \textbf{5.49 $\pm$ 1.1} & 80.17 $\pm$ 0.4 & 40.95 $\pm$ 1.7 & 43.02 $\pm$ 1.8& \multicolumn{1}{c}{\textbf{44.98 $\pm$ 1.5}} & \textbf{18.94 $\pm$ 0.2} & \textbf{28.76 $\pm$ 1.4} & \textbf{55.42 $\pm$ 0.7}\\
    
    \multicolumn{1}{c|}{GMLR (S)} & \textbf{75.54 $\pm$ 0.4} & \textbf{78.86 $\pm$ 0.4} & \multicolumn{1}{c}{\textbf{82.89 $\pm$ 0.6}} & 10.87 $\pm$ 0.3 & 6.21 $\pm$ 0.5 & \textbf{80.21 $\pm$ 0.6} & \textbf{41.68 $\pm$ 1.6} & \textbf{43.84 $\pm$ 1.7} & \multicolumn{1}{c}{44.56 $\pm$ 2.8} & 19.88 $\pm$ 0.4 & 29.48 $\pm$ 1.2 & 54.12 $\pm$ 1.5\\ \hline
    \end{tabular}
    }
    \label{tab:error-bar-real}
\end{table}

\section{Variance Experiment}
\label{sec:apx10}
\begin{figure}[h]
    \centering
    \includegraphics[width=0.6\linewidth]{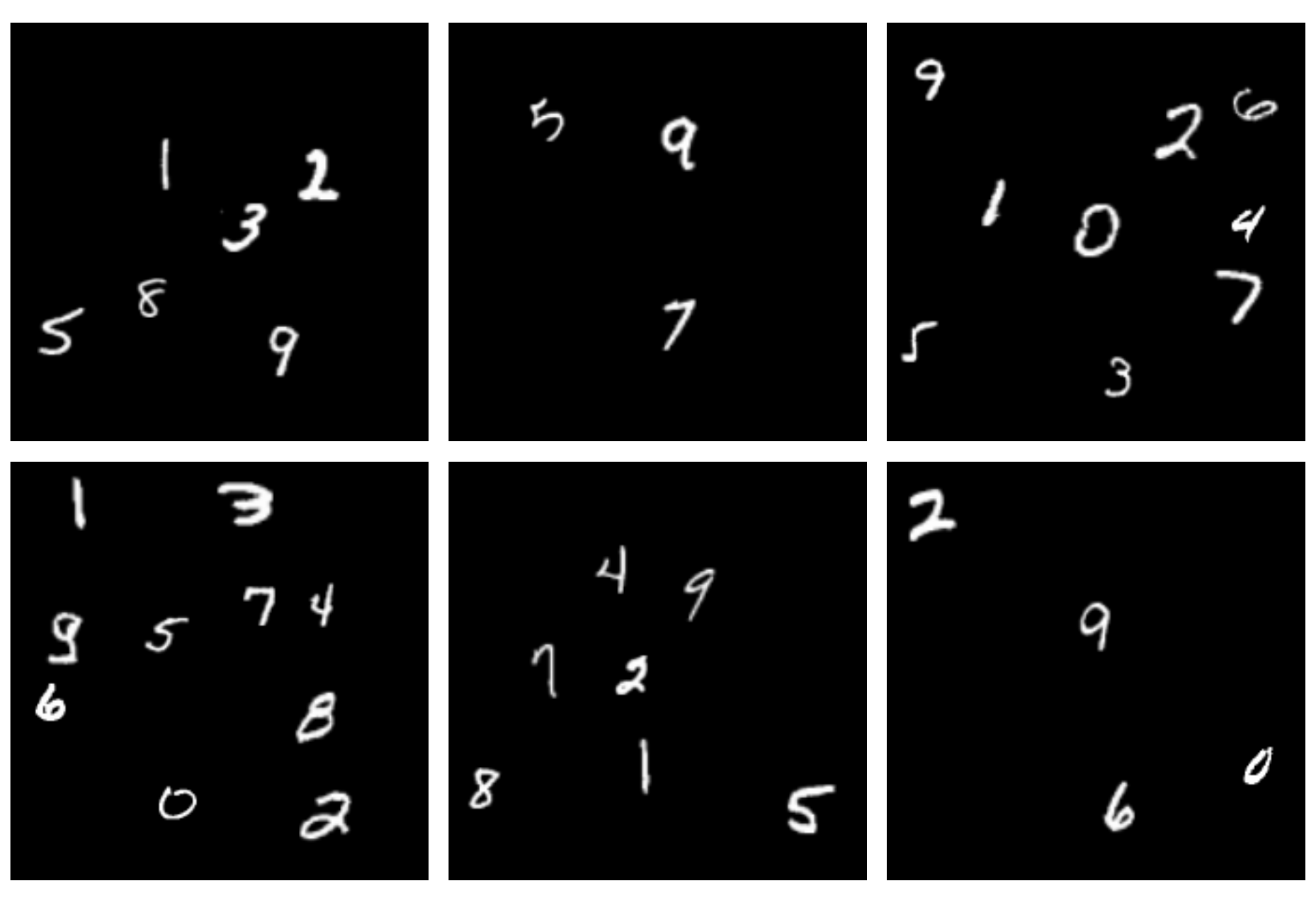}
    \caption{Sample images used in the small variance experiment.}
    \label{fig:small-variance}
\end{figure}

The variance experiments are conducted to observe whether our method distinguishes the small changes in importance factors and is able to assign ranks accordingly. The results of this experiment ran on a similar dataset to Ranked MNIST Gray-S which is provided in Table \ref{tab:small-variance} but instead of sampling the scale from $\mathcal{U}(1, 3)$ we sampled it from $\mathcal{U}(1, 1.5)$. Superior to other baselines, GMLR-Strong captures the small differences in importance factors for both classification and ranking task. Sample images created for the small variance experiment can be seen in Figure \ref{fig:small-variance}.
\begin{table}[H]
    \caption{Quantitative results of Variance Experiment on a variation of Ranked MNIST Gray-S where there are small changes in importance factors. The annotations of scores are the same with Table \ref{tab:ranked_mnist_color}.}
    \renewcommand{\arraystretch}{1.2}
    \centering
    \resizebox{0.6\textwidth}{!}{ 
    \begin{tabular}{c|cccccc} 
    \Xhline{3\arrayrulewidth} 
    \multirow{2}{*}{Method} & \multicolumn{6}{c}{Ranked MNIST Small Change}   \\
    \Xcline{2-7}{1\arrayrulewidth}
    & \multicolumn{1}{c}{$\tau_b \uparrow $} & \multicolumn{1}{c}{$ S \rho \uparrow$} & \multicolumn{1}{c}{$\gamma \uparrow$} & \multicolumn{1}{c}{HL $\downarrow$} & \multicolumn{1}{c}{M-1 $\downarrow$} & F1 $\uparrow$ \\
    \Xhline{2\arrayrulewidth}
    \multicolumn{1}{c|}{CRPC (W)}  & 50.57 & 60.94 & \multicolumn{1}{c}{59.81} & 14.74  & 0.19 & 88.05 \\
    \multicolumn{1}{c|}{LSEP (W)} &  61.50& 70.67 & \multicolumn{1}{c}{61.61} & \underline{0.39} & \underline{0.10}  & \underline{99.64}   \\
    \multicolumn{1}{c|}{GMLR (W)} & \underline{62.18} & \underline{71.31} & \multicolumn{1}{c}{\underline{62.32}} & 0.47 & \underline{0.12}  & 99.57   \\ \hline
    \multicolumn{1}{c|}{CRPC (S)}  & 62.46 & 74.30 & \multicolumn{1}{c}{74.77} & 21.27 & \textbf{0.16}  & 83.59  \\
    \multicolumn{1}{c|}{LSEP (S)} &  92.10& 96.50 & \multicolumn{1}{c}{92.61} & 1.32 & 0.25 &  98.79 \\
    \multicolumn{1}{c|}{GMLR (S)} & \textbf{92.65} & \textbf{96.77} & \multicolumn{1}{c}{\textbf{92.86}} & \textbf{0.52} & \textbf{0.15} &  \textbf{99.52} \\ \hline
    \end{tabular}
    }
    \label{tab:small-variance}
\end{table}

\begin{figure}[H]
    \centering
    
    \resizebox{\textwidth}{!}{
    \begin{tabular}{ccccc}
    
    Ranked MNIST Gray-S & Ranked MNIST Gray-B & Ranked MNIST Color-S & Ranked MNIST-B & Ranked MNIST Small Change \\
    
    \includegraphics[width=0.2\linewidth]{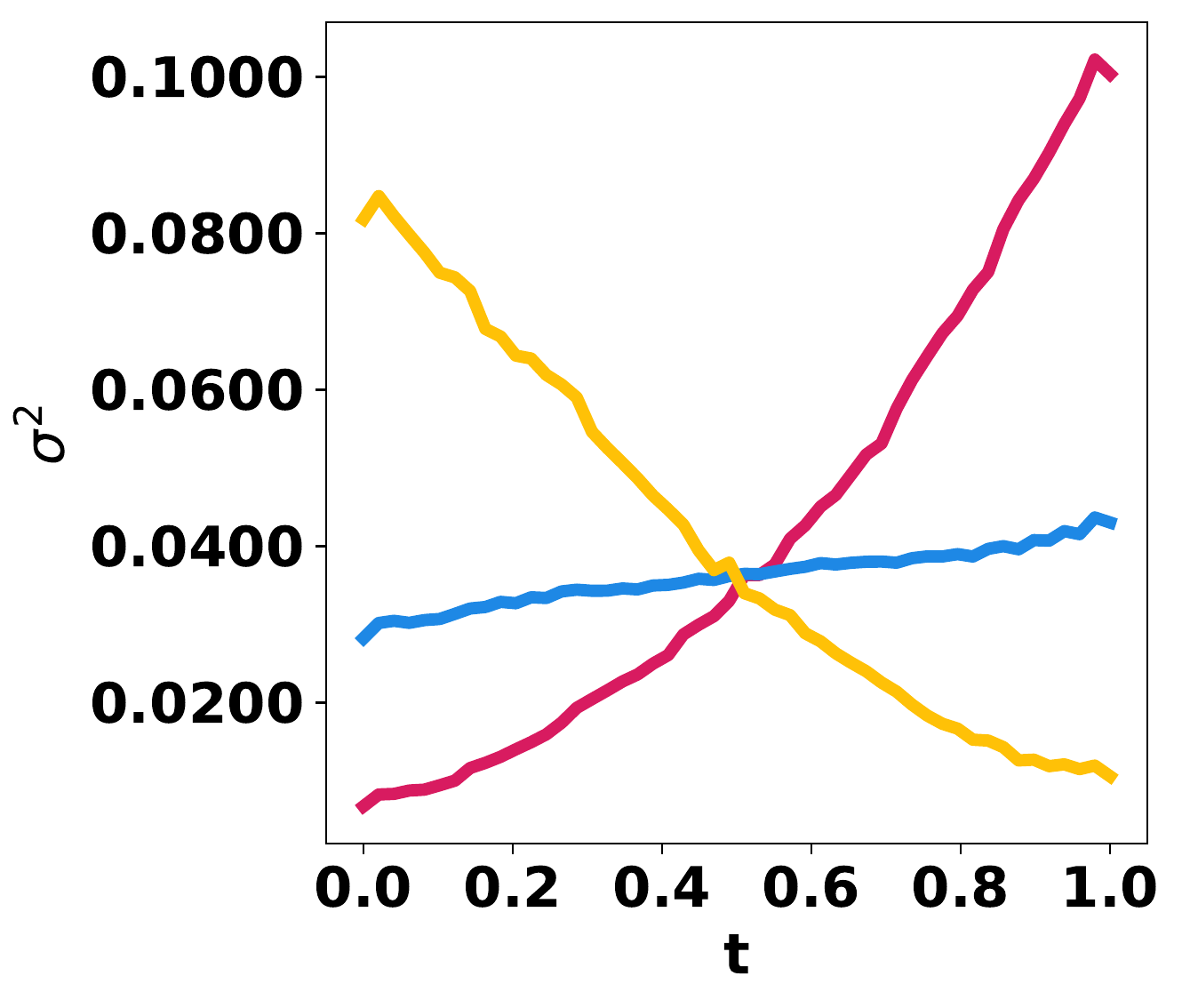} & \includegraphics[width=0.2\linewidth]{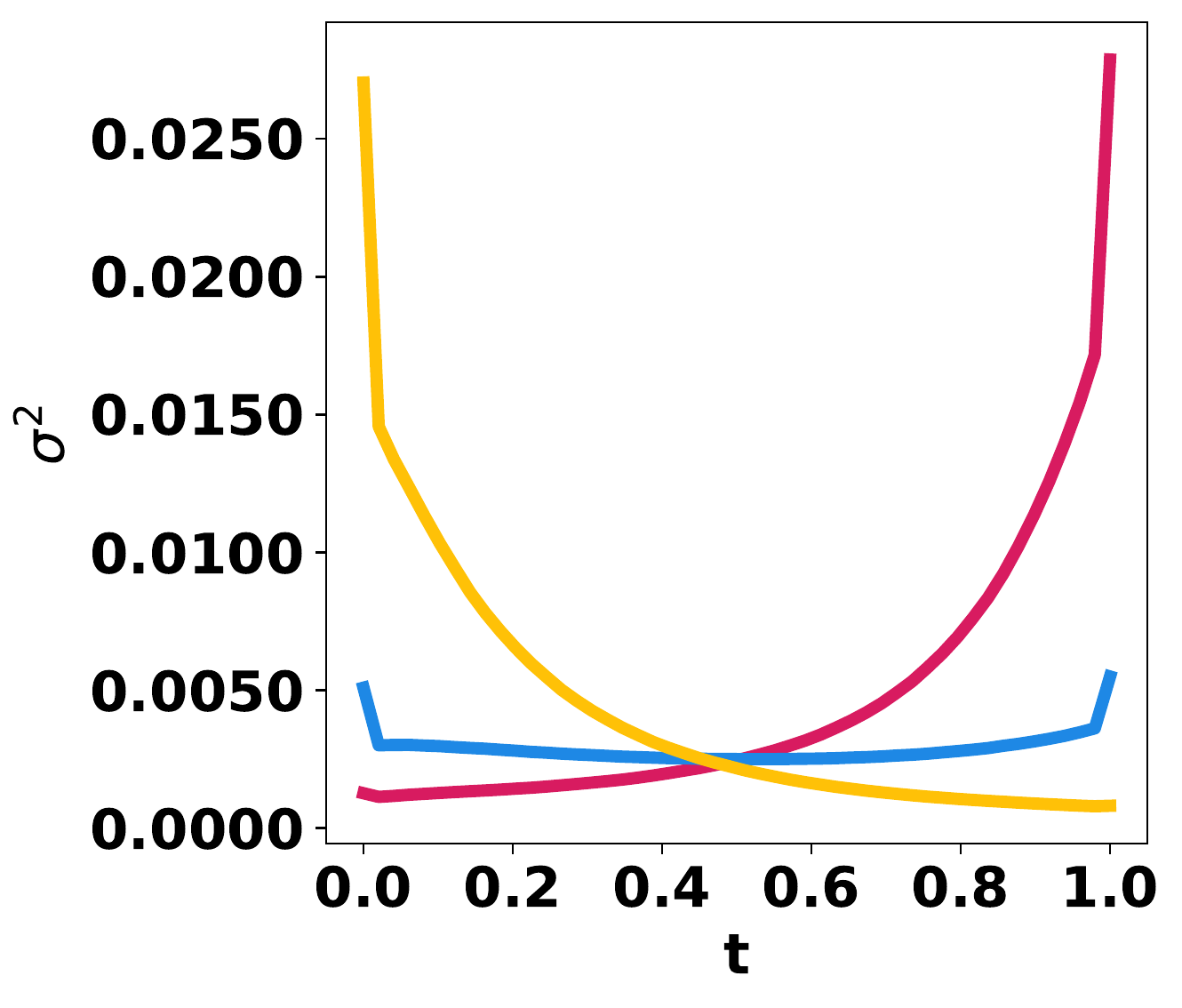} & \includegraphics[width=0.2\linewidth]{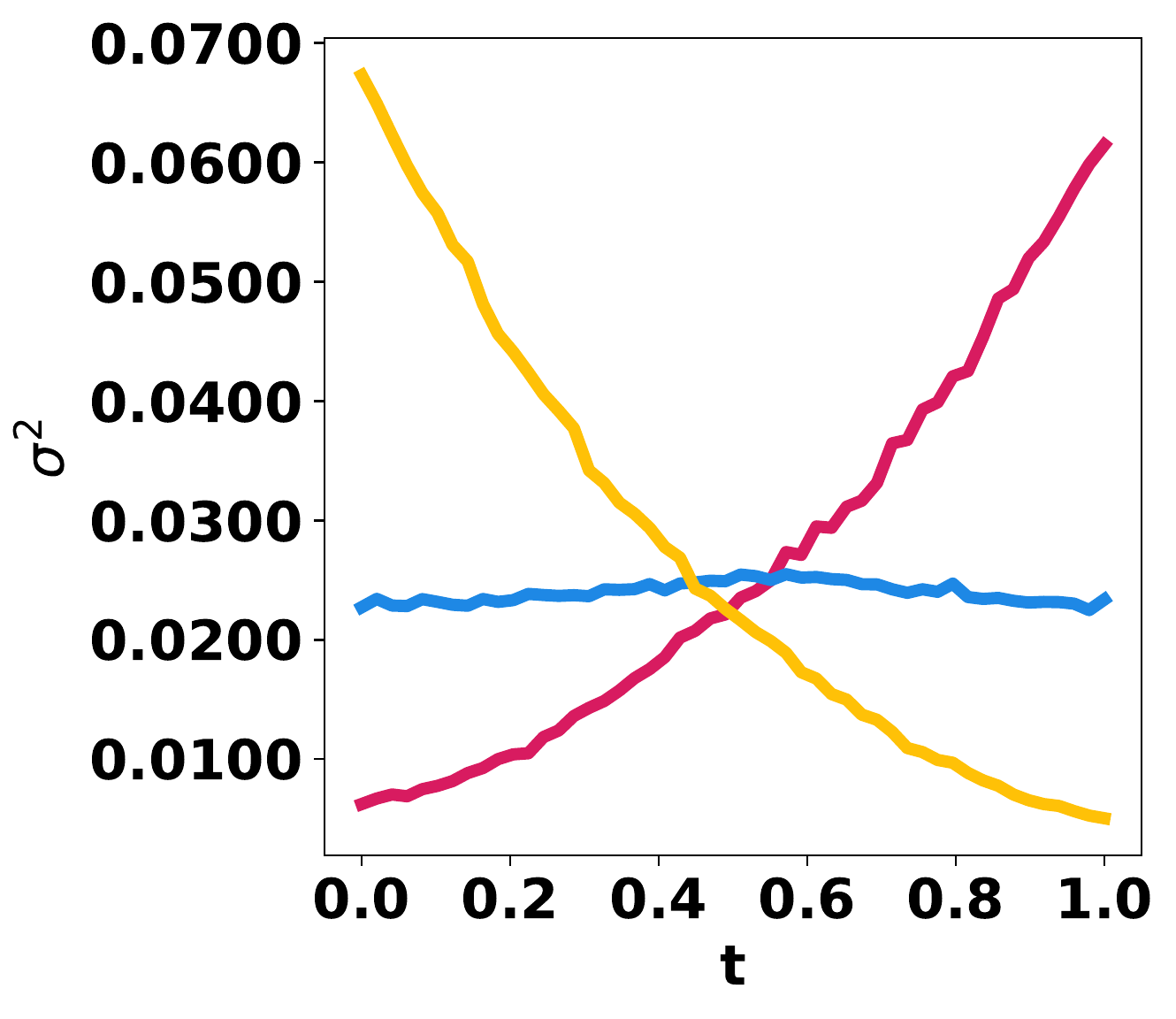} & \includegraphics[width=0.2\linewidth]{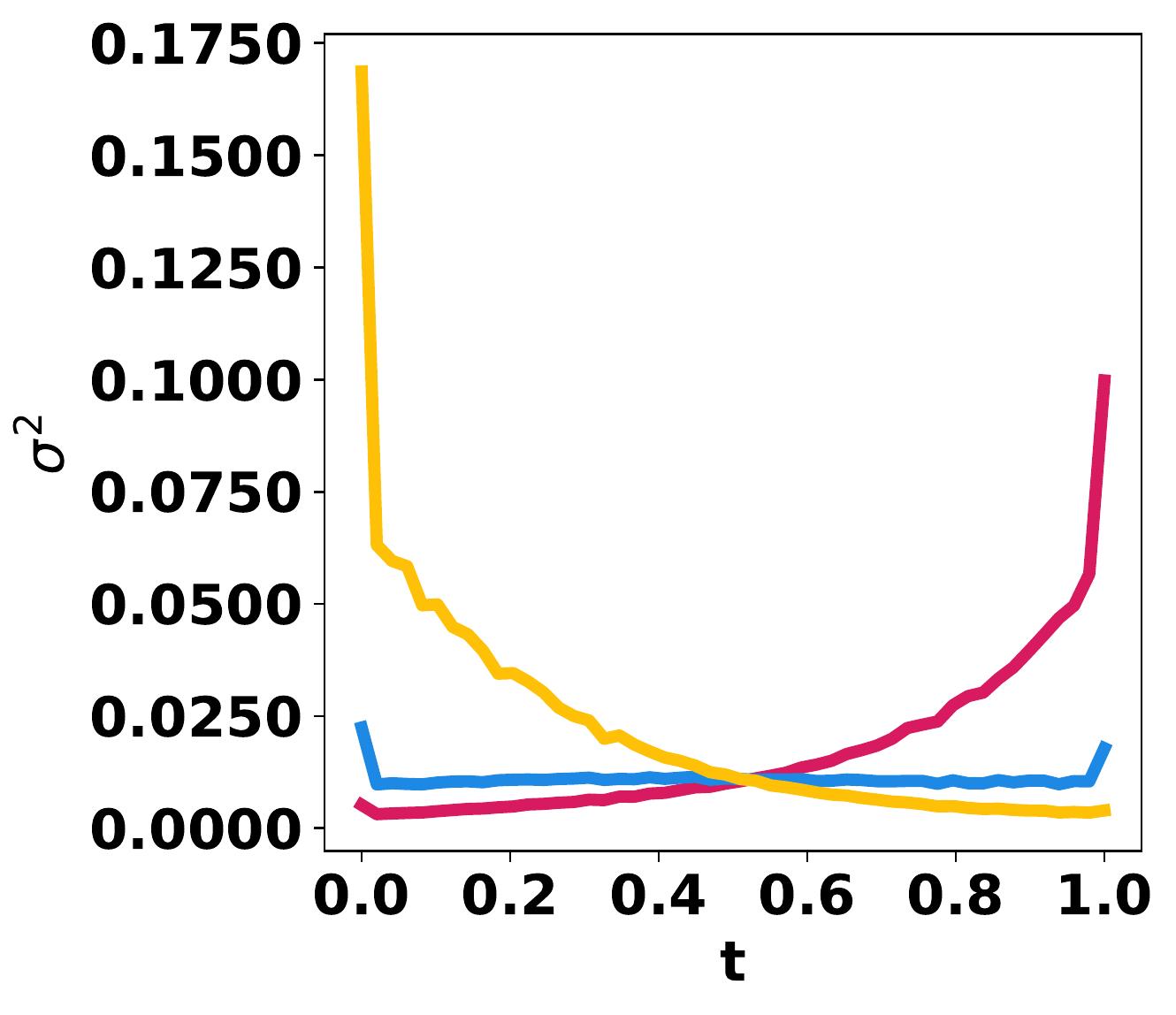} & \includegraphics[width=0.2\linewidth]{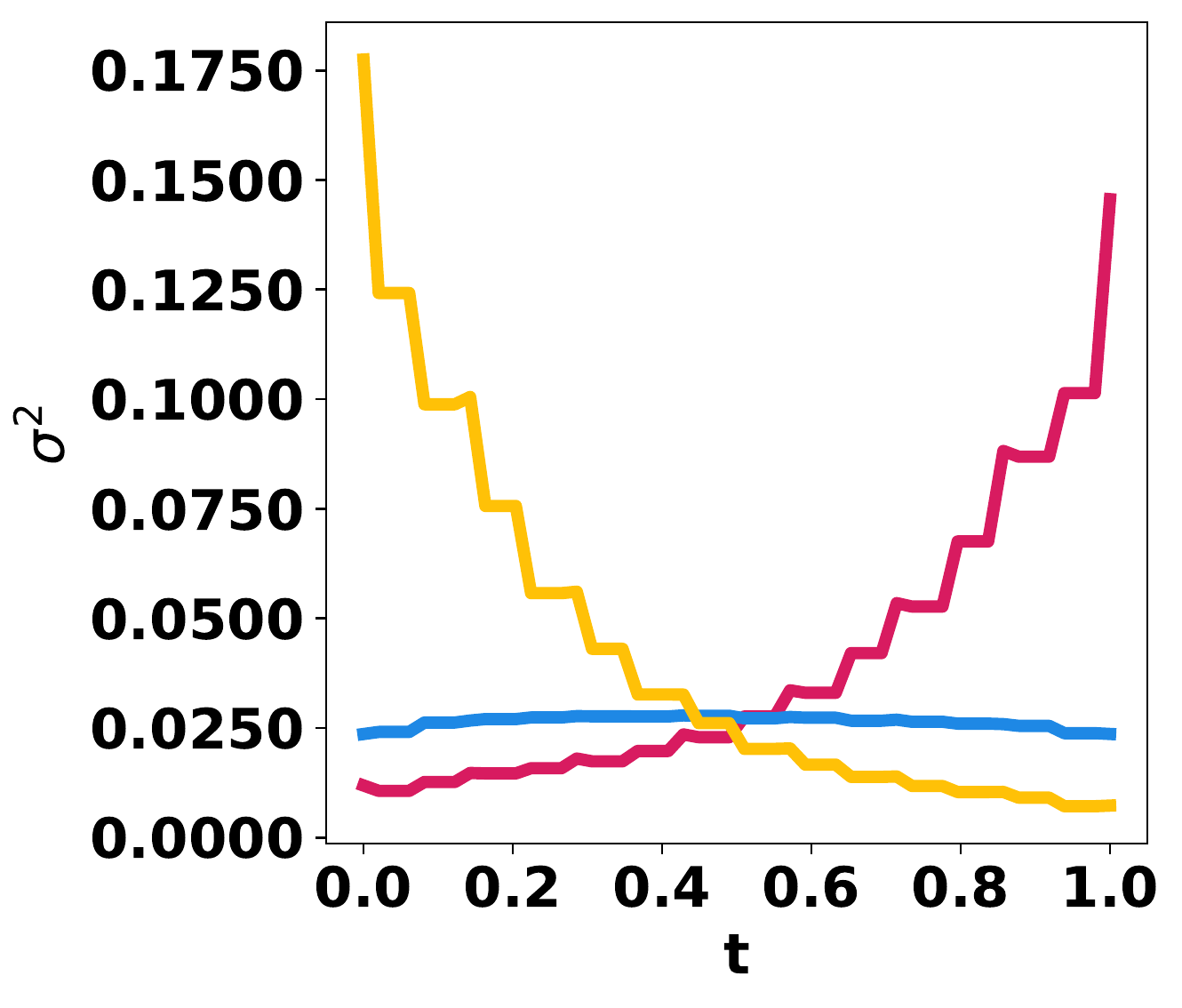}
    
    \end{tabular}
    }
    
    \caption{The change of variance for some of the experiments, conducted in Figure \ref{fig:interpolation1}, Figure \ref{fig:interpolation2}, Figure \ref{fig:interpolation3} and also the Small Change experiment.}
    \label{fig:var_interpole}
\end{figure}

We also conducted a similar experiment to the experiment given in Figure \ref{fig:interpolation1}, Figure \ref{fig:interpolation2} and Figure \ref{fig:interpolation3}, but instead of providing scores we provide variance as predicted by GaussianMLR. We provide the results in Figure \ref{fig:var_interpole} for Ranked MNIST Gray-S, Ranked MNIST Color-S, Ranked MNIST Gray-B, Ranked MNIST Color-B, and Ranked MNIST Small Change. As can be seen from the Figure \ref{fig:var_interpole}, even for the small change experiment the variance is learned in such a way that the digit with higher significance value always has higher variance. From this finding we hypothesize that this is due to a digit of high importance having a higher $\hat{\mu}$ and a digit of lower importance having a lower $\hat{\mu}$, the same percentage of error will make a higher impact on the loss for the more important digit and this error is being compensated by increasing the variance. 

\section{Extracted Significance Value Benchmark}
\label{sec:apx11}
\begin{figure}[H]
\hspace*{-1.5cm}
    \centering
    \includegraphics[width=1.2\linewidth]{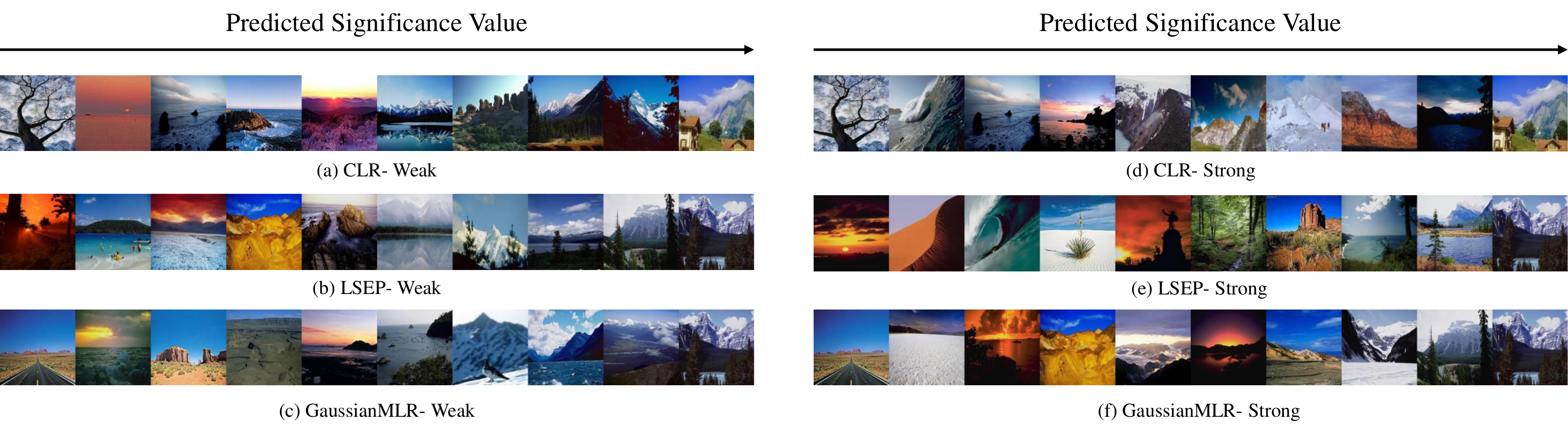}
    \caption{Mountain class benchmark for the extracted significance values.}
    \label{fig:mountain-bench}
\end{figure}

\begin{figure}[H]
\hspace*{-1.5cm}
    \centering
    \includegraphics[width=1.2\linewidth]{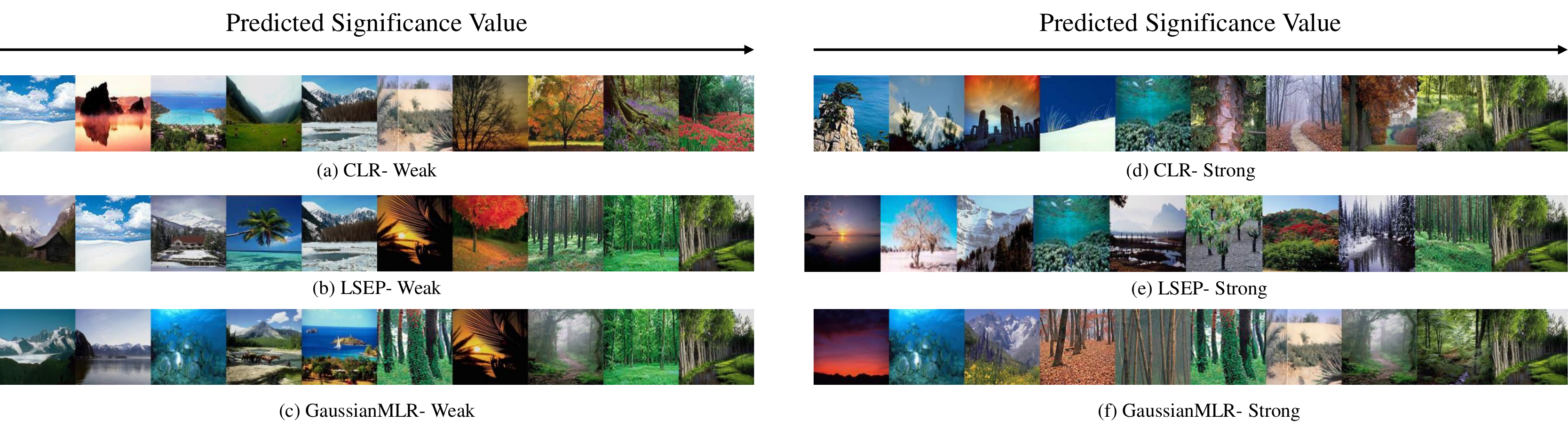}
    \caption{Plant class benchmark for the extracted significance values.}
    \label{fig:plant-bench}
\end{figure}

\begin{figure}[H]
\hspace*{-1.5cm}
    \centering
    \includegraphics[width=1.2\linewidth]{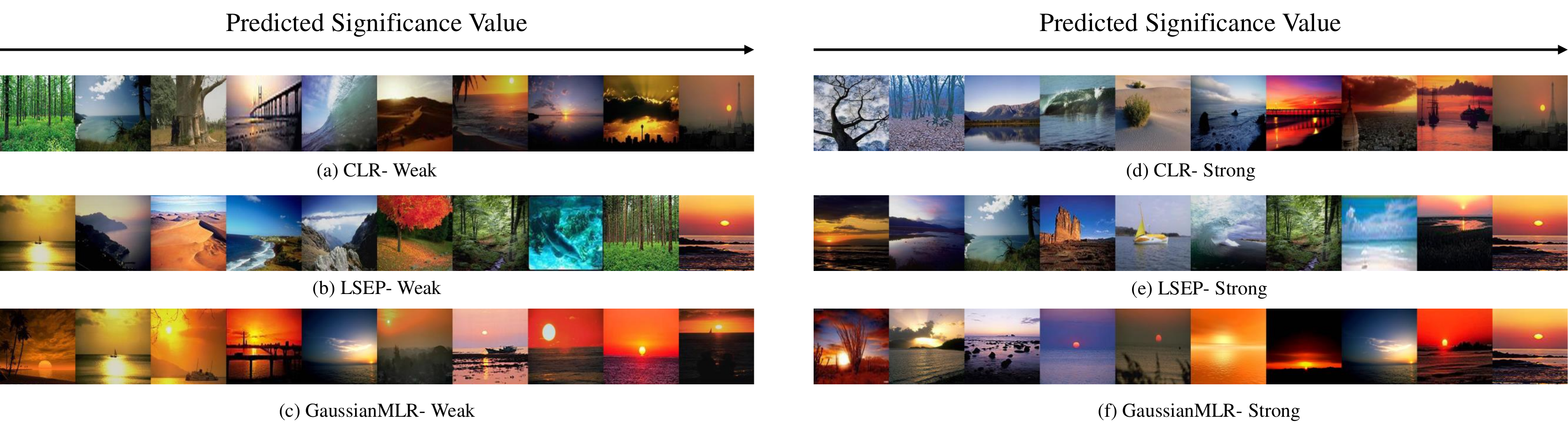}
    \caption{Sun class benchmark for the extracted significance values.}
    \label{fig:sun-bench}
\end{figure}

In order to benchmark against other baselines, we provide the results of the Extracted Significance Value experiment in Section 5.7 of the main paper for all baselines on NSID. For the three classes we provided for GMLR on the main paper, Figure \ref{fig:mountain-bench} compares the results for the mountain class, Figure \ref{fig:plant-bench} compares for the plant class and Figure \ref{fig:sun-bench} compares for the sun class. 
\newpage


\clearpage

\bibliographystyle{unsrt} 
\bibliography{references} 

\section*{Checklist}

\begin{enumerate}

\item For all authors...
\begin{enumerate}
  \item Do the main claims made in the abstract and introduction accurately reflect the paper's contributions and scope?
    \answerYes{}
  \item Did you describe the limitations of your work?
    \answerYes{See Section \ref{sec:conclusion}.}
  \item Did you discuss any potential negative societal impacts of your work?
    \answerYes{See Section \ref{sec:conclusion}.}
  \item Have you read the ethics review guidelines and ensured that your paper conforms to them?
    \answerYes{}
\end{enumerate}

\item If you are including theoretical results...
\begin{enumerate}
  \item Did you state the full set of assumptions of all theoretical results?
     \answerYes{}
        \item Did you include complete proofs of all theoretical results?
    \answerYes{}
\end{enumerate}

\item If you ran experiments...
\begin{enumerate}
  \item Did you include the code, data, and instructions needed to reproduce the main experimental results (either in the supplemental material or as a URL)?
     \answerYes{All relevant  data will be provided in the supplemental material.}
  \item Did you specify all the training details (e.g., data splits, hyperparameters, how they were chosen)?
     \answerYes{All details are provided in the Appendix \ref{sec:apx2}.}
        \item Did you report error bars (e.g., with respect to the random seed after running experiments multiple times)?
    \answerYes{See Appendix \ref{sec:apx8}.}
        \item Did you include the total amount of compute and the type of resources used (e.g., type of GPUs, internal cluster, or cloud provider)?
    \answerYes{All details are provided in the Appendix \ref{sec:apx7}.}
\end{enumerate}

\item If you are using existing assets (e.g., code, data, models) or curating/releasing new assets...
\begin{enumerate}
  \item If your work uses existing assets, did you cite the creators?
    \answerYes{}
  \item Did you mention the license of the assets?
    \answerYes{See Appendix \ref{sec:apx1}.} 
  \item Did you include any new assets either in the supplemental material or as a URL?
     \answerYes{The code to reproduce the Ranked MNIST datasets is provided in supplemental material}
  \item Did you discuss whether and how consent was obtained from people whose data you're using/curating?
     \answerYes{See Appendix \ref{sec:apx1}.} 
  \item Did you discuss whether the data you are using/curating contains personally identifiable information or offensive content?
     \answerYes{See Appendix \ref{sec:apx1}.} 
\end{enumerate}

\item If you used crowdsourcing or conducted research with human subjects...
\begin{enumerate}
  \item Did you include the full text of instructions given to participants and screenshots, if applicable?
    \answerNA{}
  \item Did you describe any potential participant risks, with links to Institutional Review Board (IRB) approvals, if applicable?
    \answerNA{}
  \item Did you include the estimated hourly wage paid to participants and the total amount spent on participant compensation?
    \answerNA{}
\end{enumerate}

\end{enumerate}


\end{document}